# Data Driven Computational Model for Bipedal Walking and Push Recovery

*A thesis Submitted*
In Partial Fulfillment of the Requirements for the Degree of Philosophy

**Vijay Bhaskar Semwal**
**RS139**

Under the supervision of
**Prof. G. C. Nandi**

To

The Department of Information Technology
(Robotics and Artificial Intelligence Laboratory)
Indian Institute of Information Technology Allahabad, Allahabad.
June, 2016

# INDIAN INSTITUTE OF INFORMATION TECHNOLOGY

# ALLAHABAD

(A Centre of Excellence in Information Technology Established by Govt. of India)

## CERTIFICATE

Date: 30/06/2016

This is to certify that the work titled **"Data Driven Computational Model for Bipedal Walking and Push Recovery"** is an original piece of work done by me under the supervision of Prof. G. C. Nandi.

I certify that the submitted work has not been undertaken elsewhere and is free from plagiarism as per the plagiarism report of PRC cell of IIIT, Allahabad.

The work may therefore be accepted in fulfillment of the thesis requirements for the Degree of Doctor of Philosophy examination of IIIT- Allahabad.

**(Vijay Bhaskar Semwal)**
**Enroll. No.: RS-139**
**IIIT- Allahabad.**

**Work Certified and Recommended for Examination:**

**(Prof. G. C. Nandi)**

**Professor, IIIT-Allahabad**

## Abstract

The bipedal walk is considered as one of the most difficult tasks learned by human beings. The bipedal is more suitable than wheeled robot to work in un-structured terrains due to dexterity and ability to step over uneven surface. This is the reason for considering the bipedal walk though it is inherently unstable and daunting. The human walk is a complex task learned by human. A human baby takes almost a year for a stable gait. The robotic limbs, which imitate the human locomotion, give birth to a bipedal robot. The emergence of humanoid robot has benefited the society due to the benefits in helping the amputee to recover their gait and assistance of elderly people. The modern robots available in the market cannot walk efficiently due to the limitation of flat foot and bending knees. Such robots consume more energy and unstable in unstructured environments. That is the reason we have not seen any robot which can work outside the controlled environments like laboratories.

In this research, we have developed the data driven computational walking model to overcome the problem with traditional kinematics based model. Our model is adaptable and can adjust the parameter morphological similar to human. The human walk is a combination of different discrete sub-phases with their continuous dynamics. Any system which exhibits the discrete switching logic and continuous dynamics can be represented using a hybrid system. In this research, the bipedal locomotion is analyzed which is important for understanding the stability and to negotiate with the external perturbations. We have also studied the other important behavior push recovery. The Push recovery is also a very important behavior acquired by human with continuous interaction with environment. The researchers are trying to develop robots that must have the capability of push recovery to safely maneuver in a dynamic environment. The push is a very commonly experienced phenomenon in cluttered environment. The human beings can recover from external push up to a certain extent using different strategies of hip, knee and ankle. The different human beings have different push recovery capabilities. For example a wrestler has a better push negotiation capability compared to normal human beings. The push negotiation capability acquired by human, therefore, is based on learning but the learning mechanism is still unknown to researchers. The research community across the world is trying to develop various humanoid models to solve this mystery. Seeing all the conventional mechanics and control based models have some inherent limitations, a learning based computational model has been developed to address effectively this issue. In this research we will discuss how we have framed this problem as hybrid system.

In the first part of the thesis, we have discussed the inherent challenges associated with bipedal robot. We have also presented the overview how computational model are suitable then kinematics based model.

In the second chapter of the thesis, we have presented the analysis of the available bipedal robot technology and bipedal model.

In chapter third of the thesis, we have given the definition of the bipedal technology. We have presented all the important terminologies used with bipedal gait and push recovery.

In chapter fourth of the thesis, we have presented our innovative idea about the data collection for gait and push recovery for different real subjects. The subjects we have considered were students of our institute having 10 left and 25 right handed persons. We have captured data using indigenously developed wearable device HMCD (Human motion capture device) as well as using HLPRDCD (Human Locomotion &Push Recovery Data Capture Device).

In chapter fifth we have framed the bipedal walking as hybrid system and developed the vector fields for each seven sub phases of bipedal walk. The major contribution of the research is the development of computational walking model and generation of joints trajectories to each sub phases of gait for all the six joints (hip, knee & ankle). The model has been configured as a rocking block and various parameters have been fitted according to different subjects. We have compared the vector field generated joints trajectories with hybrid automata model and HOAP2 model. Finally we have applied our joints trajectories with HOAP2 robot. We have also presented the cellular automata for state predication of bipedal gait.

In the chapter sixthof thesis, we have classified the human gait and push recovery data using various machine learning techniques.

In the seventh chapter of thesis, we have proposed a push recovery capable hierarchically type-1 fuzzy logic controller and compared the human push recovery data with model generated data and we have proved that the fuzzy logic based controller is fast to adapt and is more generalised. It is fast and less computational intensive.

Useful conclusion based on our research experiment, limitation and future recommendation have been made in the chapter no 8.

# Acknowledgements

While writing my Ph.D. dissertation I have been extremely fortunate to be surrounded by wonderful people, whose very special contribution to this dissertation I would like to acknowledge.

First, I would like to express my deepest gratitude to my thesis supervisor, **Prof. G C Nandi**, for his extraordinary support and guidance at IIITA. I would like to thank him for introducing me to robotics, for opening my eyes to this wonderful world, and for his energy, enthusiasm and dedication which were a constant source of inspiration. I realized the importance of honest research and integrity taught by Prof. G. C. Nandi. For these and for many other reasons I will be always indebted to him.

I would like to give my very special thanks to the people closest to my heart: my family. My deepest love and gratitude to my parents for being the most wonderful parents in the world, I can never thank them enough for the motivation, support and for the many sacrifices they made so that I can achieve the best in my life.

I would like to thanks my all friends who helped me during my ups & down of Ph.D. journey. Last but not the least I bow to the God Almighty for making these studies a successful one.

**Vijay Bhaskar Semwal**
**IIIT,Allahabad**

## Table of Contents

## List of figures:

## List of Tables:

## List of Abbreviations

| | |
|---|---|
| CMP | Centroidal Moment point |
| COG | Center of Gravity |
| COM | Centre of Mass |
| COP | Centre of Pressure |
| DSP | Double Support Phase |
| SSP | Single Support Phase |
| GRF | Ground reaction Force |
| ZMP | Zero Moment Point |
| CA | Cellular Automata |
| HA | Hybrid Automata |

# Chapter 1: Introduction

## 1.1 Essence of Bipedal Robots and its applicability

BIPEDAL robots are cyber physical system. The design of bipedal robot is inspired from the human walk which must have human like movements. Human Walking is divided into two phases – swing and stance phase. The structure similar to human is considered as anthropomorphic [1]. From the very beginning Scientists and roboticists are planning to build multipurpose and efficient robots, not only to work in industry on assembly lines but also to replace humans in dirty household and other dangerous, hazardous works[2][3].

Researchers gave more importance to humanoid robots because of their resemblance to human structure allowing interaction with made-for-human tools or environments. Humans have adapted their structure in a very long evolutionary process [4]. However human structure is inherently unstable and resembles to an inverted pendulum that is why human babies take almost 10 months in learning to balance their body, whereas animals (quadruped) do the same in few hours because of their stable four-legged structure [5]. The presently available most of the bipedal robots walk with flat foot and bent knee which are more energy consuming [6].

The bipedal robot development which can walk on uneven terrain is one of the challenging fields of research. The study of bipedal walk will help in the development of more sophisticated humanoid robots. The human walk is an evolutionary process. It decays and grows with age and is based on complex coordination between motor action and muscle. So, the study of bipedal walk is important for the understanding the problem of elderly and disabled people. The human walk is the combination of different discrete sub phases with continuous dynamics therefore such behavior can be studied as hybrid system [7] [8]. The bipedal robot development industries given much needed boost for the study of bipedal locomotion.

Bipedal humanoid robots are having much importance because they can climb and climb down on stairs, can walk on narrow places, can jump and do almost all the work that humans can. However they have an unstable structure like an inverted pendulum, programming their jobs and tasking their locomotion is a high dimensional and non-linear problem. Moreover, considerable work has been done on locomotion, and to make the humanoids respond to external forces from environment and recover from fall and push-pull, but it is not so much efficient. It is expected that the human-robot interaction is and will be an area of immense research in near future. Moreover this field is full of application in studying the human gait pattern as biometric unique pattern and for development of a stable walking pattern for developing the prosthesis legs [9]. Also it can be used in push recovery study. Apart from making bipedal robots push recovery study will help the person with disability and elderly people to move with confidence and stability [10]. The push is a very common phenomena experienced by any bipedal in cluttered environment [11].

## 1.2 Motivation

The main motivation to pursue the bipedal robot is to understand human being's capability of locomotion, push recovery and implement the same capability in the design of bipedal robot. More specifically the motivation can be formulated in the following way:

1) - To be compatible with human environment bipeds are preferred even though they are inherently unstable.

2) - Study can help elderly or persons with disabilities walk with more stability and with confidence.

3) -To understand what causes humanoids to fall, and what action can be taken.

4) - Human locomotion is outcome of years of evolution, so it is worth to pay attention how they can walk with straight leg and consume less energy. Can we must the same mechanism for humanoid robots?

## 1.3 Challenges associated with Human walk

The major challenges with bipedal is energy efficient stable walking. So far, the biped robots available are of flat footed with bending knees which consume more energy and slow. To achieve the stable walk and understand it perfectly we have divided the bipedal walk into different linear sub phases. Walk is considered as moving with a moderate pace by lifting alternative foot up and down when one foot is lifting up and another foot is kept on ground [12]. Therefore one foot must be on ground at any time during walking. There are two types of walk, one is static and another one is dynamic walk. In Static walk, the projection of centre of mass (CoM) never crosses the support polygon of foot during the walk whereas during dynamic walk, the projection of CoM leaves the support polygon for some point of time. We perform the dynamic walk in our daily life [13]. The walking style for which we are familiar can be realized as a dexterous control which is essentially unstable. In dynamic walk to prevent toppling the swing leg is brought forward to avoid fall. Such strategy in a walk allows fast walking and less energy consumption for each gait [14]. The statically stable walk can be performed by first shifting body weight to foot and next stance of leg then swing the leg so that the ground contact remains at all-time. The Bipedal locomotion is very complex problem due to inherent problem of nonlinear dynamics, discretely changing in dynamics, multivariable System, underactuated System and changing environment [15][16].

The bipedal robots have following five major challenges and constraints [17][18]:

- It is highly non-linear and unstable, the classical controller cannot use directly.
- The gait cycle consists of two hybrid phases, one is statically stable double support phase and another is statically unstable single support phase. So it is requirement of suitable controller.

- The human walking has many Degrees Of Freedom (DoF) in 3-d space. The interaction between the DoF and the co-ordination of multi joints movement is required many variable and complex.
- Underactuated System: Unlike humans the bipedal robots cannot have under actuation during swing phase due to stability issue.
- Changing Environment.

## 1.4 Problem Statement

As on date due to its inherent complexities bipedal robots are not efficient to work outside laboratory environment i.e. unstructured environment and they are controlled as fully-actuated system, which is not energy efficient. The major reason of instability of the existing kinematic model is the limitation of making perfect biped model with all correct structural, frictional and other nonlinear parameters.

## 1.5 Hypothesis

We believe the problem being addressed so far using conventional mechanics based model and automated control theory can effectively be addressed using data driven computational theory. Throughout the thesis we tried to validate this hypothesis using hybrid and cellular automata theory.

## 1.6 Major Contributions of the thesis

We have developed the computational model for prediction, formal verification and analyses of joint trajectories of bipedal locomotion using theoretically enriched hybrid automata technique for modelling. The major contributions of the thesis are:

- Development of sophisticated Human Motion Capture Device (HMCD) and Human Locomotion and Push Recovery Data Capture Device(HLPRCD) devices to capture the human gait and push recovery data [19].
- Analyses of bipedal push recovery data and establish correlation between applied forces and push recovery strategies [20].
- Establishment of vector fields and development of Hybrid Automata Model for bipedal walk and generation of joints trajectories [21].
- Design of cellular automata for gait state prediction [21].
- Classification of push recovery data using deep neural learning network and comparison using other machine learning techniques [22].
- Development of a fuzzy logic based push recovery capable controller [23].

## 1.7 Aims

The aim of this research is to develop a computational bipedal model, more specifically-

- Development of a technique which can help robots to walk in unstructured environment efficiently and able to recover from an external impact.

- We studied bipedal walking as a way to understand human walking and then to use our understanding to design a better control strategy for bipedal robot.
- Verification of Computational Bipedal model using formal method of theoretical computer science.
- To understand what causes humanoids to fall, and what can be done to avoid it.
- To develop technique which can help robots to recover from push without falling?
- To validate the technique for generating walking trajectories on HOAP2 robot.

### 1.8 Why we need Bipedal robot?

- The main advantage of bipedal robot is it is resemble to human and work and walk more efficiently in human environment.
- Bipedal humanoid robots are having much importance because they can walk on stairs, narrow places, and can jump and can do almost all the work that humans can.

### 1.9 Challenges with Bipedal robots

- Though analytical model have many potential benefits like fast computation but due to inherent limitation of a bipedal like high degree of freedom, more variables, different discrete sub phases (due to DSP and SSP) it is challenging to develop a more correct and accurate human like model.
- Whenever a robot experiences external force, it has to maintain its balance in order to avoid fall. If the push is small then it maintains balance through postural balance control, on the contrary if the push is large, it will take one or more steps to recover from the push.
- Stepping the appropriate region will lead to complete stoppage for the robot. The point where robot will step for stopping is called the capture point; it is a point where robot is able to bring itself to stop in one single step. Collection of such points is called capture region. It is difficult to calculate the capture points.

### 1.10 Thesis Structure

This thesis consists the introduction in the very first chapter, which talks about the basic history and the research work done till now on the bipedal robots locomotion and the push recovery. It is discussed about the background of the bipedal robots and its evolution. Then a brief introduction about the problem, Problem formulation and the motivation is described in this chapter. The basics about hybrid automata are discussed and a critical analysis of literature has been given at the end. The whole thesis is divided into following five parts:

In the second part (chapter 2) of the thesis presents some important terminologies which are related to this thesis work and are used in this research are discussed in context of bipedal walking and push recovery.

In the first part of the thesis, we have presented the essence of bipedal robot for modern societies. Further we have discussed the inherent challenge associated with bipedal robot. We have presented the overview how computational model are suitable then kinematics based model.

In the second chapter of the thesis, we have presented the analysis of the available bipedal robot technology and bipedal model.

In chapter third of the thesis, we have given an overview of the bipedal technology with necessary fundamentals. We have presented all the important terminologies used with bipedal gait and push recovery.

In chapter fourth of the thesis, we have presented our innovative idea about the data collection for gait and push recovery for different real subjects. The subjects we have considered were students of our institute having 10 left and 25 right handed persons. We have captured data using indigenously developed wearable device HMCD (Human motion capture device) as well as using HLPRDCD (Human Locomotion &Push Recovery Data Capture Device).

In chapter fifth we have framed the bipedal walking as hybrid system and developed the vector fields for each seven sub phases of bipedal walk. The major contribution of the research is the development of computational walking model and generation of joints trajectories to each sub phases of gait for all the six joints (hip, knee & ankle). The model has been configured as a rocking block and various parameters have been fitted according to different subjects. We have compared the vector field generated joints trajectories with hybrid automata model and HOAP2 model. Finally we have applied our joints trajectories with HOAP2 robot. We have also presented the cellular automata for state predication of bipedal gait.

In the chapter sixth of thesis, we have classified the human gait and push recovery data using various machine learning techniques.

In the seventh chapter of thesis, we have proposed a push recovery capable hierarchically type-1 fuzzy logic controller and compared the human push recovery data with model generated data and we have proved that the fuzzy logic based controller is fast to adapt and is more generalized. It is fast and less computational intensive.

Useful conclusion based on our research experiment, limitation and future recommendation have been made in the chapter no 8.

# Chapter 2: Analysis of Previous Researches

## 2.1 Bipedal Robots Evaluations

The R. W., Powell in his paper titled "human-inspired hybrid controls approach to bipedal robotic walking"[24] has discussed the hybrid automata model and constraints. Domain breakdown was the important contribution of this work which we have further exploited to develop our computational hybrid automata model for our work. The Ryan. W. Sinnet et al. in their paper they have proposed the different sub-phases of gait and given the example of human walk as domain break down into different sub phases [25].

In the paper 'Planar Multi- contact Bipedal walking using Hybrid Zero Dynamics' [26] the author has presented the method for planar multi contact , multi-phase robotic walking through control and optimization techniques used by humans. Their work shows the phases of walking with different degrees of actuation like over actuated DS (Double Support), fully actuated SS (single support) and under actuated SSP. They have used partial-hybrid zero dynamics for generating walking gaits, which produced multi contact, periodic locomotion. Their work presented three domains mainly, that are heel strike, toe strike and heel lift. This was shown as the hybrid control system. Their work handled multi contact locomotion for motion transitions but all the phases of the walk were not taken into consideration for a stable walk which the present work shows for all the phases. Their work has shown that involvement of motion transition allows a robot to be in zero dynamics manifold over the domains where the degree of actuation changes. In the paper 'Motion Control of seven Link Human Biped Model' [27] the authors have developed mathematical model for the planar seven-link biped model comprising of two legs with feet, shank and thigh of both the legs and an upper body. A bipedal structure possesses seven degrees of freedom in sagittal plane. Centre of mass coordinates included in the kinematic model are used in their procedure of modeling mathematically. Authors have used Lagrange's equations for obtaining the mathematical model. Performance was investigated by conducting a simulation for this mathematically obtained Seven-link biped model. Their equations were meant for the investigation of the motion control for a seven-link bipedal model.

Benjamin Stephens and Christopher Atkenson proposed dynamic humanoid balance compliant control in their paper [28]. They used linear biped model for modeling the dynamics of balancing on two feet. They designed the orbital energy controller for achieving periodic motion like it exists in walking for this model. Also they presented the methods for applying the control to a humanoid robot that is to be controlled by torque; this included the estimation of center of mass- state and then generated the commands for feed forward torque. In their work they created the trajectory for center of mass such that our CoP(Centre of Pressure), or ZMP (zero moment point) exists within base of support always. While being in Single support, the dynamics become equivalent to a Linear inverted pendulum model. Their concept for orbital energy allowed controlling the periodic motion without the use of any internal clock. They have shown that the system converges to a limit cycle when the energy is controlled in the coronal plane. They have considered the humanoid's upper body part as the lumped mass and the Jacobean from CoM (center of mass) to the each foot relates the linear bipedal model to a humanoid robot. They have studied the different behaviors of a humanoid robot like balance and step recovery using a linear bipedal model. Their energy controller was useful in stabilizing a robot during the periodic activities.

Sung-Hee and Ambarish Goswami presented a novel method which is based on momentum techniques for maintaining humanoid robots balance in their paper [29]. They have tried to naturally deal with non-stationary and non-level grounds and different frictional properties by doing control of CoM (center of mass) and desired GRF(ground reaction force) at every foot and ground contact. They have not used the CoP and net GRF as that might be impossible to compute or difficult to compute for a non-leveled ground. Their method reduces ankle torques when on double support. The effectiveness of their method of balance control is shown by simulating different experiments on humanoid robot that included maintain the balance while the two feet were on different moving supports with distinct velocities and different inclinations. Their controller (a momentum based controller) was able to maintain a balance of humanoid on locally flat, non-level and moving ground conditions, with providing different disturbance forces. The authors have given more priority to linear momentum and not angular momentum. So a more robust controller for balancing was required which could be possible if an optimal balance can be found between the two.

Tomoya Sato et. al. proposed the generation of a trajectory in real time walking for constant body height for 3-D bipedal robot in SSP (single support phase) in their paper [30]. From ZMP equation and the swing leg trajectory, they obtained the analytical solution for the body trajectory and then based on this analytic solution; a real-time trajectory of the body is generated. The bipedal robot walked stably that is without an up or down of body height, when this body trajectory was applied on it. Also the modeling was more precise with this proposed method of modeling as compared to the modeling of conventional methods. They provided experiments and some simulations for the confirmation of the validity of their proposed method. They used a constant body height instead of CoG trajectory. In their paper the CoG trajectory gave the trajectory for CoG of an entire robot and body trajectory represented the trajectory of a body without including legs. For simulation the floor was modeled as one spring damper system with certain spring coefficient value. They simulated five types. In first type the linear inverted pendulum model was used for CoG trajectory. In second type, the linear inverted pendulum model was used for body trajectory. In third type gravity compensated inverted pendulum mode was used for body trajectory. In fourth and fifth type they used their proposed method for the body trajectory. This work was for the single support strategy phase and for double support another strategy needed to be done.

Eric R. Westervelt, et al in their book [31] presented methods for gaining stable, efficient, easy and quick locomotion in bipeds. Their book guides for the improvement of mechanical design of the future robots. This book contributes to the upcoming theory of hybrid systems. The legged models are hybrid in nature fundamentally. The book has chapters that emphasize on sound theory, where they have described different class of robots which are under consideration are described by the list of hypothesis, And also they have mentioned that how a robot interacts with walking surface impacted and the characteristics of related gait. In this the basics of bipeds and terminologies are explained. Dynamics and the related challenges related to control of Bipedal Locomotion and common difficulties are discussed in the beginning. Static instability, Limit cycle designing, Angular moment conservation is some of the challenges that are associated with the dynamic locomotion. Apart from bipedal robot locomotion the poly-pedal locomotion is also discussed like quadrupeds and other few models. Stability concerns are taken care of for an autonomous system. The passive walking is also discussed which is motivated from drive for energy efficiency. Also it is noted that many different passive walking gait shows the natural look. Powered bipeds are the bipeds that work on energy and practically every biped requires input energy. Bipedal controllers are designed to control the biped locomotion. Different biped controllers are presented and different control strategies are discussed. Basically control strategies are of two types i.e. one is time dependent and the other one is time independent amongst which the time-dependent algorithms are more popular. In the present thesis work also the algorithm is time dependent, as the time input is provided for getting the joint angles as output. The virtual constraints are discussed in different mechanical design tools. Different types of powered bipeds are there and it is been tried to develop the prototypes of non-passive bipedal robots which is primarily led by Japanese. The first biped reportedly capable of walking was WL-5 which was three dimensional with 11 degree of freedom walker constructed in Japan at Waseda University by Tsuiki and Kato in 1972. In 1980s this same group designed and developed a WL-10RD, another three dimensional walker with 12 degrees of freedom that weighed 80 Kg and was capable of a walk at near to 0.1 m/s speed. Similarly many other bipeds and other multi- pedals came into existence time to time. Hybrid robots also started taking shape in research literatures. In which the objective was to use the least possible sensing, actuation, and control for achieving the walk that is more efficient on flat land. The designed machines were based on the passive walkers, in addition to low-power drives to replace the force of gravity as an energy source. With the use of quasi- passive robots the less energy and less control hardware was required as compared to the powered robots and yet they walked naturally. Controlling of biped locomotion is a big task to achieve artificially. The core but invisible component of every non-passive bipedal structure is its control. Several control algorithms for different categories have appeared in different literatures as feedback controlled system. Many Degrees Of Freedom causes many challenges which a successful control design should be able to address in all the legged robots. Walking and running are seen as the periodic solution of a robot model, Poincare sections method is the natural means for studying the asymptotic stability of the walking cycle. But due to the complex dynamic model, this approach had a limited success. A contribution amongst many others of this book is to present control strategy that could be designed in the way which makes it easier to apply the method of Poincare on the class of biped models, and to reduce the problem of stability assessment to scalar map calculation. They have developed the hybrid controller, with applying a feedback signal that is continuous time signal which is applied in stance phase or/and in swing phases and controller parameters' event based or discrete updates are carried out on transitions between different phases. Their controller designs used two principles which are found everywhere for no hybrid systems that are attractively and invariance. The concept of invariance is extended to the hybrid systems to address the continuous phases as well as discrete switching. Closed loop full dimensional system's hybrid subsystems with Low dimensions are created by using hybrid invariance. These hybrid sub-systems with lower dimension are also known as HZD (Hybrid zero dynamics). The attractively meant that the trajectory of a closed - loop system which is full dimension converged locally and sufficiently fast to that of any hybrid zero system dynamics that restricted the stability and existence of running motions and periodic walking to study of Hybrid zero dynamics. It turned out that Hybrid zero dynamics' Poincare map was one dimensional.

The development of RABBIT test bed, a joint effort of different French research laboratories including mechanical engineering, robotics and automatic control has been presented. A rotating bar ensures the lateral stability of RABBIT, hence only a 2D motion is considered in the sagittal plane. This prototype captures main difficulties which are there in the non-linear system: variable structure, under-actuation, state jump. Through a robot's full dynamic's detailed study including impact phases an asymptotically stable walk could be achieved. RABBIT had simplest mechanical structure that was representative of walking leg of human. The goal for initiating the RABBIT project was of demonstrating the existence of stable walking is not necessarily needs actuated ankles, which is why the RABBIT does not had a feet. Without drive ankle, light legs can be developed, which is more effective for walking and jogging. If a robot can achieve a stable walk or running in a wide range of speeds on a flat surface, then the ankle's actuation must be justified based on the improved traction plus walking surface with better adaptability on even surfaces or to facilitate shocks from affected leg on the ground. The under-actuation must be explicitly addressed in the design of feedback control as for the case of without feet; our zero moment point principle will not be applicable, which will lead to development of novel stabilized feedback methods. A mechanism was required to be designed that would allow to enable walking as well as running, for RABBIT project. It was desired from the robot that it performs anthropomorphic gaits, so The RABBIT architecture must have minimum four links that are at least a hip and also two knees. For carrying a load the robot must have a torso, which will make it to total five links. Hence the RABBIT had the mechanism possessing seven degrees of freedom and had four degrees of actuation. When both the legs are straight and together, for being in an upright posture, the tip of torso was at 1.43 meter and the hip was 80 cm above of the ground. Total mass of the RABBIT was 32 Kg. A torque speed curve for each joint was provided by these calculations as the function of running and walking speed. For a broad range of running and walking speeds this analysis made it possible to find the complete operating needed for each motor and then reach its required size. Then these specifications were matched with off- shelf components, for both the gears and motors reducers. The designing of RABBIT was done in such a way to let it be able to walk at an average speed of 5 km per hour and for running it should be at least 12 km per hour.

Another test bed was EARNIE, which was designed by Ryan Bockbrader, Jim Schmiedeler, Eric Westervelt and Adam Dunki-Jacobs at The OSU(Ohio State University between September 2005 and January 2006). The motivation behind constructing and designing ERNIE was providing an educational and scientific platform for the development of new control strategies for Bipedal locomotion at OSU. One foot, knee for both legs and torso in RABBIT was the inspiration for ERNIE's general morphology. But there were many unique features in EARNIE's mechanical designing. This impacted a range of experiments which could be conducted for design important and implementation and design.

This test bed had modular legs, which enabled to change the length of legs, leg end and the joint offsets with least redesigning. Hence the modularity facilitated the study about robot asymmetry and walking with feet, etc. The actuators for this test bed were located in torso which reduces mass which is near to the centre of mass of the robot. This way they got the lighter legs which allowed the use of small sized motors. Parallel compliance was suggested to be easily joined at this test bed's knees. Its joints had low friction relatively, which was there in RABBIT's joints because of harmonic drives. EARNIE was designed for walking on treadmill for continuous walking due to the restricted lab space. ERNIE could either walk on ground or on the treadmill, its boom was attached to the wall and the height of attachment can be adjusted, however, that fixing of boom with the wall prevented use of counter balance.

The concept of point feet is discussed, that if the legs are terminated in points then no actuation would be possible consequently at the finish of stance leg. The degree of actuation provides a large amount of complexity in a bipedal system. With point foot during the single support phase the systems is under-actuated unlike fully actuated. A biped system is always under-actuated while it is in running gait's flight phase. A flight phase is also referred as ballistic phase and when the robot is in flight phase the robot had additionally two different degrees of freedom which are associated with a horizontal and other one the vertical movement of centre of mass which is on sagittal plane. In real world the bipedal robots have feet, the model with point feed is of interest for developing simplified model but for practical robots it is not misleading. If a human walk is taken as de-facto standard with which a biped walk is compared, then current robots which walk with flat foot, needs improvement. Particularly, toe roll towards single support phase's end need to allow as a part of gait design. But since that lead to under-actuation it was not allowed and that cannot be treated for the quasi-static stability criteria like Zero moment point and trajectory tracking based control design philosophy. It is explained that a swing phase or the single support phase is the phase of locomotion when only one leg gets in contact of the ground. Opposite to that a double support phase is a phase when both feet are in contact with the ground. While only a single leg gets in contact with the surface then the contacting leg is known as stance leg and other one is known as swing leg. So, this way the walking is explained as the alternating events of double and single support phases, with the need to put the swing leg strictly anterior to the stance leg that is at impact and the movement or rearrangement of horizontal component of centre of mass of a robot to be strictly monotonic. The assumption implicit in the description was that the foot was not slipping when it was in contact of the ground. The end part of the leg was referred as foot even when sometimes it did not have links that constitute a foot. The running was described as the phenomenon of alternating phases that is of single support, single legged impact and flight, with an additional provision of the stance leg that should not occur on former stance leg and should be on former swing leg. It had been noted that while in the flight phase, the idea of swing phase was ambiguous. Gait hypothesis for walking and running had been done previously in this.

Impact model hypothesis was done which said that the impact occurs if a swing leg touched the ground. Many of the rigid impact models were discussed in the literature and each was used for obtaining generalized velocity expression after an impact of swing leg with that of the surface where the robot model walked. This impact was instantaneous, and that resulted in no slipping or rebound of swing leg. It was observed that in the case when the model walked then at the point of impact, our stance leg was lifted from ground without any interaction. But in case of running, during the moment of impact, our former stance leg was not in connect with the ground. Also the actuators could not generate an impulse and that's why could be ignored at the impact. It was also observed that the instantaneous change may occur in the velocity of robot when the impulse force is applied, but there configuration did not show any instantaneous change. Dynamic model for walking was developed mathematically for studying the walking gait of the bipedal that satisfied gait hypothesis and the robot hypothesis. Assumption was that the inertial reference frame was given and was oriented in standard form w.r.t gravity. Hypothesis said that the surface was flat and without any loss of generality it was assumed that the height of the ground was zero w.r.t. inertial frame. Their model for swing phase had a closer similarity to the pinned kinematic open chain. And it was assumed by hypothesis that the gait was symmetric so that doesn't mattered if which leg was pinned. Their dynamic model was easily obtained by Lagrange method that is why we used randomly which is not attached.

T. A. Henzinger [32] mentioned that the hybrid automaton was the formal model for varied analog and continuous systems. A hybrid automata system is dynamic system having both the discrete and continuous parts. Control graphs, initial conditions, jump conditions, floe conditions for a system are required to be defined for the development of a hybrid system model.

One of the major challenges with bipedal is energy efficient stable walking. So far, the bipeds available are flat footed with bending knees which consume more energy and are also slow. To achieve the stable walk and understand it perfectly we have divided the bipedal walk into different linear sub phases. The three basic strategies are used in a bipedal walking that are Heel contact, flat foot contact, push off or heel off followed by limb swing. The preliminary segmentation is done to define cycle for a subject and the duration of cycle is calculated to determine if they have 2 phases of the cycles or not. Study of biped locomotion can be illustrated on a complex and simplified dynamic model. In this paper 'On the stability of biped locomotion' M. Vukobratovic et al have brought the stability of a bipedal locomotion in focus. Study of gait dynamics and mechanism are related to problem of stability will help in the development of biped robot which helps the researcher to develop the artificial locomotion for disabled persons. In the paper 'Modeling and control of constrained Dynamic systems with application to biped locomotion in the frontal plane' [33] several motions in the vicinity of the vertical stances are taken into consideration and the necessary and important feedback gains are derived for a three link biped model. Nonlinear simulations are carried out to partially verify the results. But this model allowed the computation of the forces as an alternative to the sensing. The system controlling had been done with no force feedback and only state feedback was applied. For a walk to be completely stable for any system the mechanism of push recovery is also needed to be incorporated into a system so that it can walk on any kind of terrain and in different type of external conditions like any external force applied on the body due to any cause. Human push recovery study is a must for the better understanding of a bipedal system. Humanoid push recovery [34] is studied with the exploration of three basic strategies for the recovery mechanism these are 1) using ankle torques, 2) moving internal joints, and 3) taking a step. This model was made for the analysis of human balance and locomotion. In the paper 'Push Recovery by stepping for humanoid robots with force controlled joints'[35] the Model-based feed forward controls are added to achieve full body step recovery control for robots with force-controlled joints. COM dynamics model and step planning has been used for the achievement of this full body step recovery control. With this the Re-planning is initiated after each touchdown. But it could not directly address footstep rotation or cross stepping.

We perform the dynamic walk in our daily life. The walking style for which we are familiar can be realized as a dexterous control which is essentially unstable. In dynamic walk to prevent toppling the swing leg is brought forward to avoid fall. Such strategy in a walk allows fast walking and less energy consumption for each gait. The statically stable walk can be performed by first shifting body weight to foot and next stance of leg and then swing the leg so that the ground contact remains at all-time. For realizing the bipedal walking there is a time series data of different joint angles for desired walking which is called walking pattern. For each joints there is rhythmic pattern associated for each gait cycle [36].

Bipedal locomotion is not as easy as it seems. It is a complex and difficult task due to inherently unstable structure, high non linearity, varying dynamics and control steps during different sub phases of gait. It shows a hybrid nature due to discrete and continuous natural of walk i.e. an under actuated response during single support phase (swing phase) and over actuation during double support phase (stance).

The human walking is the combination of the discrete and continuous dynamics so it can be modeled as hybrid system. To design the correct and exact model of bipedal locomotion, it is required to include all the discrete nonlinear sub phases. Researchers have developed the kinematics based model which superimposed the control strategy to control different sub phases but this is not valid for nonlinear nature of walk. Seeing the complexities there we propose to use hybrid automata to model the time-series data of gait pattern obtained from human walking. Then the model can be used for synthesizing gait cycle data for morphologically similar robots. The model also should have error correcting block, if any provide stable walking gait data n for hip, knees as well as heel for the execution level controller to follow the trajectories.

### 2.1.1 Background of Bipedal Robots

In 1967 the study on artificial hands and arms began which incorporated the technological strength gained by developing active prosthesis that started three years back. The earlier studies were initially aimed at developing only the machines that perform manual labor in place of persons and they focused the development of mechanical artificial hand. The aim had been to develop the robots that can perform the work as intelligently and as nicely as manual skilled labors.

The Lower limb model named WL-1 as shown in figure 2.1 was developed in 1966-1967. It was an artificial lower limb which was made on basis of locomotion of the lower limb's analysis. This resulted in creation of bipedal robot locomotion's fundamental functions. A Master/Slave type Walking Machine WL-3 as shown in figure 2.1 was developed in 1968 to 1969 which was a mechanical model for lower limbs. This had the servo-actuator that was electro-hydraulic and was controlled by master slave technique. It could produce human-like movement for a stance and the swing phase. Also it was able to sit and stand up.

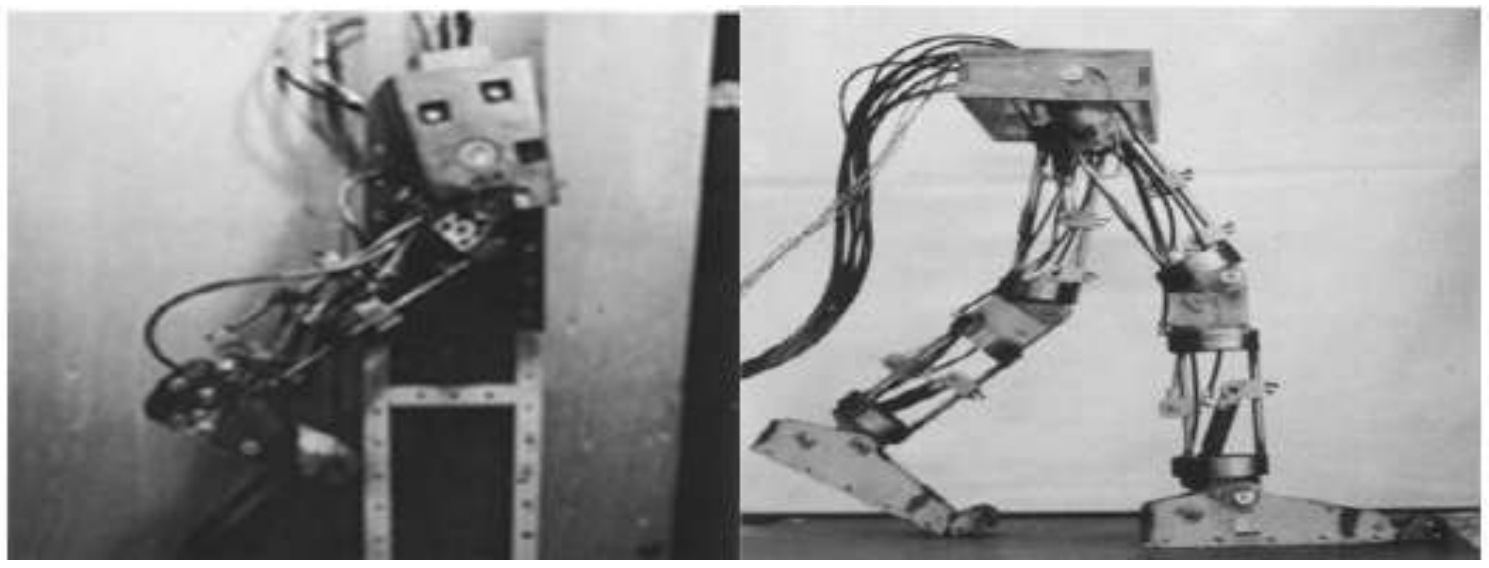

**Figure 2-1: WL-1, WL-3 models developed in1966-1969 [37]**

In 1969 WAP-1 as shown in figure 2.2 was introduced with Artificial muscles made up of rubber were attached. This was a pneumatically- activated anthropomorphic pedipulator WAP-1. These artificial rubber muscles were attached which worked as actuators. By training playback control on its artificial muscles, the bipedal planar locomotion was achieved.

Artificial muscles that were pouch types were introduced in 1970 in WAP-2.It were the second model where the powerful artificial pouch type muscles were utilized as the actuators. By implementing the pressure sensors below the soles, automatic posture handling was acquired. The WAP-3 was developed in 1971 which was a light weighing model for bipedal walking. It was a refined model for the WAP-2. These had the ability to carry their center of gravity over frontal plane, hence it was able to walk on the flat surface and also ascend and descend the slope or staircase, and it can also turn while walking. Figure 2.2 shows the pictures of the models WAP-1, WAP-2, WAP-3 below.

Here a memory based controller was directing the WAP-3 and the PWM was driving the actuators. It was the first time when an automatic three- dimensional bipedal walking was realized by WAP-3.

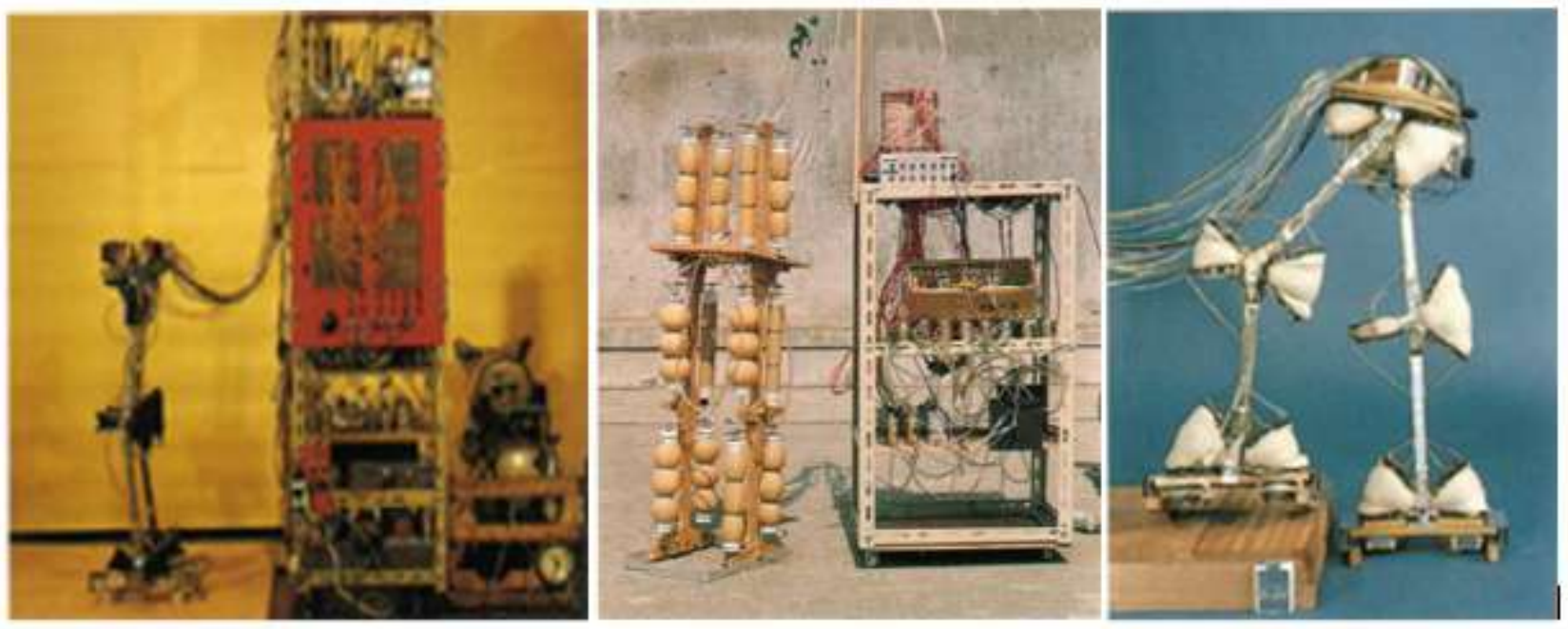

**Figure 2-2: WAP-1, WAP-2, WAP-3 models developed in1969-1971 [37]**

The static walking was realized by a heavy model WL-5. This was developed in 1970-1972. It was controlled by a mini Computer. This had the body that can laterally bend. With this feature it was able to move the center of gravity of its structure on the frontal plane. Through the use of minicomputer it was able to perform bipedal walking automatically and it got the ability of changing its direction in which it was walking. In WABOT-1 (45sec/step) these WL-5 were used as lower limbs. Figure 3.3 shows below the WL-5 bipedal model.

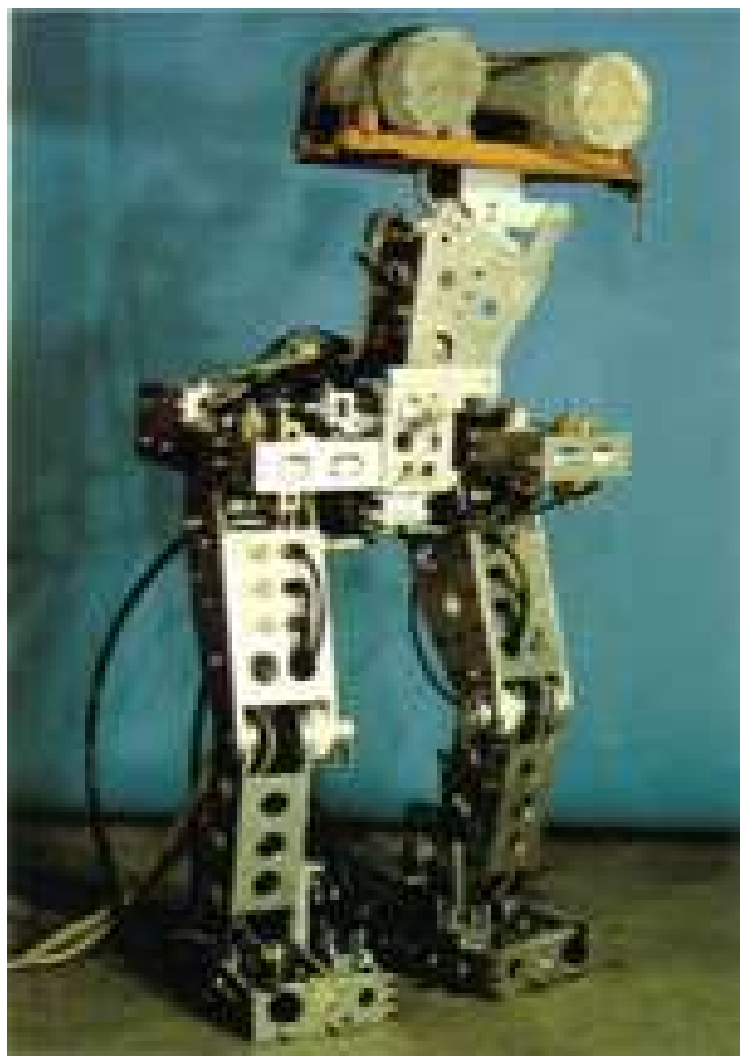

**Figure 2-3: WL-5 models developed in1970-1972 [37]**

The quasi-static walking was realized in 1979-1980 with WL-9DR as shown in figure 2.4. It was the first time such walk was realized in world by this model WL-9DR. This model used 16 bit microprocessor instead of the minicomputer as the controller for its working which enabled the versatile control. Total no. of points that WL-9DR's sole touched on the floor were raised from 3 to 4. This made mathematical solution for a particular walk pattern even easier to attain (10sec/step). Then after this another model of WL series was the refined type of WL. It was WL-10, 10R which was constructed in 1982 – 1983. In this model the rotary type servo actuators (RSA) were introduced and the use of carbon fiber reinforced plastic (CFRP) was there in its structural parts. This model WL-10R as shown in figure 2.4 had an increment of degree of freedom that was added to the hip joint at its yaw axis. This addition of degree of freedom at yaw axis enabled the WL-10R to acquire the functionality of lateral walking, forward walking, backward walking which is called the plane walking (4.4 sec/ step) and turning.

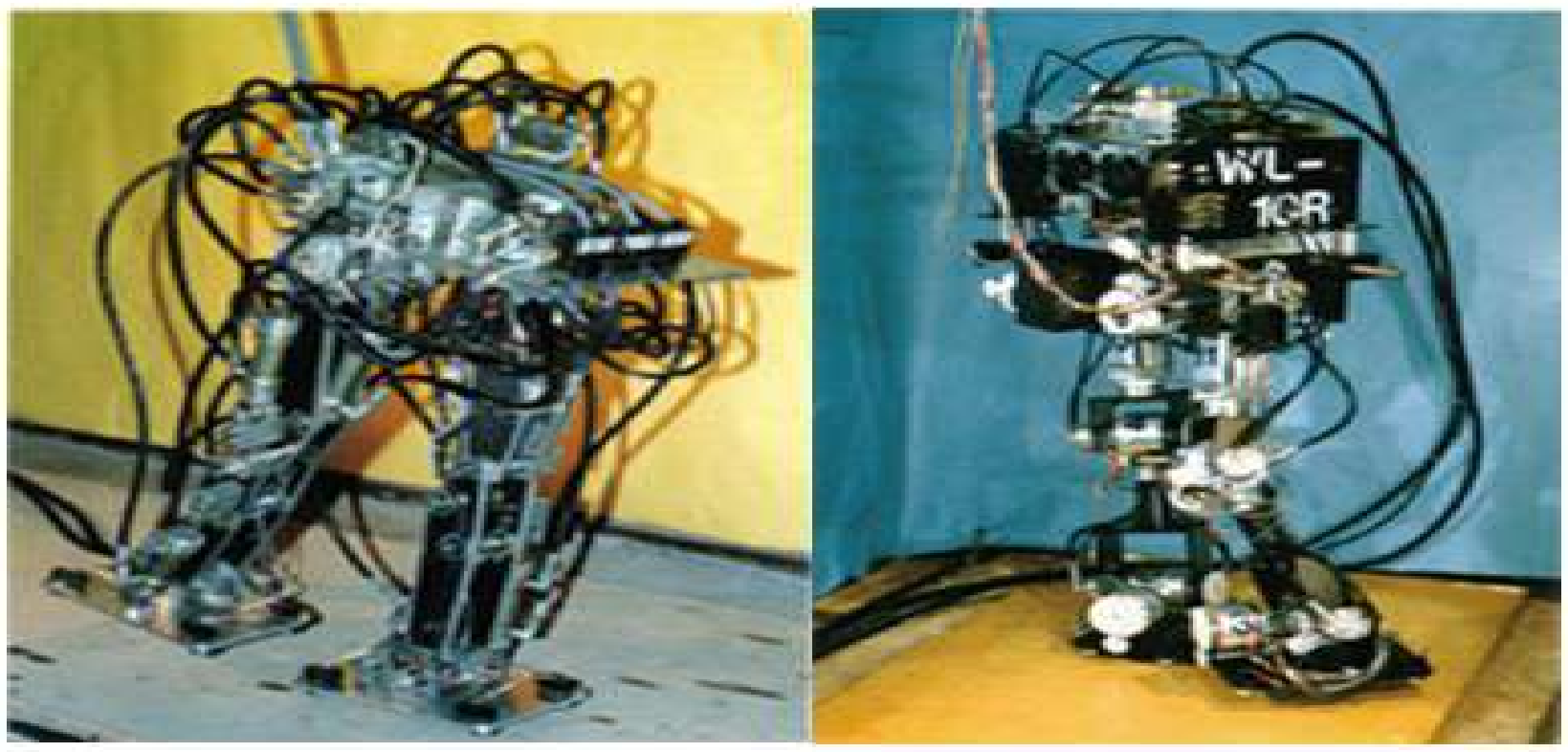

**Figure 2-4: WL-9DR, WL-10R models developed in1978-1983 [37]**

Dynamic walking was realized in 1984 with a model WL-10RD, This model was the refined version of WL-10R, this model consisted of WL-10R torque sensors which were attached to ankle and hip joint, this addition of torque sensors allowed the flexible controlling of a change-over phase i.e. transition from standing on a one leg to the position of standing on another leg by using torque feedback. This was the successful dynamic complete walking which was realized first of its kind in the world (1.3 sec/step) [37]. Figure 2.5 shows the WL-10RD model as follows.

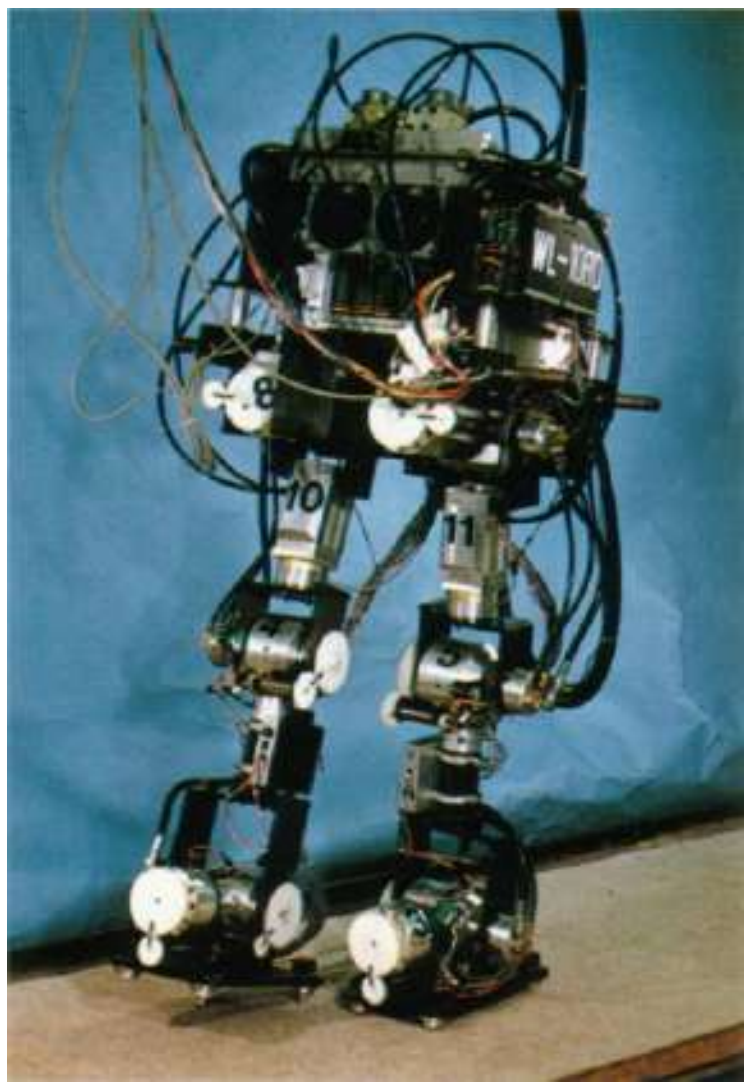

**Figure 2-5: WL-10RD models developed in 1984 [37]**

Link mechanisms can be designed using Computer aided design systems. In 1979 a simulated system and mechanism design was made for dynamics analysis. This study aimed at designing machines rapidly and effectively. This system was used to display the movement of machine, graphically by just inputting shape of the robot's parts, its external forces and mechanical structure. In 1980-1981 the Computer aided design system was developed for artificial limbs. It was an interacting computer system that was developed for the aid in designing artificial limbs. Limbs were simulated as the link mechanism with multiple degrees of freedom. After that the force for each joint was calculated by providing their trajectory or it determined each joint trajectory by providing external forces. Figure 2.6 shows the images for simulation of Bipedal walking existed in 1979 and Design for manipulator developed in 1981.

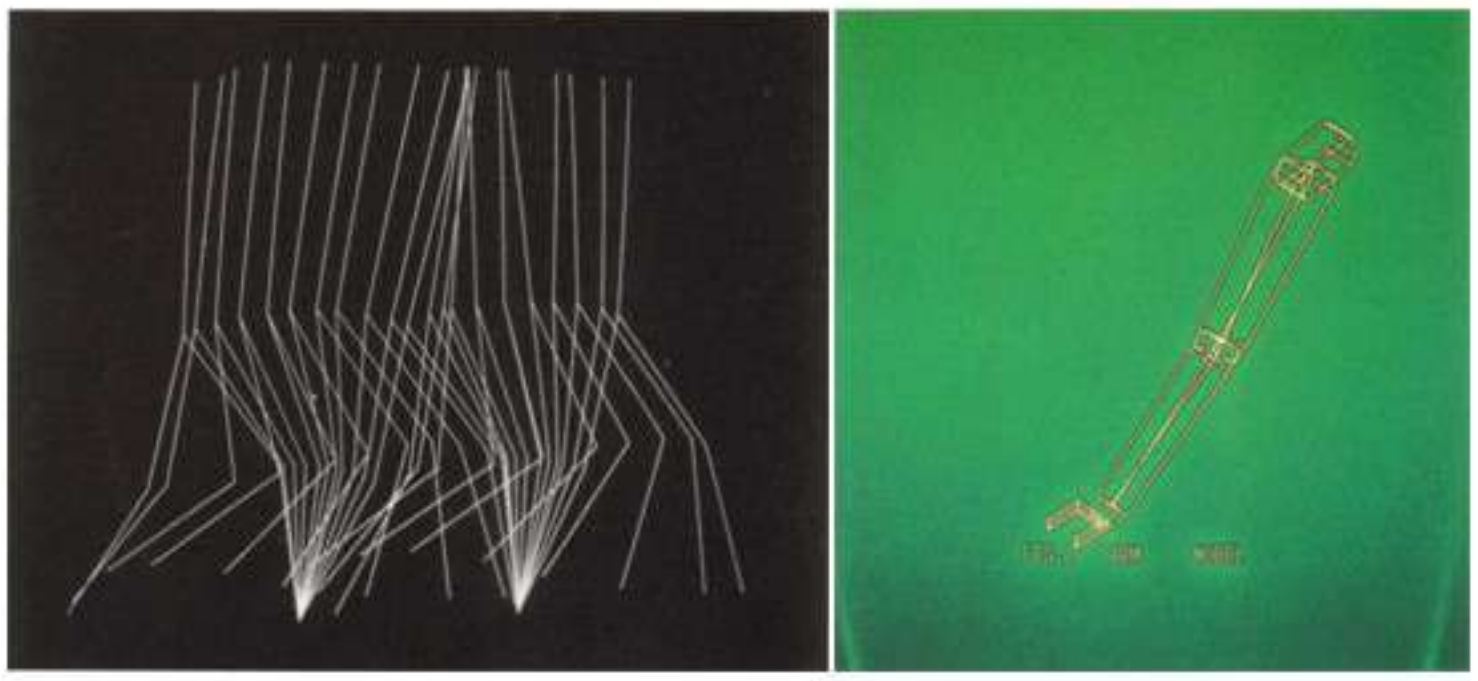

**Figure 2-6: Simulation of Bipedal walking and Design for manipulator[37]**

In 1982- 1983, the computer aided design system was developed for robotics. It was the interactive software which was constructed for automatically creating and solving the modeled and dynamic equations of the robot with arbitrary degree of freedom. This system was made so that it can be applied on a link mechanism that had a control system. Another computer aided composition system was Walk Master-2 which was for walking pattern that was developed in 1984. The figure 2.7 given below shows the Link model for bipedal that developed in 1982. And Figure 2.8 shows the different phases of Walk Master-2 that was developed in 1982.

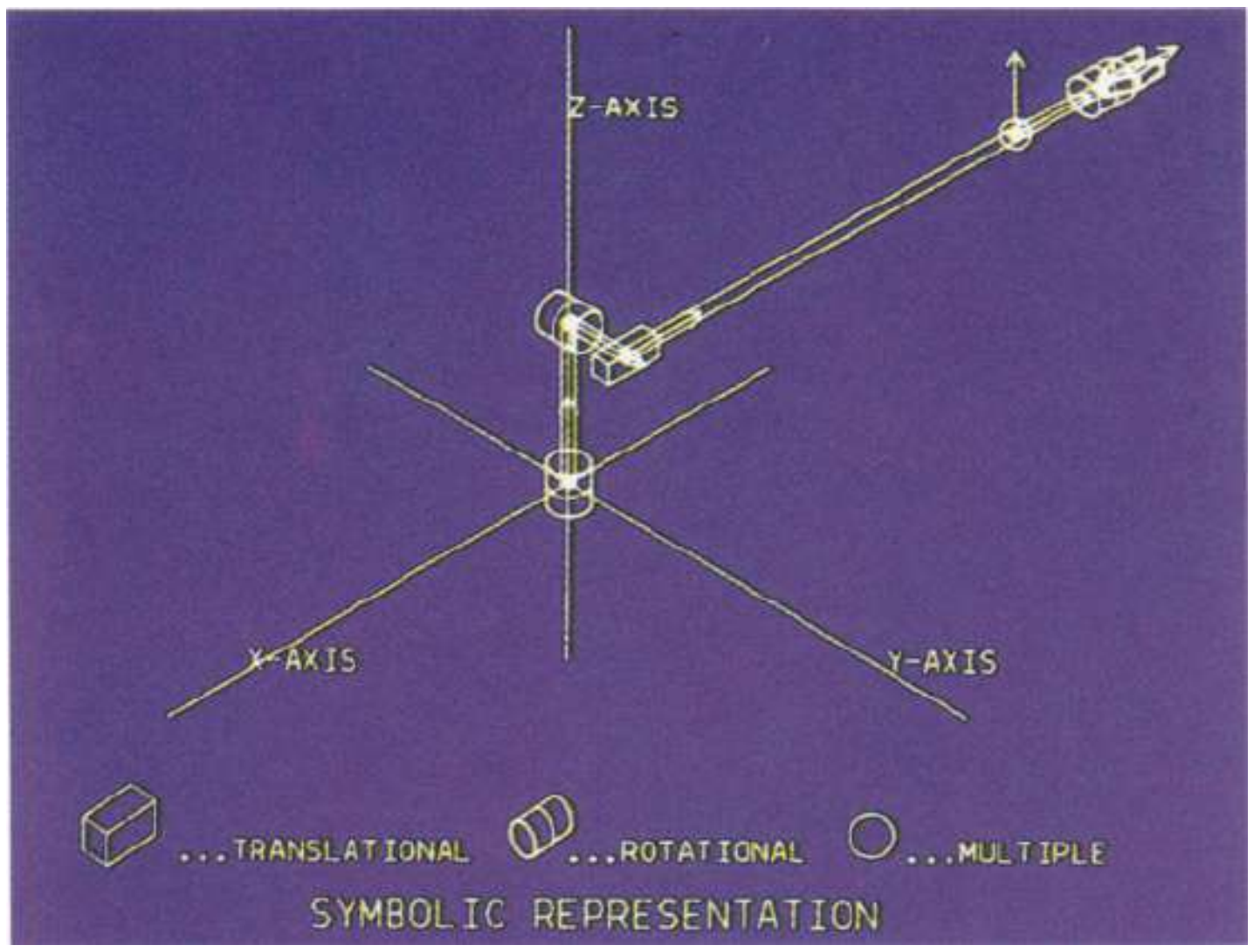

**Figure 2-7: Link model for bipedal (1982)**

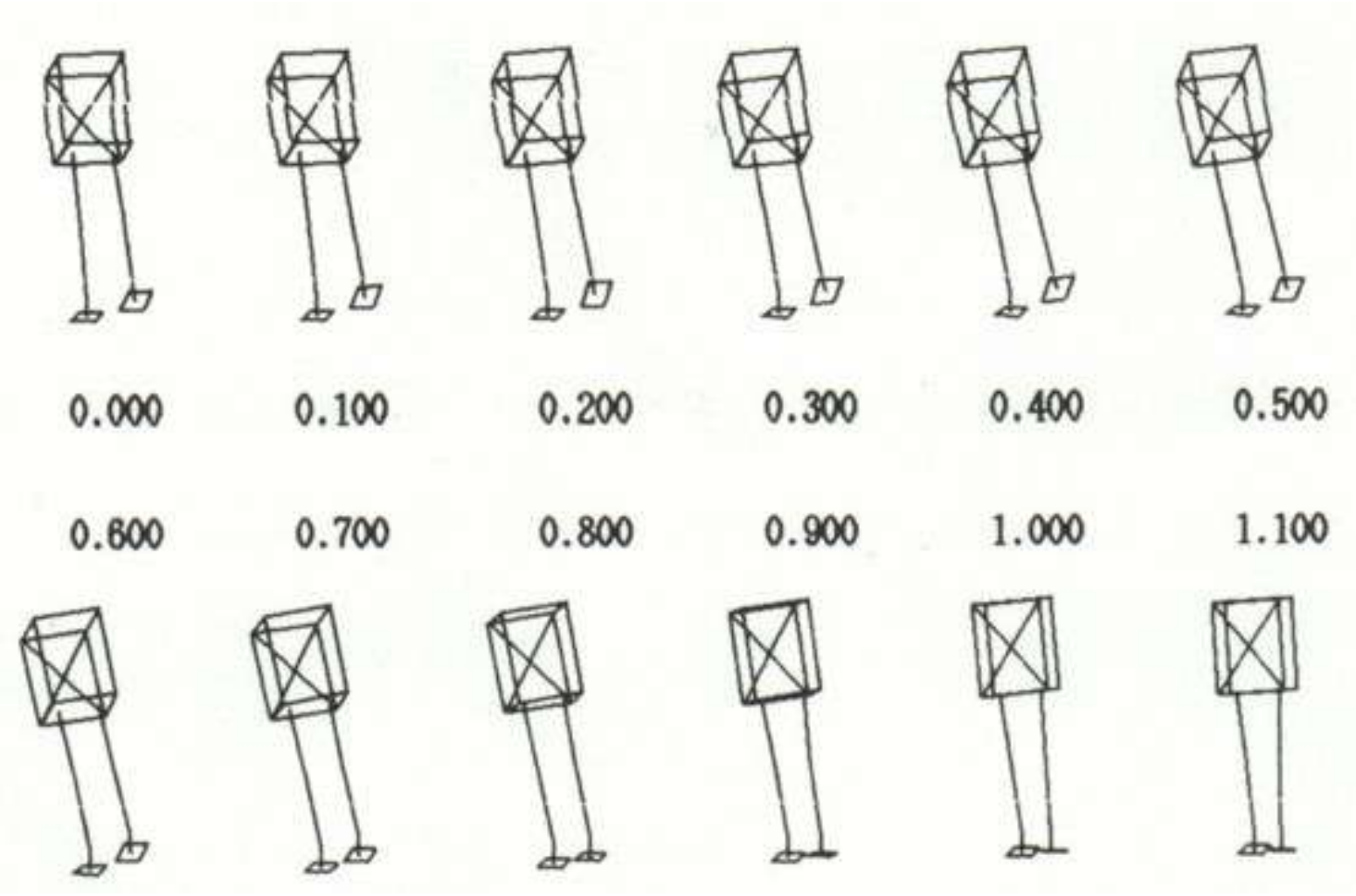

**Figure 2-8: Walk Master-2 (1984)[38]**

This interactive system was made for the personal computer for analyzing and composing the walking pattern of the robot. This system enables analysis of Zero moment point (ZMP) when the biped was walking and composition of walking pattern that is combined with a characteristic of actuator of robots over three dimensional graphics. The figure 2.9 below shows the Walk Master-2 ZMP (zero moment point).

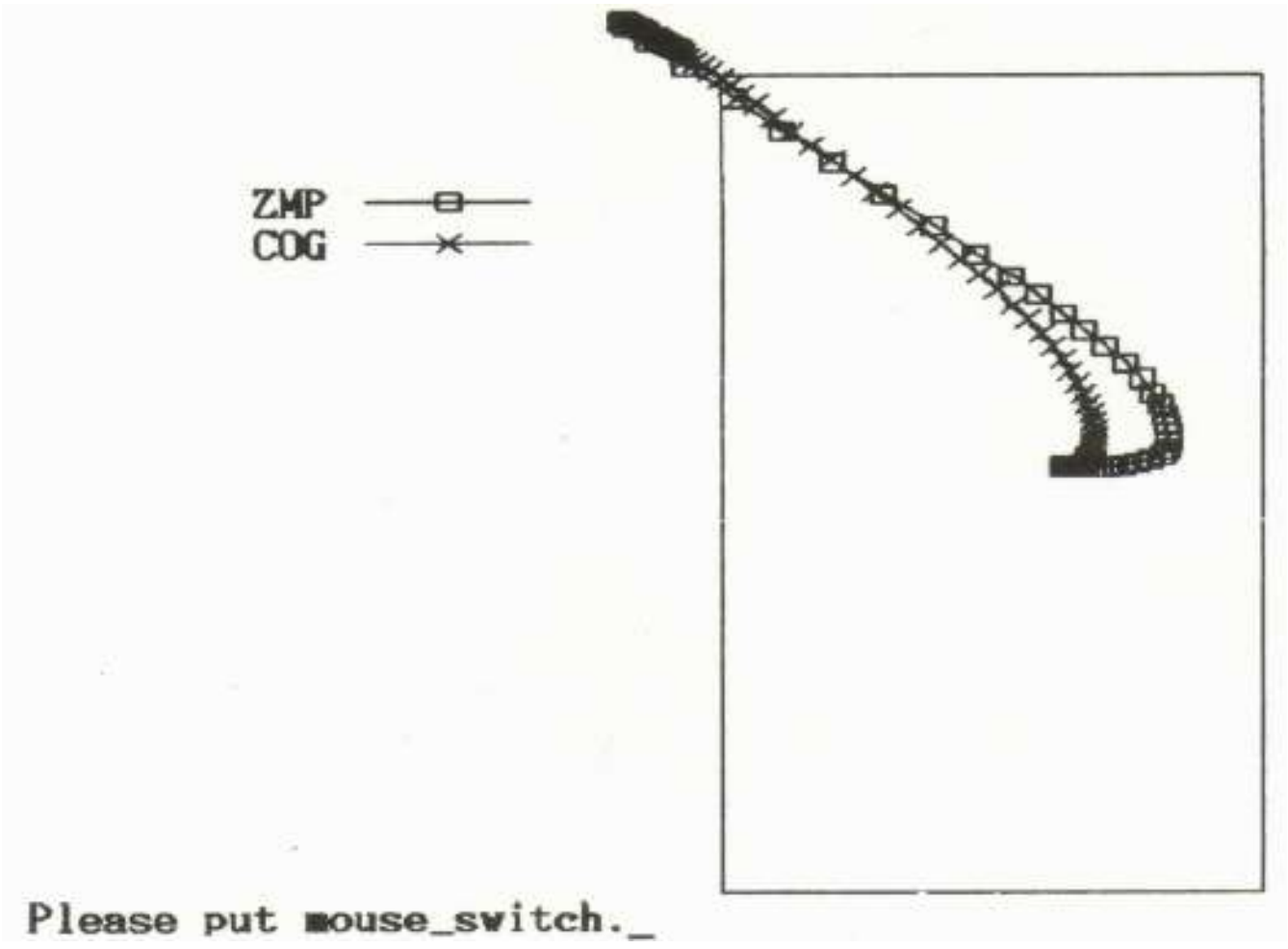

**Figure 2-9: Walk Master-2 ZMP (1984)[38]**

## 2.2 Bipedal Model for Stability against External Perturbation

All entire existing models are based on kinematic based model. Following are the bipedal models for push recovery-

### 2.2.1 Linear Inverted Pendulum (LIP) with Fly Wheel model:

To avoid the external force human beings used to take the one or two steps. To achieve this goal, capture points and the capture region are computed so that humanoid can come to a complete stop by stepping. The strategy robot will choose totally depends upon the intersection between the support base and the capture region. To introduce inverted pendulum strategy as well as lunging fly wheel model has been used. Lunging plays an important role in helping our body to come to stop with or without taking a step [33]. Figure 2.10 is the model of LIP with fly wheel.

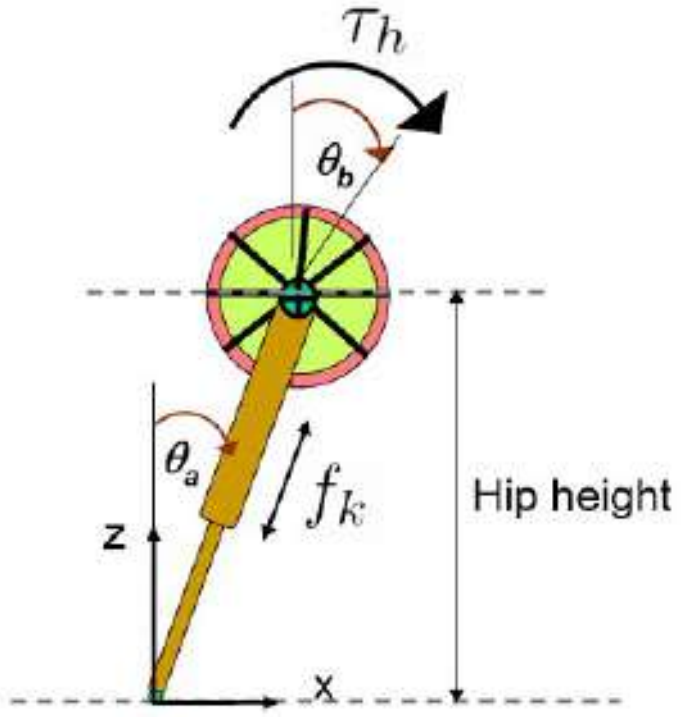

**Figure 2-10: Linear Inverted Pendulum (LIP) [39]**

### 2.2.2 3D Linear Inverted Pendulum Model:

This model has a 3 D pendulum whose motions are constrained to move along an arbitrarily defined plane. This model takes account of orbital energy as the governing parameter for the stability [39]. Motions of this model are captured in two different planes (Sagittal and Lateral). Two separate controllers have been used to generate motions in two different planes and generation of walking trajectories has been simplified. The below figure 2.11 is 3D-LIPM model.

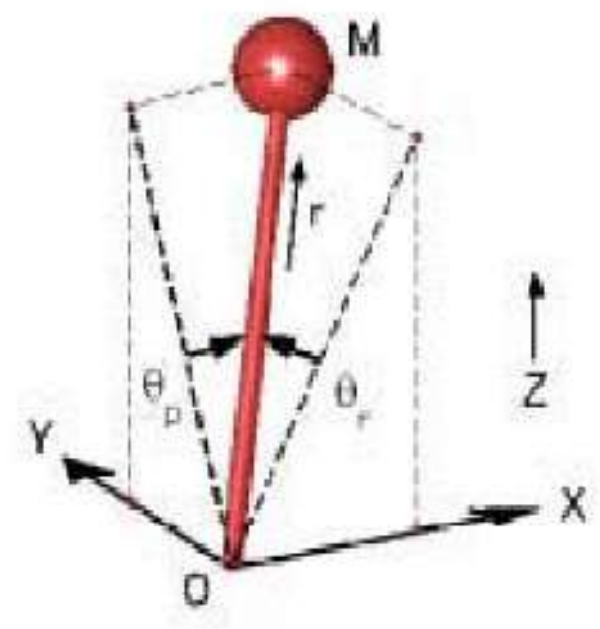

**Figure 2-11: 3D Linear Inverted Pendulum Model [39]**

### 2.2.3 Linear Inverted Biped Model (LIBM):

This model is composed of two legs for balancing; this model is an extension of linear inverted pendulum model as it behaves like inverted pendulum in single support phase, and superimposition of two LIPM's in double support phase. Then a controller is derived for the orbital energy and this model can be used to make quick decisions to recover balance as well as planning high level walking trajectories. Figure 2.12 is the LIBM with two legs.

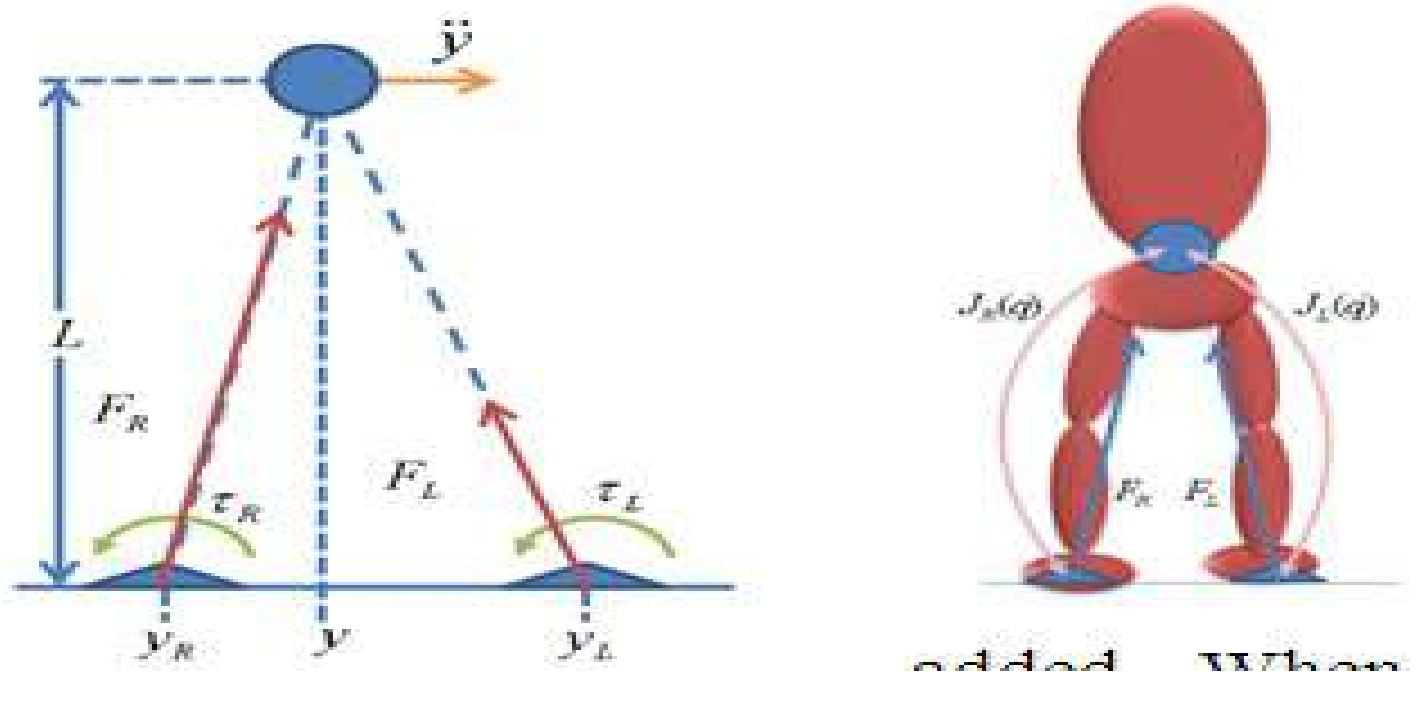

**Figure 2-12: Linear Inverted Biped Model (LIBM) [39]**

### 2.2.4 LIPM with learning module:

In this approach, a learning module has been added. By adding this learning module the offset to the step the robot takes to compensate the error while calculating the capture point has been tried using both ONLINE and Offline approach [49]. After sufficient amount of learning it was found that this model showed more robustness to external pushes than that without learning.

### 2.2.5 Planar Rimless Wheel Model:

This model consists of only two spokes which are attached to the point mass at the centre corresponding to the centre of mass of the robot. These two spokes represent the two legs of the robot. A reference stepping location is computed as humanoid is modeled as a passive rimless wheel with two spokes such that stepping on the location leads to a complete stop of the wheel at the vertically upright position.

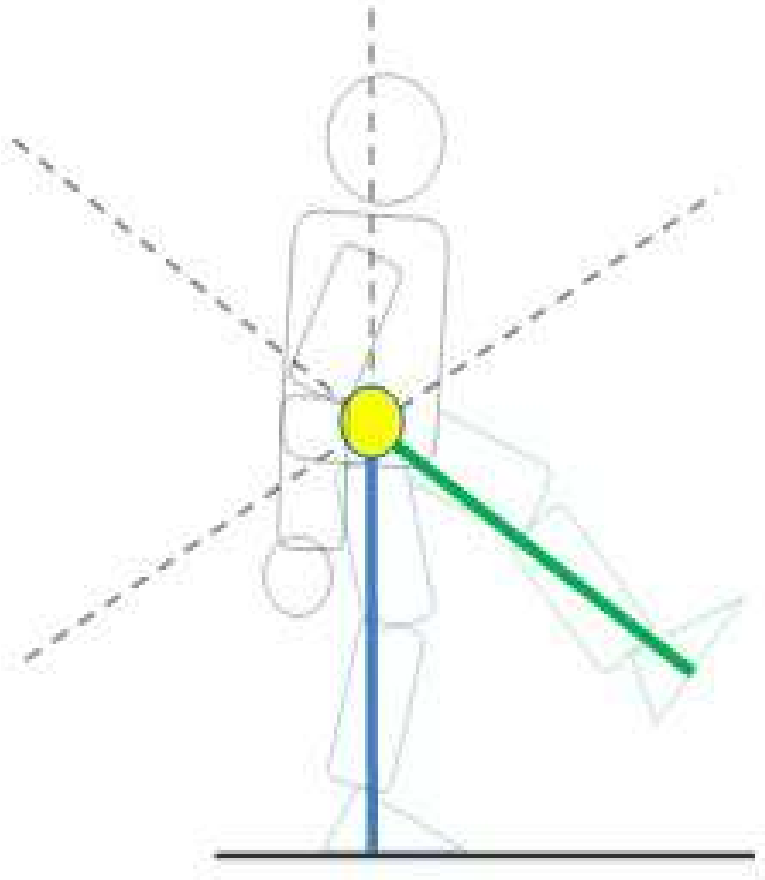

**Figure 2-13: Planar Rimless Wheel Model [40]**

Strategies: The three basic strategies that are used in push recovery are:

(1)COP Balancing (Ankle Strategy): In this strategy centre of pressure under the foot is changed to apply reverse torque about the centre of gravity of robot. It is suitable for small magnitude of push.

(2) CMP Balancing (Hip Strategy): This strategy is also called lunging. This strategy is used in case of slightly large magnate push.

(3)Stepping (Change of support Strategy): It is the final action that a robot can take in case of external push\pull when the magnitude is large and robots need to take step to maintain stability.

These strategies are applied top to bottom depending upon the magnitude of external force. ( Fig 2.14 ).

Concept of centroidal Angular Momentum: We also use lunging reaction torque to prevent us from falling. Few models used a Fly wheel added to LIPM with centroidal Angular Momentum to generate a reaction torque. This Fly wheel [40] generates reverse torque which is similar to the Hip strategy humans' use.

Concept of LIBM: Few models [40] used super impositions of two LIPMs which is very much similar to Compass model. In this model, reverse torques at each foot is applied to maintain stability that resembles Ankle strategy human use. This is very useful in recovering from the impact of small forces.

### 2.2.6 Concept of Linear and Angular momentum:

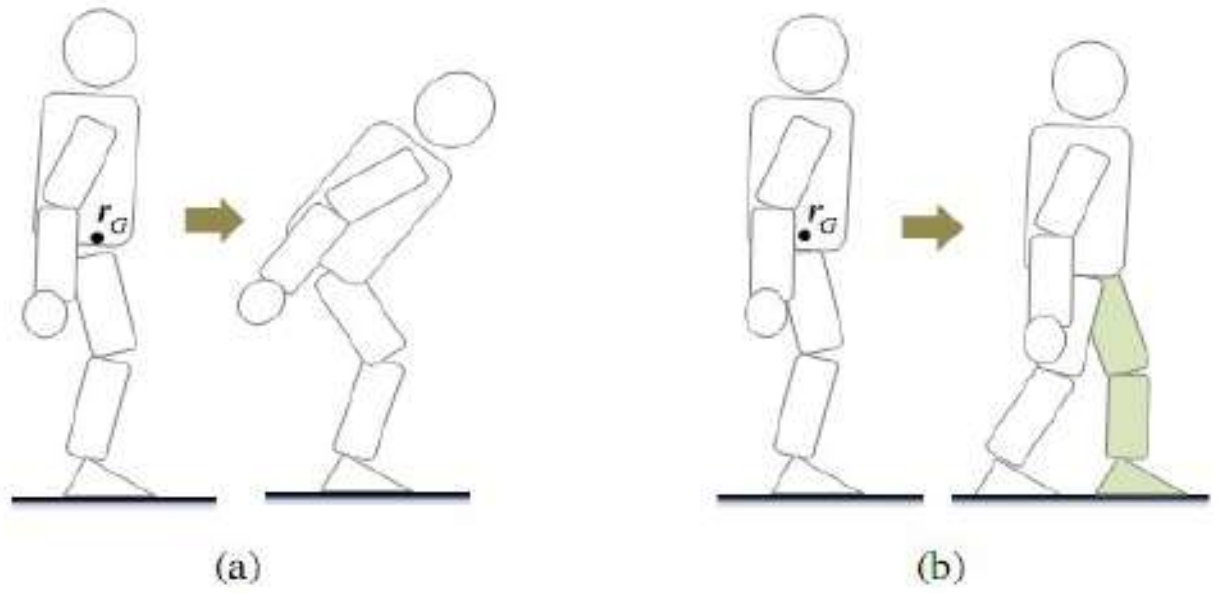

**Figure 2-14: Different Push Recovery Strategy [41]**

This paper proposes a unified strategy of postural balance and reactive stepping integrated with momentum based controller. This enables us to control the angular momentum of humanoid. This helps robot to maintain balance by making postural adjustments and hence avoid stepping.

Generalized Foot Placement Estimator [GFPE]: GFPE is proposed as a reference point for stepping on level and non-level grounds by modeling robot as rimless wheel. GFPE is chosen so that the COM will stop vertically over the stepping location. Robot is controlled in such a way that the offset while stepping to the point due to the difference between dynamics rimless wheel model and actual robot leg swing is compensated by the controller. GFPE computation also computes the duration of stepping for which GFPE remains stationary until the swing spoke touches ground, which helps in designing trajectories of the swing leg. This paper also paid attention to the anchor point, about which the pendulum rotates. Practically humanoids have a non-zero contact area, while pendulum models have a point contact.

# Chapter 3: Important Terminologies

## 3.1 Bipedal locomotion

The Bipedal locomotion makes the human being more capable of walking over uneven terrain. Different animals and humans walk in their own unique manner and have a unique pattern called gait [42]. This bipedal locomotion is a complex task to learn, that's why it takes a long time for a new born to learn walking on two legs whereas on the other hand the animals that walk on four legs have their infants walking just after their birth. This gait pattern for a human bipedal walk though is complex but when compared with the gait pattern of other quadrupeds it shows the simplest gait pattern, which is the reason why scientists and researchers are trying to apply this bipedal locomotion on the machines and robotic structures. Also the bipedal locomotion is robust and is very helpful in walking in the uneven terrains. If this bipedal locomotion is achieved in the machines in the similar way as humans have, then the artificial bipedal structures can be used in the medical help, for those people who have lost their legs or are challenged by birth. Apart from medical advancement it can also be used in other technology advancements like space exploration or mine detections etc. where it is dangerous to send humans. Bipedal locomotion is an interesting topic for researchers and has shown wide possibilities of advancements in different fields of studies.

Bipedal locomotion has some key terms like gait, gait pattern, gait cycle, phases, etc. to explain the process of bipedal locomotion in a better way.

## 3.2 Different Anatomical plane:

There are three planes in which one can visualize the human walk (Fig 3.1) [43].

1. Sagittal plane: Divides the whole body in two parts i.e. left and right.
2. Transverse plane: Divides the whole body in two parts i.e. top and down.
3. Coronal Plane: Divides the whole body in two parts i.e. front and back.

To visualize and represent the human walk we used the Sagittal plane.

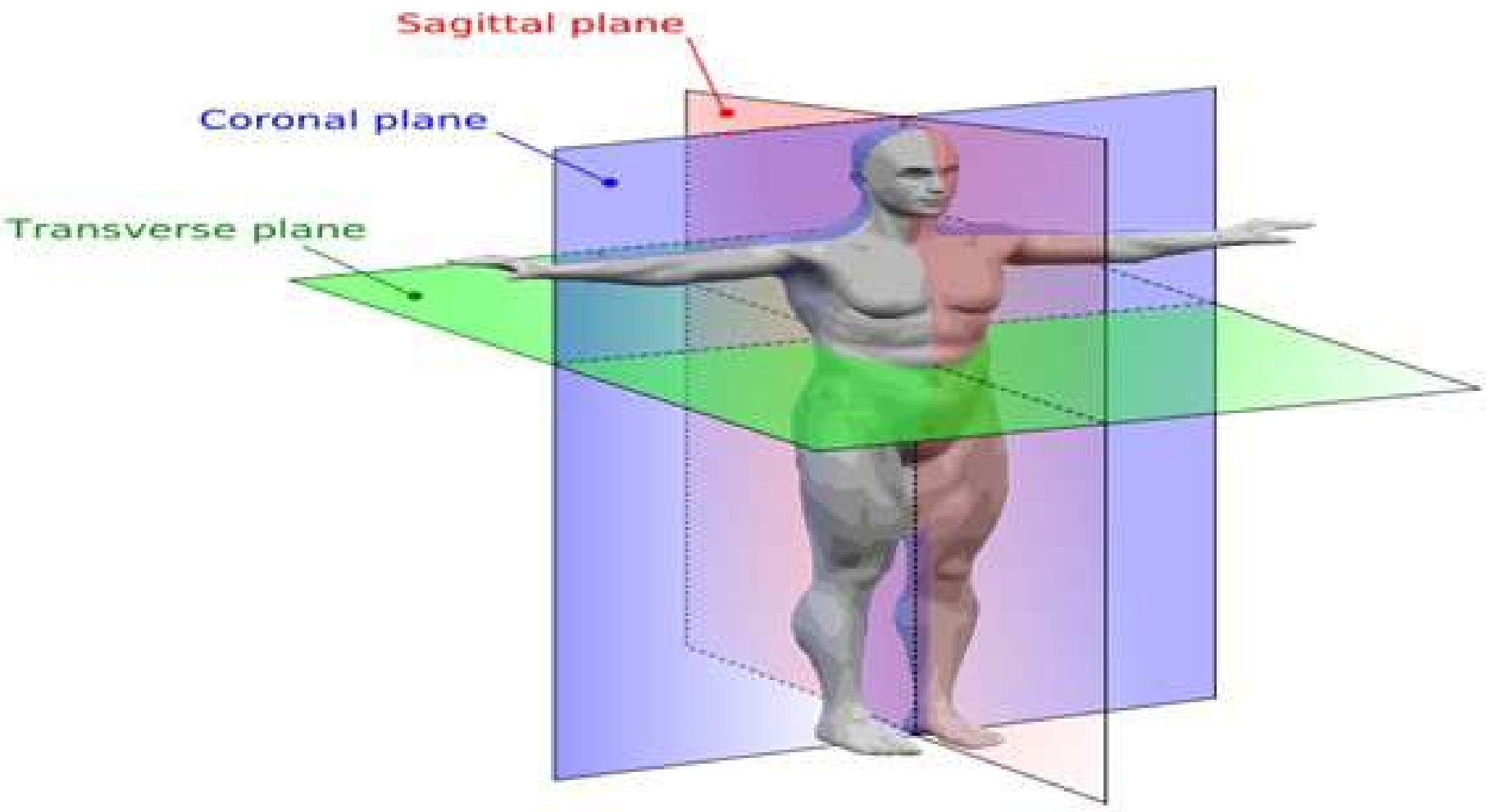

**Figure 3-1: Different Anatomical Plane of Human [51]**

## 3.3 Gait cycle

A single succession of operation of one leg is called as a gait cycle and a single gait cycle is also called as a stride. The gait cycle is extended from a heel strike to heel strike of one leg. Each animal species has their own gait pattern; it depends on various factors like terrain, speed, and energy efficiency etc. Amongst the entire gait pattern for different animals the gait pattern for a human bipedal is found to be the sophisticated. Figure 3.2 is the foot trajectories of both left and right foot.

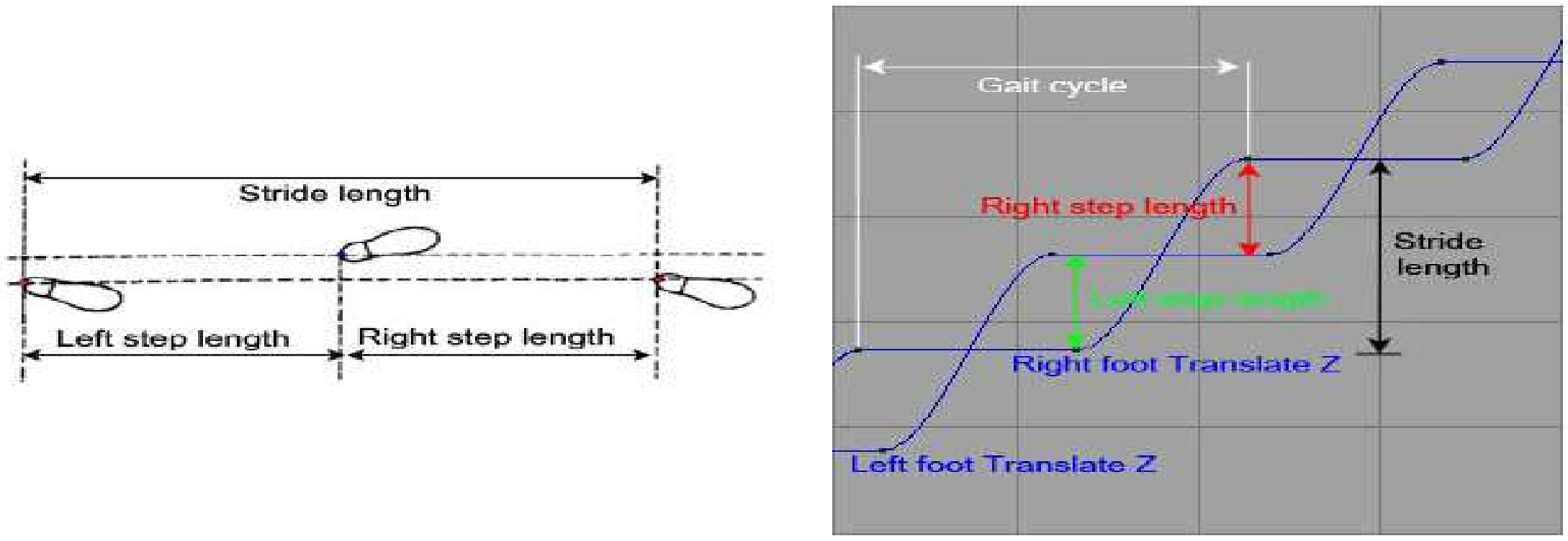

**Figure 3-2: The stride length and foot translation trajectory of left and right leg during human walk [43]**

In a basic gait cycle, the movements are divided into how many times and when the foot is touching the ground or it is on the ground (that's the stance phase which take up to 60% of a complete gait cycle) and when the foot is lifted off the ground (that's the swing phase which takes the remaining 40% of the complete gait cycle). The double support phase (DSP) occurs when both the feet are on the ground, the single support phase (SSP) occurs when one foot is on the ground. A human gait and stride cycle comprise broadly two phases, i.e. the stance phase which is the discrete form of phase and the swing phase which is the continuous form of phase. The figure 3.3 shows the percentage distribution of different phases in a gait cycle [51]. And figure3.4 shows the positions of the leg at any particular phase of gait cycle.

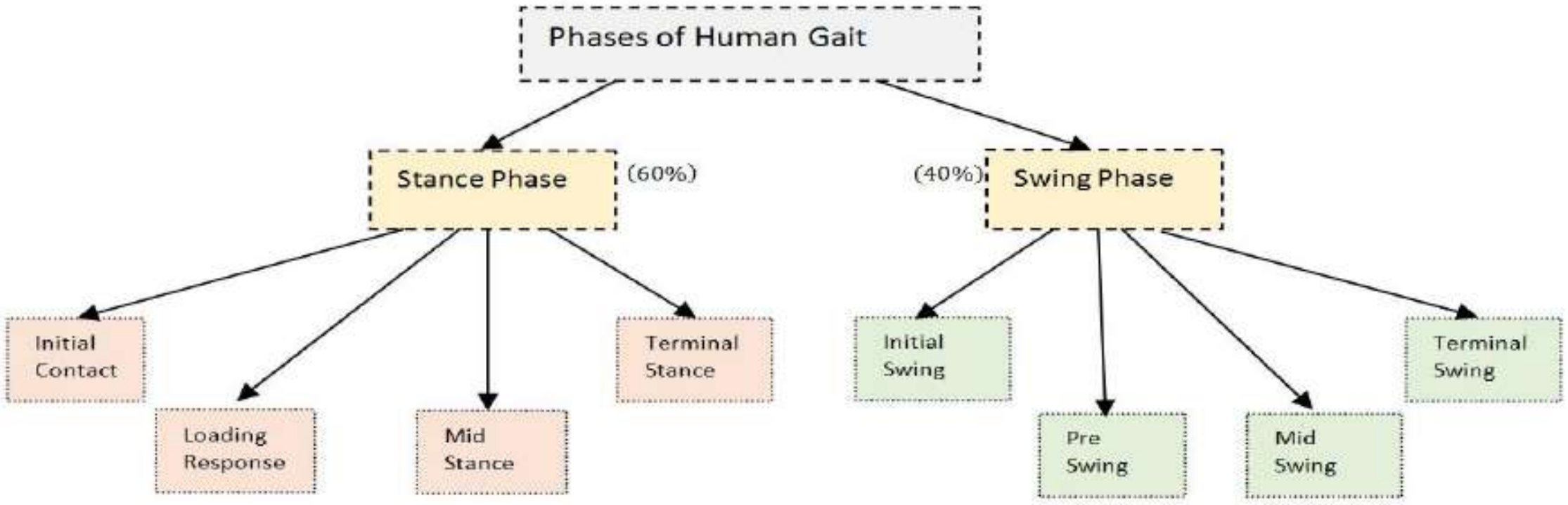

**Figure 3-3: The Breakdown of Human Gait in the Various Discrete Sub Phase**

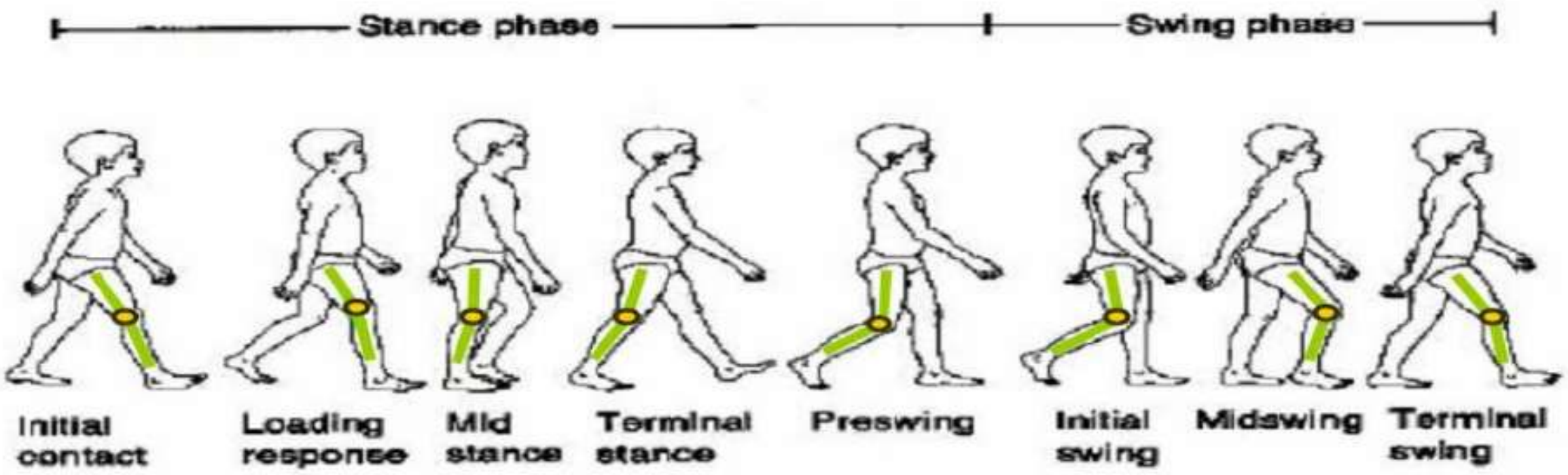

**Figure 3-4:Breakdown of Human Gait into different Discrete Sub Phases [14]**

## 3.4 Phases of Gait cycle

There are basically two phases in the human gait cycle that enables a walk for any biped structure. These are:

- Stance phase
- Swing phase

The gait cycle is sometimes referred as walking cycle. The gait cycle extends from heel strike to heel strike of one leg and includes the swing and stance phase [14]. Broadly gait is divvied into 7 different sub phase refer the Figure 3.5. It is the breakdown of human gait in the various discrete sub phases.

### 3.4.1 Stance Phase

The stance phase represents the major portion of gait cycle which is almost 60% of one gait cycle. This phase provides a means of progression and it requires stability. The stance phase also provides a means of energy conservation. Stance phase is further divided into the five parts or phases that are, initial contact (0 - 2%), loading response (0 - 12%), mid stance (12 - 30%), terminal stance (30 - 50%), and pre-swing (50 - 62%).

The loading response starts with initial contact that is the point when the foot contacts ground. During initial contact the heel strike will strike with the ground, the ankle will be neutral and knee will be extended. During loading the energy conserving mechanism takes place, throughout early mid stance the knee and hip stability is maintained.

Usually the heel portion of the foot contacts the ground first. But in patients who shows the pathological gait patterns, their entire foot or the toe part contacts the ground first [28]. By mid-stance the ankle is neutral and knee is extended again, double support that is when both the feet are in contact with ground happens for about10% of the overall gait cycle. Mid stance begins with a contra lateral toe off and it ends with the centre of gravity when it is directly over reference foot. In this phase and the early terminal stance phase the CoG lies truly over base of support and these are the only times when this happens in the gait cycle. In terminal stance phases the toes that have remained neutral, now extend towards the ankle. In this the centre of Gravity is over supporting foot which ends when contra lateral foot touches the ground.

In the last sub phase of Stance phase the pre-swing initiates at the contra lateral initial contact which ends at the toe off, this is around the 60% of the complete gait cycle. So pre swing corresponds to second period of double limb in the gait cycle.

### 3.4.2 Swing Phase

It is the phase of the gait cycle when the foot is off the ground and is swinging in air, swing phase follows just after the stance phase, it’s a continuous phase and it subdivided in to three parts initial swing (62–75%), mid swing (75–85%) and the terminal swing (85–100%).

The initial swing starts at toe off which continue till maximum knee flexion occurs. Maximum knee flexion can be 60 degrees angle between the joints. Mid swing is the phase when the transition from maximum knee flexion occurs until the tibia becomes perpendicular or vertical to the ground.

Terminal swing starts at the point when the tibia is straight or vertical and it ends back at initial contact or loading response. Figure 3.5 represents the all the 7 discrete sub phases of our model.The positions shown in the figure are the original poses obtained from our models output.

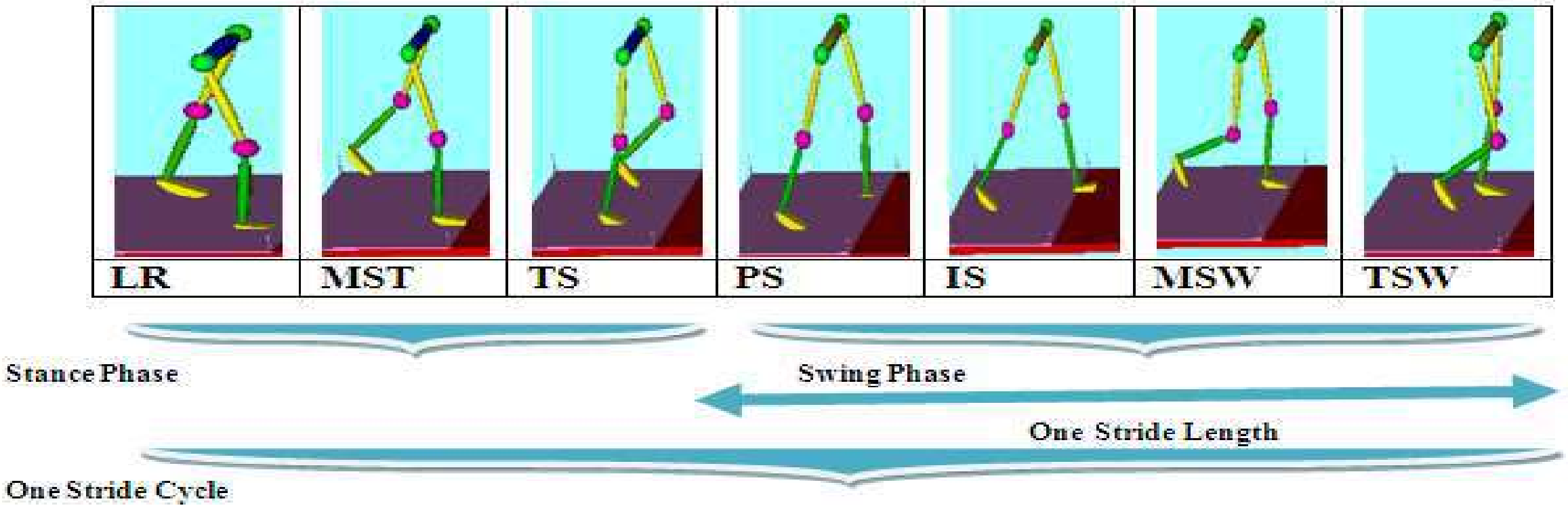

**Figure 3-5: Different Gait sub phases of Our Model**

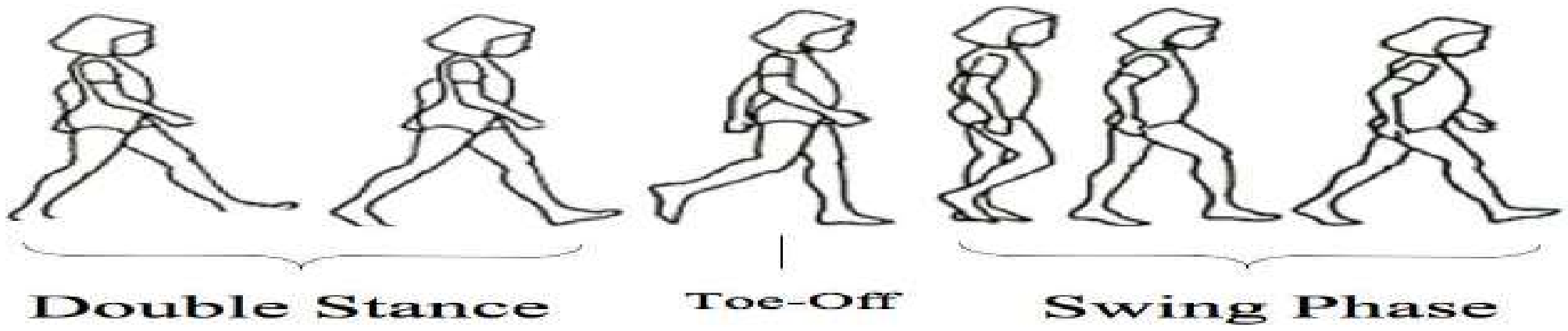

**Figure 3-6: DSP (Double Support Phase) and SSP (Single Support Phase) during Normal walk [97]**

## 3.5 Biomechanics of Humanoid robot:

The strike with ground will affect the knee and hip joints. As we increase the speed the stance phase decrease and swing phase get increase. Steps length is the length of alternative foot strike. There are two types of steps left and right foot step. Stride length is the length when same foot strike again on ground. There are following important observation about Biomechanics of Bipedal robot:

- Researchers have studied bipedal as under actuated system. There is no actuation at knee joints. This type of configuration can be studied as an inverted pendulum.
- The DSP will arise when the swing leg i.e. free flight foot of the bipedal will hit the ground and support foot changed the immediately and rotation continue.
- This configuration makes DSP zero. From human biomechanical observation the DSP is 20% of total cycle of walk [44].
- To design the human similar robot we need to consider the DSP during bipedal locomotion.

Figure 3.6 represents the DSP and SSP during one gait cycle. Figure 3.7 is percentage wise decomposition of human gait whereas Figure 3.8 and 3.9 represent the DSP during running and normal walk.

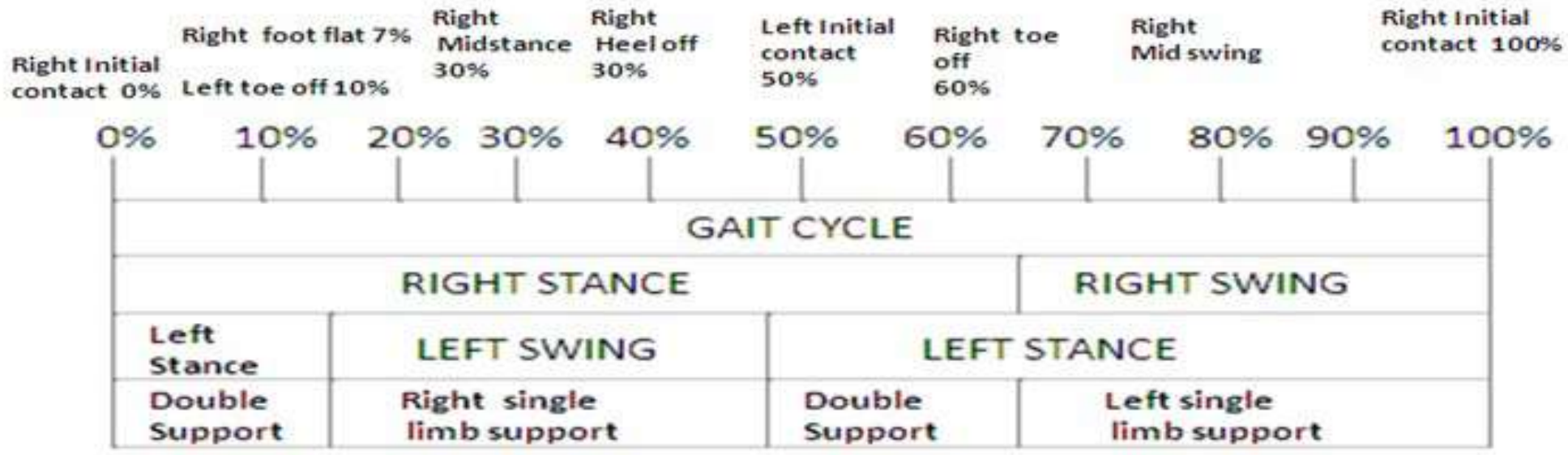

**Figure 3-7: Human GAIT different Phase [44]**

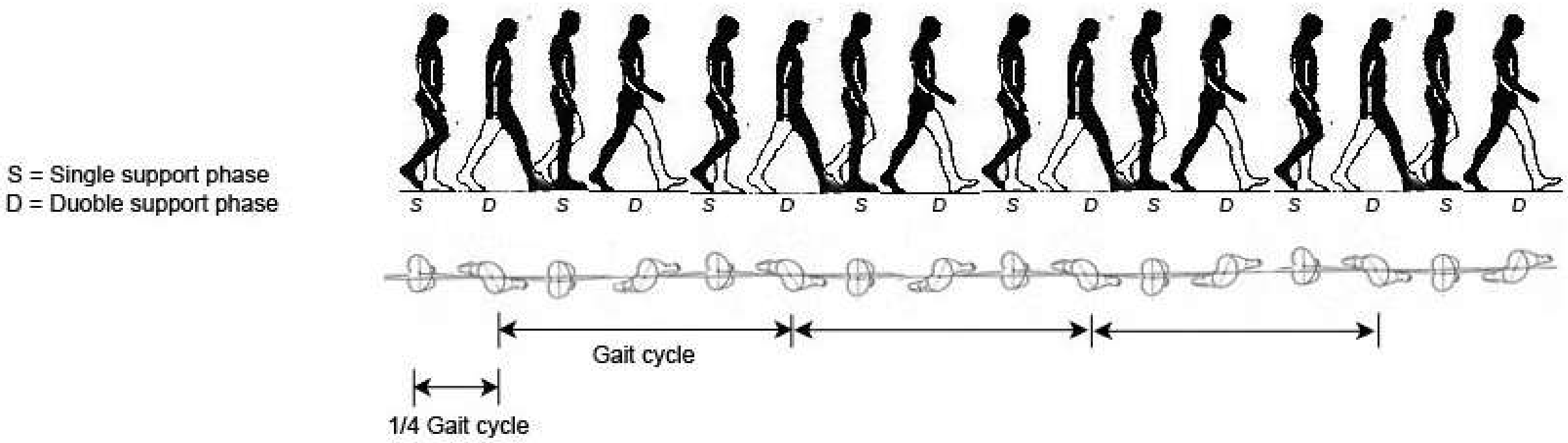

**Figure 3-8: Double and Single Support Phase during One Gait Cycle [45]**

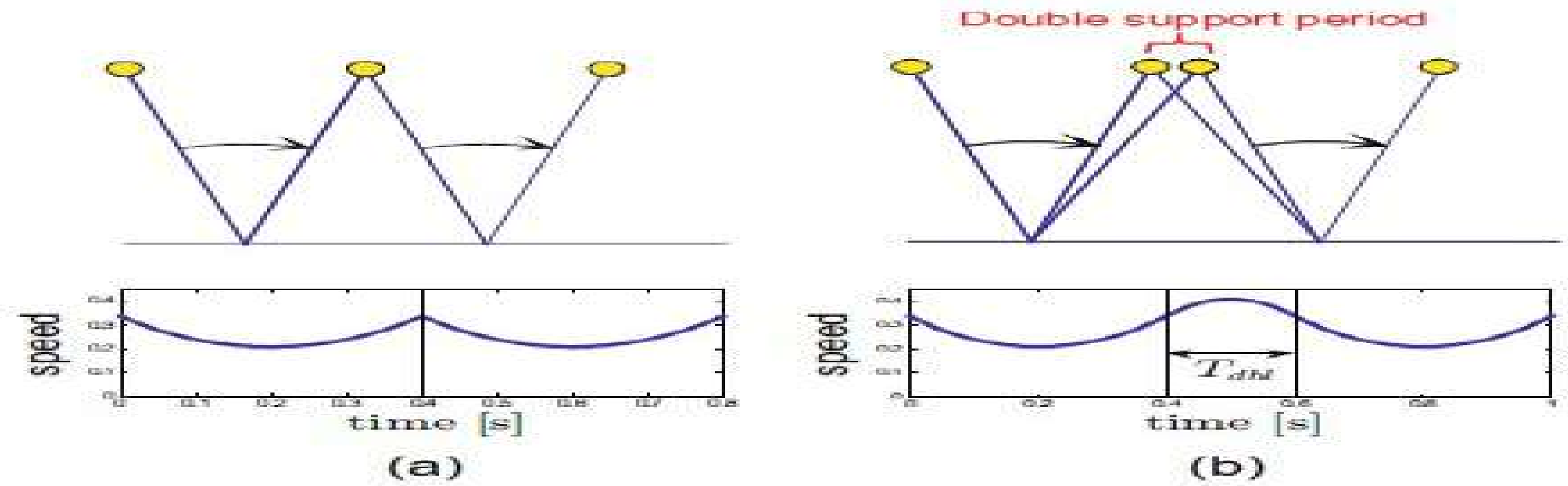

**Figure 3-9: Double Support Phase during Running and Normal Walk [45]**

## 3.6 Static and Dynamic Walk & Foot Translation

Static Walk: The projection of CoM will pass through the support polygon.

Dynamic Walk: The projection of CoM will not pass though the support polygon. Figure 3.10 is the projection of CoM during static and dynamic walk.

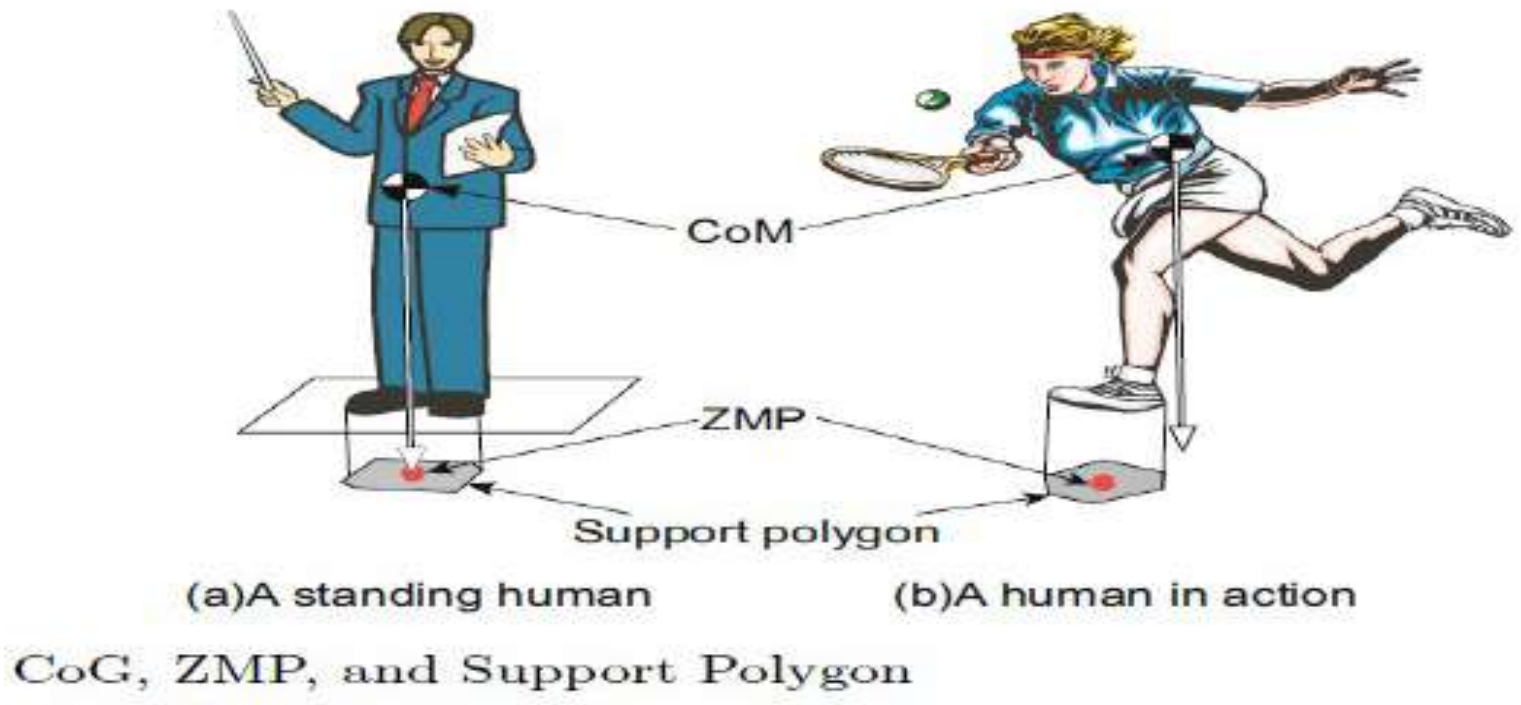

**Figure 3-10: Static & Dynamic Walk [10]**

When the foot is in a swing phase the foot is moving above the ground; when the foot is in a stance (support) phase, the foot is staying on the ground. Fig 3.11 is the representation of foot translation of left and right foot.

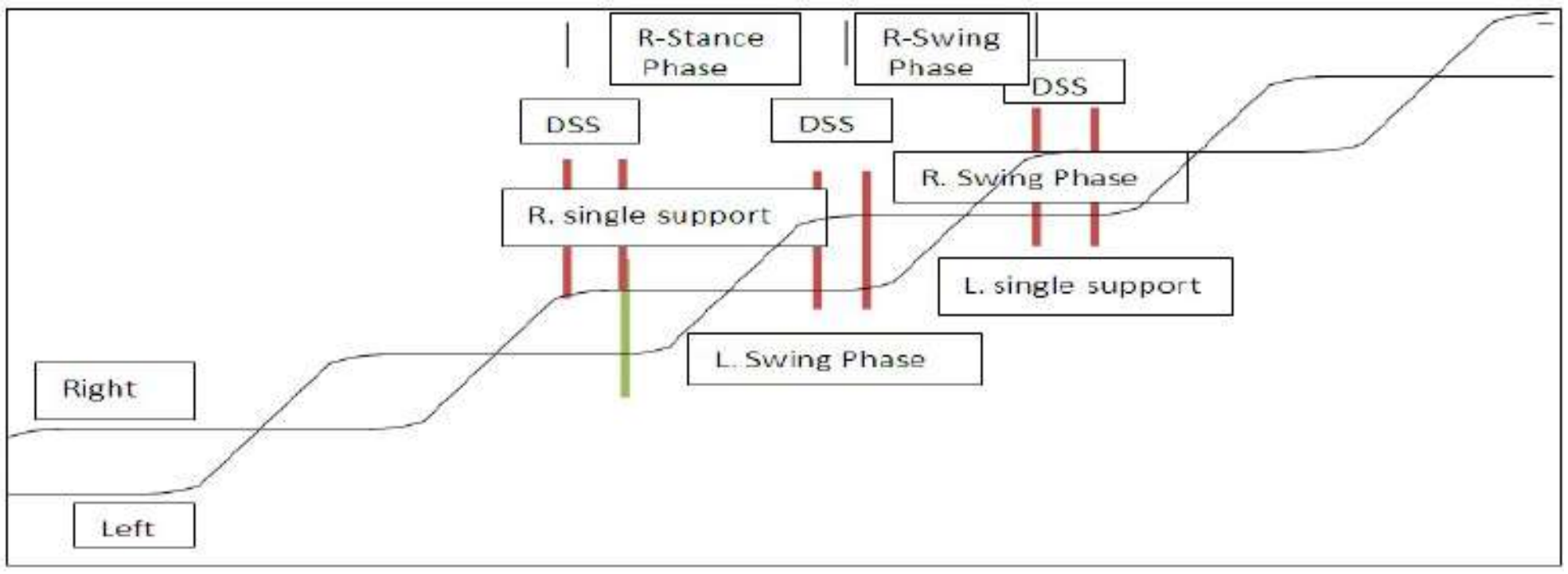

**Figure 3-11: Foot translation.**

## 3.7 Push Recovery

It is a very important behavior for bipedal robot to safely maneuver into cluttered environment without damaging itself. So, the bipedal robot must have the push recovery capability to safely maneuver in a dynamic real environment. Push recovery is the ability of human beings to recover from applied unknown external force with the support of other limbs. It requires strong co-ordination between human brain, spinal cord and other sensory organs, which govern motor action of human body. Figure 3.12 shows the push recovery strategies for different magnitude of forces.

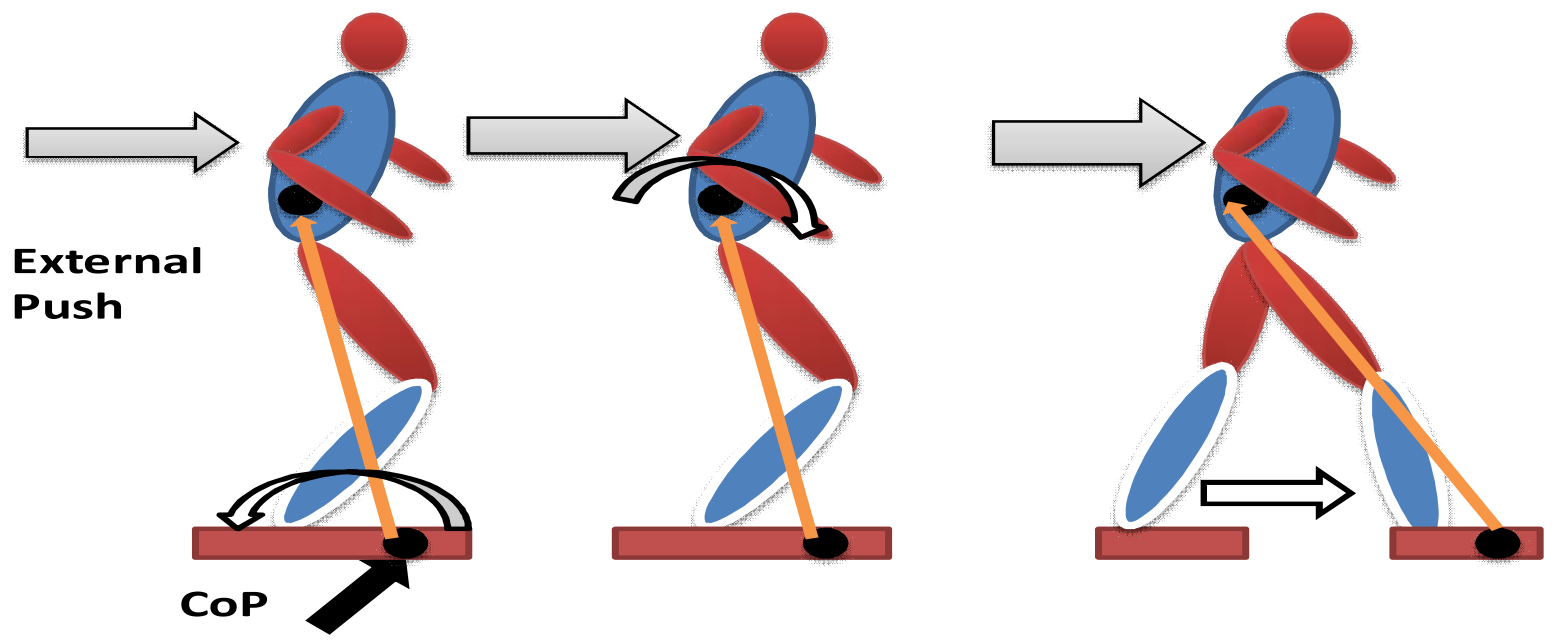

**Figure 3-12: Push Recovery Strategy [46]**

Whenever a robot experiences external force, it has to maintain its balance in order to avoid fall. If the push is small then it maintains balance through postural balance control, on the contrary if the push is large, it will take one or more steps to recover from the push. Stepping the appropriate region will lead to complete stoppage for the robot. The point where robot will step for stopping is called the capture point; it is a point where robot is able to bring itself to stop in one single step. Collection of such points is called capture region. When to take step: the robot is able to recover from the external force without taking step if a capture point is situated within the convex hull of the foot support area. Figure 3.13 showing the capture region [47].

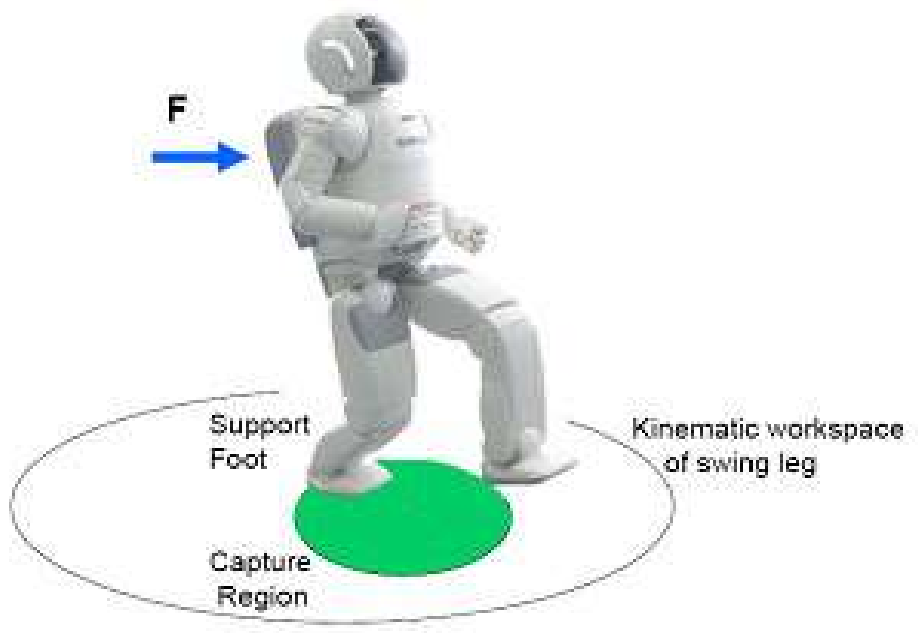

**Figure 3-13: Capture Region of Bipedal during Stepping [47]**

Where to take step: if the base of support regains an intersection with the capture region after taking a step, robot is able to bring itself to stop.

Failure: The humanoid will fail to recover from a push in one step if the capture region in its entirety lies outside the kinematic workspace of the swing foot. In this case the robot must take at least two steps in order to stop, if it can stop at all.

Push Recovery as explained by Benjamin Stephen [38] in his paper 'Humanoid push recovery', the simple models developed so far are extended to large push recovery. These models are used to develop analytic decision surfaces which will be the function of reference points like – center of mass, center of pressure, centroid moment point. This will predict whether a fall is inevitable or not. In push recovery there are mainly three strategies involved – (a) using ankle torques, (b) moving internal joints and (c) taking a step [47] [48].To make a machine learn the push recovery behaviour is an even more complicated task. Push recovery is an important technique that needs to be developed for an effective use of Bipedal humanoid robot in the day today works and in the mixed type of environment and also in house hold applications [49][50].There are three different strategies for push recovery as following:

- Ankle Strategy
- Hip Strategy
- Stepping Strategy

### 3.7.1 Ankle strategy

In ankle strategy as the name suggests the torque is applied at the ankle in opposite direction of force or push applied so as to prevent fall. The center of mass always stays within base of the support. The ankle strategy displaces the center of mass slightly when a standing posture is disturbed. The ankle strategy is realized only through ankle torque. A human body reacts with this strategy when a body experiences a small push on the back.

### 3.7.2 Hip strategy

In Hip strategy a forward lunge takes place to generate torque at the hip. In this strategy the center of mass displacement from the vertical is minimized by applying the torque on the hip mainly. A human body reacts with this strategy when a stronger push is applied on the back. With the stronger push on the back, the balance is maintained by applying bending in the hips.

### 3.7.3 Stepping strategy

In the hip strategy a step is taken in forward direction to prevent fall. This step is taken when the above two strategies fail to work to maintain the body balance. When this acting force or push is even stronger than the push or force applied in the above two strategies then for maintaining the balance, the base of support has to be changed. This is done by taking a forward step.

## 3.8 Research Questions & Important Terms:

### 3.8.1 Why Bipedal and what we expect from bipedal?

Bipedal is more suitable structure to work in human environment. The bipedal can avoid the obstacle and climbs the stair case more easily compare to quadruple robot. The bipedal is supposed to work in 4-D (Dull, Dirty, Dangerous and Difficult) environments. So, it is more suitable structure for such kind of jobs. The main benefits to accept bipedal are;

- Dexterity, Ability to step uneven terrain.
- Assistant for elderly people
- Replace the human beings during war
- Biometric Identification

- Help Amputee to recover from gait.

### 3.8.2 Why Hybrid Automata?

Response: A hybrid automaton is a language tool to describe the system which has two phases continuous dynamics and discrete switching logics. It helps in modeling the systems in real time. A hybrid automaton is studied from a dynamic systems perspective. The human walk is combination of different discrete sub phases and each sub phases has continuous dynamics. So to model a bipedal system as a part of dynamic system that is to model it as hybrid automata is of great relevance towards the study of bipedal walking [53].Necessary and sufficient conditions are derived for existence and uniqueness of the system.

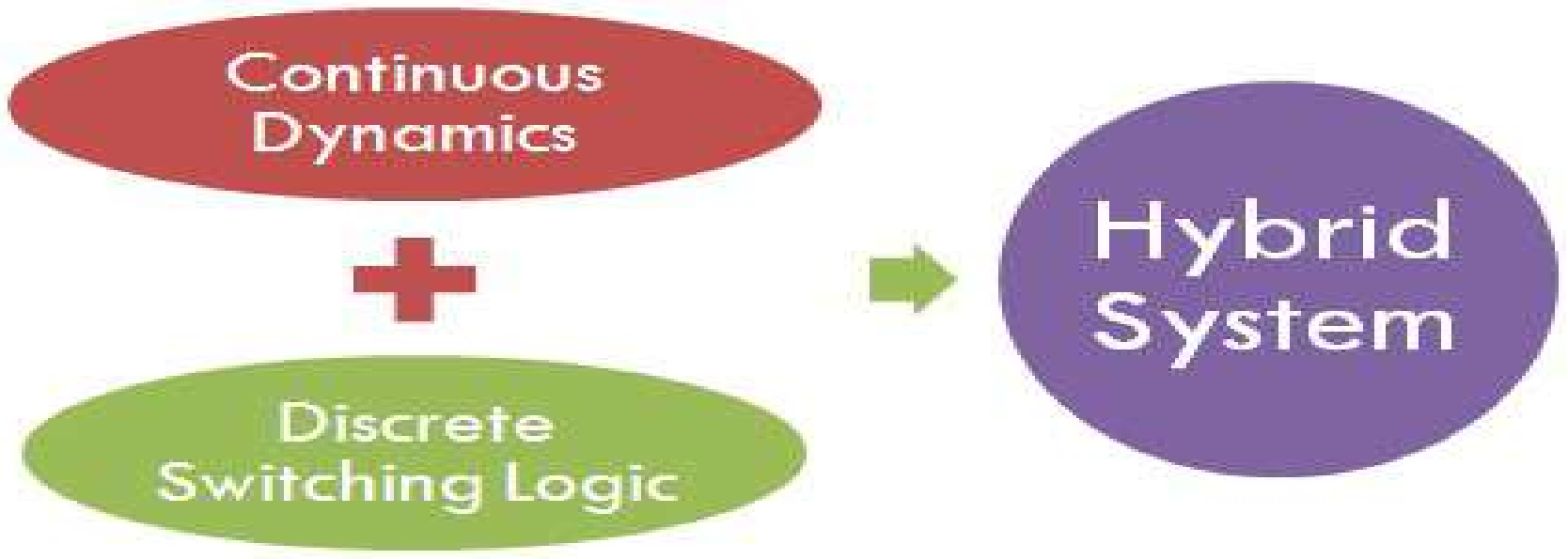

**Figure 3-14: Hybrid Automata Model**

### 3.8.3 What is Computational Model?

A computational model is used to study the complex behaviours of any complex system with the help of computer simulation. It is a real data driven. The figure 3.15 shows the base model for the verification of hybrid automata equations, here l1 and l2 are the length of thigh and shank respectively and l3 is foot length.

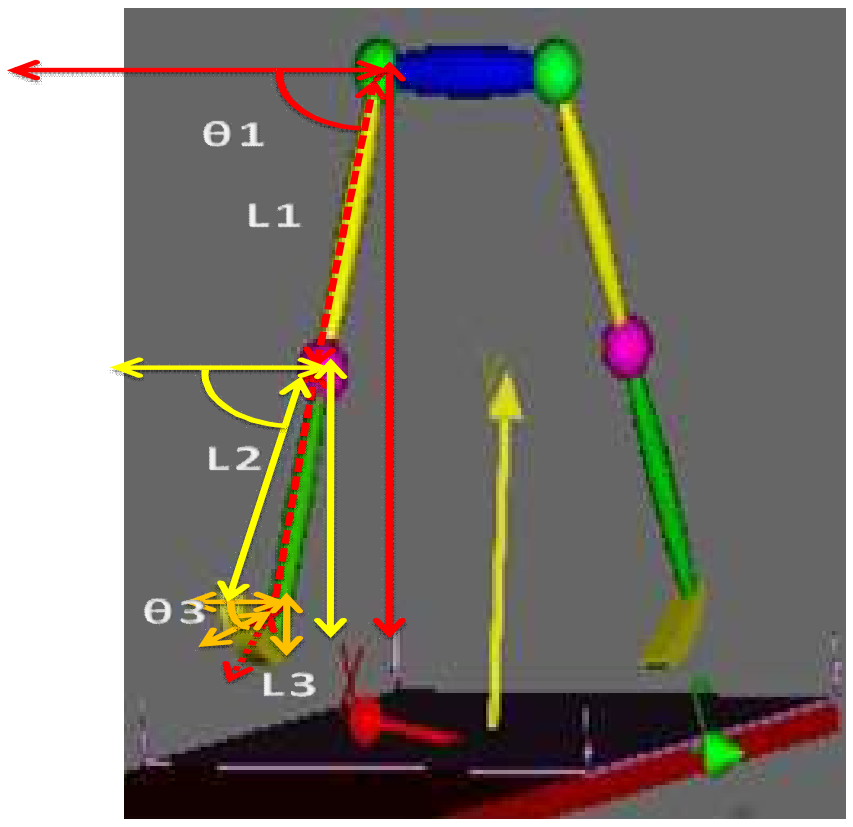

**Figure 3-15:Universal Computational Hybrid Automata Model**

Table 3.1below shows the detailed list of the input parameters for the generation of this computational model.

**Table 3:1: Input Parameter of Universal Base Model**

| Variable | Details about Variable | Type of Variable |
|---|---|---|
| L1 | Thigh Length(m) | Input |
| L2 | Shank Length(m) | Input |
| L3 | Foot length(m) | Input |
| m1 | Torso Mass(g) | Input |
| m2 | Shank Mass(g) | Input |
| m3 | Swing Mass(g) | Input |
| g | gravitational force | Input |
| Gamma(γ) | Slope of Plane | Input |

### 3.8.4 Why Computational Model?

The existing humanoid Robot moves keeping many constraints into consideration like contact of sole and ground, calculation of ZMP, Maintenance of different discrete sub phases. ZMP based humanoid cannot work when the terrains are not flat. Though analytical model have many potential benefits like fast computation but due to inherent limitation of a bipedal like high degree of freedom, more variables, different discrete sub phases (due to DSP and SSP) it is challenging to develop a more correct and accurate human like model. So these limitations lead to development of computational model [9]. Here we have designed the hybrid automata for human walk. The whole gait cycle is decomposed into 7 different sub phases and for each sub phase 6 equations are generated corresponding to left and right leg's Hip, Knee and Ankle joints.

### 3.8.5 Finite State machines:

Finite state machine diagrams are the important part of a Hybrid automata model. A finite state machine or just a state machine diagram is mathematical model for computation that is used to design computer programs or a computation model. It can be visualized as the abstract machine which can be existing in one of the finite no. of states. The machine can exist in only one state at a particular given point of time. The computational model consists of sets of finite states, a start state, the inputs and transition function. This functions maps the current states and input symbols to the next states. The computation for the model begins with the input value in the very start state. Then according to the transition function it turns to the next state. This might sound a bit complicated but actually in reality it is quite simple. Finite state machines are largely used in computer program designing, but it finds an extensive use in different other fields also, like biology, engineering, linguistics, and other sciences which are able to recognize sequencing.

### 3.8.6 Why it should be used for biped locomotion?

As we know a Hybrid automata is mathematical tool for modeling a dynamical system that merges the specifications for continuous and discrete behaviors of a dynamical system So, it is an appropriate model for designing a human bipedal trajectory in terms of hybrid automata vector fields because a human bipedal locomotion trajectory is also a combination of discrete and continuous states and a stable walk can be obtained using hybrid automata mathematical model. This bipedal trajectory can be very precisely obtained using hybrid automata methodology and is very appropriate tool for such dynamical system's designing.

### 3.8.7 Formulation of problem - How to model Hybrid Automata

To design learning based hybrid automata model for humanoid bipedal locomotion, a real time data is collected for walking as well as push recovery and the torque equations for each joint angle are found out. Human has a particular gait pattern for its walk. This gait cycle has mainly two phases that are swing phase and stance phase. The swing and stance phases are further sub divided into different sub phases. These are initial contact, loading response, mid stance, terminal stance, pre swing, initial swing, mid swing, terminal swing. With these sub phases a hybrid automata model is formed with different parameters defined and gait cycle data is divided into segments or phases for each joint angle of both legs. The gait data for Gait 2354 model is divided into seven phases for each joint of left and right leg in this model. The canonical equations are then formed from these divided sub phases data to give joint angles as the output. The time series data is used to find the equations of different joints trajectories using curve fitting tool in MATLAB. This model takes time values, mass and length as the input for this model. We get the output as hip knee joint angle values for left and right leg. This output is then applied on the OpenSim model for verification purpose of our hybrid automata model. The gait pattern curve for the actual gait and the output pattern generated by our hybrid automata model is compared. The error is minimized using regression. Finally it was observed that the hybrid automata model generates the same similar pattern and this strategy is universal which can be used alternatively for human walk. This model can further be helpful in the study of push recovery and artificial leg walking in different types of terrains.

### 3.8.8 What is a Vector field

A vector field plays an important role in this research work. By finding vector fields for different joints for each sub-phase of the gait cycle we can find the joint angle value of each joint of a biped at different instance of time. This vector field is a function of time which gives the joint angle value of any particular joint.

### 3.8.9 Role of Guard condition and Reset mapping in bipedal locomotion

The guards tell us when we are going to make the transitions from one phase to another. This model will depend on change in contact points and these guard conditions depend on the percentage of gait for each sub-phase. These changes in conditions are the guard conditions, because of which the model switches from one state to another. The contact points will define that at certain point or in certain domain which points on robot will be in contact with the ground, which will decide the phase of the gait where a robot lies in at a particular time. When the contact point will change, the robot will discretely move into the other phase of walking which will be having a different dynamic model and control. Directed cycles are made as the function of vertices and edges, the set of domains are defined, set of guards or switching surfaces are there. The time series data is divided into multiple sub-phases. The switch of one phase to another depends on the guard condition i.e. after a certain time interval (time series value) a state change occurs, and the vector fields for different joint angles are used at that particular state of the model.

Reset mapping or reset conditions tells that how the changes are affecting the states. We need to reset the values of the state after the transition from one state to another state occurs. This is the reset mapping of the states that we do by making changes in the state and its variable after transition is called reset. As a final output for a state, we need the reset conditions for the updating of state depending on certain guard conditions.

### 3.8.10 What is Limit Cycle?

Limit cycle is used to measure the stability of any non linear system with oscillations. It can tell whether the system is stable and unstable. Figure3.16 is the limit cycle curve of bipedal locomotion.

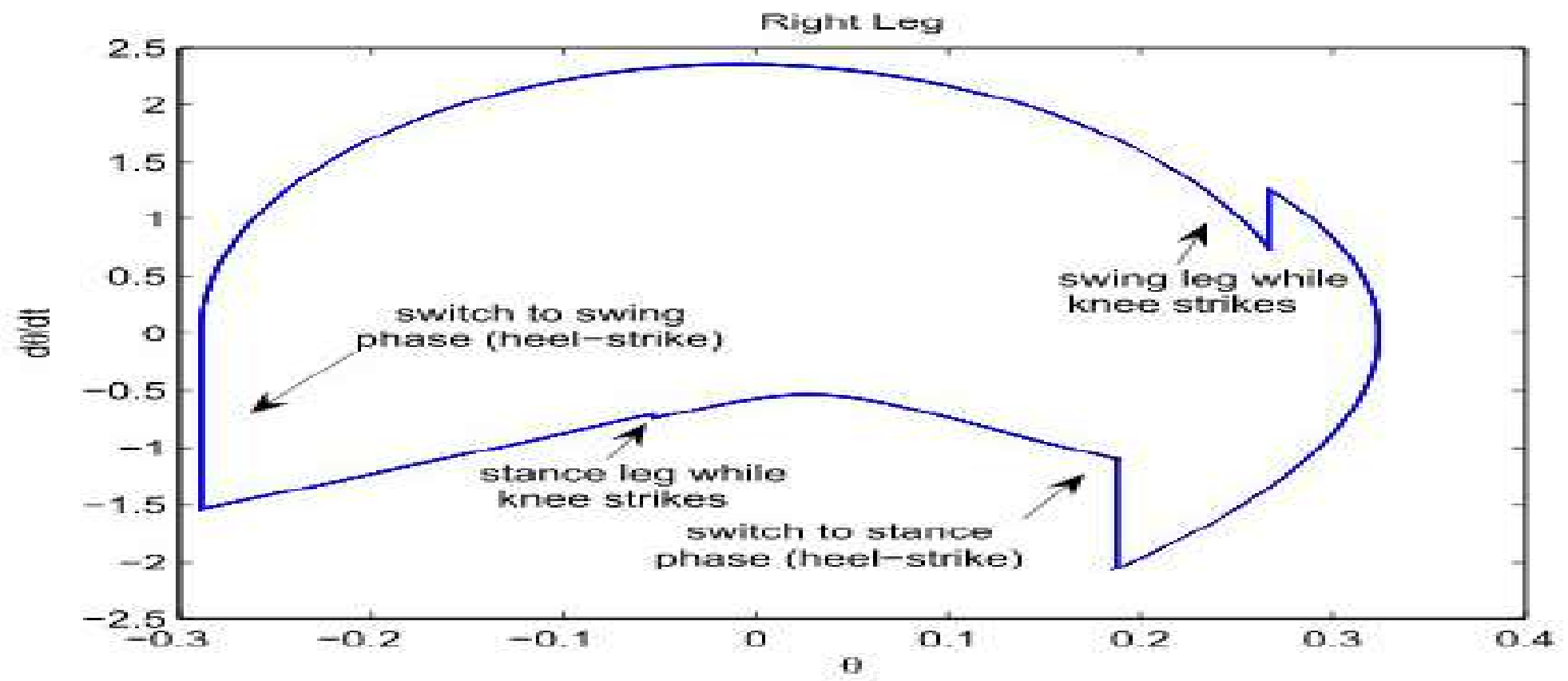

**Figure 3-16: Limit Cycle Curve for Standard Human Walk [64]**

Limitations:

- All the models based on only conventional mechanics and controls have inherent limitations.

- All models have flat foot which limits the bipedal to walk like human beings.
- Different people have different push recovery capability.
- The acquired push recovery capability, therefore, is based on learning.
- The mechanism of learning is not fully known to us. It is partial known to us.
- Researchers around the world are trying to explore this mystery through developing various models and implementing them on various humanoid robots.
- The computational model based on learning could be a reliable effective model.

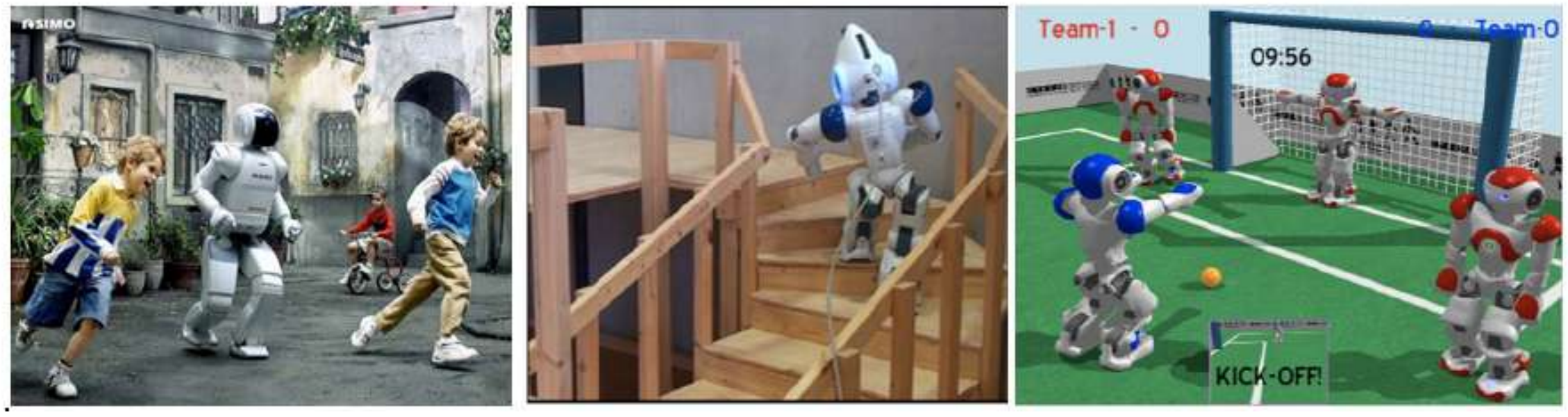

**Figure 3-17: Different Environment where bipedal are compatible**

## 3.9 Research finding, gaps and intuition

### 3.9.1 Challenges associated with human walk

One of the major challenges with bipedal is to accomplish energy efficient stable walking [55]. So far, most the biped available are flat footed with bent knees which are not energy efficient and slow. To achieve the stable walk and understand it perfectly we have divided the bipedal walk into different linear sub phases. Walk is considered as moving with a moderate pace by lifting alternative foot up and down when one foot is lifting up and another foot is putting on ground. Therefore one foot must be on ground at any time during walking. There are two types of walking, one is static and another one is dynamic walking. In Static walking, the projection of centre of mass never crosses the support polygon of foot during walk whereas during dynamic walk, the projection of CoM leaves the support polygon for some point of time. We perform the dynamic walk in our daily life [56][57]. The walking style for which we are familiar can be realized as dexterous control which is essentially unstable. In dynamic walk to prevent toppling the swing leg is brought forward to avoid fall. Such walking allows fast walk and less energy consumption for each gait. The statically stable walk can be performed by first shifting body weight to foot next stance leg and then swing the leg so ground contact will remain.

### 3.9.2 Why Hybrid automata model and Can hybrid automata act as classifier and a controller?

The Human walk is a combination of double support phase and single support phase. The DSP is 20% of entire gait and rest is SSP. The more correct representation of human walk divides the whole gait cycle into 7 different sub phases [58] [59]. The hybrid automata are evolved from time automata. The time automata is extension of Deterministic finite state automata in which states are finite with extra parameter constant time interval evaluation where as in hybrid automata the sequence of state appear into continuous evaluation of time. The hybrid automata represent the finite set of control steps with continuous dynamics. The continuous variables evolved according to an ordinary differential equation [60]. The systems jump the control states change when the continuous variables cross the certain threshold. The human walk is a combination of such discrete phases and continuous variable so we develop a computational model for walking using hybrid automata [61-64].Challenges: The challenges faced by biped during walking are as following:

- Switching between the swing and stance phases: The human walk consists of swing and stance phases. Each time the leg switches between these two phases. So it becomes important to select proper switching points as if the swing leg will left too early , the energy required for switching is not sufficient
- Swing Leg: During walking swing leg needs to leave the ground contact and subsequently work as stance in immediate forward step in time. In case of uneven terrain it becomes the complex task to keep orientation of leg with respect to ground.
- Stable trajectory of CoM:-The major challenge during swing phase is to maintain the balance of body by leaning forward and backward. The balance of body could be disturbed due to inertia during impact to ground. As CoM trajectory always lies in upper position of hip.
- Support of trunk
- Lateral Stability
- Control of forward velocity

## 3.10 Summary

The humanoid locomotion is a combination of double support phase and single support phase. Due to this configuration the humanoid behave like underactuated system during swing phase and over actuated system during double support phase. Which makes it difficult to develop the robot exact like human? The standard division of human walk into 7 discrete sub phases with continuous dynamic gives the actual glimpse of human walk. In our model the foot is not flat and we are configuring the model accordingly to particular person dynamics. In future work we will take care about dynamic stability.

How the hybrid automata are connected with polynomial trajectories? A hybrid automaton is a language tool to represent the behaviour of such system which has two main components continuous dynamic and discrete switching logic. We have explained the behaviour of human gait using the 7 discrete sub phases and generated the polynomial equation for each sub phases and configured our hybrid automata model based on subject and provide polynomial equation as input to each joint trajectory.

There is very clear correlation between the hybrid automata and the connected polynomial trajectories. The hybrid automata provide the language tool to modeling and design of any engineering system with continuous dynamic and discrete switching logic. It makes bipedal humanoid walking as a perfect example of hybrid automata. We have developed the hybrid automata equation to produce the joint trajectories for robot. We have compared the captured joint trajectory with other stable motion trajectory.

Finite state machine technique used to implement hybrid automata is classically done in humanoid robotics due to the switching contact. However it is usually not mention as hybrid automata because it commonly refers to a term in control theory to prove stability.

# Chapter 4: Experimental Setup and Data Collection

---

## 4.1 Deliverable:

In this research we will present the data collection techniques for both gait and push recovery data for different subjects. The subjects we have considered 10 left and 25 right handed people. We have captured human gait and push recovery data using several experiments with the help of indigenously developed wearable device HMCD (Human motion capture device) as well as mobile phone based HLPRDCD (Human Locomotion and Push Recovery Data Capture Device) as manifestation of joints (hip, knee and ankle) angle. We have considered the different terrain (flat & inclined) and different walking speeds (normal and brisk).

## 4.2 Proposed Method:

### 4.2.1 Experimental Setup and Data Acquisition:

We have designed the wearable mobile phone suit with accelerometer embedded named Human Locomotion and Push Recovery Data Capture Device (HLPRDCD) to capture the different joint (Hip, Knee and Angle) angle data which is the manifestation of push recovery and locomotion. Later we have fused the collected data into three directions x, y, z and we converted the captured values into biped's configuration space using inverse kinematics. Figure 4.2 shows the subject using HLPRDCD device to capture data. Earlier we developed the HMCD suit [3] to capture the push recovery data of human. We capture the real human data using indigenously developed wearable suit Human Motion Capture Device (HMCD) suit [1]. Figure 4.1 is the HMCD suit wear by left hand subject for capturing push recovery data. We captured the different joint angles change (knee, hip, and ankle) which is manifestation of push recovery. The control reverse torques can be computed for the joints in a bipedal humanoid using the equations 2:

$$\tau = M(\theta)\ddot{\theta} + C(\theta, \dot{\theta}) + G(\theta) - (2)$$

Where: M (θ) -inertial torque, $C(\theta, \dot{\theta}) +$ - centripetal and coriolis forces, G(θ) - gravitational force. HMCD have following parts six potentiometers (100 KΩ) for six joint Left and right Hip, Knee, and Ankle. When a force is applied, each joint move with certain angle, aluminum link with potentiometer attached in each joint will also move with certain degree (0° to 300°).These force and angle is measure in each reading by using FSR (force sensor) and potentiometer respectively. It gives value in digital counts (within range 0 to 999 due to limitation of potentiometer as it can rotate only 300 degree) from which we can further convert it into angular values using the given formula [6], given by equation (3) and (4).

$$Angle(Degree) = \left[(\theta - \theta_0) * \frac{300}{100}\right] (3)$$

where Θ is the observed and Θ0 is the initial joint angle value

$$Force(Newton) = \left[(f) * \frac{9.8}{100}\right] \quad (4)$$

Where, f = force in digital counts.

Force is applied from only one direction, due to limited area of force sensor, the range of FSR digital count is 1N to 100N and the sensing area of FSR 3105 is 14.5 cm². We used a wooden hammer like structure to stimulate push.

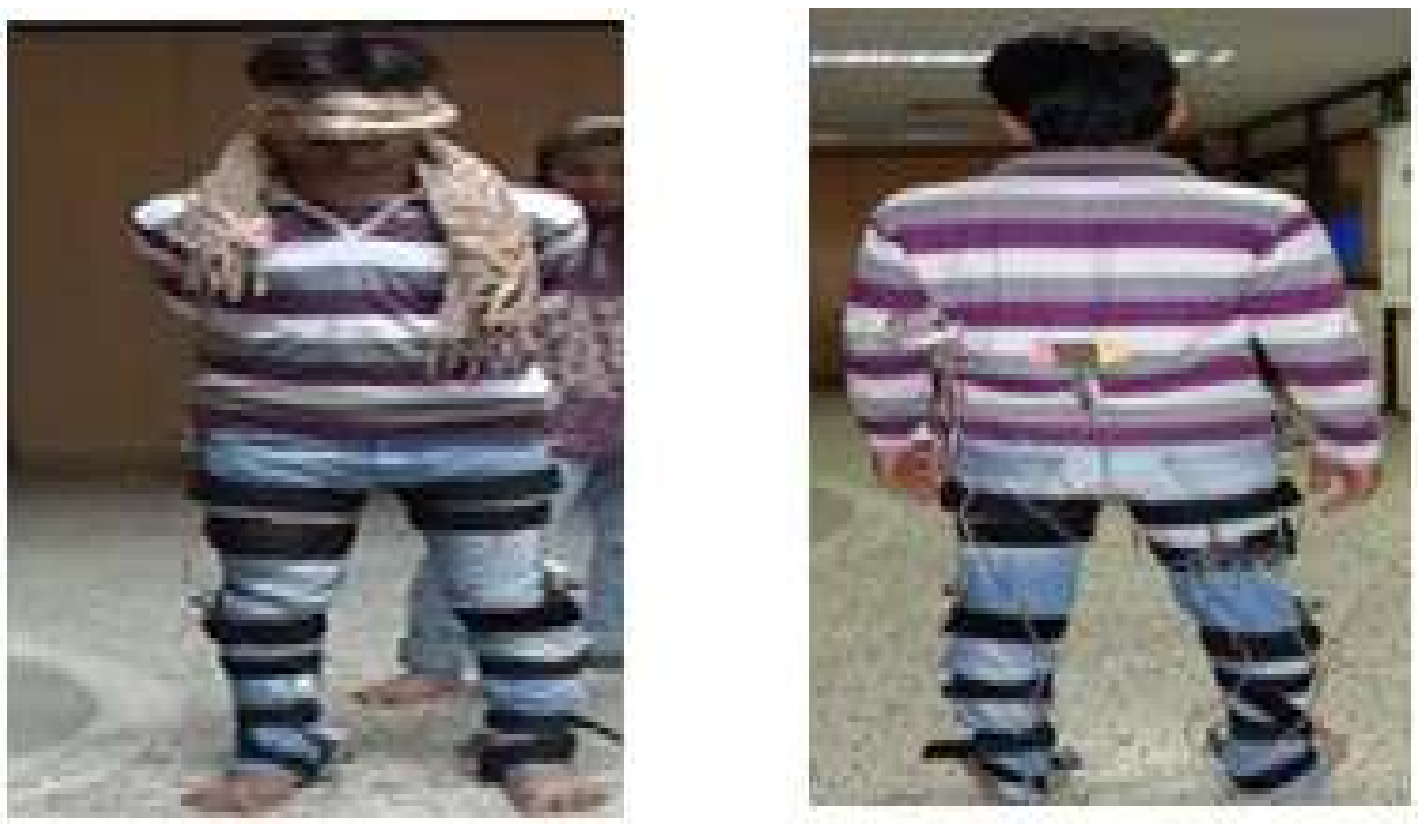

**Figure 4-1: Person wearing HMCD Suit (a). Frontal (b). Back view**

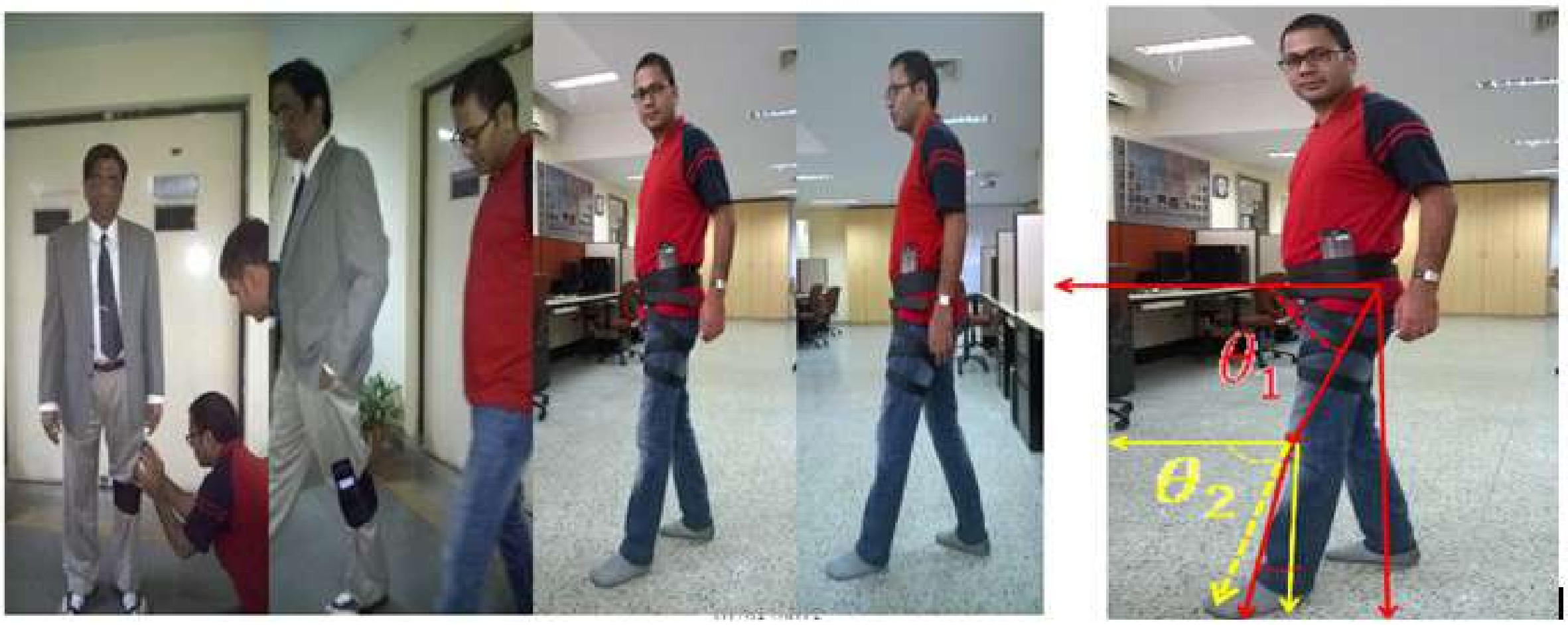

**Figure 4-2:Subject with HLPRDCD Suit (a) Frontal (b) Back view**

### 4.2.2 The process of the data acquisition for different subjects

The real data of human gait pattern towards push forces has been taken in eight different ways:

1. Open eyes with lunging technique in the static mode
2. Open eyes without lunging technique in the static mode
3. Open eyes with lunging technique in the dynamic mode
4. Open eyes without lunging technique in the dynamic mode
5. Closed eyes with lunging technique in the static mode
6. Closed eyes without lunging technique in the static mode
7. Closed eyes with lunging technique in the dynamic mode
8. Closed eyes without lunging technique in the dynamic mode

### 4.2.3 Data Correction and Smoothing:

Steps for data Collection:

1. Mounting HMCD and HLPRCD devices at appropriate joints.
2. Push recovery and locomotion is the manifestation of changes in the 6 joints angle values. We have collected data of six joints angle for 25 right and 10 left handed persons for the different magnitude of force by our developed suits. As push recovery is different for right and left handed person.
3. To remove the noise performed the Zero Correction of data by subtracting the first row from all time series data.
4. Applied the cubic spine to smooth the data.
5. Finally Error correction.

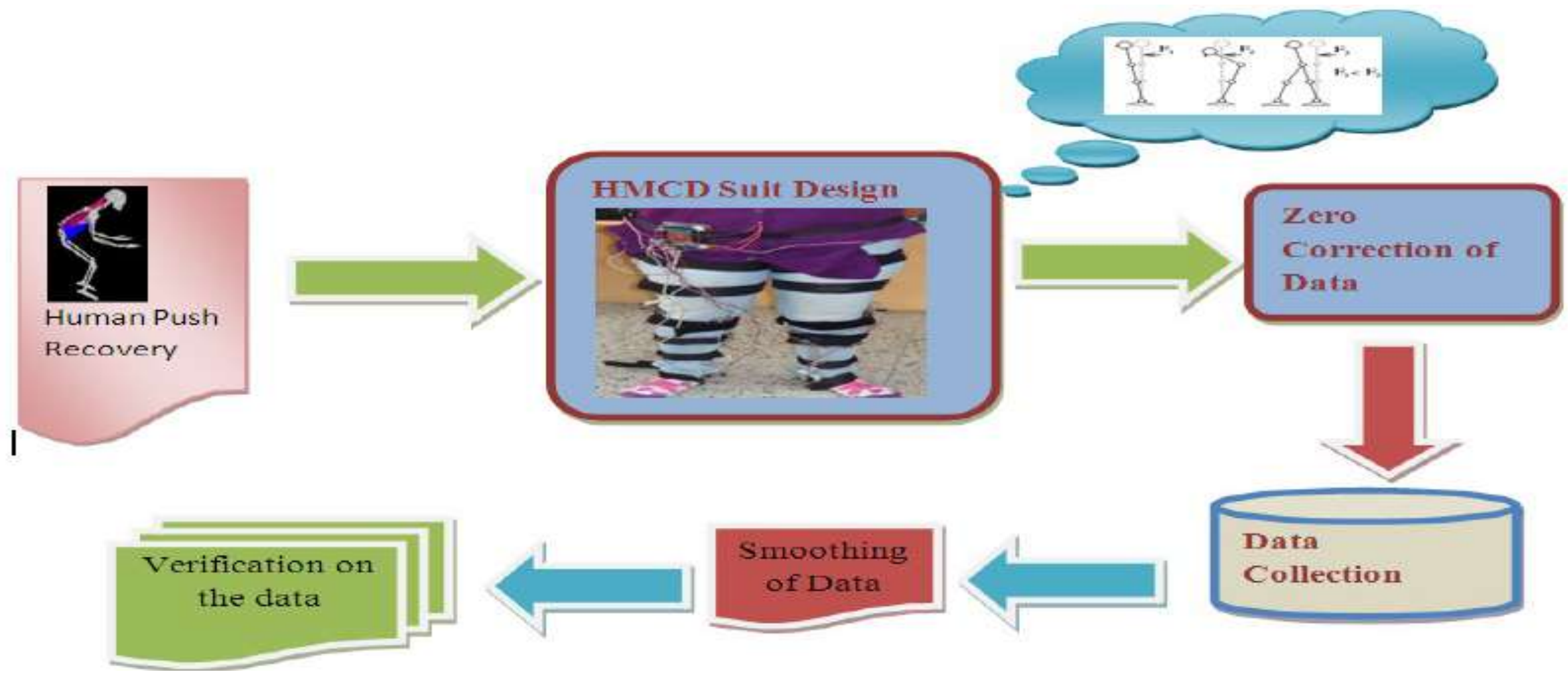

**Figure 4-3: Sequence of steps for push recovery experiment**

The ideal curve of hip, knee and ankle joints will be disturbed due to applied force from behind. To make it smooth we have applied the cubic spine. The initial reading of all the sensor is not zero in beginning due to noise, so we have subtracted the all the value by performing the zero correction. The above Figure 4.3 represents the entire process of data acquisition. The algorithm1is the converting accelerometer data into θ.

---

**Algorithm1**: Extracting data from accelerometer to biped's configuration space using inverse kinematics - Complexity $\theta(n)$

---

***Input***: $LeftKnee_x(x[n]), LeftKnee_y(y[n]), lengthofthigh(l_t),$
$lengthofshank(l_s), constant(k1) \& constant(k2)$
***Output***: $\theta_1[n], \theta_2[n]$
***Initial***: $l_t \leftarrow 5;\ l_s \leftarrow 4;$
***Begin***

    ***fori*** $\leftarrow$ **1**: ***n***

$$tmp[i] = \frac{x[i]^2 + y[i]^2 - l_t^{\,2} - l_s^{\,2}}{2 \times l_t \times l_s}$$

    ***Endfor***

    $tmp_{max} = \max(tmp)$ *% where 'max' will find the maximum*

***fori*** $\leftarrow$ **1**: ***n***

$$cos\theta_2[i] = \frac{tmp[i]}{tmp_{max}}$$

$$sin\theta_2[i] = \sqrt{1 - (cos\theta_2[i])^2}$$

$$k1[i] = l_t + l_s \times cos\theta_2[i]$$

$$k2[i] = l_s \times cos\theta_2[i]$$

$$\theta_1[i] = atan2(y[i], x[i]) - atan2(k2[i], k1[i])$$

$$\theta_2[i] = atan2(sin\theta_2[i], cos\theta_2[i])$$

    ***Endfor***

***EndBegin***

---

## 4.3 Results and Accuracy Calculation

### 4.3.1 Analysis of Bipedal Locomotion for Different Joints (Using HMCD)

Ideally a person has oscillatory motion in hip which is almost sinusoidal, has two sharp humps for ankle and two humps in knee joints curve for normal walking pattern without external force [57].It is a complex co-ordination of muscles and nerves i.e. motor system, sensory organ (sensation) and other parts of the brain which help co-ordination [58][59][60].

We have captured the data for normal walking of different subjects using HMCD and HLPRCD device. Below curves represent the gait pattern of right hand persons in three different planes. There is 180° phase difference between both the left and right curve. The plot for hip joints in sagittal plane shows oscillatory motion and swing in the angle of hip joint. We have found some very good observation and results after studying the gait cycle in depth. The observations are as follow: The peak point of vertical oscillator (about 2 inches) can be achieved when the unilateral weight is maximum and the lower extremity is in full extension. This configuration can be achieved during the near right mid stance of single support limb and mid swing of another non-weight limb. The lower point is reached when the distance is maximum between both feet i.e. when both feet are on the ground, one foot at toe off and another one foot at heel strike. This posture we called Double Support Phase (DSP) and it appear for almost 20% of complete gait cycle [64]. The DSP period would be zero during running. The below figure 4.4 represents the ideal curve for hip, knee and ankle joints. The next below curve represents the left and right hip joints curve in different plane. Refer the figure 4.5.

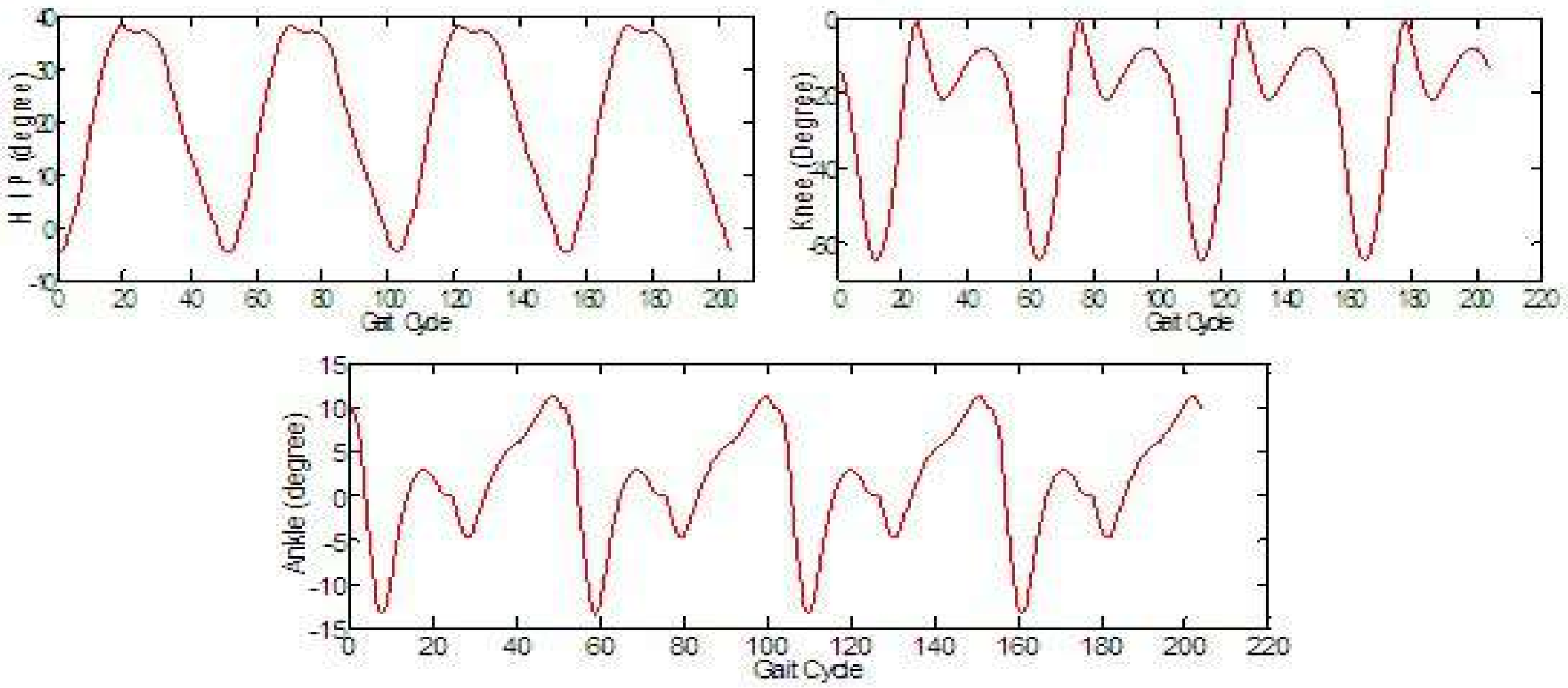

**Figure 4-4:An ideal gait pattern of human (a) hip, (b) knee and (c) ankle respectively**

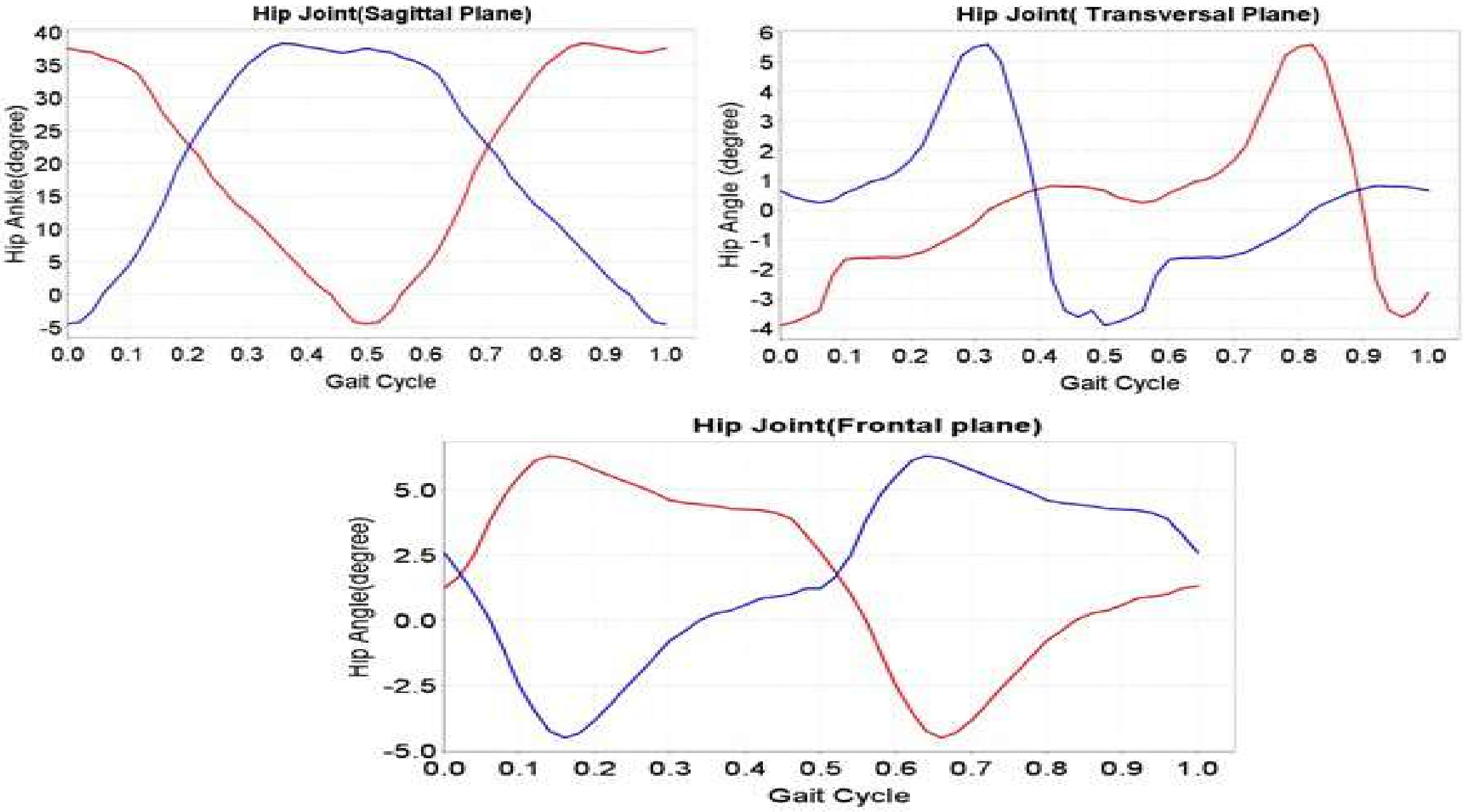

**Figure 4-5: Left and Right Hip Joint Curve for different plane (a)- Sagittal (b)-Frontal (c)-Transversal plane**

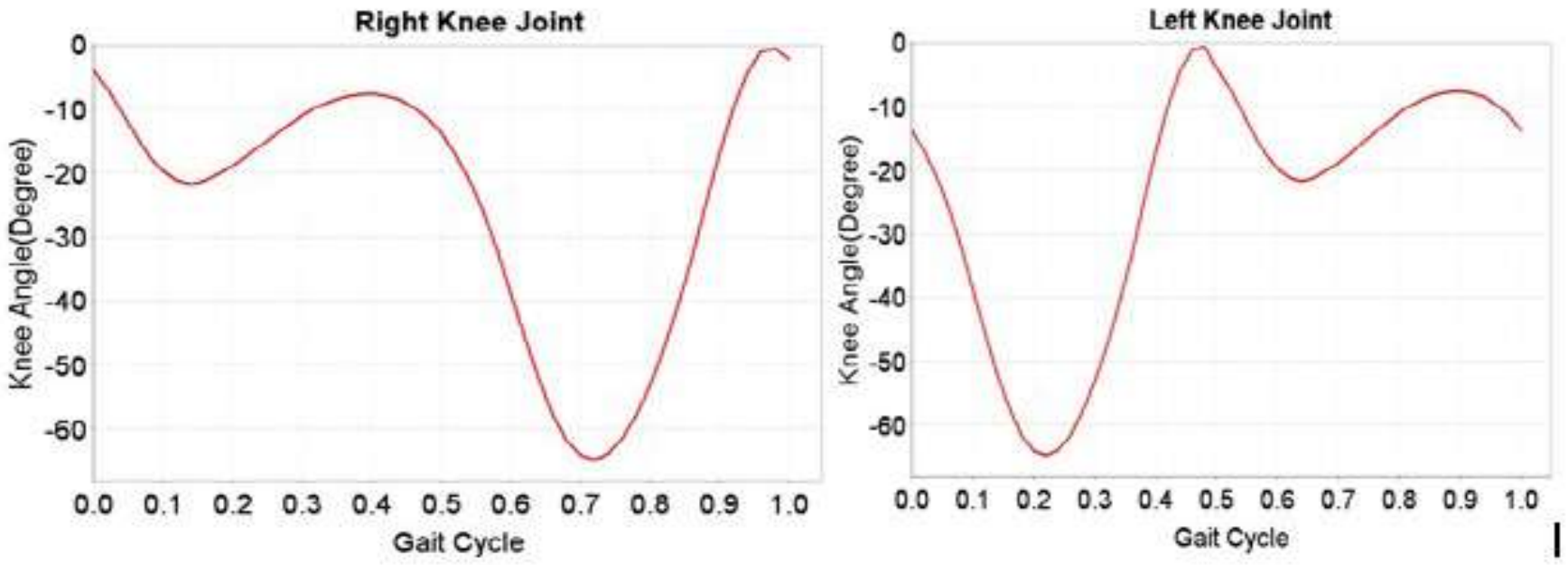

**Figure 4-6: Left and Right Knee Joint Curve**

Figure 4.6 represents the curve for Right and Left Knee joint; there is one small bump and the big bump for right leg and reverse for right leg. This explanation shows that the high point can be reached during between heel strike and toe off and lowest point during right mid stance.

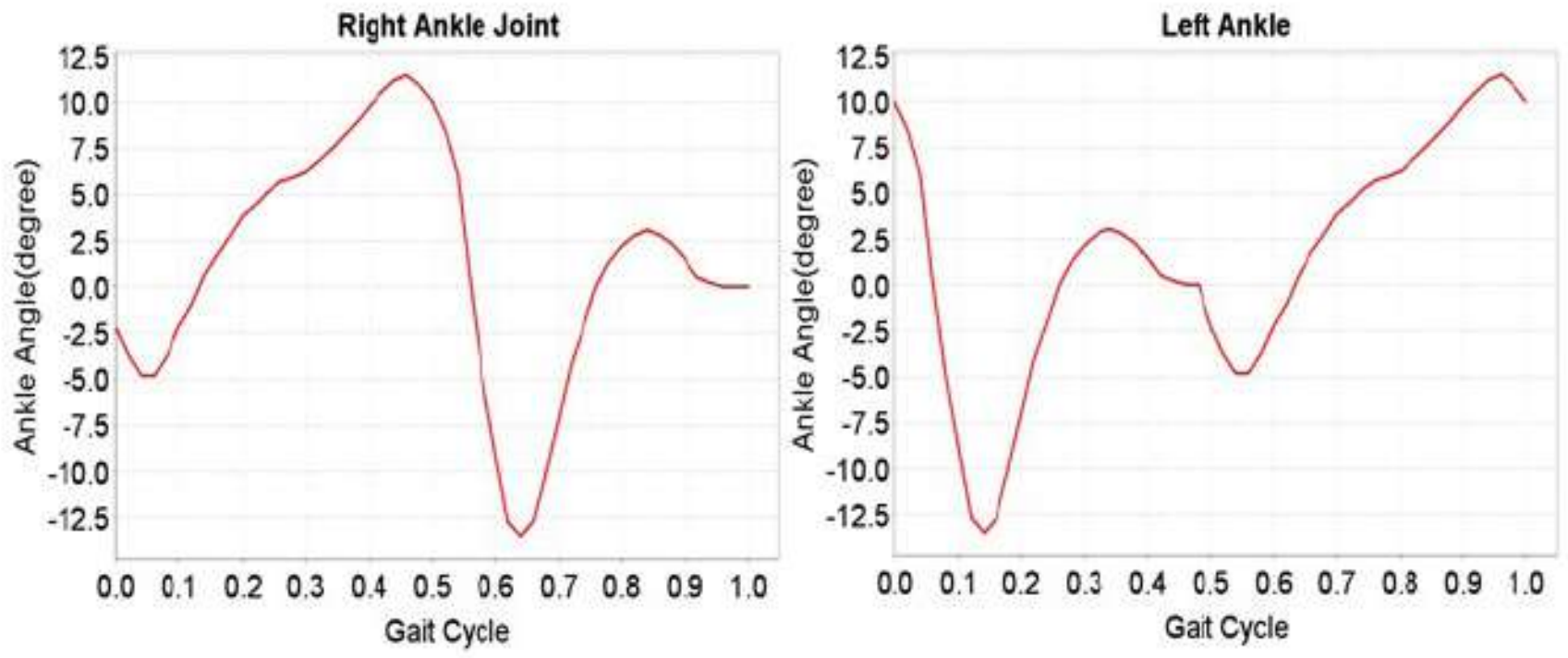

**Figure 4-7: Left and Right Ankle Joint Curve**

Figure 4.7 is the plot for Right and Left Ankle joint curve, which shows one big bump then a small bump for right leg and reverse for right leg. The above explanation shows that the high point reached during between right mid stance and lowest point during toe off and heel strike. All the three curve represents the complete description of the bipedal locomotion and verification of data capture for different subject and model with ideal curve.

### 4.3.2 HLPRCD Captured Data:

The observed leg joint curve for right handed person's right and left leg is shown in Fig. 4.8. In our experiment, results have been noted for both left and right handed subjects with specific setting i.e. closed eyes without hand movement to produce robot like environment. Fig. 4.8 shows the graph of angle variation of hip and knee (in degree) versus time duration for a right handed person .In given graph we can see an oscillatory motion of hip, very close to ideal one[65][66]. This can further be used in the analysis of crouch or abnormal subject [67]. It is one step more advance of our previous research toward more sophisticated device with less error and more accuracy.

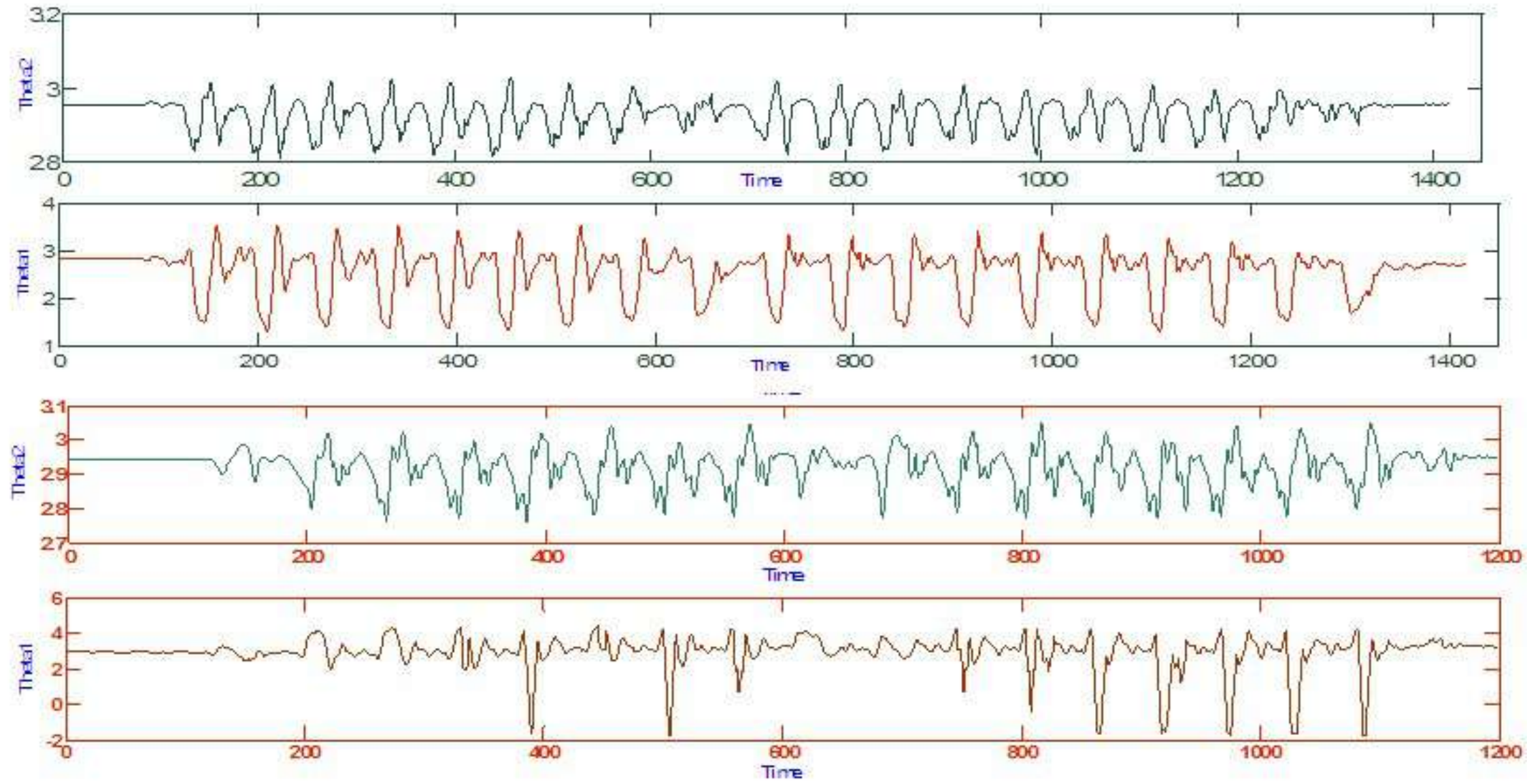

**Figure 4-8: Observed Left Hip, left Knee and Right Hip, Right Knee Joint curve for Right and Left Leg**

## 4.4 Towards developing a Computational model for Bipedal Push Recovery

The human being can negotiate with external push up to certain extent reactively. Grown up persons have better push recovery capability than human kids and also the professional wrestlers acquire better push recovery capability than normal human being. The acquired push recovery capability, therefore, is based on learning. However, the mechanism of learning is not known to us. Researchers around the world are trying to explore this mystery through developing various models and implementing them on various humanoid robots. All the models based on conventional mechanics and controls have inherent limitations. We believe appropriate computational model based on learning will be able to effectively address this issue. Accordingly we have collected extensively humanoid push recovery data using our innovative idea of exploiting the accelerometer sensor of smart phone. Through our experiments we have studied the human push recovery by fusing data at feature level using physics toolbar accelerometer of android interface kit. The subjects for the experiments were selected both as right handed and left handed. Pushes were induced from the behind with close eyes to observe the motor action as well as with open eyes to observe learning based reactive behaviors. A LVQ (Learning Vector Quantization) based classifier has been developed to identify the coordination between various push and hip and knee joints [65].

## 4.5 Push Recovery Result for Different Subject

Figure 4.9 describes the accuracy curve for the two push recovery strategies for different population size using LVQ (Learning Vector Quantization [66][67].

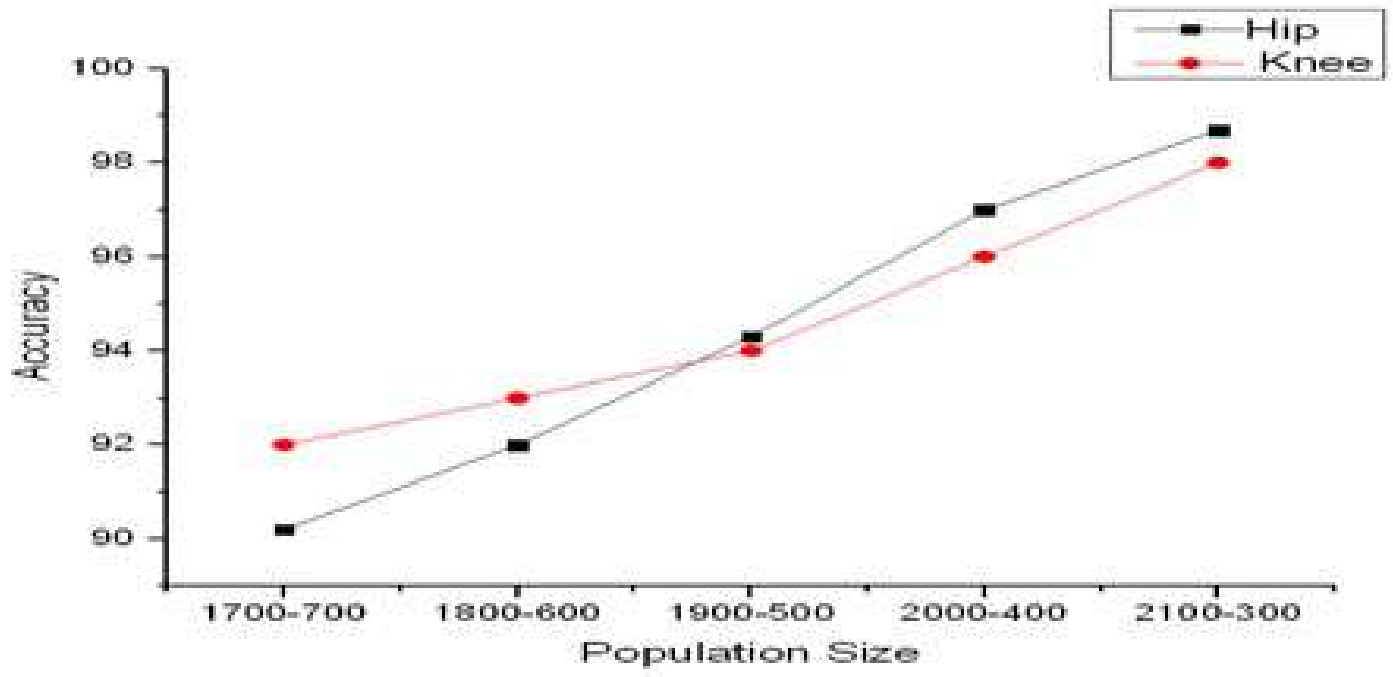

**Figure 4-9: Accuracy of classifier over different strategy**

The below all the figure are for push recovery curve for different subject. We have plotted for total 6 subjects, out of which 2 are left handed and 4 are right handed persons. The time is measured in millisecond.

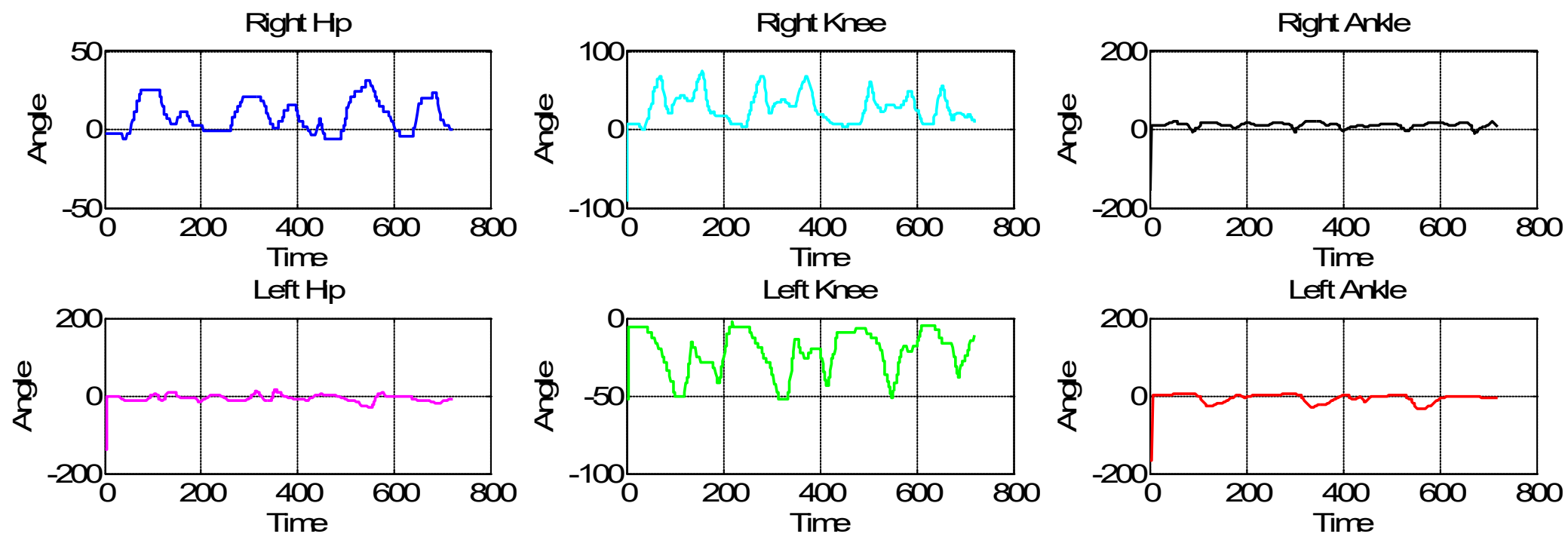

**Figure 4-10:Subject1 right handed person Push recovery plot for all six joint**

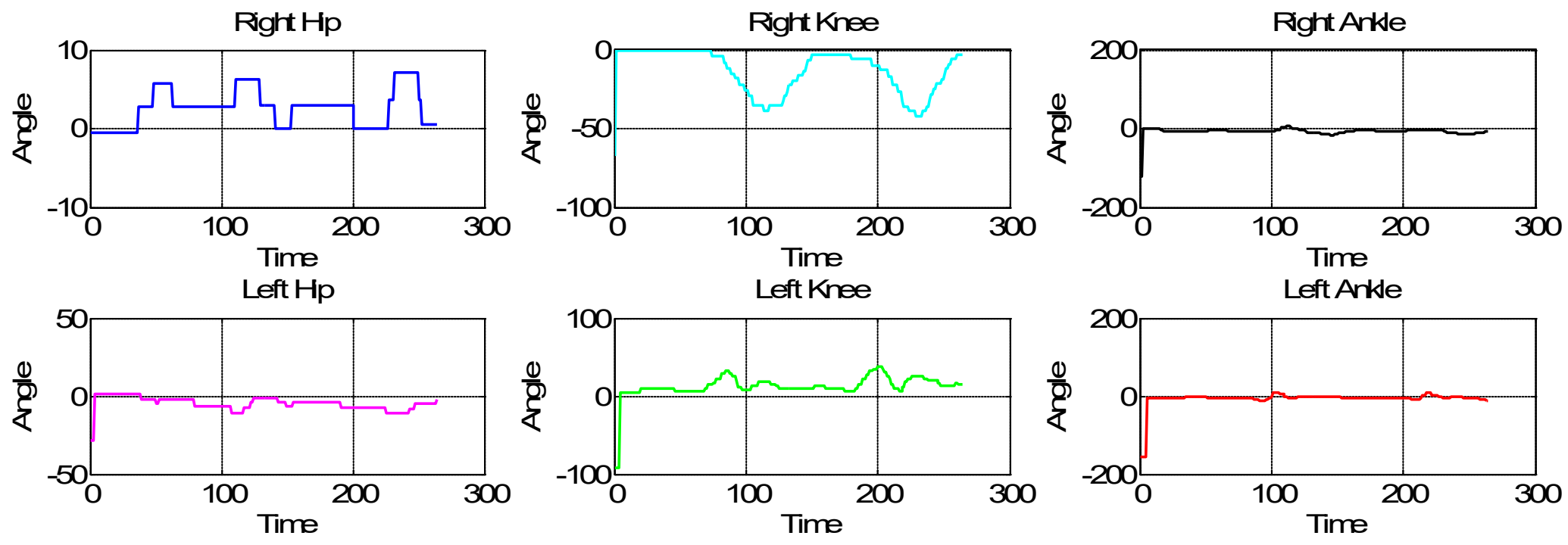

**Figure 4-11:Subject2 right handed person Push recovery plot for all six joint**

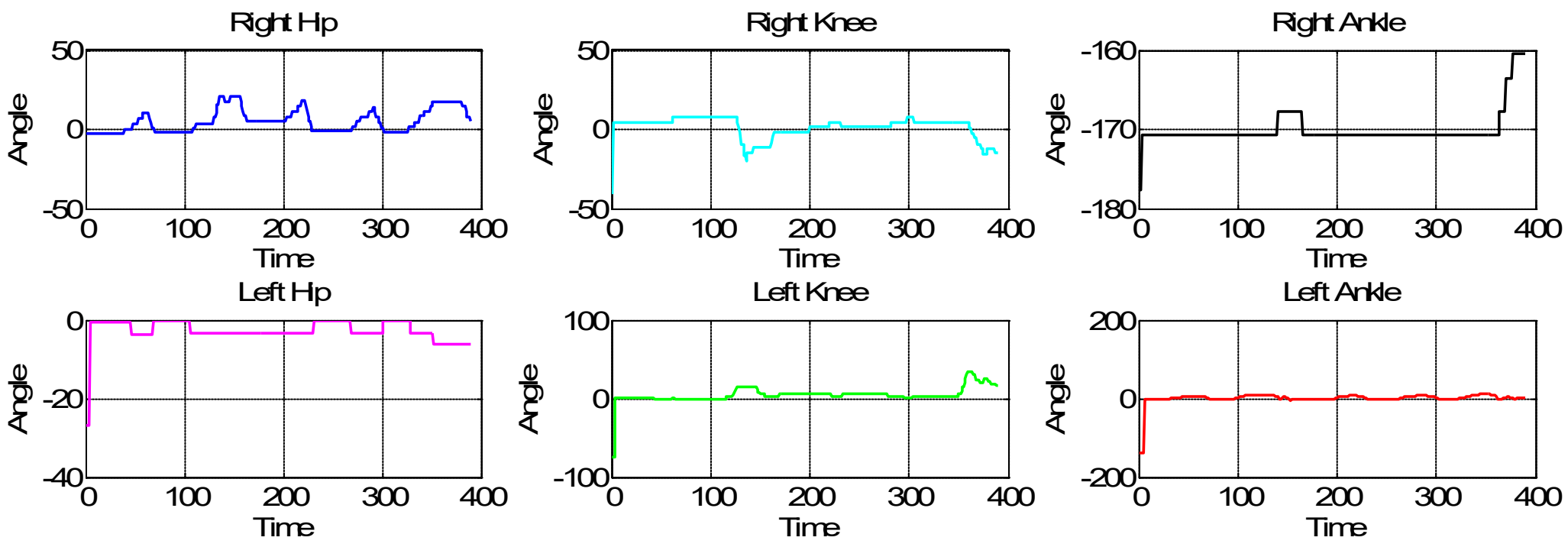

**Figure 4-12:Subject3 right handed person Push recovery plot for all six joint**

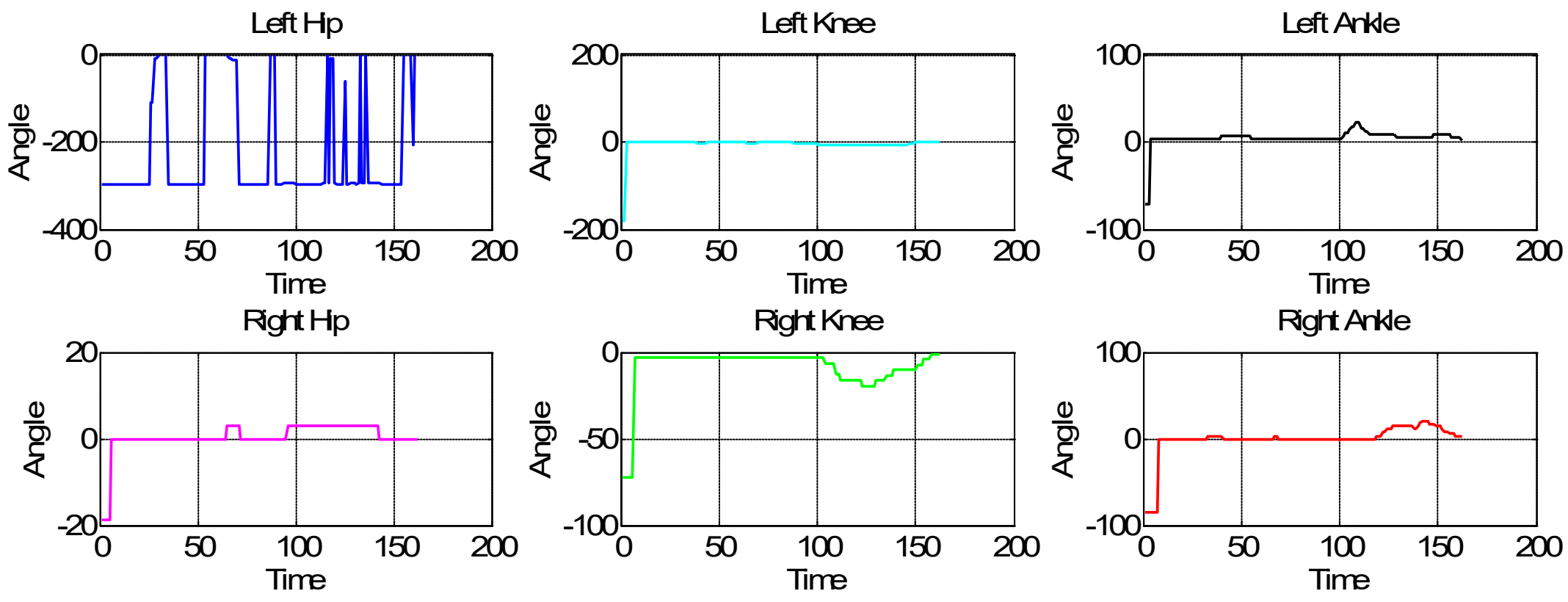

**Figure 4-13:Subject4 right handed person Push recovery plot for all six joint**

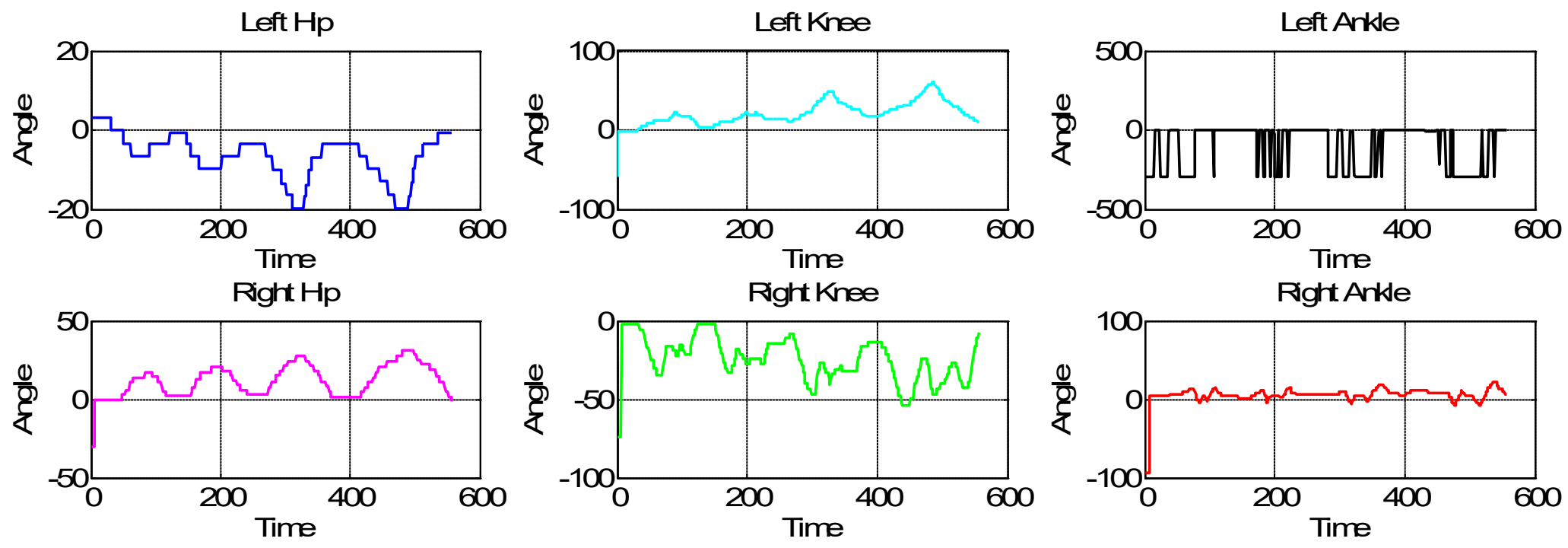

**Figure 4-14:Subject1 Left handed person Push recovery plot for all six joint**

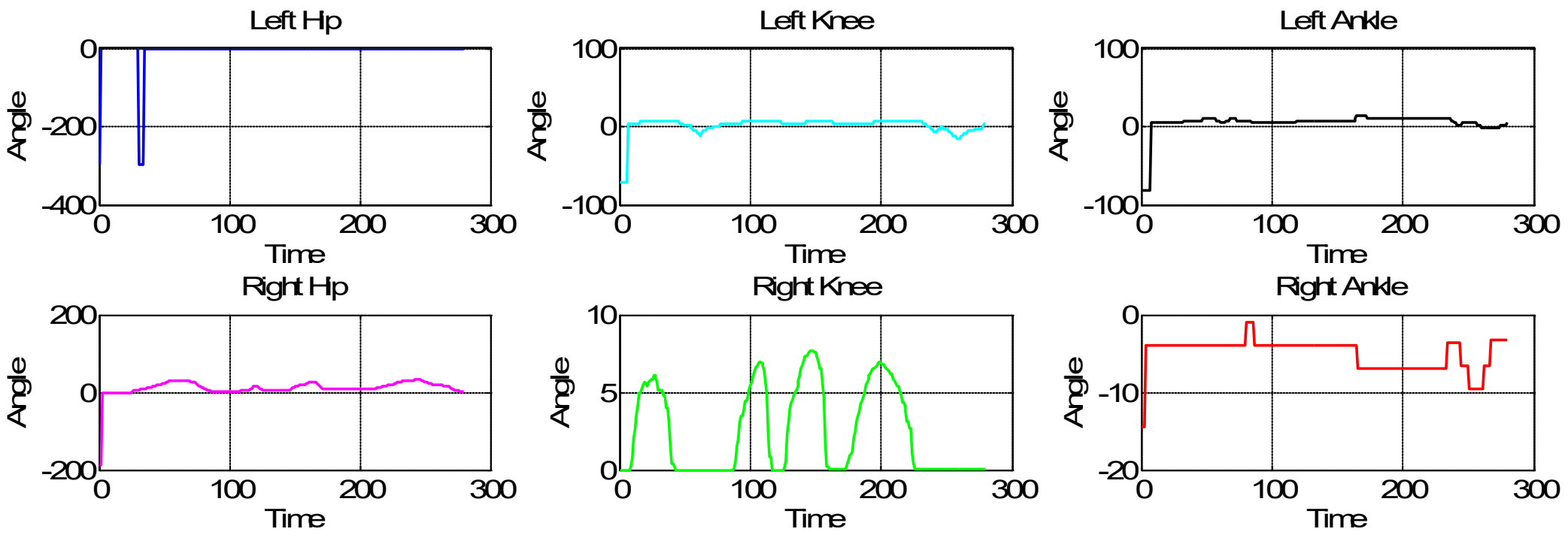

**Figure 4-15:Subject2 left handed person Push recovery plot for all six joint**

## 4.6 Summary

The result is an important contribution toward our hypothesis the push recovery is a software engineering problem.

### 4.6.1 Analysis:

1. Push Recovery is depend on age, sex, weigh, height i.e. based on human anatomy.
2. In push recovery pattern of human corresponding knee and ankle joint angles are inversely proportional to each other.
3. A push recovery pattern in human depends on whether a person is left handed or right handed.
4. Knee is the most active joint in push recovery of human.
5. Push recovery is a software engineering problem instead of a hardware engineering problem.
6. We analyzed the gait pattern of human that left side joints are more active than right side joints of a right handed person while in case of left handed person, right side joints are more active.
7. This swing phase is critical since only one leg needs to carry the entire body load stably and balancing is difficult like inverted pendulum balancing.
8. The load transfer between these two phases is inherently nonlinear.

### 4.6.2 Challenges

1. The data collected through HMCD is not ideal gait pattern of human due to noise.
2. The size of force sensor is only $14.5cm^2$ .
3. FSRs are not reliable due to the excessive noise; requiring day-to-day calibrations. It seems that the data collection rate is low.
4. Difficult to analyses the push recovery for ambidextrous person.
5. In the phidget kit the Digital counter of sensor having a range of 0 to 999 only.
6. As the controller has totally symmetric control for left and right parts of its body.
7. It is impossible to make a humanoid balance without closed loop control for torque commands, at least in servos.

### 4.6.3 Physical Observation of Human Gait:

1. The CoM is positioned always above the hip and the pattern of motion of the COM is sinusoidal in nature both in the vertical and the horizontal plane.
2. Each gait cycle have two phases swing and stance, both occur alternately i.e. when one leg in swing phase another will be in stance phase and reverse.
3. Each and every person have unique gait signal whereas the phases and their response time in general are almost common to all persons.
4. This periodicity in gait signal has to be exploited in order to achieve tangible results.
5. The motion signal of the knee is highly non-linear due a double hump which is noticed in the knee signal. This makes things much more difficult to model. All the difficulty arises from this non-linear signal.

This chapter presented a process of data collection though wearable device. The next chapter of this thesis will discuss about development of computational model using Hybrid automata.

# Chapter 5: Modeling Bipedal Locomotion Trajectories Using Hybrid Automata and Cellular Automata

The analysis of human gait and push recovery data presented in previous chapter 4 motivated us to develop the data driven computational model. The computational model is morphological similar to human being. The human walk is very efficient and effective. The human similar data can be used for bipedal robot development which can overcome the problem of kinematical base model presented in chapter 3.

## 5.1 Background

The modelling of the joint trajectories of a biped locomotion correctly is a highly challenging problem due to the complex nature of biped locomotion. In human as well as in biped humanoids the periodic walking pattern which is known as gait, consists of both discrete (known as stance phase) and continuous phases (known as swing phase). In this research work we present a computational model for prediction, formal verification and analyses of joint trajectories of bipedal locomotion using powerful theoretical computer science framework which is known as automata theory-more precisely hybrid automata technique. Human walk is the combination of 7 different discrete sub-phases [68]. To develop the human like bipedal robot, the walk cycle is divided into 7 discrete sub phases [69]. Each sub phase has its own continuous dynamics. To express the phase trajectories more accurate the hybrid automata is proposed. The bipedal walk is configured as the rocking block model [70]. During DSP (Double Support Phases) it is vertical rectangular plane and during left, right leg swing it is configured as tilt of rectangular rocking block in left and right direction. In this research we have configured the bipedal robot as rocking block before and after impact. The novelty of work is the configuration of bipedal walk as rocking block and development of hybrid automata [71].

## 5.2 Deliverable

In this research we have configured the bipedal walk as rocking block then we have designed the vector fields for all the six joints (Hip, Knee & Ankle) of bipedal walking model. The bipedal gait is the manifestation of temporal changes in the six joints angles, two each for hip, knee and ankle values and it is a combination of seven different discrete sub phases. Developing the correct joint trajectories for all the six joints was difficult from a purely mechanics based model due to its inherent complexities. To get the correct and exact joint trajectories is very essential for modern bipedal robot to walk stably. By designing vector field correctly we are able to get the stable joint trajectory ranges and able to reproduce angle ranges from theses designed vector fields. This is purely a data driven computational modeling approach which is based on the hypothesis that morphologically similar structure (Human-robot) can adopt similar gait patterns. To validate the correctness of the design we have applied few hybrid automata generated joint trajectories to HOAP-2 bipedal robot which could walk successfully maintaining its stability. The vector field provides joint trajectories for a particular joint. The results show that our data driven computational model is able to provide the correct joints angle ranges which are stable.

### 5.3 Hybrid Automata:

The human walk is the combination of different discrete sub phases and continuous dynamics [72]. A system with discrete switching logic and continuous dynamic is known as hybrid system. To represent the hybrid system we need language tool which we called hybrid automata [73].

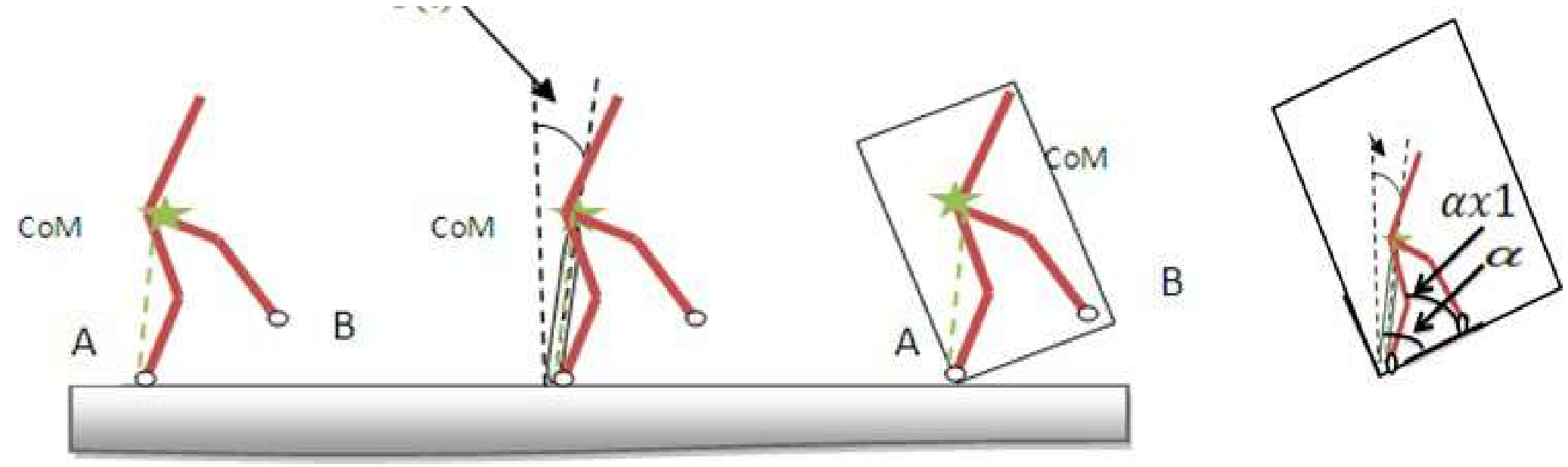

**Figure 5-1: Rocking Block/ Inverted Pendulum equivalence**

The bipedal walk is a complex and daunting task due to inherently unstable structure, high non linearity, varying dynamics and control steps during different sub phases of gait [3]. It is hybrid in nature due to discrete switching and continuous dynamics of walk i.e. under actuation during single support phase (swing phase) and over actuation during double support phase (stance). To design the correct and exact model of bipedal walk, it is required to include all the discrete non linear sub phases. Hybrid Automata is a language tool which is used for modeling and analyzing for many real time systems. There are a lot of real time systems where hybrid system widely used. It is interaction between the continuous dynamics and discrete switching logic [74]. Such system arises naturally in numerous engineering problems. The hybrid system paradigm has been implemented successfully for address engineering problem like air traffic control [75], automotive control [76], process control, highway systems [77] and bipedal robot [78]. These applications lead to the development of theoretical and computational tool for verification, simulation and modeling, analysis and controller synthesis for hybrid systems. Fundamental properties of hybrid systems, such as existence and uniqueness of solutions, continuity with respect to initial conditions, etc., naturally attracted the attention of researchers fairly early on. The majority of the work in this area concentrates on developing conditions for special classes of hybrid systems: variable structure systems, piecewise linear systems.

The trajectory of actual gait and hybrid automata based model generated trajectory are compared and errors are minimized [78]. It is observed that the hybrid automata based bipedal locomotion control strategy is universal and make robot move like the trajectories of human walking. The human motion joint trajectory cannot be directly used for robot due to difference between in physical structure, movable range, maximum moving speed, degrees of freedom (DOF). The method uses a finite state machine switching from one state to another according to pre-defined schedule [79] [80]. The schedule is based upon an ad-hoc set of intervals. The joints trajectories are defined by polynomials that changes according to the state of the finite state-machine. It is indeed a hybrid automaton. The physical structure of robot cannot keep its dynamic balance if the robot can track the appropriate trajectories. The constrained knee joint motion is very complex because of the interaction between ground and legs. To adopt the captured motion, the robot is required to satisfy a number of constraints simultaneously. The universal hybrid automata model has been configured as rocking block according to both leg and the data is collected for individual subject and the model is adjusted according to the individual subject mass, weight and physical conditions [81]. The computational model is data driven and adjusted with similar morphological structure of human. It has advantage of avoiding the problem of kinematics solution.

### 5.4 Rocking Block

In this research we have configured the human walk as rocking block. The rocking block is the configuration where the solid structure moves to and fro around the pivot stably. During double support phase we are in condition when both limbs on ground. The next phase is the swing phase in which the one foot will take step ahead and will become support leg for another leg so this type of shifting can be configured as rocking block. It is important to configure the biped before reaching DSS. The duration of DSS is 20% of entire gait [82]. Figure5.2 is the depiction of CoM as rocking block and Figure 5.3 is depiction of human walk as rocking block during left swing leg, DSP and right swing phase.

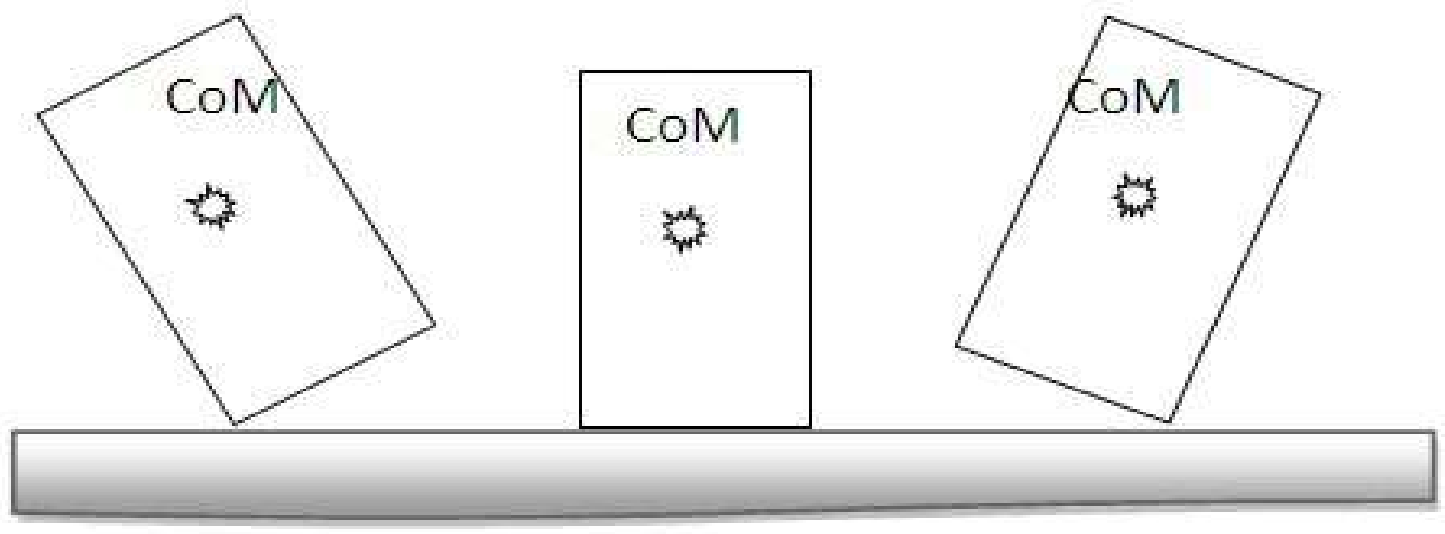

**Figure 5-2: Three impact evolution**

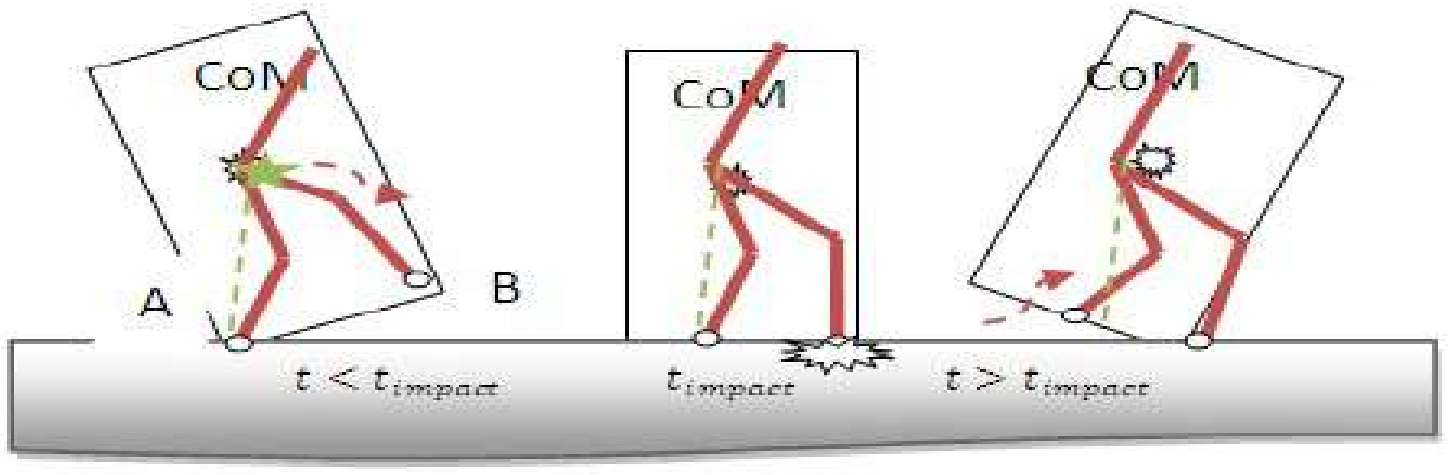

**Figure 5-3: Transfer of support foot**

Rocking block is the rectangular block with certain height and weight. The Center of Mass (CoM) is the point on the whole body mass lies.

## 5.5 Development of Hybrid automata Model of Human Gait:

Before developing the formal description of our biped model, it is necessary to provide the details breakdown of human gait into different sub phases and various parameter associate with it. Figure 5.4 is the depiction of our model as rocking block.

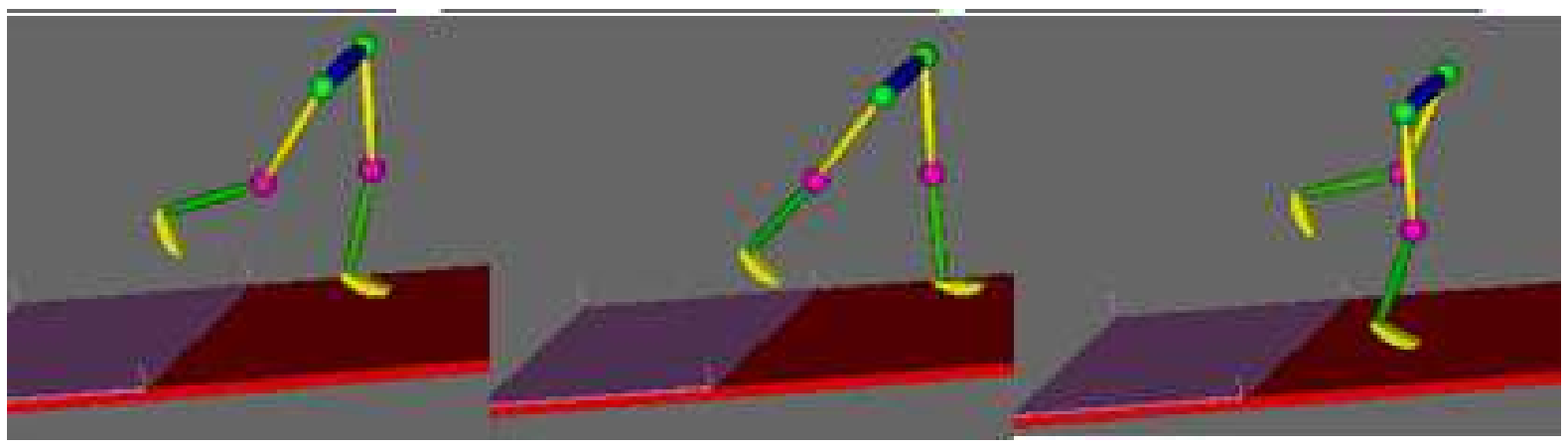

**Figure 5-4: Transfer of support foot**

## 5.6 Proposed Methodology

1- Studying the OpenSim biped models.

2- Dividing gait data into different phases.

3- Generating the hybrid automata vector field equations for each phase and joints.

4- Applying vector field equations to get human like motion using gait data.

5- Verifying the hybrid automata model generated.

The methodology used here is, starting with human data and then looking for various behaviors incorporated in human walking, by representing data in a general yet a simple form to make it yield a function. Using this anthropomorphic representation, we aim to design a nonlinear controller for robots. Along the lines of a biologically-inspired control [19], we constitute a canonical walking function incorporating all other output data which defines a solution to system.

### 5.6.1 Studying the OpenSim biped models

We have studied the both leg and gait2354 models of OpenSim model for this work. The models show the walking pattern on the simulated human leg for these two different gaits. The gait data of both the models has been studied and used in this work. The gait data for the both leg model and gait 2354 model in OpenSim are different and each model's single gait cycle has different no. of data sets. Both the leg model has 51 time series data in a cycle and the gait 2354 model has 73 time series data for a single cycle. This data works properly on their respective models when the .mot file for that corresponding model is loaded in OpenSim for their respective model. The file contains joint angle values for the different joints like left hip, left knee, right hip and right knee. Also for each joint the velocities are given in separate columns. So the model moves with the defined velocity at different parts of time with different joint angle values. The curve for each joint angle at any point of time can be plotted in the OpenSim toolbox and also the speed of the motion of the model can be easily adjusted to notice the minute changes happening while the model is running on these mot files. Also we can change the mass and lengths of the model's limbs and muscles.

### 5.6.2 Generating the hybrid automata vector field equations for each phase

The divided data for different phases of the gait is then used for generating equations in MATLAB. This is our vector field in the tuple of hybrid automata model. The equations are generated using curve fitting tool box in MATLAB. The data is plotted for a certain range of time values and the curve is fitted for that. After curve fitting we get the polynomial equation of different orders which depends on the accuracy of fitting of the curve. The equation can be of sinusoidal form as well, but since our data is divided in such a way that the nonlinear curve of the complete gait cycle is divided into parts which are nonlinear but not sinusoidal in nature. The general curve for hip joint for a single gait cycle is sinusoidal in nature and for knee it is sinusoidal with double humps. So if we take a generic equation for a single gait cycle as a whole for a joint angle then the equation will be sinusoidal but here with divided phases of gait cycle it is appropriate to generate the polynomial equations of degrees varying from second order to fifth order. This is our vector field for our hybrid automata model.

### 5.6.3 Developing Computational model based on Hybrid Automata

The bipedal model is developed in WEBTOS frame work first to test the hybrid automata equations generated. So further generate the joint angle values for the left hip, left knee, right hip and right knee joints. The input provided to the vector fields is the time series values at a certain fixed interval of some 20 msec. This data is applied on each joint of the bipedal model developed in WEBTOS. The OpenSim model is created using MATLAB software. To generate the model the mass and length values of right and left thigh and shank is provided as input. Then the generated model is tested for the vector fields. This generic model is our hybrid automata model which has been used for the verification purpose. The hybrid automata vector fields are then used for generating joint angle values for all the seven phases and four joints that are left and right knee and hip joints for bipedal walking model. These values are then clubbed together to form the complete gait cycle and are saved in .mot file which is to be applied on this computational model.

### 5.6.4 Applying vector fields to get Human like motion using gait data

The vector fields defined in the previous steps for this hybrid automata model is of ultimate importance for the purpose of generating joint angles for different phases. The Vector is a quantity which has magnitude as well as direction. The vector field is simply a function which is used to assign a vector to a point in the plane i.e. a point in space. It is mainly useful for representing various types of force fields and velocity fields. For generating vector plot, we have used MATHEMATICA tool. The vector is able to generate the stable joint angles trajectories. These vector fields further used to generate the data for each joint angle i.e. Right and left hip and knee and then this data is applied on our hybrid automata model in OpenSim. This data is saved in .mot file format and then it is loaded on OpenSim on the corresponding model.

### 5.6.5 Verifying the hybrid automata model generated for Bipedal walk

The hybrid automata model used for verification is generated using MATLAB and the thigh and shank mass and length are the inputs for generating this model. It is an OpenSim stick model which has curved foot unlike the flat foot used usually by the researchers. This computational model is now verified using the data generated from the hybrid automata vector fields and the curve for the gait pattern of this computational model is generated and compared with the actual curve for the gait pattern of the original model. Also the gait data of another different model is applied on our computational hybrid automata model to verify the accuracy of our model. This model has curved foot unlike other models which use flat foot. This curved foot provides more robustness and prevents fall while walking on an irregular terrain.

## 5.7 Hybrid Automata parameters and description [73]

A hybrid automaton is used to describe the dynamic system that evolves with time and have both discrete and continuous phases. The hybrid automata parameters are explained as below:

$\mathrm{H} = (\mathrm{Q}, \mathrm{X}, \mathrm{f}, \mathrm{Init}, \mathrm{D}, \mathrm{E}, \mathrm{G}, \mathrm{R})$, where

Q Finite set of statei. e. discrete variable

X Finite set of continuous variables

$\mathrm{f}: \mathrm{Q} \times \mathrm{X} \rightarrow \mathrm{T}\,\mathrm{X}$ vector field

$\mathrm{Init} \subseteq \mathrm{Q} \times \mathrm{X}$ se tof initial states

$\mathrm{D}: \mathrm{Q} \rightarrow \mathrm{P}(\mathrm{X})$a domain

$\mathrm{E} \subseteq \mathrm{Q} \times \mathrm{Q}$ set of edges;

$\mathrm{G}: \mathrm{E} \rightarrow \mathrm{P}(\mathrm{X})$ guard condition

$\mathrm{R}: \mathrm{E} \times \mathrm{X} \rightarrow \mathrm{P}(\mathrm{X})$reset map

**An hybrid automata model consists of**:

Q $\mathbf{x1},\ \mathbf{x2},\ \mathbf{x3},\ \mathbf{x4},\ \mathbf{x5}, \mathbf{x6}, \mathbf{x7}$

X $\boldsymbol{\theta} = \ \boldsymbol{\theta}\mathbf{h}\ ,\quad \boldsymbol{\theta}\mathbf{k}\ ,\quad \boldsymbol{\theta}\mathbf{a}\ , \mathbf{x}$

$f: Q \times X \rightarrow TX$ $\mathbf{X\theta}$

$Init \subseteq Q \times X$ $\mathbf{0}, \mathbf{t1}, \mathbf{t2}, \mathbf{t3}, \mathbf{t4}, \mathbf{t5}, \mathbf{t6},$

$D: Q \rightarrow P(X)$ $-23.7532 < D\boldsymbol{\theta}\mathbf{h} < 20.06656, -68.3523 < D\boldsymbol{\theta}\mathbf{k} < 3.2090,$
0<**x**<1.6

$E \subseteq Q \times Q$ $\mathbf{LR} \rightarrow \mathbf{MST}, \mathbf{MST} \rightarrow \mathbf{TS}, \mathbf{TS} \rightarrow \mathbf{PS}, \mathbf{PS} \rightarrow \mathbf{IS}, \mathbf{IS} \rightarrow \mathbf{MSW}, \mathbf{MSW} \rightarrow \mathbf{TSW}, \mathbf{TSW} \rightarrow \mathbf{LR}$

$G: E \rightarrow P(X)$ $\mathbf{LR} \rightarrow \mathbf{MST} \text{ if } \mathbf{x} > 0.5$

$\mathbf{MST} \rightarrow \mathbf{TS} \text{ if } \mathbf{x} > 0.\mathbf{733}$

$\mathbf{TS} \rightarrow \mathbf{PS} \text{ if } \mathbf{x} > 0.\mathbf{9833}$

$\mathbf{PS} \rightarrow \mathbf{IS} \text{ if } \mathbf{x} > \mathbf{1.1167}$

$\mathbf{IS} \rightarrow \mathbf{MSW} \text{ if } \mathbf{x} > \mathbf{1.2667}$

$\mathbf{MSW} \rightarrow \mathbf{TSW} \text{ if } \mathbf{x} > \mathbf{1.4333}$

$\mathbf{TSW} \rightarrow \mathbf{LR} \text{ if } \mathbf{x} > \mathbf{1.600}$

$R: E \times X \rightarrow P(X)$ $\mathbf{LR} \rightarrow \mathbf{MSTx} = \mathbf{x1}$

$\mathbf{MST} \rightarrow \mathbf{TSx} = \mathbf{x2}\ , \mathbf{TS} \rightarrow \mathbf{PS}\quad \mathbf{x} = \mathbf{x3}$

$\mathbf{PS} \rightarrow \mathbf{ISx} = \mathbf{x4}, \mathbf{IS} \rightarrow \mathbf{MSW}\quad \mathbf{x} = \mathbf{x5}$

$\mathbf{MSW} \rightarrow \mathbf{TSW}\quad \mathbf{x} = \mathbf{x6}\ , \mathbf{TSW} \rightarrow \mathbf{LR}\quad \mathbf{x} = \mathbf{x7}$

**Different variables used are:**

$\theta$ = joint angle and a function of time **x**

$x$ = time value to be given as input

**£** = small value of time to be incremented in $x$ in each phase

**Init** = initial condition for each phase,

**t = 0,** $\mathbf{t_1, t_2, t_3, t_4, t_5, t_6}$, **x=** $\mathbf{x_1,\ x_2,\ x_3,\ x_4,\ x_5,\ x_6,\ x_7}$

**f($x$)** = function of **x** giving output value as joint angle values for each time in each phase

**Different notations for phases used are:**

**LR ->**Loading Response

**MST ->**Mid Stance

**TS ->**Terminal stance

**PS ->**Pre Swing

**IS ->**Initial Swing

**MSW ->**Mid Swing

**TSW ->**Terminal Swing

**Hybrid Automata and Execution:** A hybrid automaton is used to describe the dynamic system that evolve with time and have both discrete and continuous phases.

$H = \{Q, X, f, Init, D, E, G, R\}$

$Q = \{q_i\}$: $\mathbf{Q} =$ Set of States, in our case the cardinality of the set is seven
$Q = \{IC, LR, MSt, Tst, PSw, ISw, Msw, Tsw\}$
A Hybrid automaton H is a collection H = (Q, X, f, Init, D, E, G, R), we have X={x1, x2} where x1 represents the angle the left leg makes with the vertical (as a fraction of $\alpha$) and x2 represents the block's angular velocity. Refer the figure 5.1.

$$f(left, x) = \begin{pmatrix} x2 \\ \frac{\sin(\alpha(1+x1))}{\alpha} \end{pmatrix} \quad (5)$$

$$f(right, x) = \begin{pmatrix} x2 \\ \frac{\sin(\alpha(1-x1))}{\alpha} \end{pmatrix} \quad (6)$$

$D(left) = \{x \in R^2$:. : x1 <= 0}, D (right) = {x ∈ $R^2$ :x1 >= 0}
Where D(left) and D(right) are the domain of left and right leg.
Init = {Left} * {x ∈ R^2 : [-1 <= x1 <= 0] ∩ [cos($\alpha$(1+x1)) + (($\alpha$x2)^2)/2 <= 1]
U{Right} * {x ∈ R^2: [0 <= x1 <= 1] ∩ [cos($\alpha$(1-x1)) + (($\alpha$x2)^2)/2 <= 1}
G(left, right) = {x ∈ R^2 :(x1 = 0) ∩ (x2 >= 0)}
G(right, left) = {x $\epsilon$ R^2 : (x1 = 0) ∩ (x2 <= 0)}
R(left, right, x) = R(right, left, x) = $\{[\begin{matrix} x1 \\ rx2 \end{matrix}]\}$
The algorithm 2 is for data generation using hybrid automata equations.

---

**Algorithm 2**: Gait Data generation using Hybrid Automata Equations:

---

$\boldsymbol{Input}$: $TimeConstant(t_c), PolynomialCoefficient(p_n)$
$wherenisorder, Error(er)$, Time (t), Time_LR ($t_{lr}$),
Time_MST ($t_{mst}$), Time_TS ($t_{ts}$), Time_PS ($t_{ps}$), Time_IS ($t_{is}$), Time_MSW ($t_{msw}$), Time_TSW ($t_{tsw}$)
$\boldsymbol{Output}$: $\theta_{LH}[i], \theta_{RH}[i], \theta_{LK}[i], \theta_{RK}[i], \theta_{LA}[i], \theta_{RA}[i]$
$\boldsymbol{Initial}$: $t_c \leftarrow 0.0167$; $l_s \leftarrow 4, p_i where i = 1,2,3,..,n$ ; Per= [10, 30, 50, 60, 73, 87,100]/100;
//Time wise division of one gait cycle into different sub phases
$t_{lr}$=t * per [1]; $t_{mst}$=t*per [2]; $t_{ts}$=t*per [3]; $t_{PS}$=t*per [4]; $t_{is}$=t*per [5]; $t_{msw}$=t*per [6]; $t_{tsw}$=t*per [7];
$t_{sp}$ = [$t_{0,}$lr, mst, ts, ps, is, msw, tsw];j=0;k=0;
$\theta_{K=}[\theta_{LH,}\theta_{RH}, \theta_{LK}, \theta_{RK}, \theta_{LA}, \theta_{RA}]$
$\boldsymbol{Begin}$
// *for loading response*
$\boldsymbol{forx \leftarrow t_{sp}[j]: t_c: t_{sp}[j+1]}$
$tmp[i] = p_i \times x^i + er,$
$where i = 1,2 \ldots n, n - degree of polynomial$
$j = j + 1;$
$\boldsymbol{Endfor}$
//Reverse Engineering
$\boldsymbol{fori \leftarrow 1: n}$
$\theta_K[k] = tmp[i]$
$\boldsymbol{Endfor}$
$\boldsymbol{EndBegin}$

---

## 5.8 Hybrid Automata Finite State machine diagram

A finite state machine or just a state machine diagram is mathematical model for computation that is used to design computer programs or a computation model. It can be visualized as the abstract machine which can be existing in one of the finite no. of states [29]. The machine can exist in only one state at a particular given point of time. The computational model consists of sets of finite states, a start state, the inputs and transition function. This functions maps the current states and input symbols to the next states. The computation for the model begins with the input value in the very start state. Then according to the transition function it turns to the next state. This might sound a bit complicated but actually in reality it is quite simple. Finite state machines are largely used in computer program designing, but it finds an extensive use in different other fields also, like biology, engineering, linguistics, and other sciences which are able to recognize sequencing. Finite state machines are also expressed visually as state transition diagram of finite state machine diagrams. This diagram is used for showing all the possible states, the inputs to that and their output. Each and every state is represented in a separate block and the transitions are represented using arrows. Transition conditions have to be mentioned for each transition happening between the states [30]. Also in some cases the finite state machine can yield no outputs as well.  It is simply a mathematical abstraction used for designing algorithms for a complete or partial system.

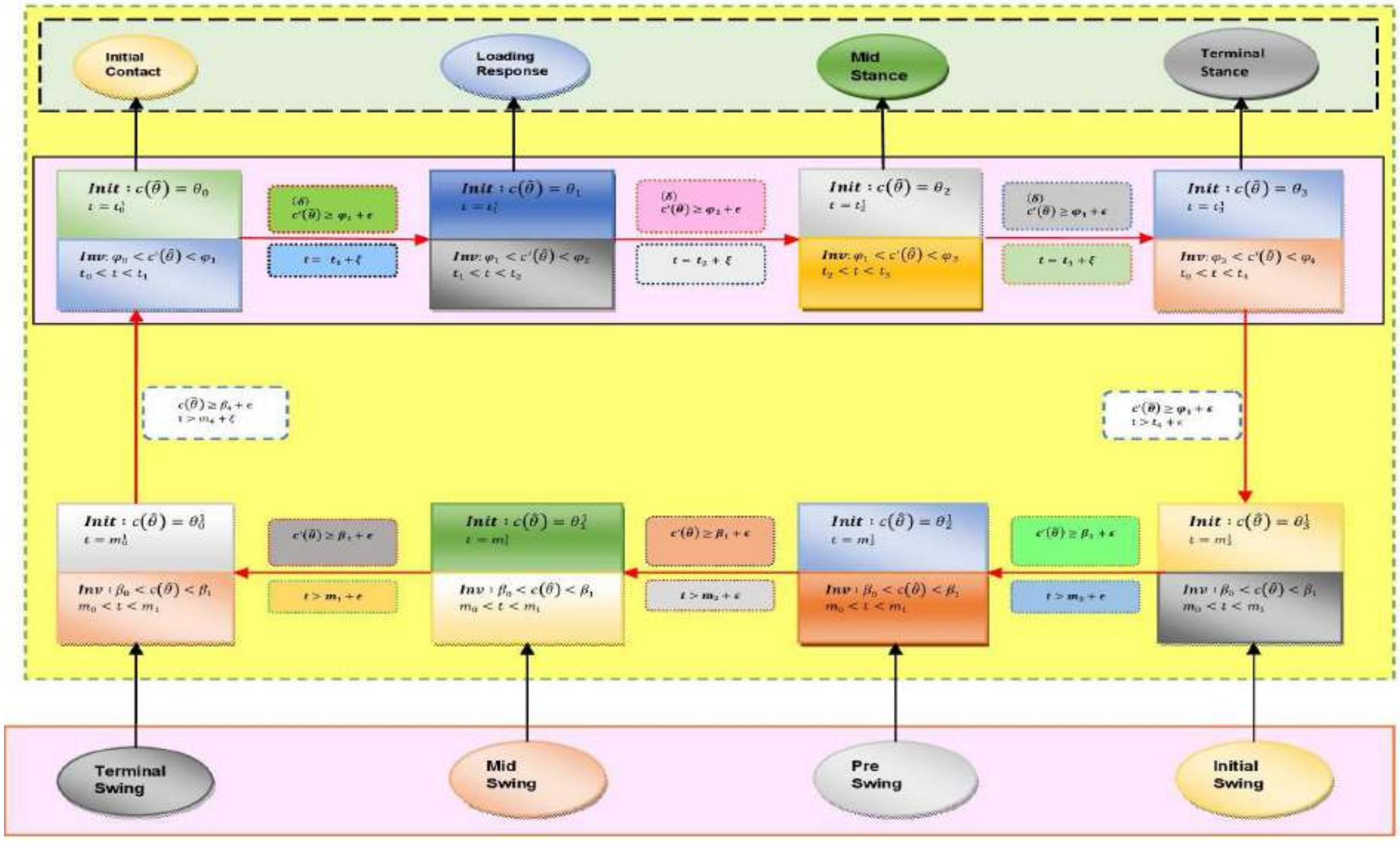

**Figure 5-5: Details of the Automata Implementation**

Figure 5.5 is the finite state diagram for our Hybrid Automata model with different variables and initial conditions.

## 5.9 Details Methodology:

### 5.9.1 Developing Vector Field $\mathbf{f}: \mathbf{Q} \times \mathbf{X}$

Developing vector field function **(VF)** with proper G: and R: is one of the critical part of our research. The vector field has been developed using captured gait data from human and through multiple regression analysis. We have used 60% of all captured data for evaluating f: and 40% data for validation. Table 5.1 shows the vector field parameter f: for left hip joint during different sub phases of gait. Here Error (the difference between captured data and regression model data) is adjusted constantly to converge it towards zero. The Pi represents the coefficients associated with degrees of polynomial. Here seven degrees of polynomial has been used with continuous manifold.Later we have configured the hybrid automata dynamic walker model for individual subject and compared the joint trajectories (hips and knees) with same individual's model developed on OpenSim walking model named as gait2354. The hybrid automata vector field functions for all the joints and for all the seven phases are as follows:

**Table 5:1: Vector field f: for all the seven phases for Left hip joint**

f(x) = $p_1x^4 + p_2x^3 + p_3x^2 + p_4x + p_5 + error$ {LR,MST,TS}
f(x) = $p_1x^3 + p_2x^2 + p_3x + p_4 + error$ {PS,IS,MSW,TSW}

| | LR_LH | MST | TS | PS | IS | MSW | TSW |
|---|---|---|---|---|---|---|---|
| *error | +2.8 | 0 | +1 | -.2 | +.05 | -.4 | -.3 |
| $p_1$ | -4.061e+04 | 1.14e+04 | -8069 | 1148 | -27.79 | 986.1 | -722.2 |
| $p_2$ | 7.201e+04 | -2.723e+04 | 2.867e+04 | -3915 | 407.5 | -4430 | 3374 |
| $p_3$ | -4.774e+04 | 2.461e+04 | -3.842e+04 | 4431 | -865 | 6529 | -5351 |
| $p_4$ | 1.393e+04 | -1.001e+04 | 2.315e+04 | -1645 | 516.4 | -3148 | 2878 |
| $p_5$ | -1513 | 1528 | -5312 | 0 | 0 | 0 | 0 |

NOTE: * Error indicates the difference between the actual human data and data generated through hybrid automata. It is must to include this error in hybrid automata. Error is adjusted constant value to compensate the error. The table 5.2 gives the Hybrid automata vector field function for Right Hip joint for each sub phase of the human gait cycle.

**Table 5:2: Vector field f: for all the seven phases for Right Hip joint**

| f(x) = p1*x^3 + p2*x^2 + p3*x + p4 +Error { LR,MST,TS ,PS,IS,MSW,TSW} | | | | | | | |
|---|---|---|---|---|---|---|---|
| | LR_RH | MST | TS | PS | IS | MSW | TSW |
| *error | 0 | -.4 | -.4 | +.5 | -.8 | +.2 | -2.5 |
| $p_1$ | 642.3 | -2732 | 1475 | -1203 | 1437 | -2162 | -1382 |
| $p_2$ | -1288 | 5192 | -3917 | 3914 | -4788 | 9538 | 6100 |
| $p_3$ | 760.2 | -3247 | 3356 | -4328 | 5219 | -1.382e+04 | -8827 |
| $p_4$ | -119.1 | 687.6 | -915.4 | 1614 | -1870 | 6575 | 4188 |

The table 5.3 gives the Hybrid automata vector field function for left knee joint for each sub phase of the human gait cycle.

**Table 5:3: Vector field f: for all the seven phases for Left knee joint**

| $f(x) = p_1x^4 + p_2x^3 + p_3x^2 + p_4x + p_5 + error$ {MST,TS,PS }<br>f(x) = p1*x^3 + p2*x^2 + p3*x + p4 +Error { LR,IS,MSW,TSW } | | | | | | | |
|---|---|---|---|---|---|---|---|
| | LR_LH | MST | TS | PS | IS | MSW | TSW |
| *error | 0 | -.3 | 0 | -1 | -6.1 | -1 | +0.2 |
| $p_1$ | 2458 | -2.173e+04 | -1139 | 8138 | 2435 | 1757 | -555.6 |
| $p_2$ | -3383 | 5.163e+04 | 3383 | 2.899e+04 | 1.09e+04 | -6405 | 2571 |
| $p_3$ | 1598 | -4.648e+04 | -2122 | -3.678e+04 | 1.571e+04 | 7591 | -3892 |
| $p_4$ | 258.8 | 1.874e+04 | -1106 | 1.938e+04 | -7356 | -2911 | 1915 |
| $p_5$ | 0 | -2842 | 926.9 | -3508 | 0 | 0 | 0 |

The table 5.4 gives the Hybrid automata vector field function for right knee joint for each sub phase of the human gait cycle.

**Table 5:4: Vector field f: for all the seven phases for Right Knee joint**

f(x) = $p_1x^4 + p_2x^3 + p_3x^2 + p_4x + p_5 + error$ {LR,MST,TS}
f(x) = $p_1x^3 + p_2x^2 + p_3x + p_4 + error$ {IS,MSW}
f(x) = $p_1x^2 + p_2x + p_3 + error$ {PS,TSW}

| | LR_RK | MST | TS | PS | IS | MSW | TSW |
|---|---|---|---|---|---|---|---|
| *error | +1.7 | -1 | +1.8 | +.07 | +1 | +0.5 | -1 |
| $p_1$ | -5.103e+04 | 1.835e+04 | 3539 | -66.25 | -1798 | 9973 | 1402 |
| $p_2$ | 8.963e+04 | -3.649e+04 | -1.442e+04 | 188.1 | 5842 | 4.122e+04 | -4290 |
| $p_3$ | -5.816e+04 | 2.387e+04 | 2.174e+04 | -132.7 | -6279 | 5.642e+04 | 3213 |
| $p_4$ | 1.685e+04 | -5134 | -1.433e+04 | 0 | 2224 | -2.561e+04 | 0 |
| $p_5$ | -1921 | -6.756 | 3459 | 0 | 0 | 0 | 0 |

The table 5.5 gives the Hybrid automata vector field function for left ankle joint for each sub phase of the human gait cycle.

**Table 5:5: Vector field f: for all the seven phases for Left Ankle joint**

f(x) =$p_1x^3 + p_2x^2 + p_3x + p_4 + error$ {LR,TS, PS,IS,MSW,TSW }
f(x) = $p_1x^4 + p_2x^3 + p_3x^2 + p_4x + p_5 + error${MST}

| | LR_LA | MST | TS | PS | IS | MSW | TSW |
|---|---|---|---|---|---|---|---|
| *error | 0 | -0.25 | +3.3 | +3.9 | +2.1 | -1.8 | +0.6 |
| $p_1$ | -1210 | -706.8 | -7514 | 4112 | -5056 | -2875 | 590.5 |
| $p_2$ | 1677 | -337 | 2.025e+04 | -1.342e+04 | 1.801e+04 | 1.211e+04 | -2735 |
| $p_3$ | -753.4 | 2017 | -1.81e+04 | 1.454e+04 | -2.136e+04 | -1.695e+04 | 4242 |
| $p_4$ | 113.7 | -1404 | 5358 | -5235 | 8430 | 7882 | -2202 |
| $p_5$ | 0 | 289.5 | 0 | 0 | 0 | 0 | 0 |

The table 5.6 gives the Hybrid automata vector field function for right ankle joint for each sub phase of the human gait cycle.

**Table 5:6: Vector field f: for all the seven phases for Right Ankle joint**

f(x) $= p_1x^2 + p_2x + p_3 + error$ {IS}
f(x) $= p_1x^3 + p_2x^2 + p_3x + p_4 + error$ {LR,TS,PS,MSW,TSW }
f(x) $= p_1x^4 + p_2x^3 + p_3x^2 + p_4x + p_5 + error$ {MST }

| | LR_RH | MST | TS | PS | IS | MSW | TSW |
|---|---|---|---|---|---|---|---|
| *error | 0 | +2.5 | -0.1 | 0 | 0 | +1 | 0 |
| $p_1$ | 1325 | 2.99e+04 | 538 | 228.9 | -43.13 | 8742 | 160.4 |
| $p_2$ | -2024 | -7.249e+04 | -1543 | -699.1 | 129 | -3.652e+04 | -1316 |
| $p_3$ | 993 | 6.539e+04 | 1491 | 729.4 | -80.58 | 5.062e+04 | 2976 |
| $p_4$ | -155.6 | -2.602e+04 | -478.5 | -251.8 | 0 | -2.328e+04 | -2048 |
| $p_5$ | 0 | 3853 | 0 | 0 | 0 | 0 | 0 |

## 5.10 Hybrid Automata Model comparison and verification

The OpenSim Dynamic walker Hybrid Automata model designed in MATLAB is tested by using different scenarios of walking using different model's gait data. The model is first tested using the data generated from our hybrid automata equations. We have named this model as Hybrid automata model. After applying the data through the .mot file in OpenSim for this model, it was observed that the model showed the stable walking pattern for our equations generated data, this clearly verify the correctness of our equations. Then for the verification of the model, this model is loaded with the gait data of original OpenSim's gait2354 model's data. It was observed that the model was able to walk with this data as well in the similar manner. This was also confirmed by comparing the phase plot for hip joint and knee joint curves for both the cases. The verification was done by using the gait data of human both leg model on our hybrid automata model in OpenSim. With this data also we were able to obtain the proper walk on this model and the curves for hip and knee joint were confirming the model's validity. The position of our stick model at different phases has been shown in this figure, and each position is labeled. Figure 5.6 shows the different phase of gait depicted by our model based on the pattern generated through hybrid automata model for normal walk.

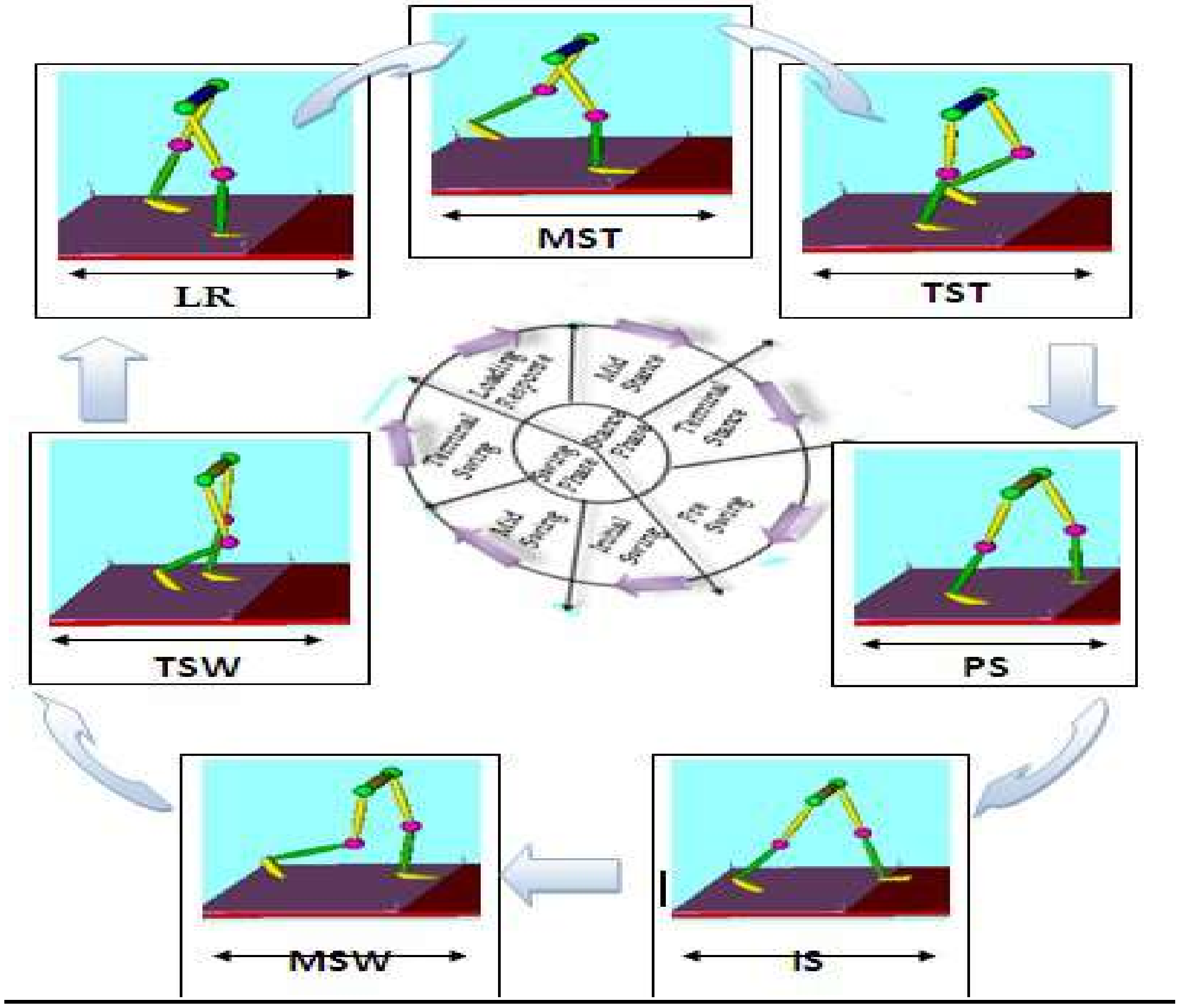

**Figure 5-6: Phase wise output of the hybrid automata model in OpenSim showing different positions for all the seven phases.**

## 5.11 State Prediction of Human Gait using Cellular Automata (Deliverable):

In this research work we have explored the cellular automata [83] [84] for prediction of bipedal gait states. A cellular automaton is a theoretical tool which uses to predict the next state based on certain set of rule and neighbor condition. We have even incorporated the behavior, interaction, priority (BIP) [85] model to understand the complexity of bipedal walk. Due to non-linearity in bipedal walk it is very complex to understand. We have designed the model which will predict the next gait state of bipedal model based on the previous two neighbor states. We have designed for normal walk model. The state prediction will help to correctly design the bipedal walk. The normal depend on next two states and have total 8 states. We have considered the current and previous state to predict next state. So we have formulated 16 rules using cellular automata, 8 rule for each leg. The priority order maintained using the BIP as if right leg in swing phase then left leg will be in stance phase. To validate the model we have applied these on HOAP2 model. We have explored the trajectories and compares with another gait trajectories. We have generalized our cellular automata based prediction model as universal. The model will be able to predict the next state based on current state and previous model. Total 8! Permutation is possible. The model is able to predict the state on any terrain. In this case we have considered only states in terms of joints angle value we would not referred the terrain.

### 5.12 Developing Cellular Automata (CA)

CA is discrete dynamic systems. CA's are said to be discrete because they operate in finite space and time and with properties that can have only a finite number of states. CA's are said to be dynamic because they exhibit dynamic behaviors. Equation 1 is the representation of state prediction. Where S(t) represents the current state. S is the set of all possible discrete states of our gait model for us it is 8.

$\text{S: Finite set of state i. e. discreter variable}$

$$\mathrm{S} = \{\mathrm{IC}, \mathrm{LR}, \mathrm{MS}, \mathrm{TST}, \mathrm{PSW}, \mathrm{ISW}, \mathrm{MSW}, \mathrm{TSW}\} \quad \text{- (7)}$$

$$\mathrm{S}(t+1) = \{\mathrm{S}(t), \mathrm{S}(t-1)\} \quad \text{- (8)}$$

Consider the state S={1,2,3,4,5,6,7,8} so we have assumed 8 discrete state here. In this work we have considered 8 discrete as 8 neighbors.

#### 5.12.1 Building automatic component

We have used a bottom-up approach to build the system model, first constructing each atomic component and modeling their behavior with Cellular automaton then combining these atomic components into composite components. The interactions between these atomic and composite components are captured through the well-defined semantics of algebra of connectors and primarily the causal chain type of interaction. The priorities set certain restrictions on the type of interactions and resolves possible deadlock scenarios. Each leg is decomposed into three atomic components (hip, knee and ankle). Thus, we have six atomic components in total, three for each leg and two composite components (each individual leg). The behavior of each atomic component is described by a six state (three stance states and three swing states) cellular automaton. In our model the left leg starts in the stance phase by default, so the right leg will invariably start from the swing phase. Each of the atomic components of the left leg passes through the three states of the stance phase (initial contact, mid stance, terminal stance) in sync and then a phase transition between the left leg and right leg occurs that is, the left leg goes into swing phase. After that each component of the left leg passes through the three states of the swing phase (initial swing, mid swing and terminal swing) after the swing phase is over the left leg comes back to the stance phase and the cycle is repeated [86][87]. The atomic components of the left leg are shown here ankle, knee and hip. We have assumed binary state of movement of atomic components of a leg (Ankle, Knee, Hip) is either in motion or in rest. So we consider binary stage 0 and 1 for each component. 0 represents atomic components are in rest and 1 represents atomic components are in motion. Since, there are three atomic components and each have two state either 0 or 1. So, there will be a total of eight states (Refer Table 5.8).During Locomotion human each leg passes through eight sub phases [88] [89].

### 5.12.2 CA Rules

Here we have written 16 CA rules to determine the state of atomic components of one leg with the help of second leg. All the states are represented using 4-bit stream. First bit represent the leg that if the fourth bit is zero it represents left leg whereas if the fourth bit is 1 it represents the right leg. Other three bits represents the sub phases of that leg. It will be among one of the eight states so there will be a total of 16 rules. 1000 can be seen as two parts 1+000(leg + Sub phase) which means right leg is in initial contact. The neighbor row represents the state or phase of another leg whereas the rule row depicts the state of atomic components of that leg. Set of rules to determine the state of locomotion. Cellular automata rule Rule-8, universal, generalizes Rule for left and right leg during normal walk. Following are the states relation between left and right leg.

Left_Leg_Stance->Right_Leg_Swing
Left_Leg_Swing->Right_Leg_Stance
Left_Leg_IC ->Right_Leg_PSw
Left_Leg_MS->Right_Leg_Msw
Left_Leg_TS->Right_Leg_TSw
Left_Leg_PSw->Right_Leg_LR
Left_Leg_ISw->Right_Leg_MS
Left_Leg_MSw->Right_Leg_TS
Left_Leg_TSw->Right_Leg_IC

**Table 5:7 : Binary State Representation of Bipedal Gait 8 states**

| Number | 7 | 6 | 5 | 4 | 3 | 2 | 1 | 0 |
|---|---|---|---|---|---|---|---|---|
| Neighborhood | 111 | 110 | 101 | 100 | 011 | 010 | 001 | 000 |
| Rule Result | TS | MS | IS | PS | TS | MS | LR | IC |

**Table 5:8: Cellular Automata state prediction for Left leg**

| Number | 7 | 6 | 5 | 4 | 3 | 2 | 1 | 0 |
|---|---|---|---|---|---|---|---|---|
| Neighbor | 0111 | 0110 | 0101 | 0100 | 0011 | 0010 | 0001 | 0000 |
| Rule Result | 1011 | 1010 | 1001 | 1000 | 1111 | 1110 | 1101 | 1100 |

**Table 5:9: Cellular Automata state prediction for Left leg**

| Number | 15 | 14 | 13 | 12 | 11 | 10 | 9 | 8 |
|---|---|---|---|---|---|---|---|---|
| Neighbor | 1011 | 1010 | 1001 | 1000 | 1111 | 1110 | 1101 | 1100 |
| Rule Result | 0111 | 0110 | 0101 | 0100 | 0011 | 0010 | 0001 | 0000 |

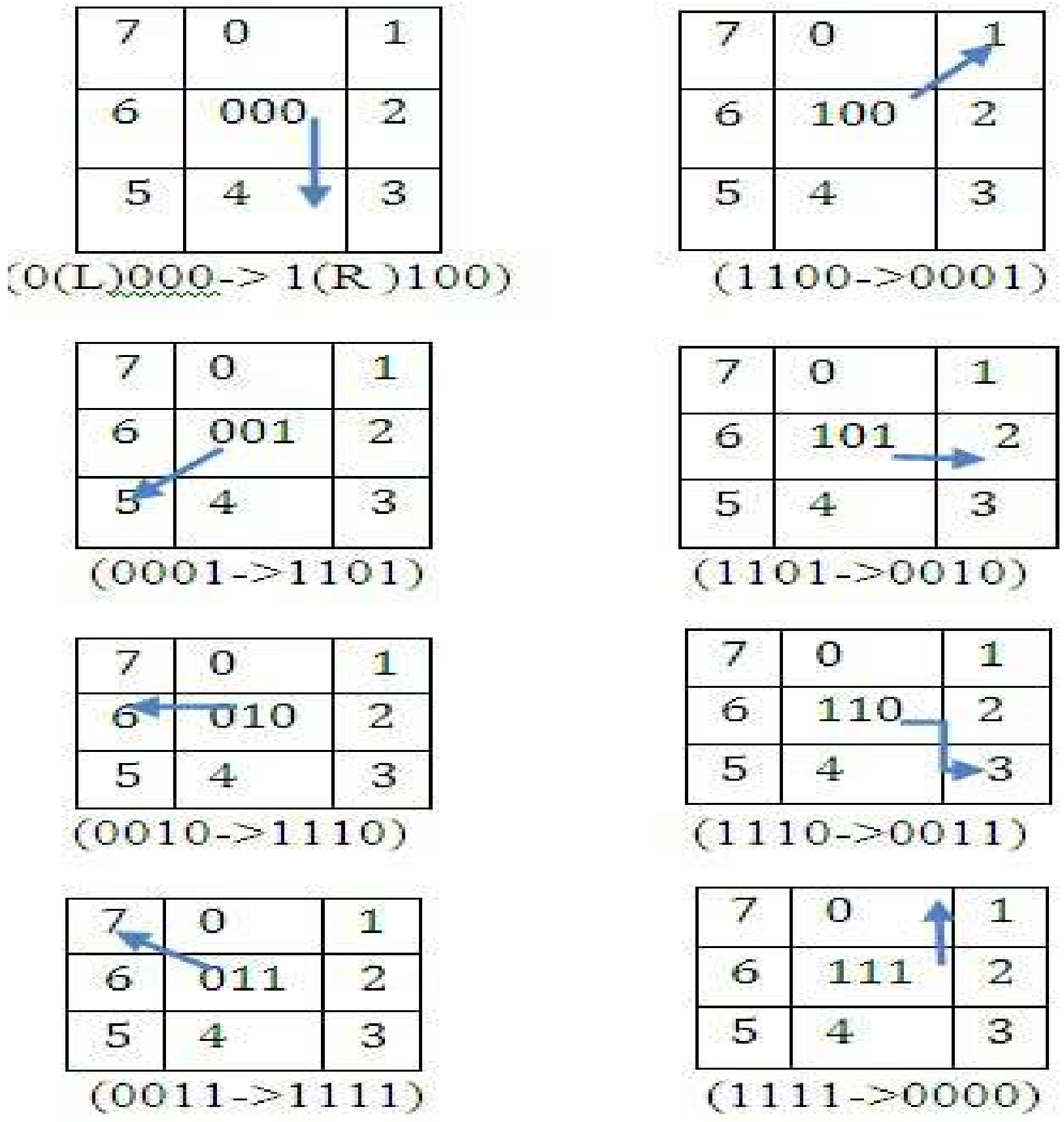

**Figure 5-7: Transaction of leg state using CA**

The unique approach to model the human gait presented here. It is able to model the normal human gait within a negotiable degree of error. Here we have written 16 CA rules to determine the state of atomic components of one leg with the help of second leg. All the states are represented using 4-bit stream. First bit represent the leg that if the fourth bit is zero it represents left leg whereas if the fourth bit is 1 it represents the right leg. Other three bits represents the sub phases of that leg. It will be among one of the eight states so there will be a total of 16 rules. 1000 can be seen as two parts 1+000(leg + Sub phase) which means right leg is in initial contact. The neighbor row represents the state or phase of another leg whereas the rule row depicts the state of atomic components of that leg.

## 5.13 Results and Discussions

### 5.13.1 Vector Field representation for all the six joints (Hip, Knee and ankle)

The Vector is a quantity which has magnitude as well as direction. A vector field is an assignment of a vector to each point in a subset of space. It is mainly useful for representing various types of force and velocity fields. For generating vector plot, we have used MATHEMATICA tool. Similarly we draw the vector plot for every left and right joint (hip, knee and ankle). The vector is able to generate the stable joint angles trajectories. The length of arrow indicates the magnitude of force and moving arrows shows direction of force. Number of arrow indicates that the strength of force. Take a moving point at any position in field, and then you can see the force acting on that moving point when it moves from one location to another. Figure 5.8 to 5.13 are the vector field representation of corresponding to left ankle, right ankle, left knee, right knee, left hip and right hip. This further can be used to generate the range of stable joints trajectories to make robot walk. The equations are the part of hybrid automata model referred as vector fields. The vector fields are then used for getting the joint angle as the output from the time series input data of accurately divided separate data for each phase. This data for each phase is then joined to get the data for a complete gait cycle for the corresponding joint angles. It is shown that this model works for the data generated from the developed hybrid automata vector fields and also it works for the human both leg model's data, which verifies that the model is a generic one. The model generated joint trajectory data enabled a biped robot walking in a stable manner.

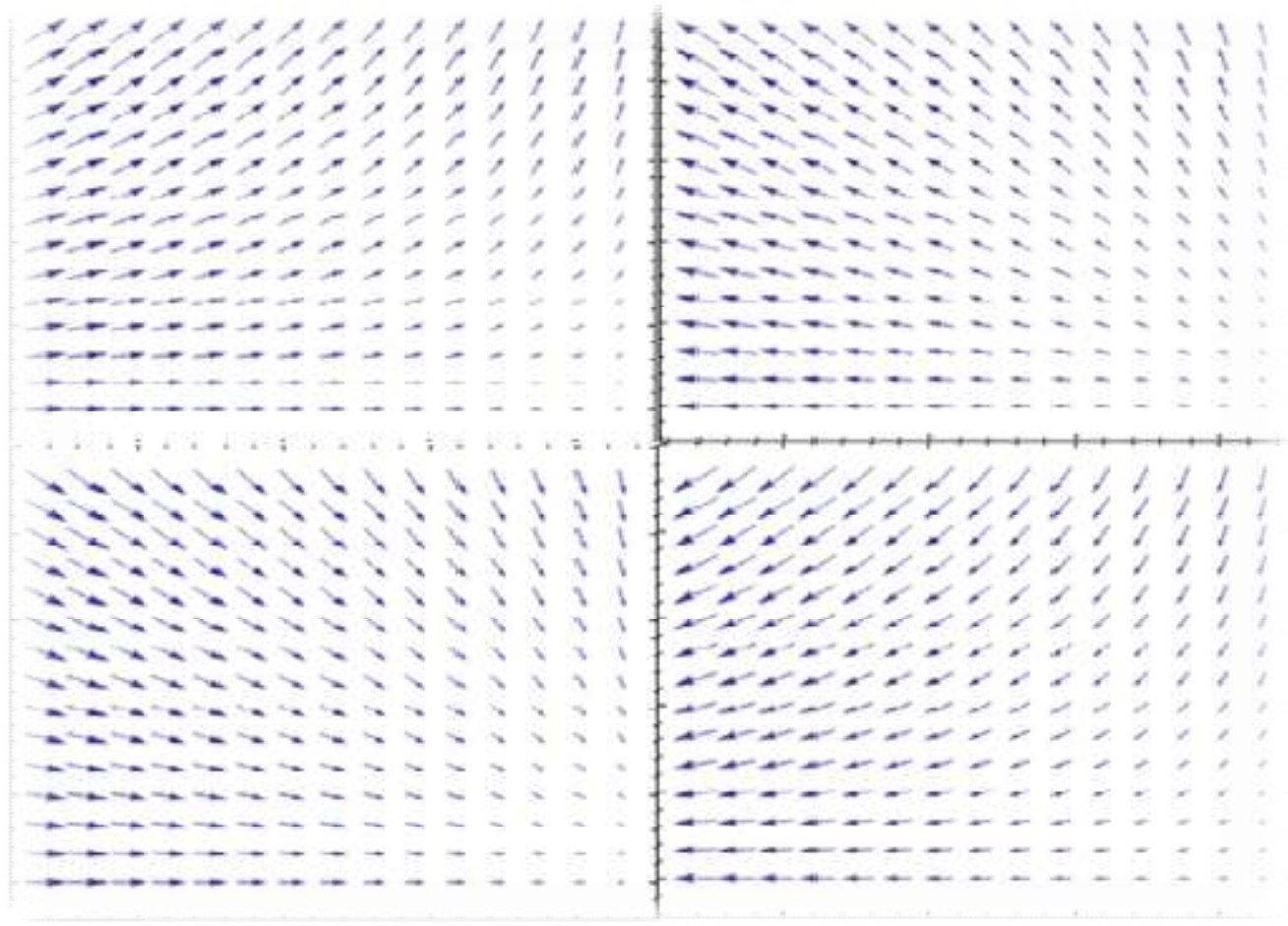

**Figure 5-8: Vector Plot for Left Ankle**

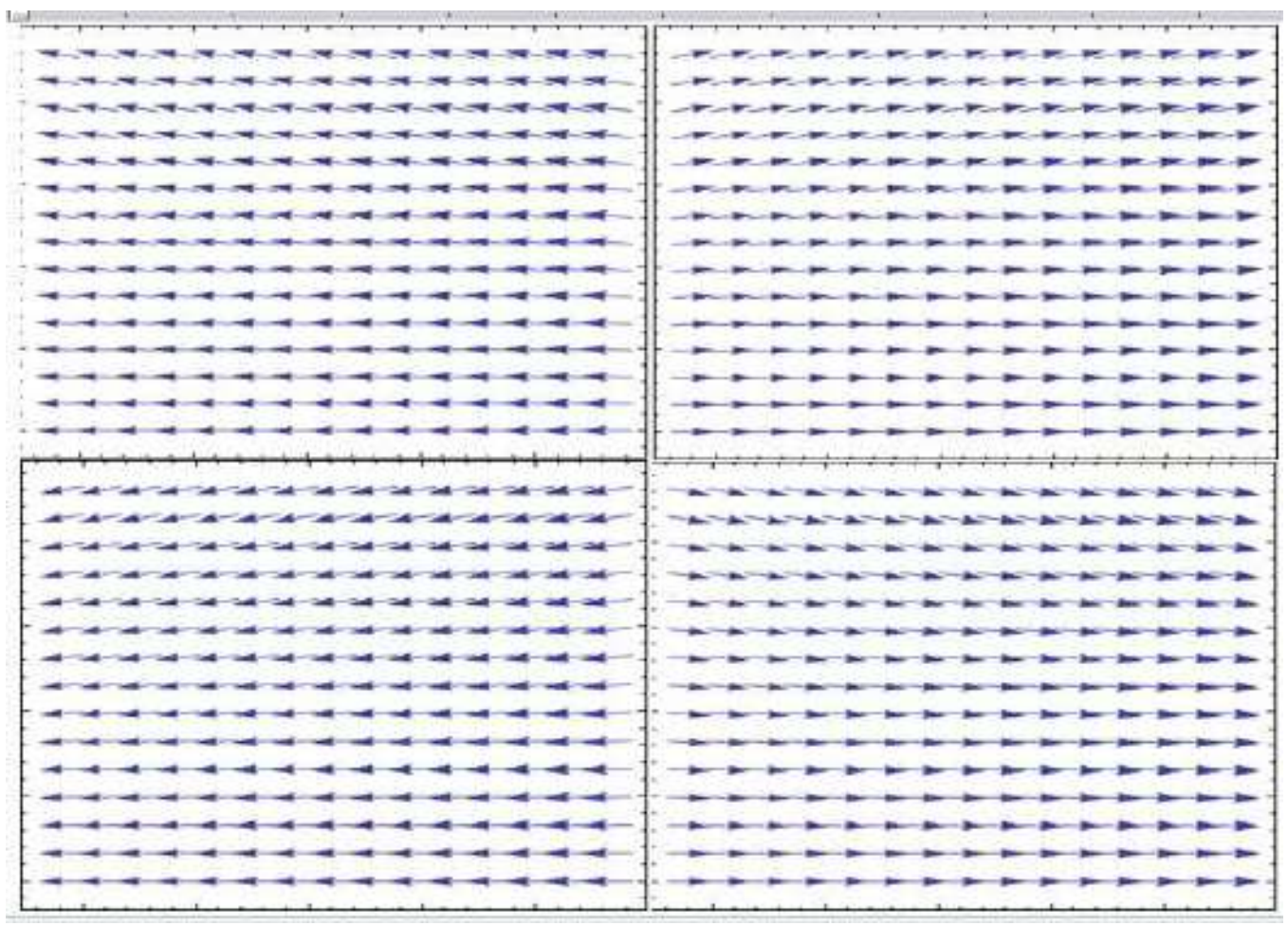

**Figure 5-9: Vector Plot for Right Ankle**

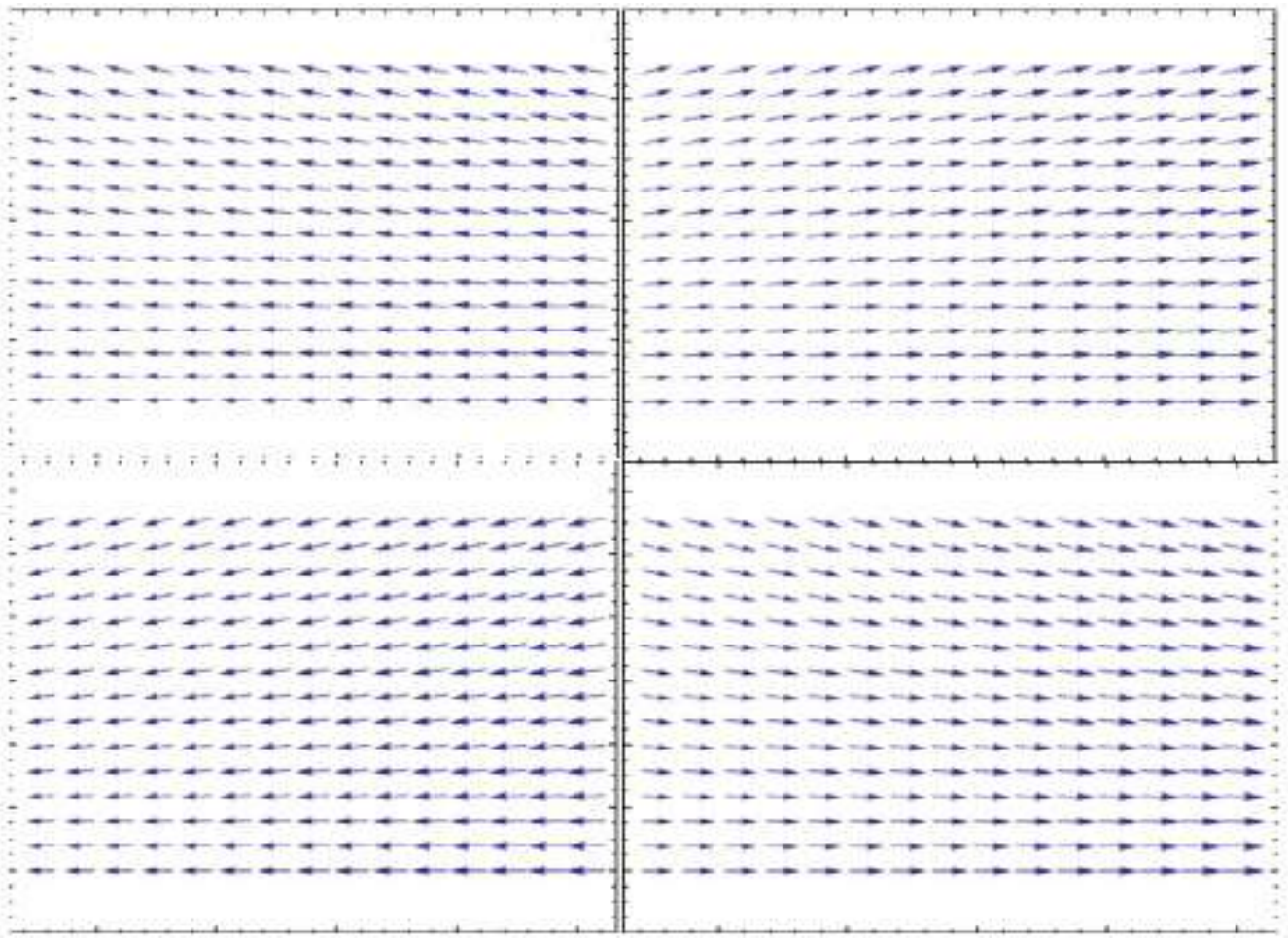

**Figure 5-10: Vector Plot for Left Hip**

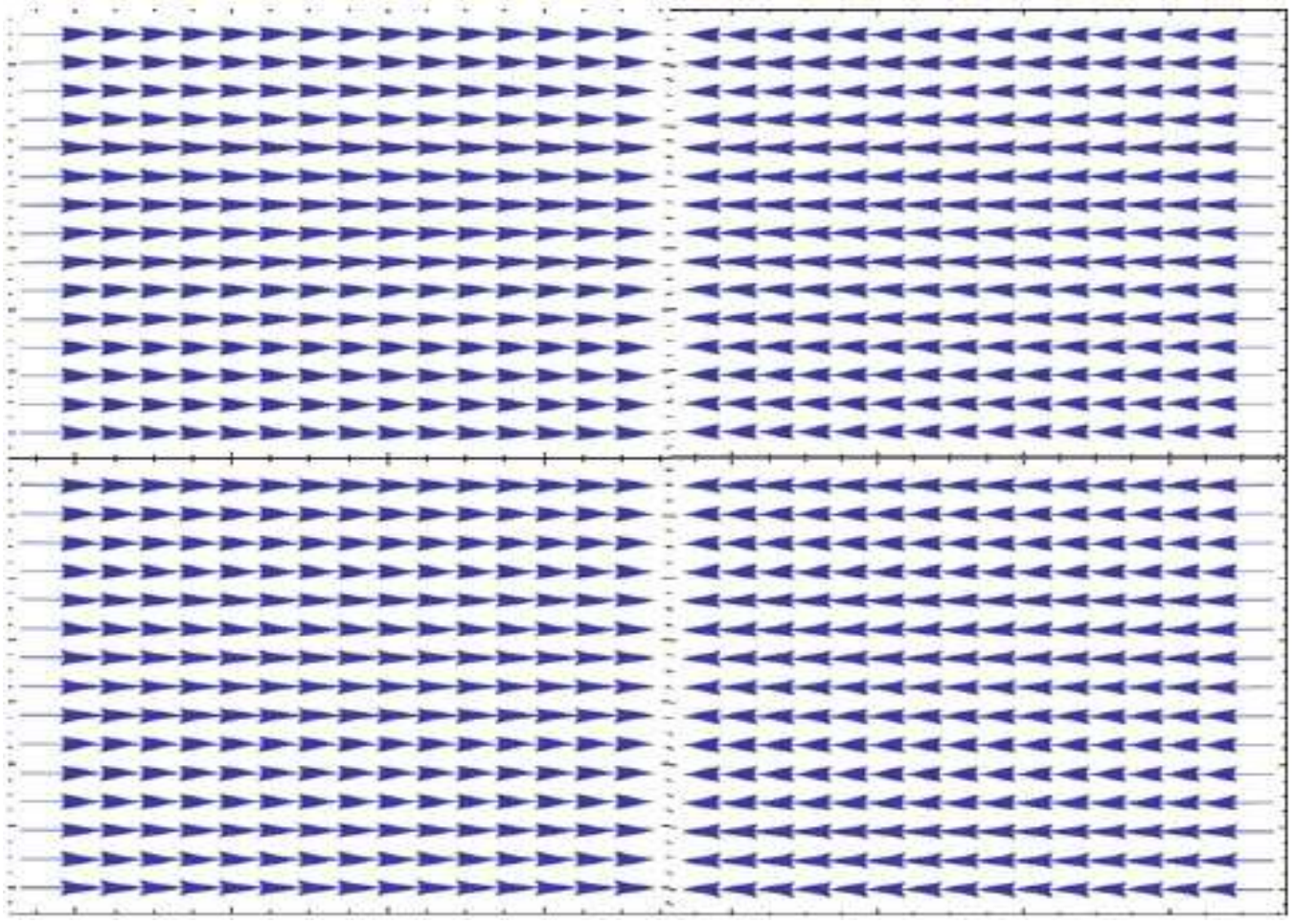

**Figure 5-11: Vector Plot for Right Hip**

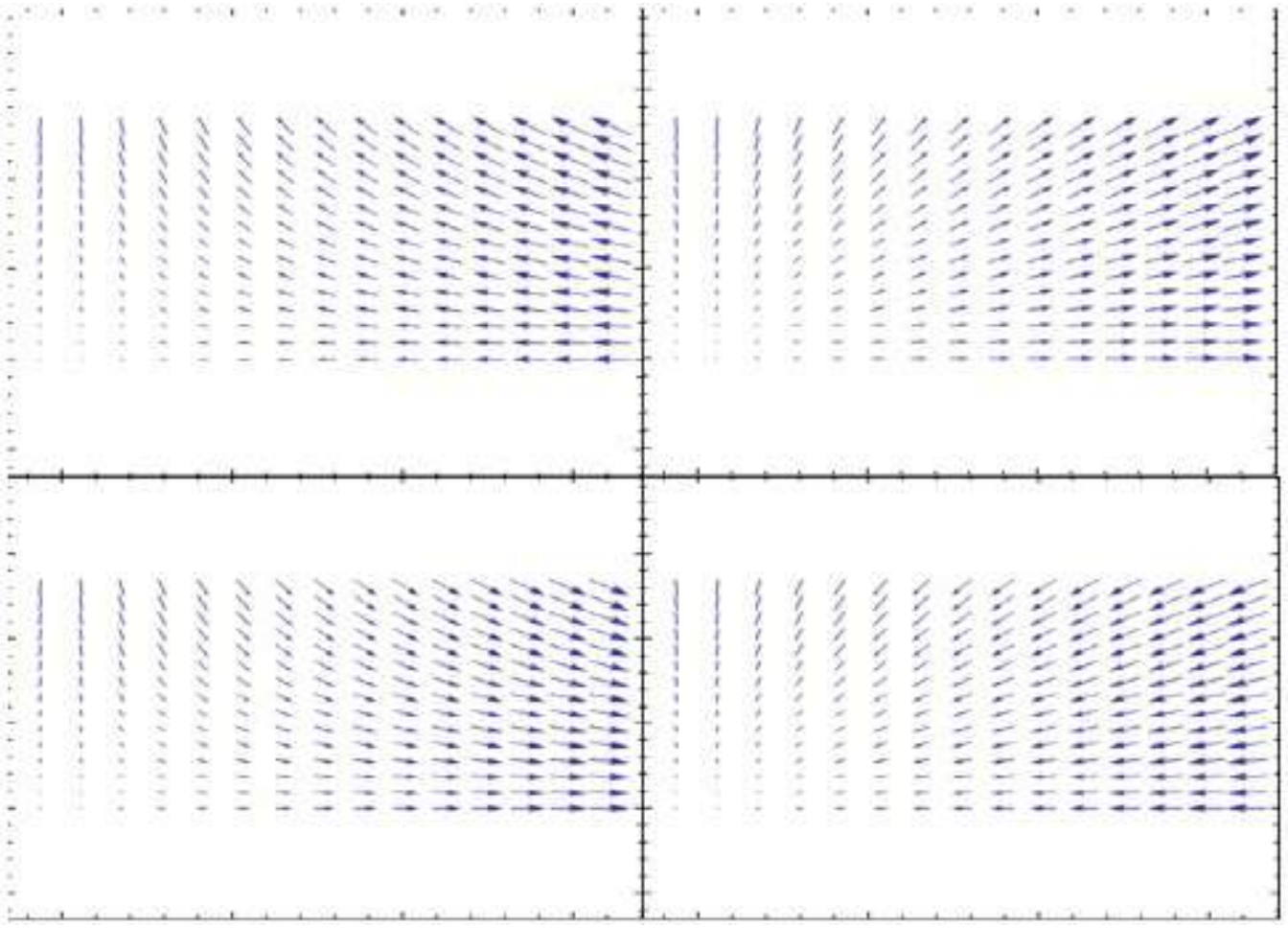

**Figure 5-12: Vector Plot for Left Knee**

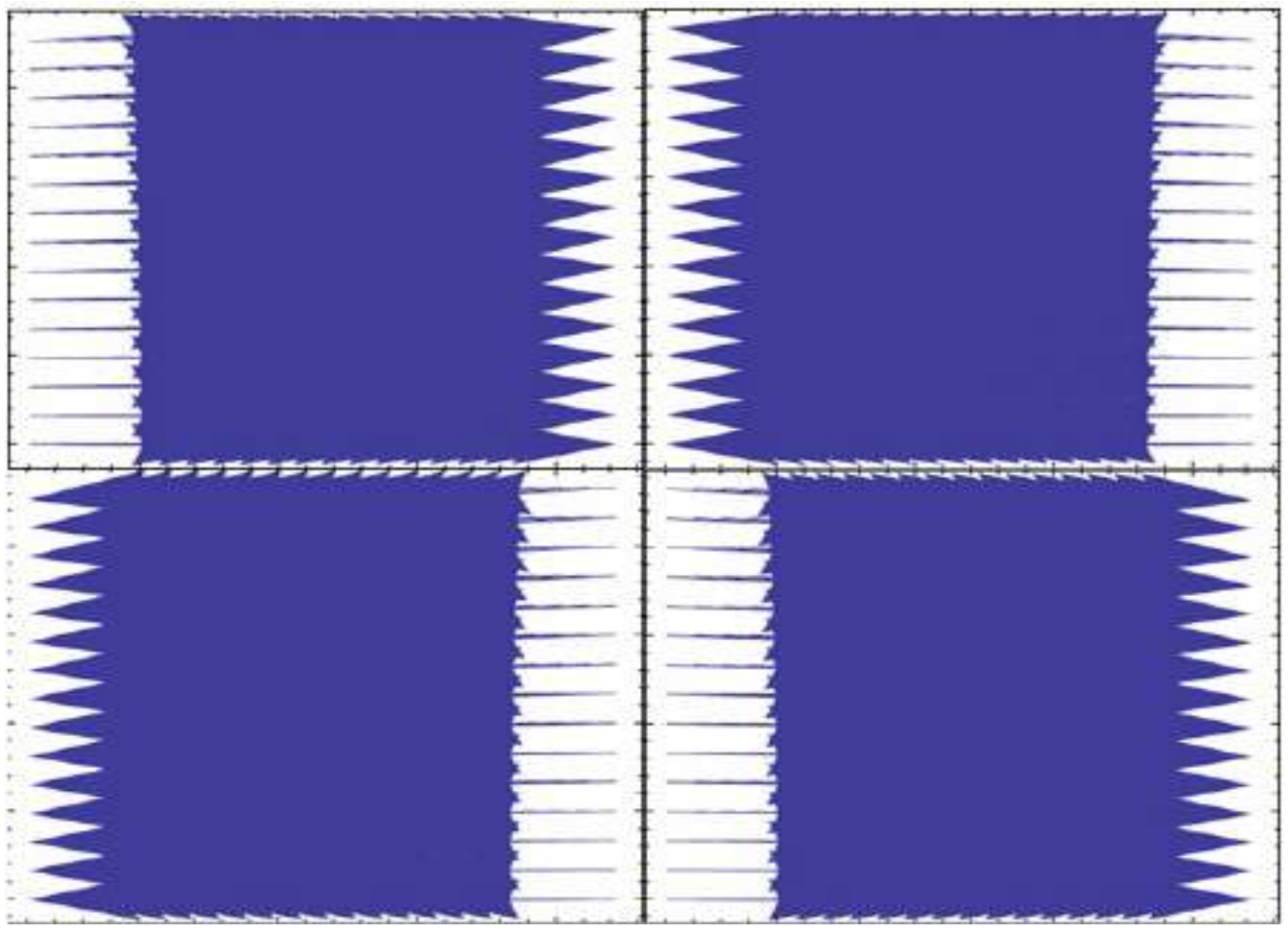

**Figure 5-13: Vector Plot for Right Knee**

Table 5.10represents the all possible stable joint angle range for all the six joints. It shows that between the ranges the trajectory will stable and ranges are depicted on curve. The x axis is the time axis and y is the joint angle value.

**Table 5:10: Joint angle range during different sub phase**

| | $\theta_{lh}$ | $\theta_{rh}$ | $\theta_{ln}$ | $\theta_{rn}$ | T |
|---|---|---|---|---|---|
| Initial Phase | $-7.576 \leq \theta_{lh} < -17.0965$ | $20.0072 \leq \theta_{rh} <$ 19.27827, | $-3.418 \leq \theta_{lk} < 1.861712$ | $-54.948 \leq \theta_{rk} < -19.8455$ | $0 \leq t < 0.1269$ |
| Loading Response | $-17.0965 \leq \theta_{lh} < -23.1957$ | $19.27827 \leq \theta_{rh} < 18.62818$ | $-1.861712 \leq \theta_{ln} < -1.079407$ | $-19.8455 \leq \theta_{rn} < 0.23332$ | $0.1269 \leq t < 0.021837$ |
| Mid Stance | $-23.1957 \leq \theta_{lh} < -19.6602$ | $18.62818 \leq \theta_{rh} < 20.77033$ | $-1.079407 \leq \theta_{ln} < -21.5189$ , | $0.23332 \leq \theta_{rn} < -19.62$ | $0.021837 \leq t < 0.026161$ |
| Terminal Stance | $-19.6602 \leq \theta_{lh} < 5.516299$ | $20.77033 \leq \theta_{rh} < 10.65248$ | $-21.5189 \leq \theta_{ln} < -65.9002$ | $-19.62 \leq \theta_{rn} < -19.6561$ | $0.026161 \leq t < 0.030395$ |
| Pre Swing | $5.516299 \leq \theta_{lh} < 19.08572$ | $10.65248 \leq \theta_{rh} < -1.41037$ | $-65.9002 \leq \theta_{ln} < -60.7186$ | $-19.6561 \leq \theta_{rn} < -11.0189$ | ,$0.030395 \leq t <$ 0.034623 |
| Initial Swing | $19.08572 \leq \theta_{lh} < 19.45794$ | $-1.41037 \leq \theta_{rh} < -12.9959$ | $-60.7186 \leq \theta_{ln} < -17.7753$ | $-11.0189 \leq \theta_{rn} < -5.18571$ | $0.034623 \leq t < 0.039607$ |
| Mid Swing | $19.45794 \leq \theta_{lh} < 17.33661$ | ,$-12.9959 \leq \theta_{rh} <$ $-21.4385$ | $-17.7753 \leq \theta_{ln} < 1.594863$ | $-5.18571 \leq \theta_{rn} < -5.89577$ | $0.039607 \leq t < 0.046046$ |
| Terminal Swing | $17.33661 \leq \theta_{lh} < 18.11716$ | ,$-21.4385 \leq \theta_{rh} <$ $-18.1146$ | $1.594863 \leq \theta_{ln} < -14.3932$ | $-5.89577 \leq \theta_{rn} < -28.6648$ | $0.046046 \leq t < 0.057367$ |

## 5.14 Values of range

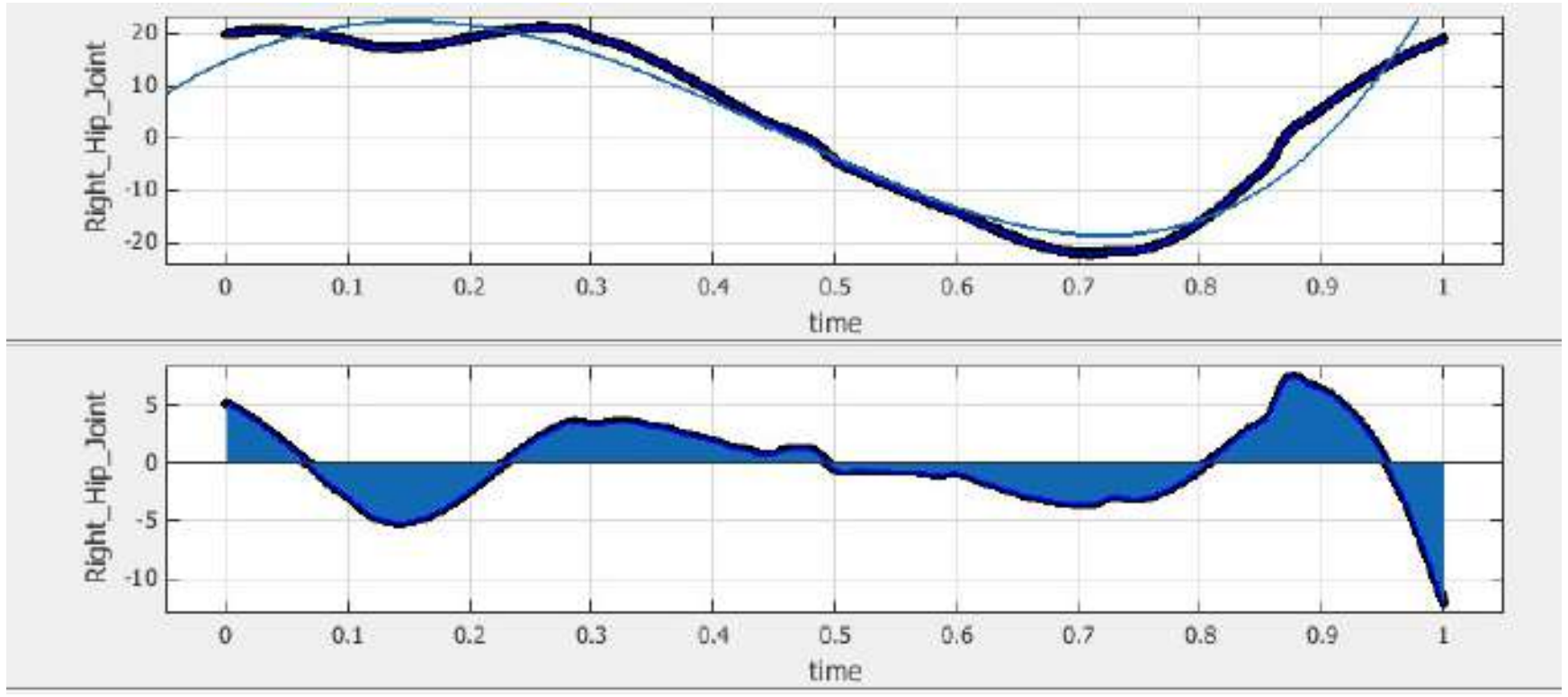

**Figure 5-14: The possible range of Right Hip joint trajectory where robot can walk**

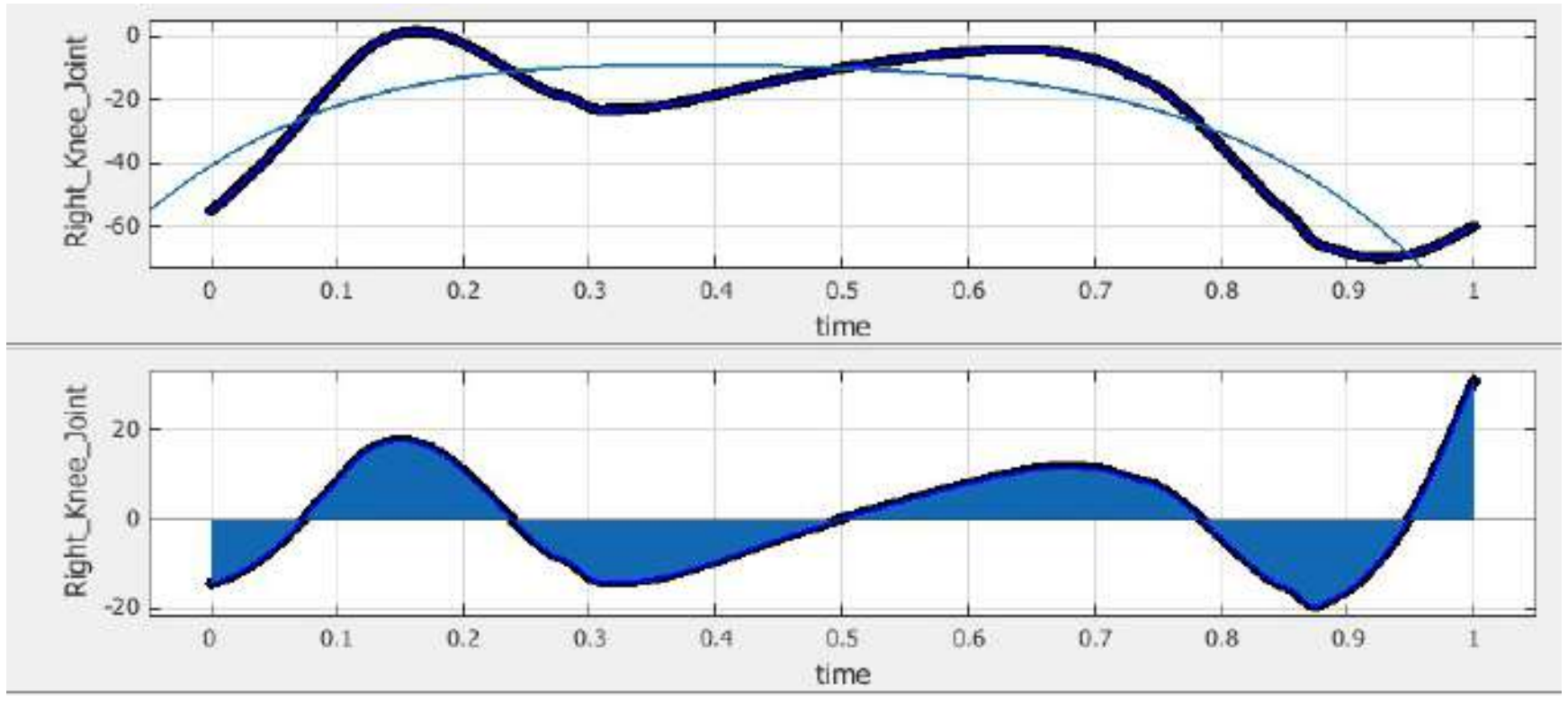

**Figure 5-15: The possible range of Right Knee joint trajectory where robot can walk**

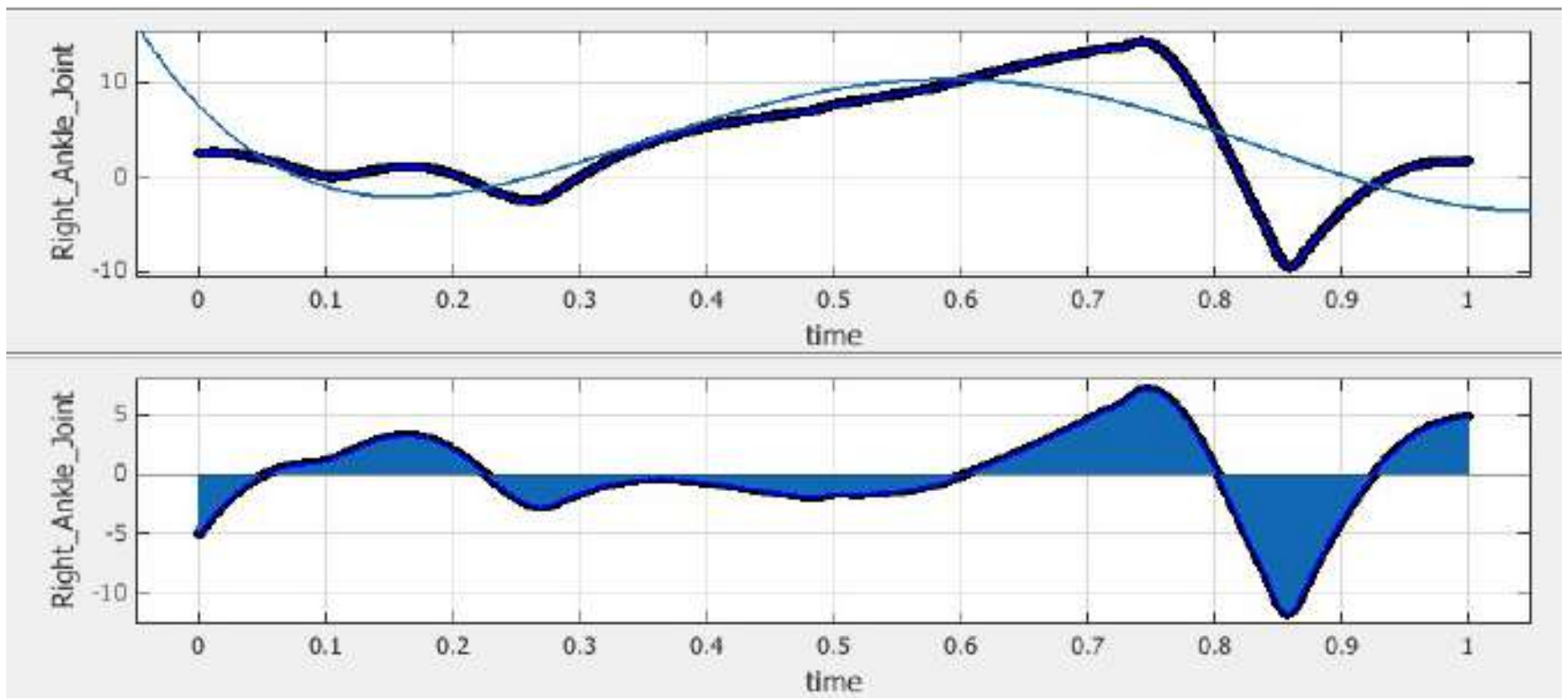

**Figure 5-16: The possible range of Right Ankle joint trajectory where robot can walk**

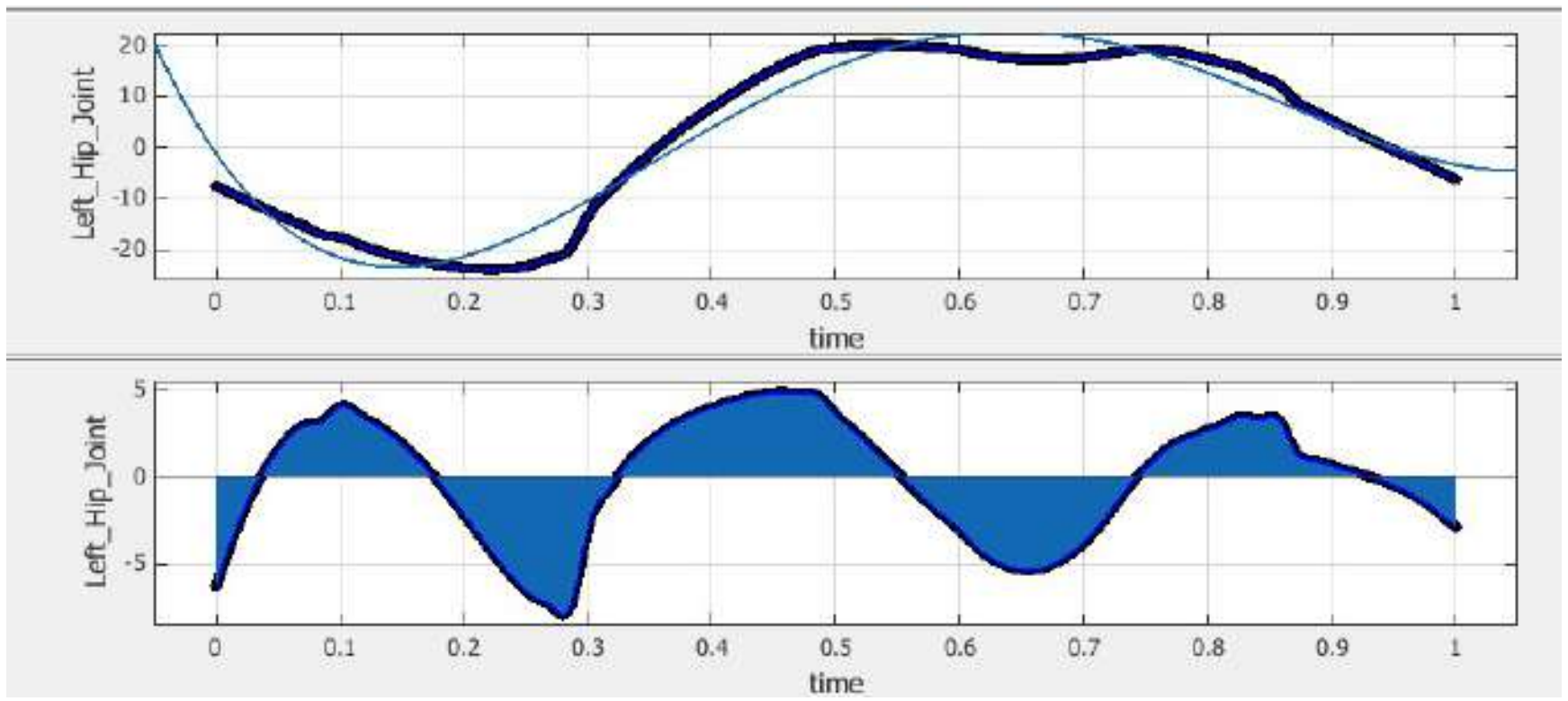

**Figure 5-17: The possible range of Left Hip joint trajectory where robot can walk**

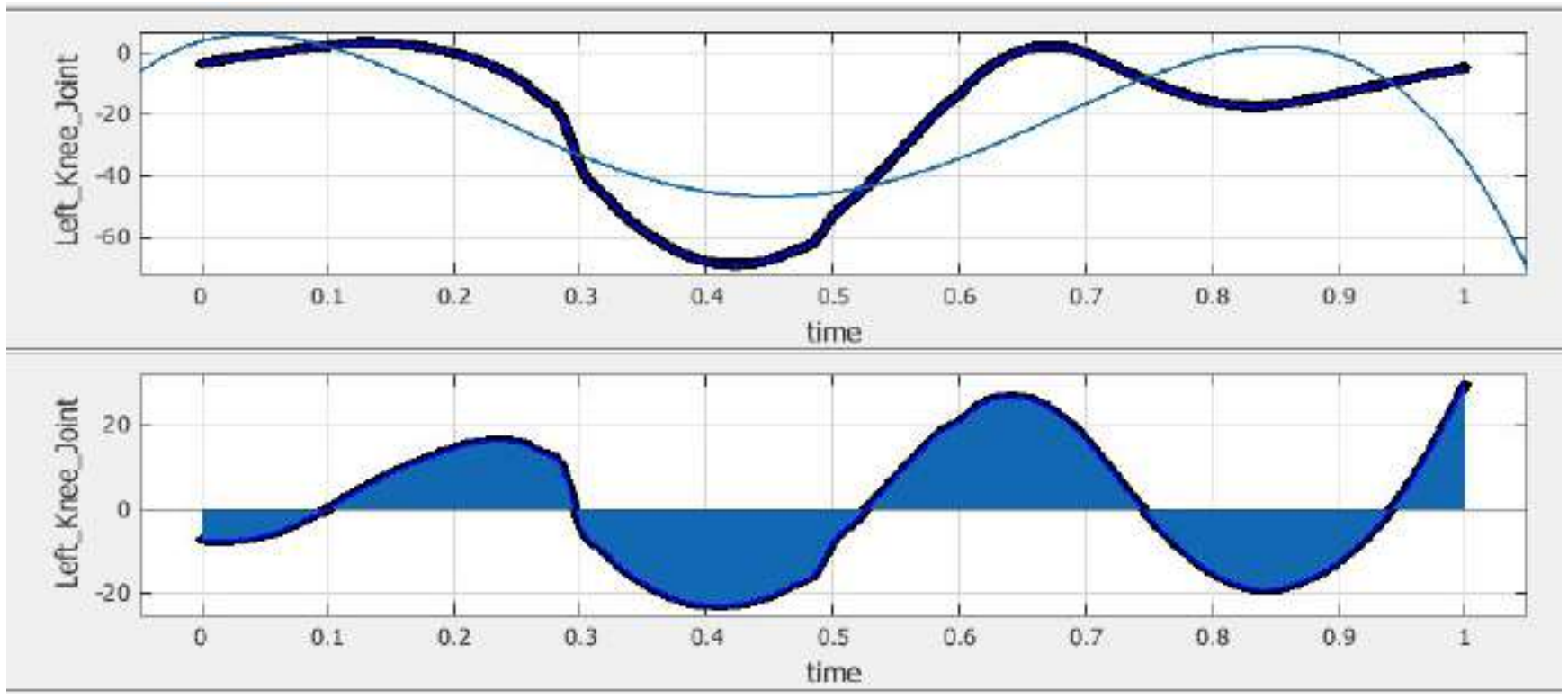

**Figure 5-18: The possible range of Left Knee joint trajectory where robot can walk**

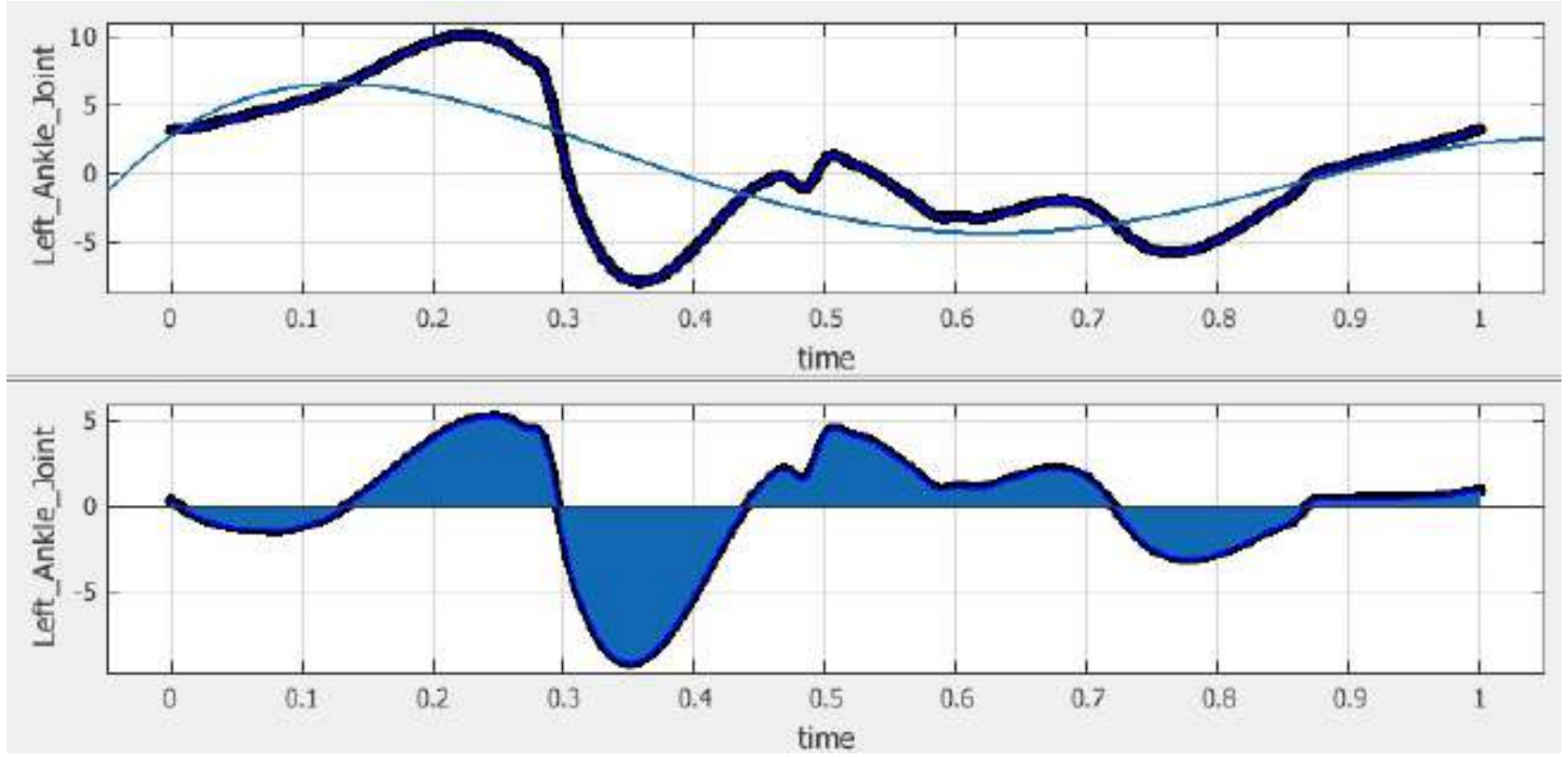

**Figure 5-19: The possible range of Left Ankle joint trajectory where robot can walk**

To Avoid Over fitting we have fitted the equation of 4th order polynomial and 2nd order polynomial and subtracted both. It will provide the possible joint angle range. We can calculate the $E_{in}$ as the difference of 4th order polynomial and 2nd order polynomial.

$$E_{in} = E_4 - E_2 \qquad - \quad (9)$$

Where $E_4$is the 4th order polynomial fitting and $E_2$2nd order polynomial fitting the difference between this ranges we can proposed the solution in which range the joint angle value can vary. Figure 5.14 to 5.19 are the joints angle range of the right hip, right knee, right ankle, left hip, left knee and left ankle. Similarly we have calculated for rest joints.

## 5.15 Toward universal computational model

This section describes the joint trajectory comparison of our hybrid automata model i.e. hybrid automata dynamic walker with GAIT2335 model of OpenSim. It is an important development to prove that hybrid automata based computational model can act alternative for other model and it is universal model. Figure 5.20 shows the comparison between the OpenSim model Gait2354 model, our developed hybrid automata model and normal human gait for all the different joints i.e. right and left hip, knee and ankle. It is observed that all the three have same pattern so hybrid automata based model can be used as an alternative for other model.

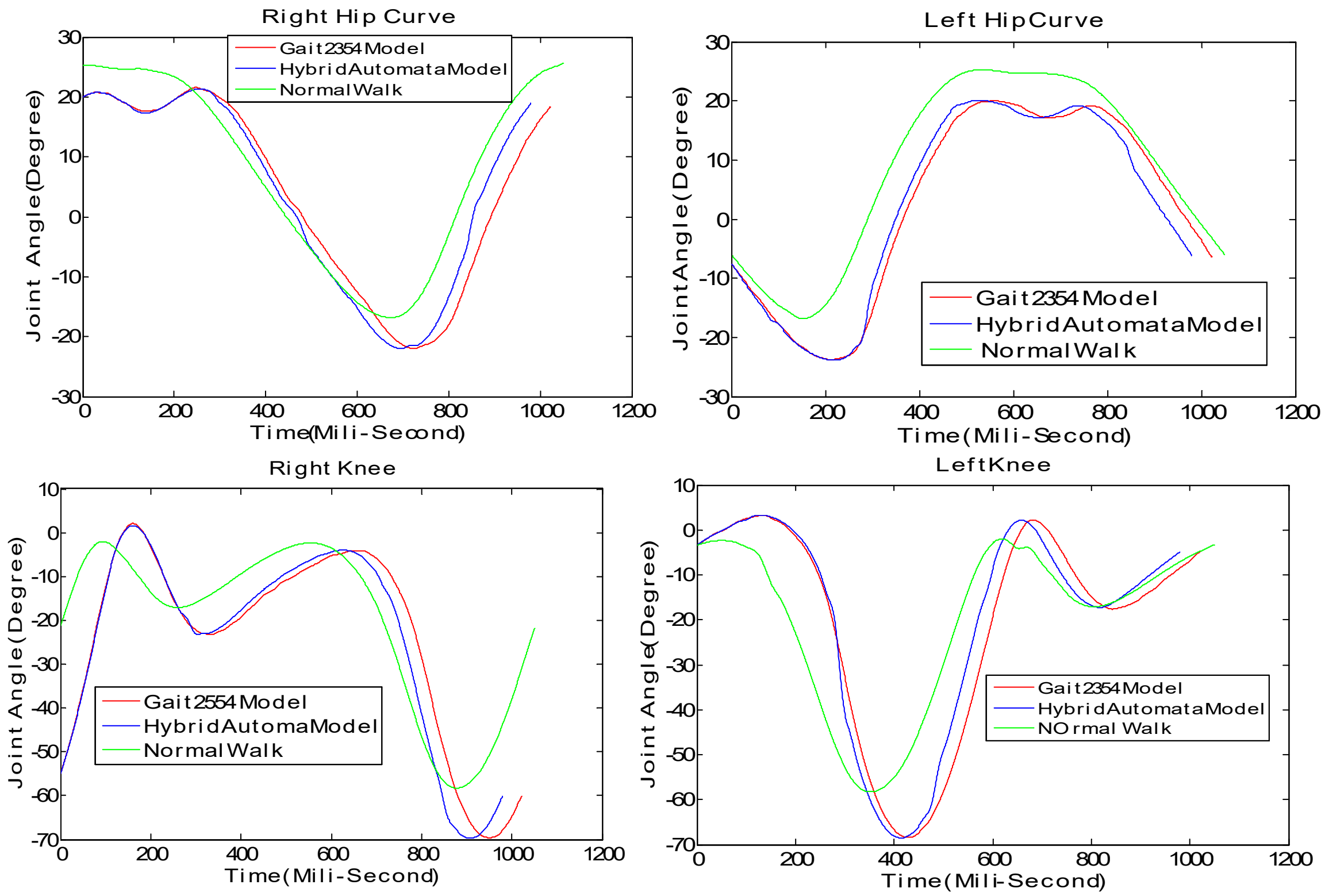

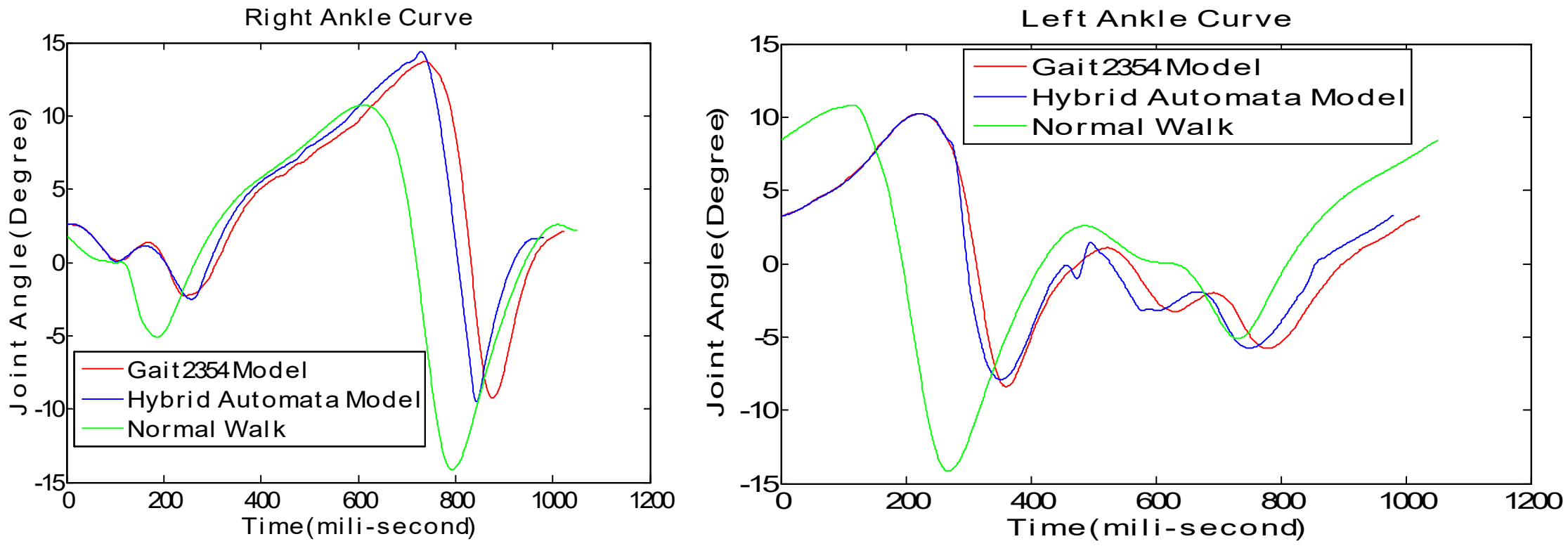

**Figure 5-20:Gait Pattern of different joints for Gait 2354 model, hybrid automata model and normal walk**

Figure 5.21is the stick diagram of our hybrid automata bipedal walk model where as Figure 5.22 is the stick diagram for Gait2354 OpenSim model. We have considered the shank length (l1=.4 cm), thigh length (l2=.4 cm) and foot length (l3=.1cm)

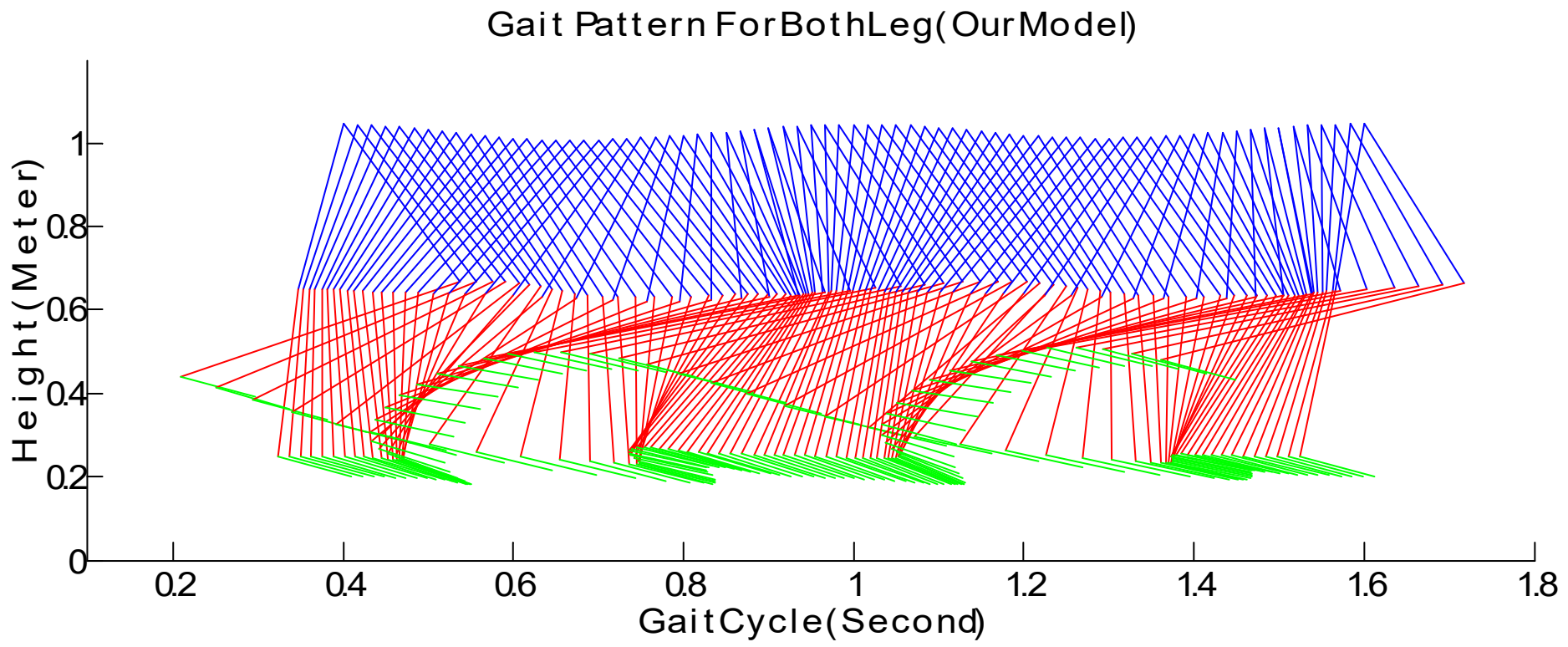

**Figure 5-21: Stick diagram for both leg gait pattern of our model**

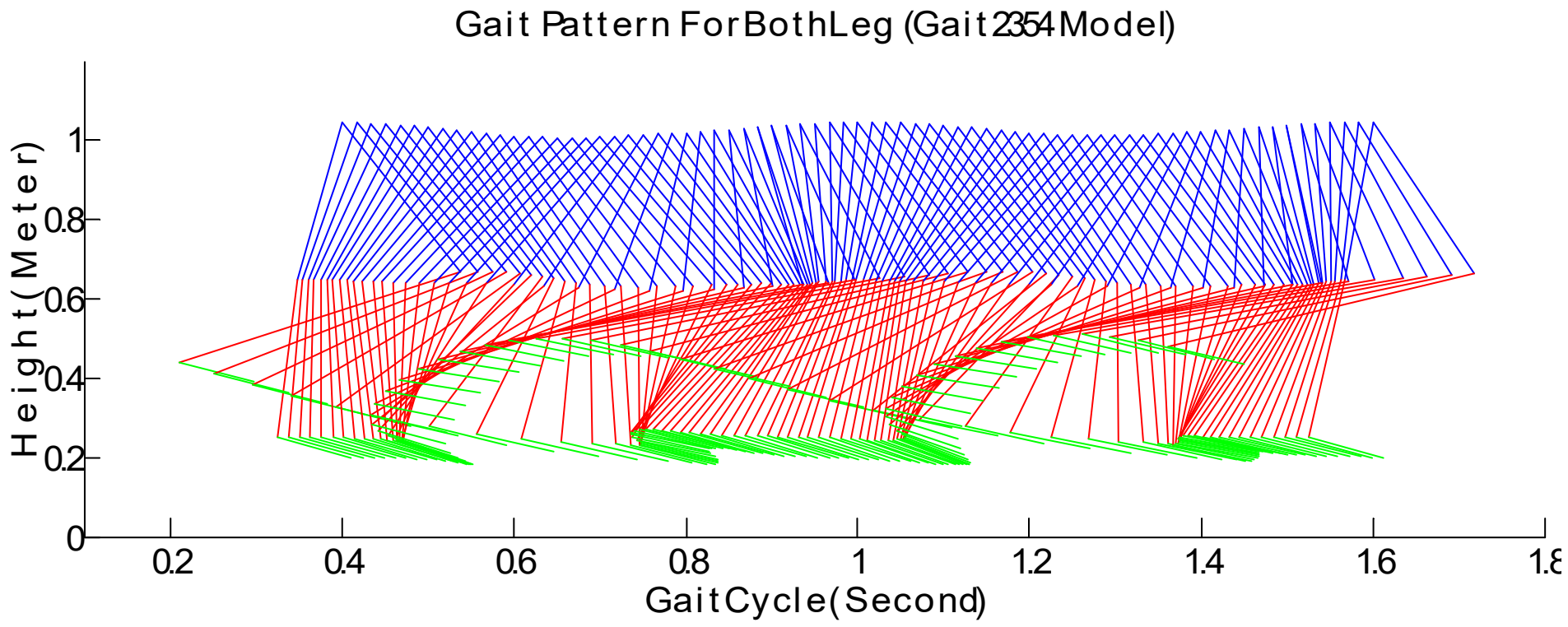

**Figure 5-22: Stick diagram for both leg gait pattern of model Gait2354**

Limit cycle curve for left and right hip/knee is shown in figure 5.23 and 5.24 left and right knee and left and right hip respectively.

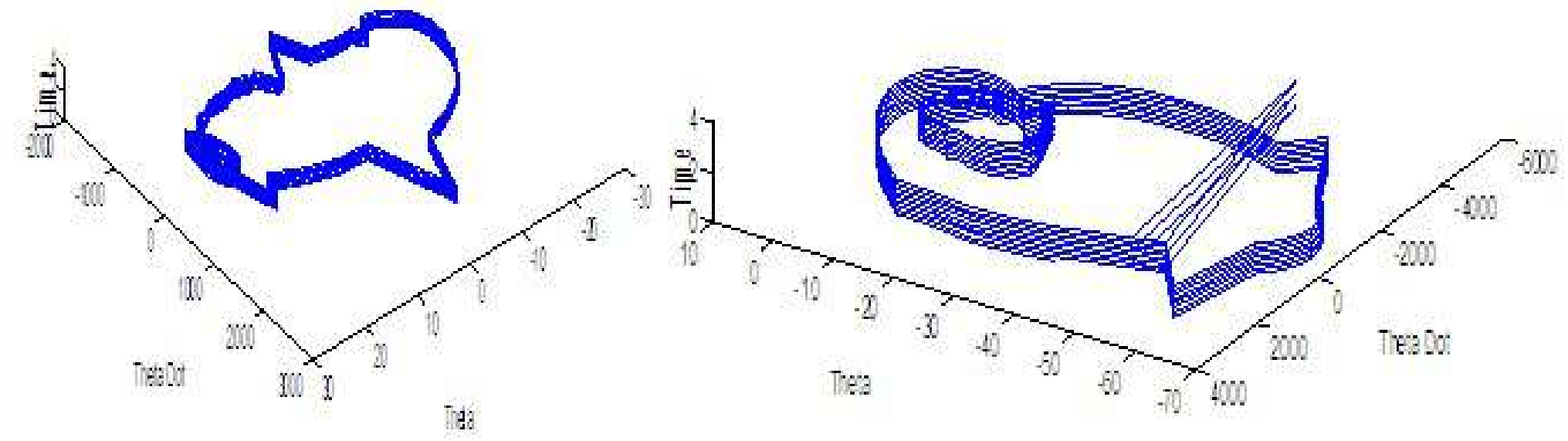

**Figure 5-23: Limit Cycle curve for Left and Right Knee**

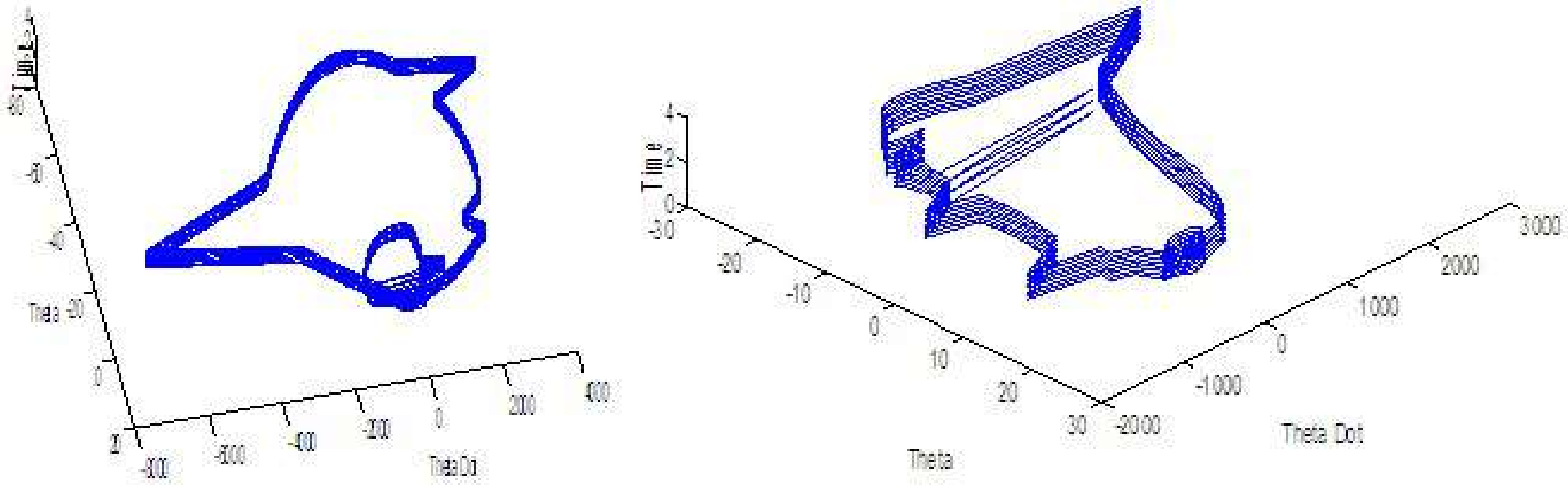

**Figure 5-24: Limit Cycle curve for Left and right Hip.**

## 5.16 Comparison of different joints trajectories:

Figure 5-25 and 5-28 is the joints trajectory comparison of left and right hip and knee Green: Vector field generated Joint trajectory 1 Black: Vector field generated Joint trajectory 2, Red: HOAP2 Joint trajectory Blue: Hybrid automata model Joint trajectory

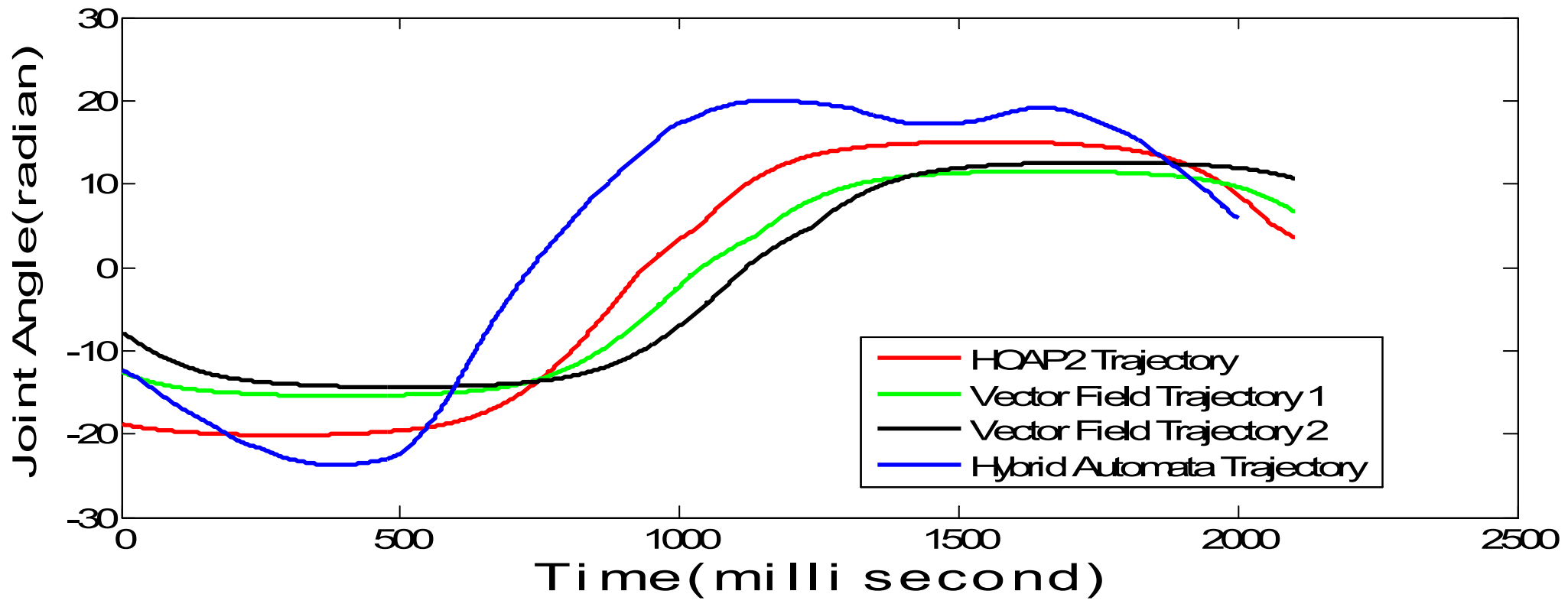

Figure 5-25: Comparison of Left Hip joints trajectories

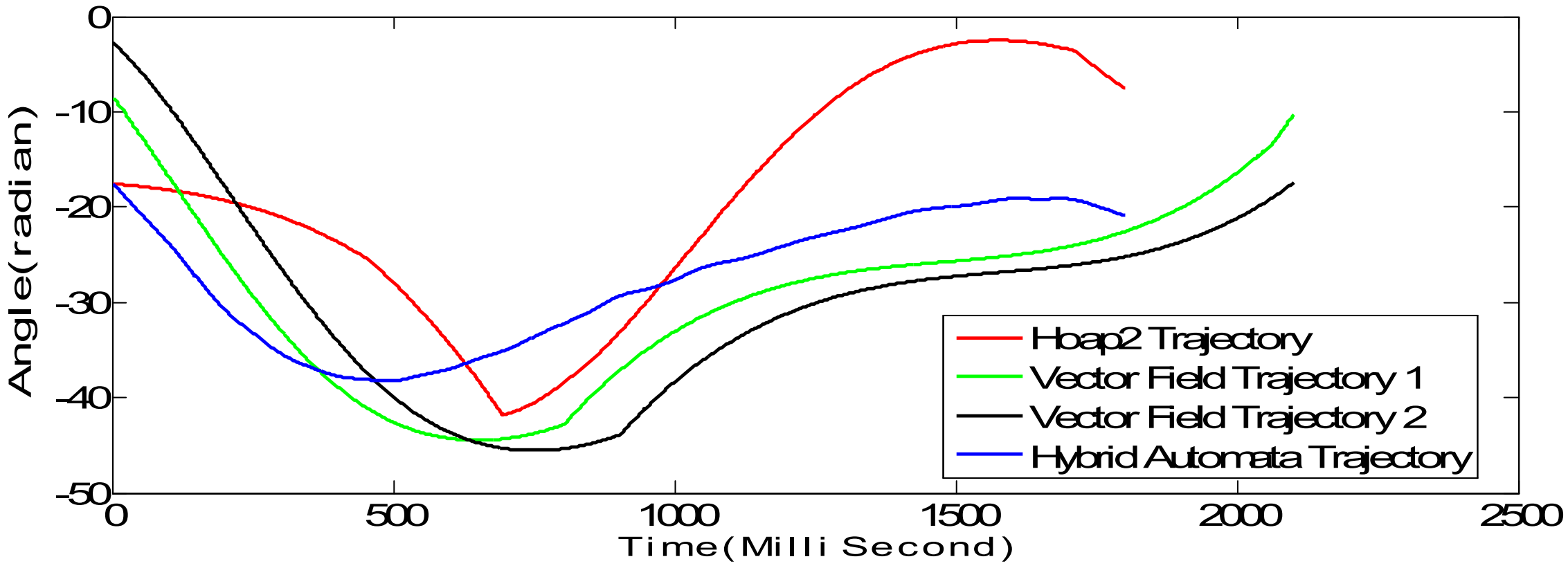

Figure 5-26: Comparison of Left Knee joints trajectories

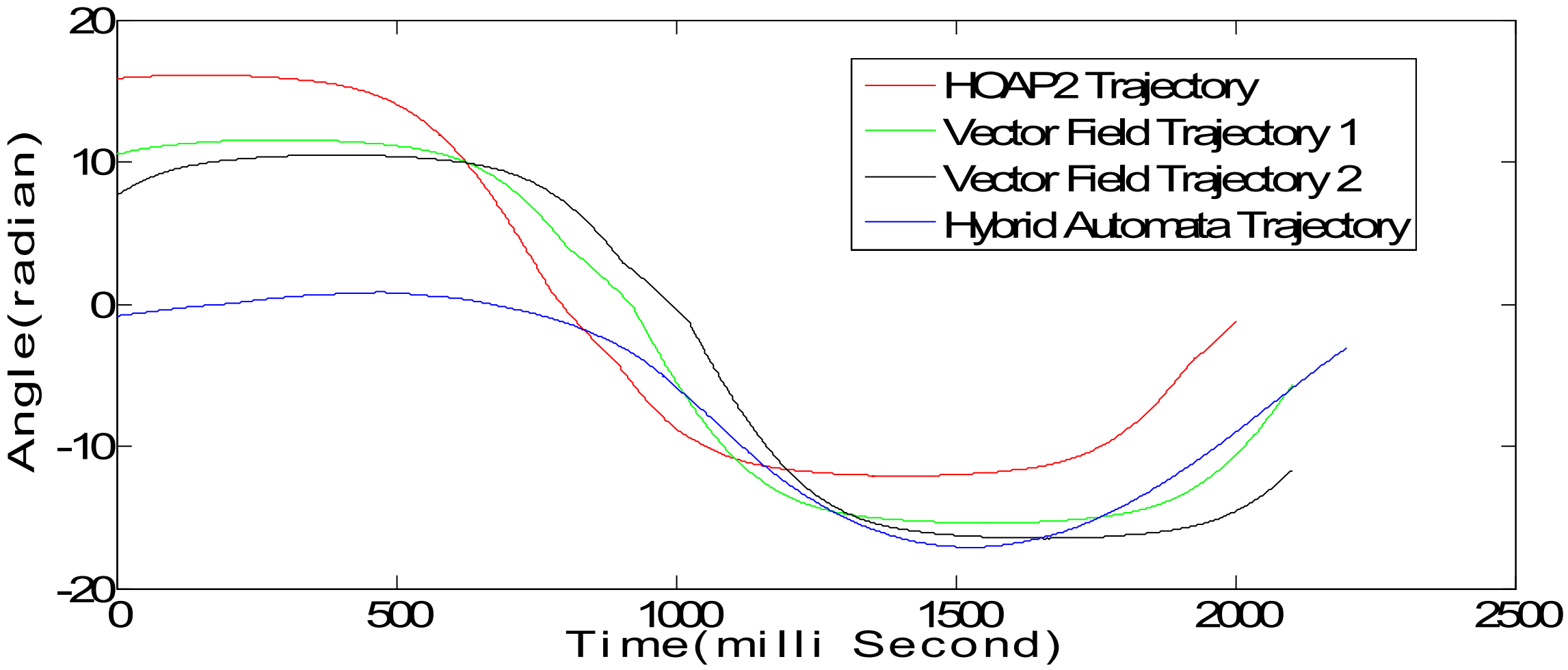

Figure 5-27: Comparison of Right Hip joints trajectories

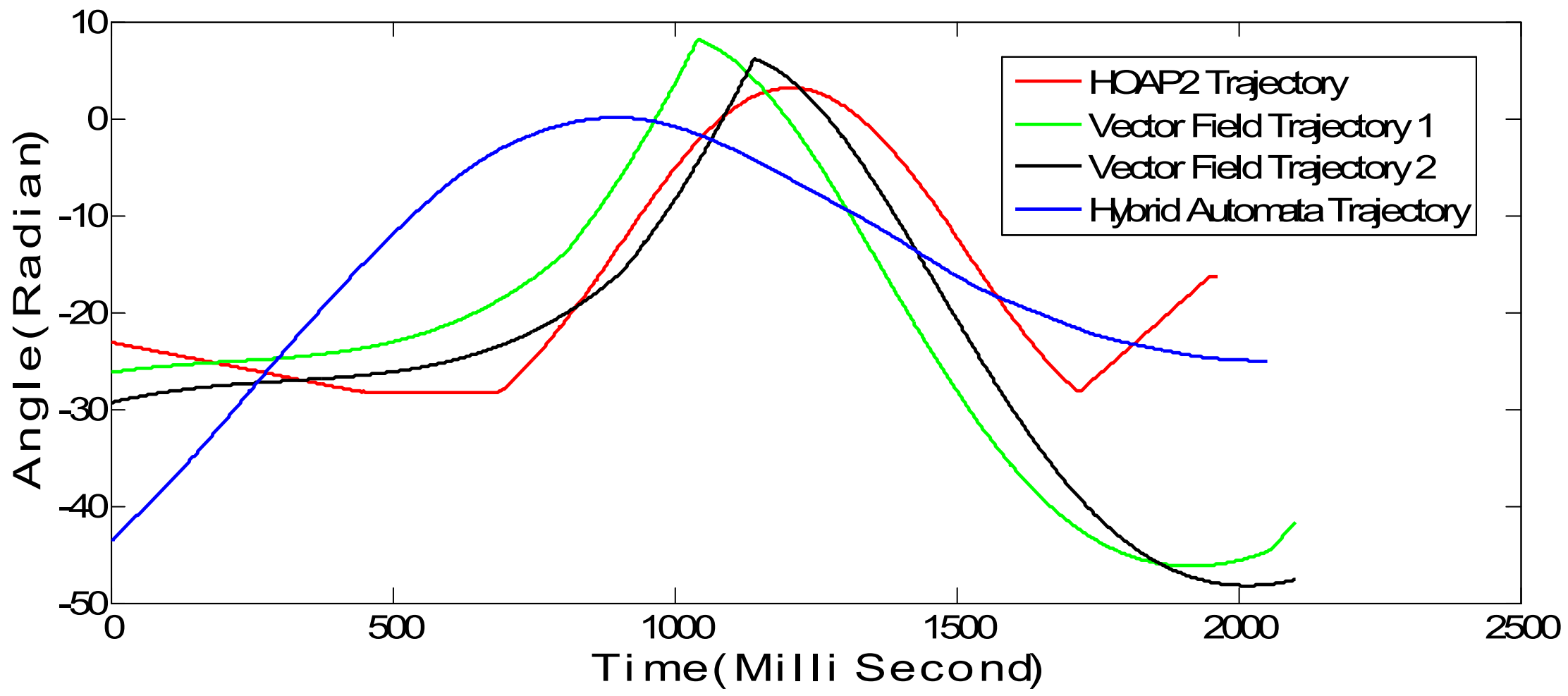

**Figure 5-28: Comparison of Right Knee joints trajectories**

## 5.17 OpenSim Simulation Results

Figure 5.29 shows the OpenSim Simulation result of our dynamic walker hybrid automata model for data generated through our hybrid automata equations, figure 5.30 shows the OpenSim Simulation result of our dynamic walker hybrid automata model for human Both Leg gait data in OpenSim. Figure 5.31shows the OpenSim Simulation result of our dynamic walker hybrid automata model for original gait2354 model's gait data. The curve in these images satisfy the gait cycle curve for the hip joint which is sinusoidal in nature and also the knee joint with sinusoid pattern having double humps.

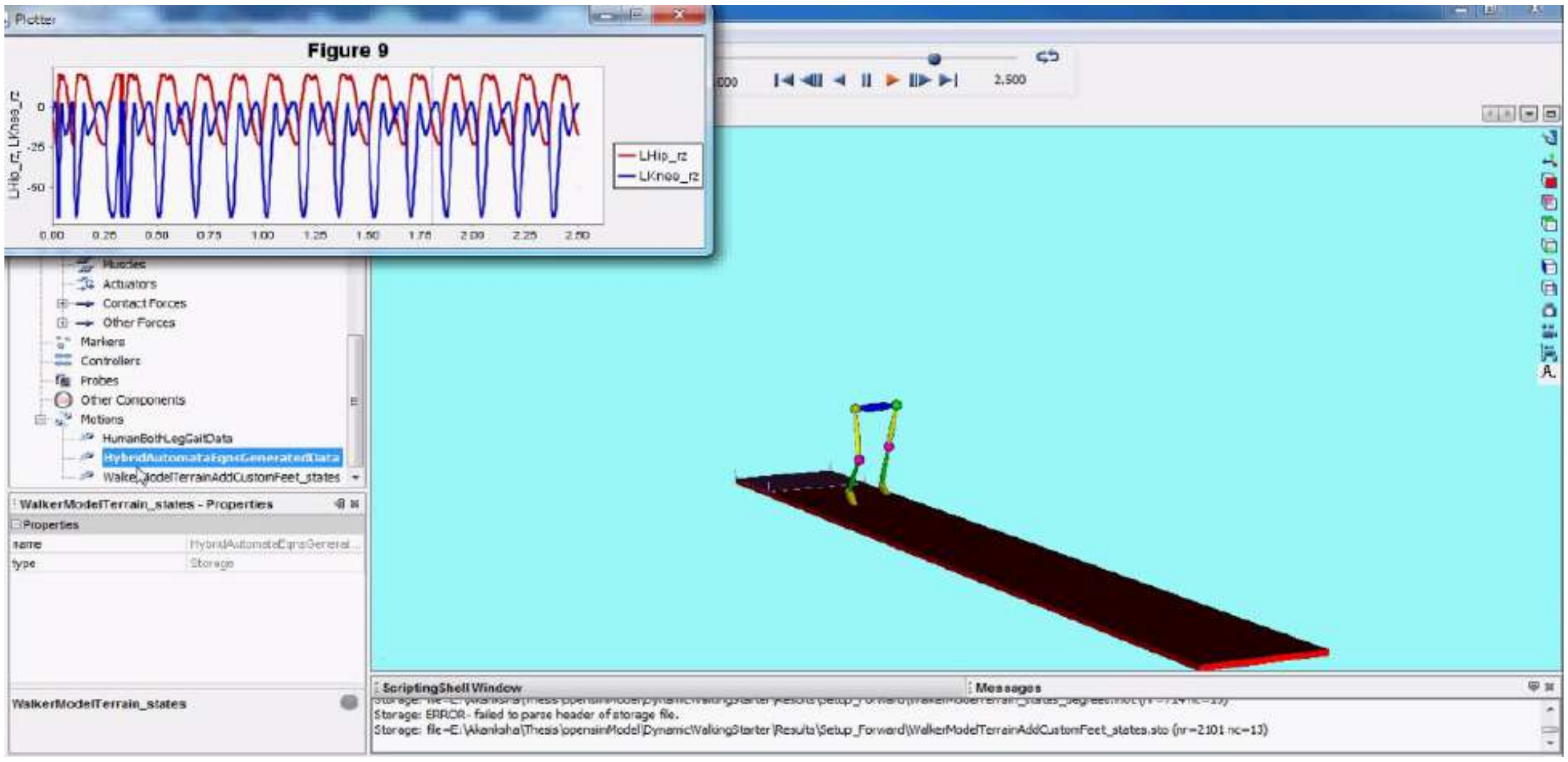

**Figure 5-29: OpenSim Simulation result of dynamic walker model for data generated through hybrid automata equations.**

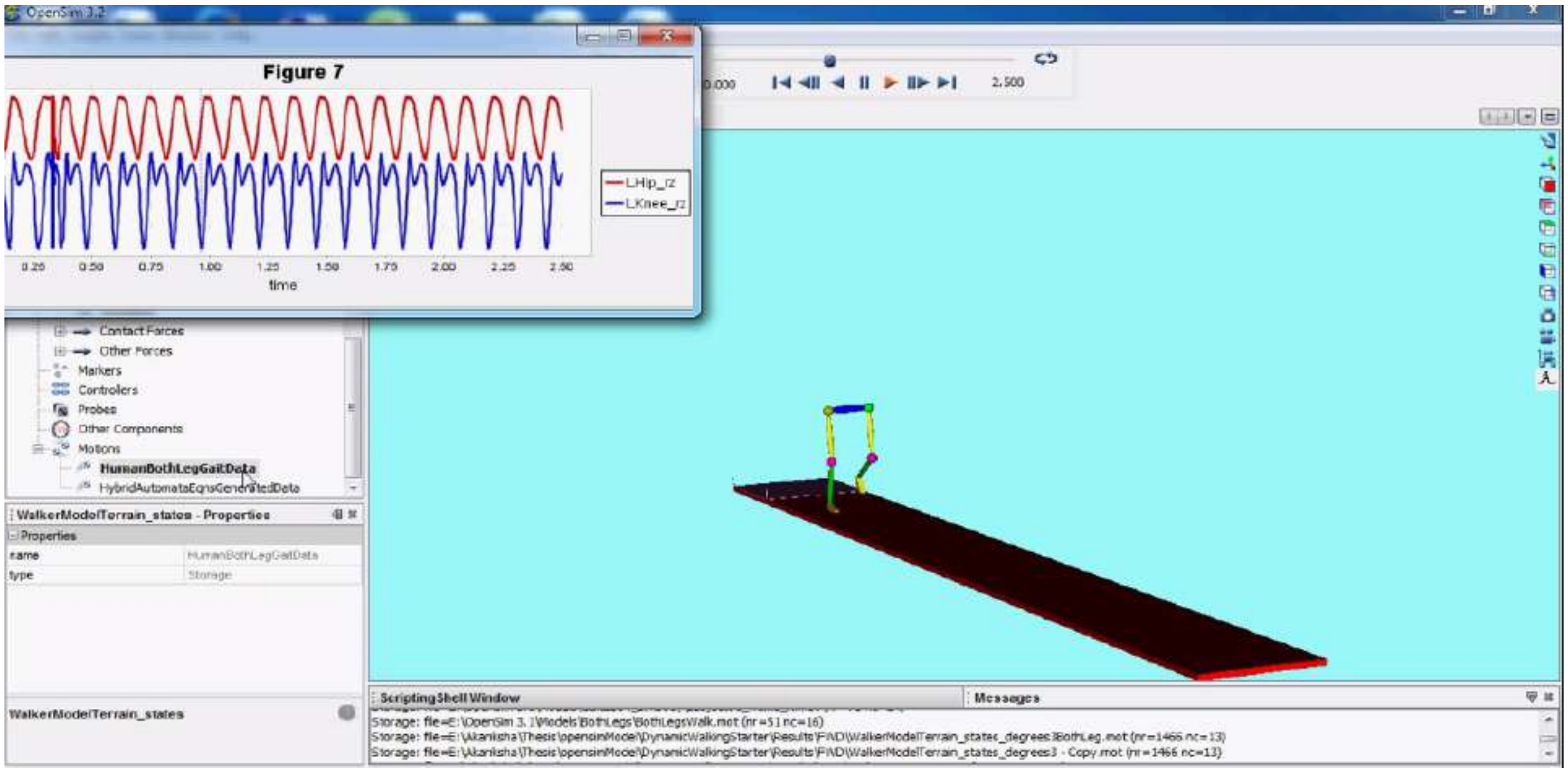

**Figure 5-30: OpenSim Simulation result of dynamic walker models for human Both Leg gait data in OpenSim.**

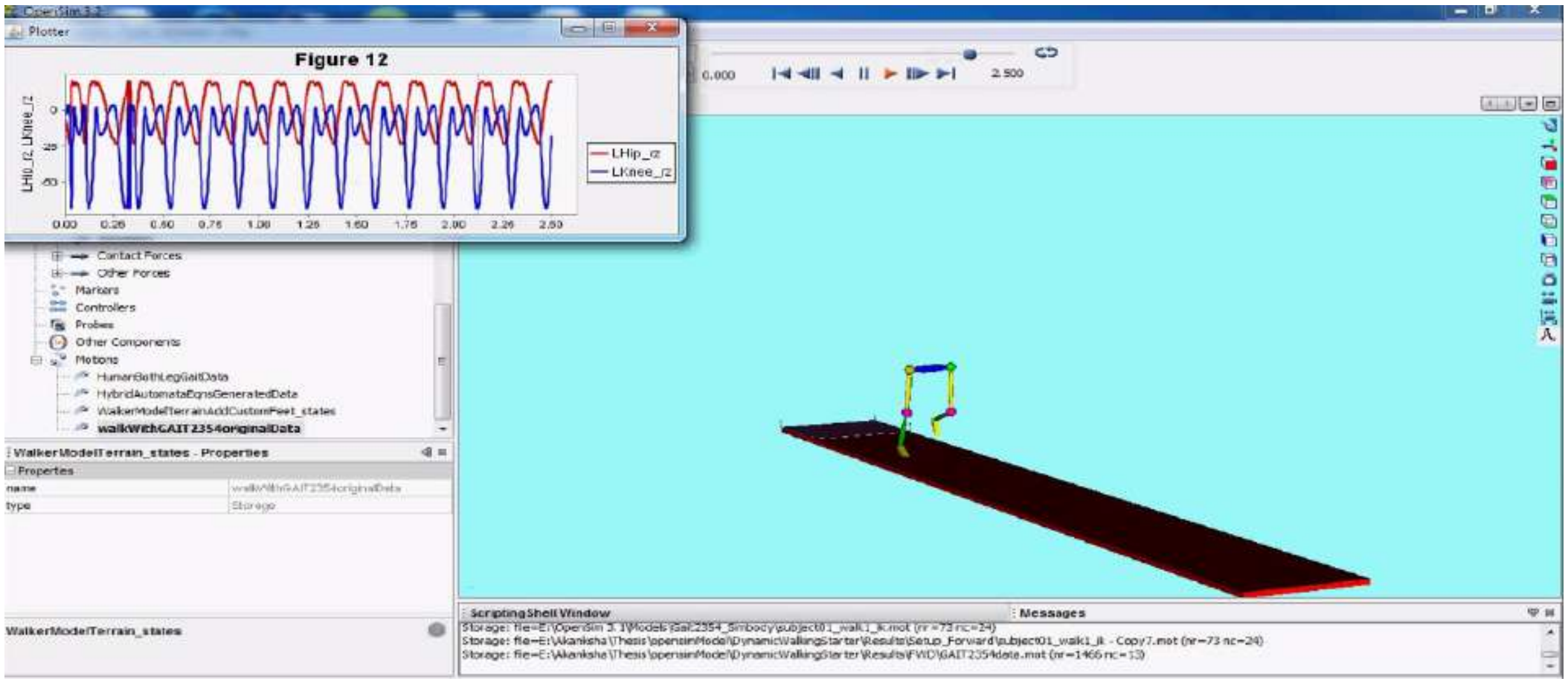

**Figure 5-31: OpenSim Simulation result of dynamic walker model for original gait2354 models gait data**

The below figure 5.32 is the left and right leg state prediction using CA rule on opensim bipedal model.

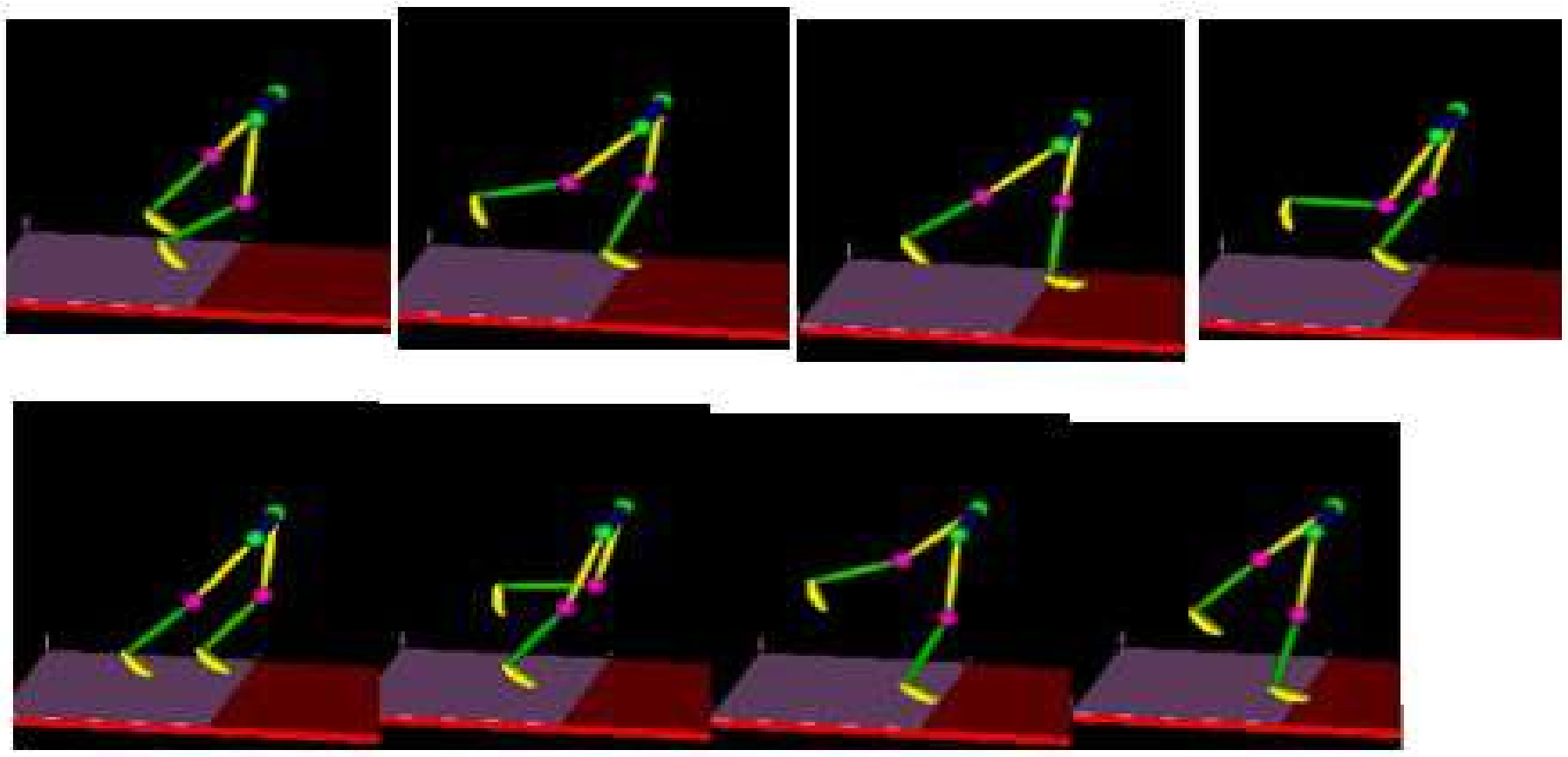

**Figure 5-32: For Left and Right Leg state prediction using CA rule**

## 5.18 Verification of Vector field using HOAP2 model

The figure 5.33 is the HOAP2 model walking using joint trajectories generated through designed vector field.

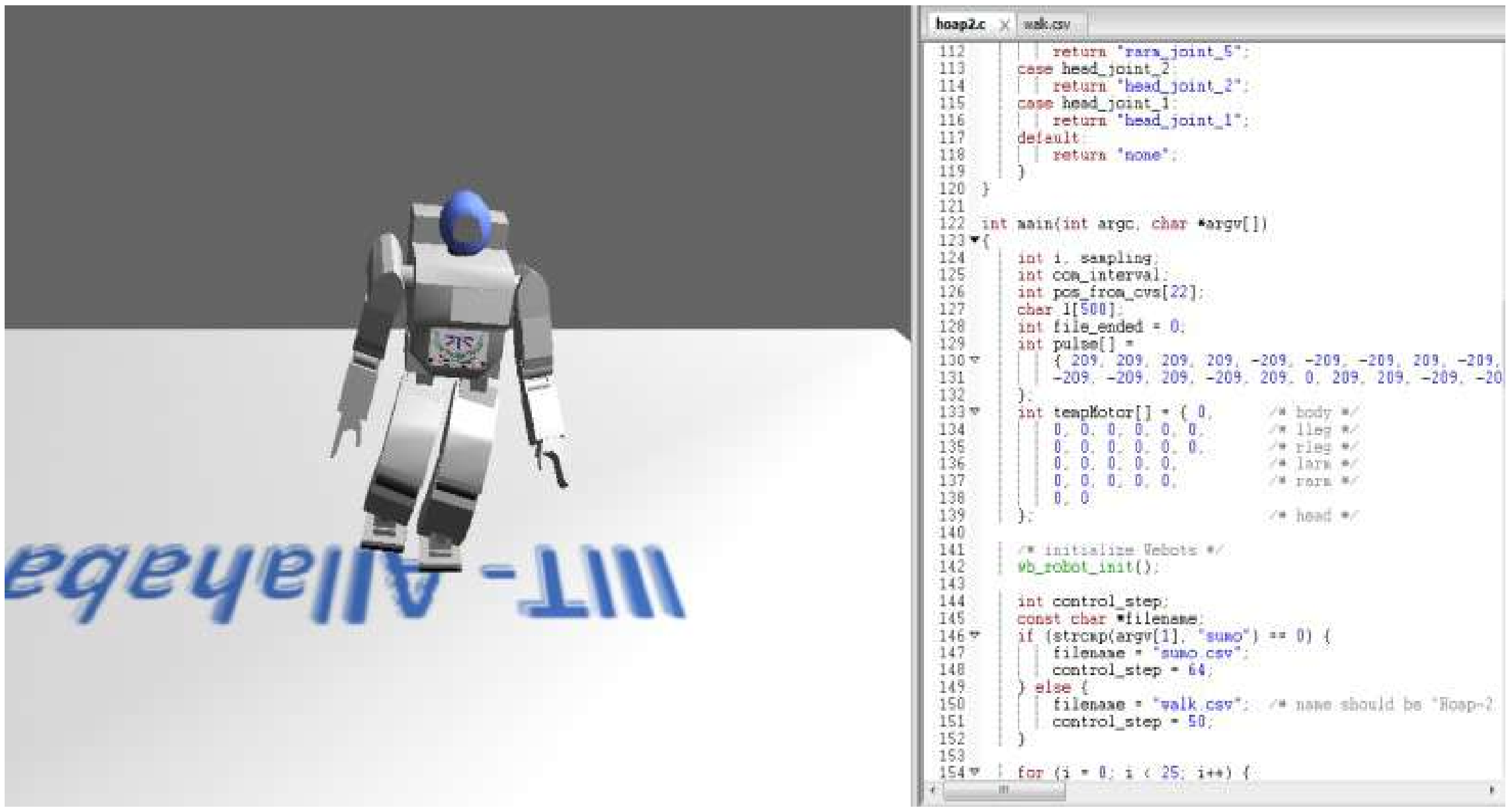

**Figure 5-33: HOAP2 model walking through designed vector field.**

## 5.19 Summary

We presented an approach of modelling joint trajectories of biped locomotion using hybrid automata. The strong theoretical framework of hybrid automata provides the necessary scalability and generic character of the model making it suitable for generating walking trajectories of any correctly configured biped robots. Since the vector field for generating joint trajectories are designed using actual normal human gait captured data, they are inherently stable. To achieve this, designing appropriate vector fields f: was critical and that is the major contribution of the work. Other contributions include developing a biped model which can accommodate any normal human legs including ankle and feet. Comparison of our hybrid automata based model with the OpenSim model gait 2354 shows the correctness of the development. We have generated the 2 possible joints trajectories and applied it to HOAP2 robot. The HOAP2 robot is perfectly able to walk. We have also presented the CA model for predication of bipedal gait states.Later we have tested these trajectories to the HOAP-2 bipedal robot and have shown that these are stable configurations. Hybrid automata vector fields are presented separately in tabular form for each phase of gait cycle and for each joint angle. The vector field is designed for each six joints the left hip, left knee, left ankle, right hip, right knee and right ankle joints. These vector fields generated temporal joint angle data which could make a biped robot walk stable.

# Chapter 6: Classification of Human Gait and Push Recovery data for computational model

In the previous chapter 5 we have generated the set of stable joints trajectories through our hybrid automata based model for different walk. The generated trajectories are successfully tested on HOAP2 model and verified using different mathematical analysis tool. It motivated us to develop the similar computational data driven model for bipedal push recovery. To focus the push recovery model we have classified the push recovery data into four categories of pushes (Small, Medium, Moderately High and High). We have also compared the hybrid automata generated trajectories for normal walk with other gait.

## 6.1 Deliverable: Classification of human gait and push recovery data

The main deliverable of this chapter is the classification of Human Push recovery and gait data collected through experiment. We have collected data for four kinds of pushes were applied (Small, Medium, Moderately High, High) during the experiment to analyze the recovery mechanism. We have classified using deep neural network and achieved 89.28% classification accuracy. In second section of this chapter we have classified gait data into four categories named normal, crouch-2, crouch-3 and crouch-4. The data is generated through the hybrid automata model and normal human being walking.

This chapter describes human push recovery data classification using features that are obtained from intrinsic mode functions (IMF) using empirical mode decomposition (EMD) on different leg-joints angle (Hip, Knee and Ankle). The joints angle data were calculate for both open eyes and close eyes subjects. The classification was performed based on these different kinds of the pushes using deep neural network (DNN) and the overall 89.28% accuracy was achieved. We have also used the Artificial Neural network (ANN) based on feed forward back-propagation neural network (FF-BPNN) and compared with the DNN. The statistical analysis tool ANOVA (Analysis of Variance) is also conducted to show the statistical significance of results. The corresponding strategies (Hip, Knee and Ankle) can be utilized once the categories of pushes (Small, Medium, Moderately High, High) were identified accordingly push recovery.

## 6.2 Introduction:

To develop the computational model based on the joint trajectory value we need features. To move towards more generic model, we need the algorithms to automatically find the 'interesting' features that disentangle the data [90]. Deep learning is a step towards identifying these 'interesting' representations. The goal is to automatically identify higher level features from low level features. Deep learning is used for the training of deep architectures such as Multi-layered perceptron which has several hidden layers [91]. Deep architecture consists of multilevel non-linear operations that transform the input data to better represent [92]. Researchers also worked for the recognition of gait for different type of walk [93] [94].
We have analyzed the push recovery pattern. The classification was performed using features (Min, Max, RMS, Shannon entropy, Log Energy, Entropy, Zero crossing rate) obtained from IMF by performing empirical mode decomposition (EMD)[93]. This analysis will help in the development of artificial limbs and development of social robots [94]. Then the collected data is classified using the ANN and DNN based classifier and after forming the 5 fold cross validation, it is obtained that DNN based proposed classifier is better than ANN. The ANOVA (Analysis of Variance) has also been performed to show the statistical significance of results [95].
We have also presented the gait data classification using different machine learning technique ANN [96], KNN [100] and K-Mean [101]algorithm and analyzed classified data. The human walk is coordinated using regular periodic motion of upper or lower body extremity; this is responsible for unique locomotion of individual. It is considered very difficult to disguise and conceal the Gait based biometric identification system. In this chapter we first selected the principle feature using KPCA then based on these feature we classified gait data into five different gait data named normal and four types of crouch and applied the ANN machine learning technique and finally point out the performance comparison.

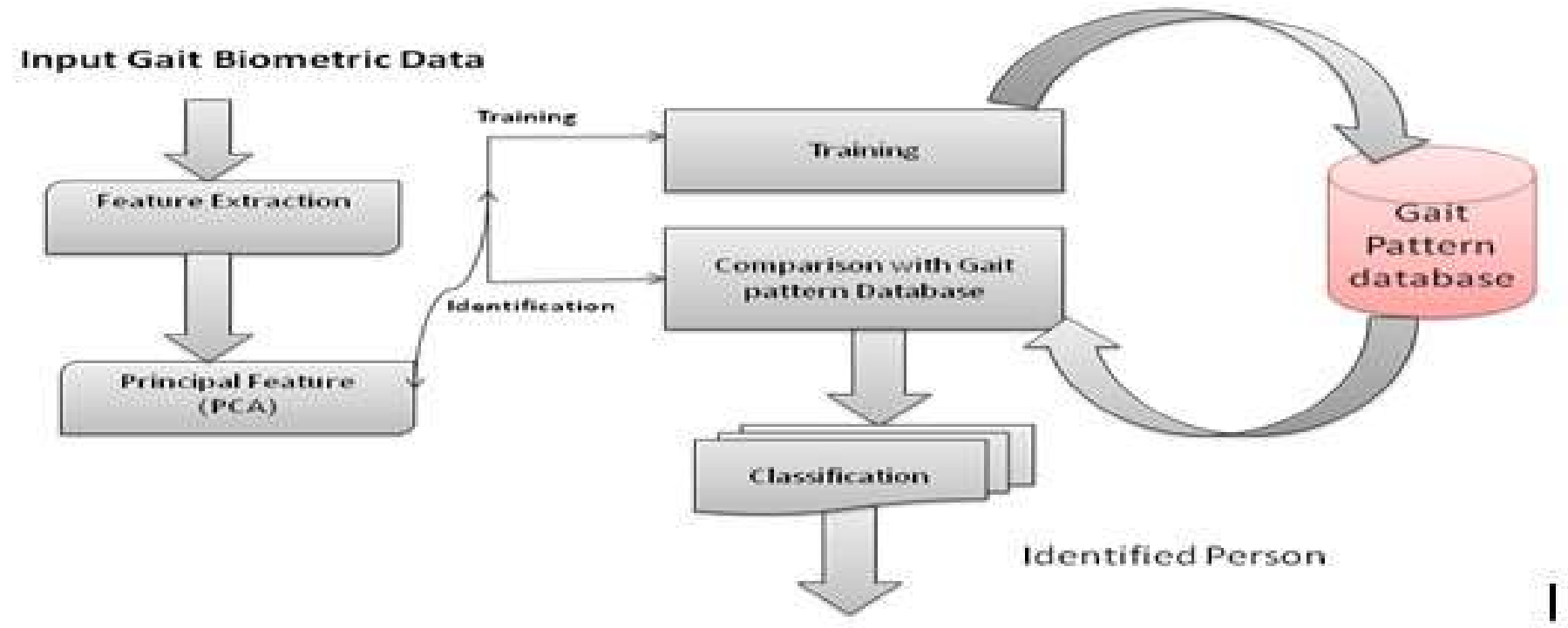

**Figure 6-1: Gait Biometric based identification design**

Fig.6.1 is the block diagram of entire process of Gait biometric identification system.

## 6.3 Methodology: Data capturing and feature extraction

Data capturing technique changes with the available modern device in the field of research. Earlier we used to use potentiometer and body sensor based Human Motion Capture Device (HMCD) till 2014. The extracted data from that device used to take several purifying techniques to get noise free useful data. The data capturing technique proposed here is innovative and reliable. The software, named "physics tool box" powered by Google enable us to represent an accelerometer sensor that works on android interface. The fig.6.2 shows the entire process of data collection. For realistic push recovery data collection the subjects were selected from different age group, weight and height. Four different kinds of pushes were applied with a known range of intensity from behind. The pushes are named as Less, Medium, High and Moderately High push. The range of less push starts with the zero Newton. The highest value of Moderately High push is the maximum value after that recover is not possible means zero sensing value. The data were collected for both left and right handed. Force sensing resistor (FSR 3105) is used to measure the reading of the applied push on the back between spinal cord and the last rib of subjects. The measured value is then converted into Newton using following formula.

$$Force\ (\ Newton) = f * 9.8/100 \qquad (10)$$

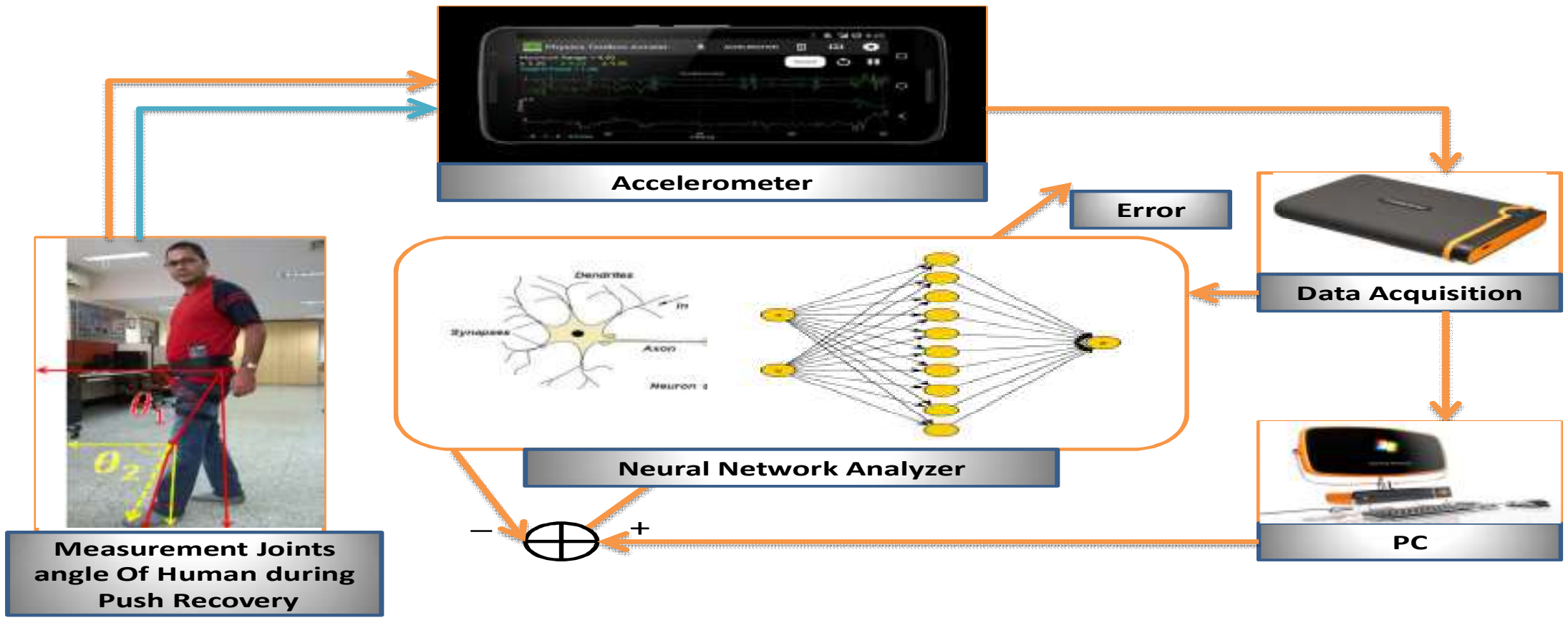

**Figure 6-2: Extracting Feature using accelerometer**

The unknown forces were applied to formulate the supervised learning with the main experimental data set. The push was applied up to few seconds to collect the data. With the known range of push the data set is arranged separately for performing the next experiment. We have modeled the leg as two link robot similar to 2-link planer manipulator. Fig.6.3 shows the representation of human leg as 2-link manipulator. The android embedded sensor gives the output in a dot comma separated values (.csv) file. The inverse kinematics was solved to get the joint angles (hip, knee& ankle).

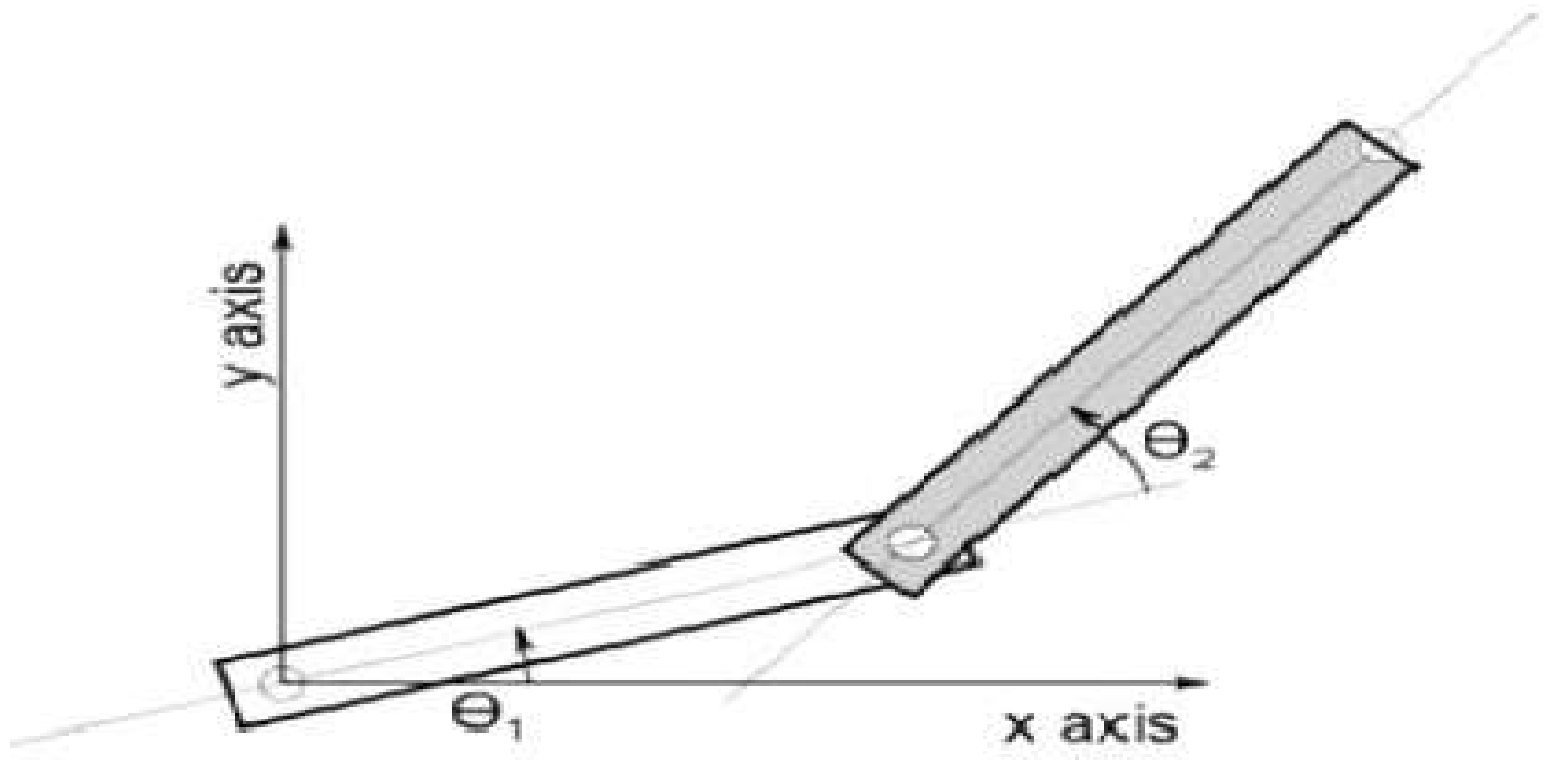

**Figure 6-3: Diagram of a two link manipulator**

## 6.4 Joint Angle Calculation

Using a simple model of 2-link planer manipulator for modeling accelerometer mounted on knee, we tried to calculate joints angles from the accelerometer reading of a mobile phone by solving inverse kinematics using Atan2 function. The equations used are given in 11 and 12.

$$\theta_1 = atan2(y, x) - anta2(k_2, -k_1) \qquad (11)$$

$$\theta_2 = atan2(\pm\sqrt{1-(\frac{x^2+y^2-l_1^2-l_2^2}{2l_1l_2})^2}, \frac{x^2+y^2-l_1^2-l_2^2}{2l_1l_2}) \qquad (12)$$

Where

$$k_1 = l_1 + l_2\cos\theta_2 \qquad k_2 = l_2\sin\theta_2$$

Here $l_1$ represents the length of the first link, and $l_2$ is the length of the second link, in this experiment these are Femur and Fibula is shown in fig.6.3. Inverse kinematics tells all the possible sets of joint angle and the geometries of link which can be used to get the orientation and position of the end effectors from the given orientation and position of the link manipulator.

The forces were applied in four different ways, i.e. Small, Medium, Moderately high and high from behind. FSR 3105-Force sensing resistor was used to measure the intensity level of push. Here 0 to 3 Newton (unit of force) is treated as less push, 3 to 6 Newton as Medium push, 6 to 9 as moderately high push and 9 to 12 Newton as High push beyond this value our subject could not recover from the push. This sensor is also used to measure ZMP based stability of biped robot. The extracted joint angles are then used in EMD process to get the intrinsic features.

## 6.5 Empirical Mode Decomposition

Empirical mode decomposition (EMD) is a very useful method for breaking down the data into finite and a very small number of components. EMD filters out the data set from a complete and orthogonal basis. The obtained component after doing decomposition is called IMF that is intrinsic mode function [102].

IMF has following two properties:-

1) Between zero crossings, it has only one extreme

2) It has zero mean

The Algorithmic 3 shows how to calculate the IMF features of data.

| Algorithm3: Shifting procedure |
|---|
| Begin<br>Identify all extrema of $\theta(t)$<br>Determine the upper envelop $u(\theta)$ from its local maxima<br>Determine the lower envelop $l(\theta)$ from its local minima<br>Obtain the mean envelop, $m(t) = [\, u(\theta) + l(\theta)]/2$<br>Subtract the mean from $\theta(t)$ , $h(t) = \theta(t) - m(t)$<br>Check whether $h(t)$ satisfies the properties of IMF<br> If (Yes)<br>  The obtained $h(t)$ is IMF<br>  Stop shifting<br>  Break<br> Else<br>  $\theta(t) = h(t)$<br>  Keep shifting<br>  Continue<br>End |

Empirical mode decomposition (EMD)is a very useful method for breaking down the signal into finite and very small number of components. For the analysis of natural signal this process is very useful, which are most of the cases non-stationary and non-linear [103].

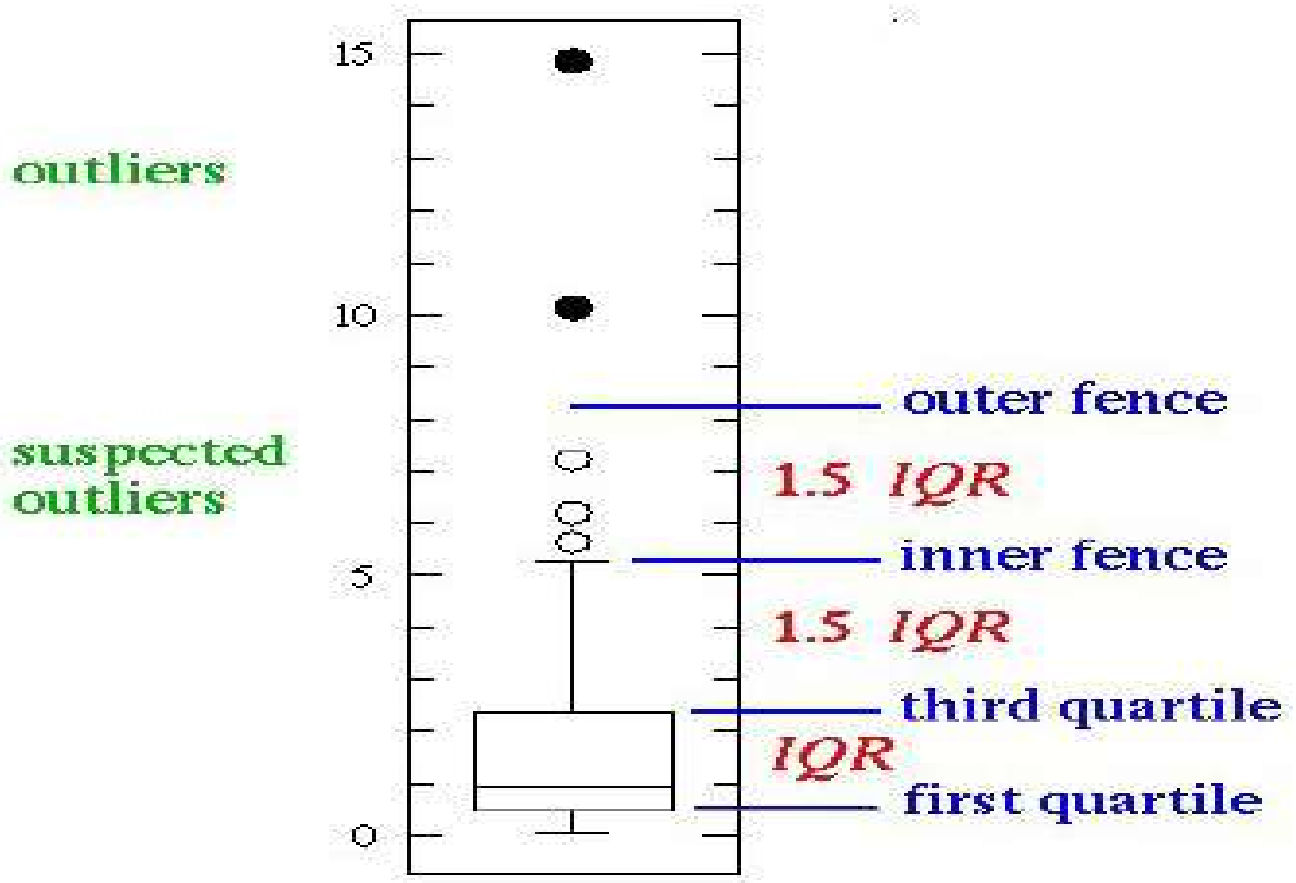

**Figure 6-4: Generalize diagram of box plot**

### 6.5.1 The Distribution of Data:

One popular technique is box plot that graphically shows the distribution of a group of numerical data through their quartiles. The quartiles are the three points of data set that divide the data set into four groups. The middle number between the median of the data set and the smallest number is called first quartile, and the median of the data set is called the second quartile, the middle number between the highest value of data set and the median of the data set is called third quartile. The middle number between third quartile and the first quartile is called inter quartile rang (IQR). The data in the range of more than three times of the inter quartile range is called outliers, and data in the range of more than 1.5 times of the inter quartile range are called suspected outliers. Fig 6.4 shows the Generalize diagram of box plot.

An IMF corresponding to every joint angle has this kind of data distribution. If the data values are well synchronized, then that data set will be considered as best data set among all. Fig.6.5 illustrates the distribution of corresponding IMF. The next step is data classification. The classification is needed to get the exact result or the exact class for any unknown pattern as input from that analysis the robot configuration will be upgraded.

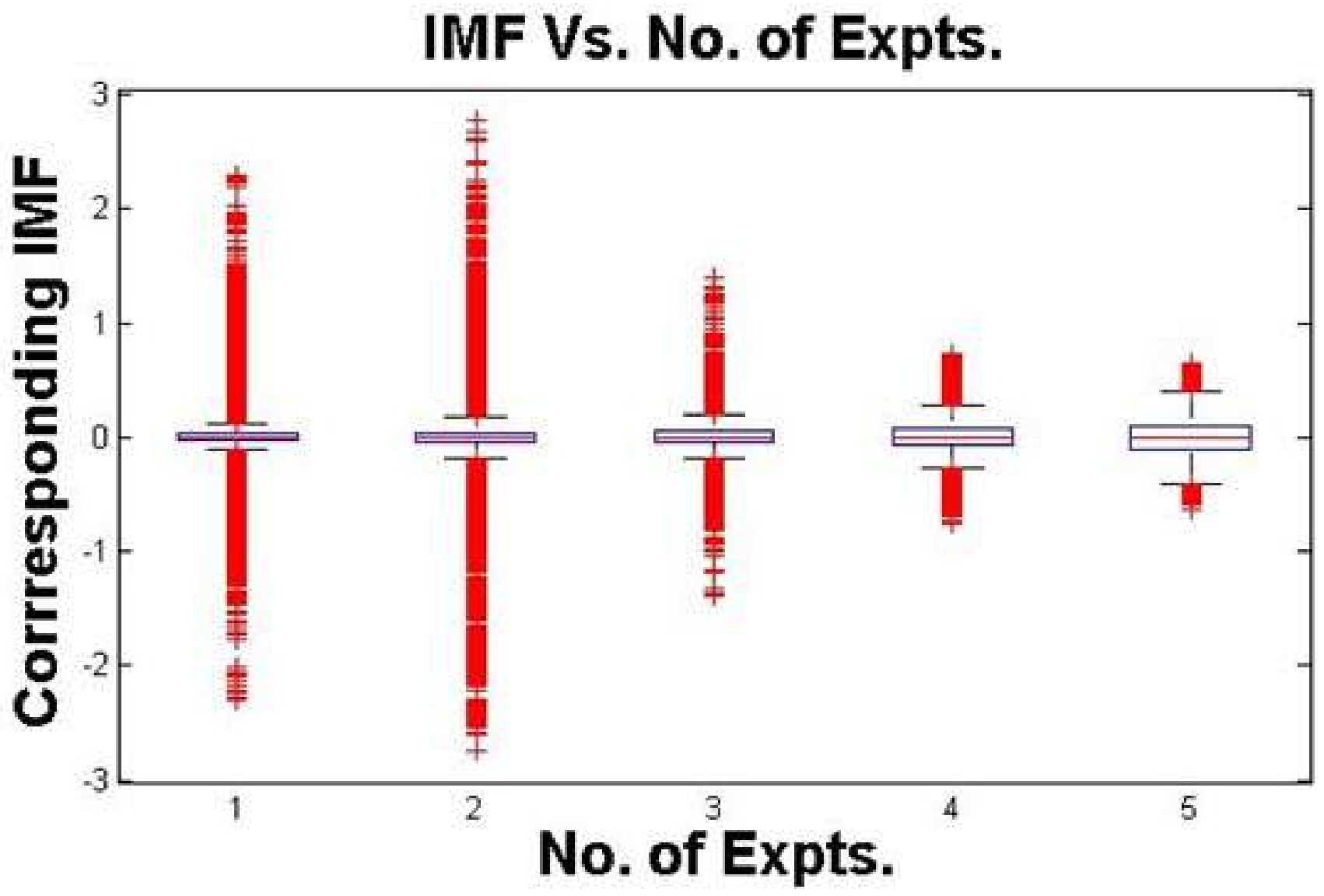

**Figure 6-5:Box plot of corresponding IMF**

| | A | B | C | D | E | F | G | H | I | J |
|---|---|---|---|---|---|---|---|---|---|---|
| 1 | 0 | -0.00033 | 0.000114 | 0.000416 | 0.000763 | 0.000377 | -0.00075 | 0.000695 | -0.00063 | 0.000568 |
| 2 | 0 | -0.00016 | 0.000304 | 7.72E-05 | -0.00055 | -0.00066 | 0.000199 | 0.000844 | -2.11E-05 | -0.00098 |
| 3 | 0 | -0.00033 | -0.00011 | 0.000423 | 0.000989 | 0.001362 | 0.00145 | 0.001261 | 0.000815 | 0.000124 |
| 4 | 0 | -0.00018 | -0.00023 | -0.00018 | -5.53E-05 | 0.00012 | 0.000319 | 0.000516 | 0.000686 | 0.000802 |
| 5 | 0 | -0.00034 | -0.00054 | -0.00061 | -0.00057 | -0.00044 | -0.00025 | -4.72E-06 | 0.000274 | 0.000567 |

**Figure 6-6:Corresponding IMF of every joint angle**

## 6.6 The Statistical Feature selection

The selection of feature is very important for a particular data set. The machine learning algorithm performs based on appropriate statistical feature. Therefore, good quality of feature always gives the best result. In equation 13, $d_{min}$ is the minimum value of a particular data set.

$$d_{min} = p \in x_i \quad (13)$$

And in equation 14, $d_{max}$ is the maximum value of a particular data set.

$$d_{max} = q \in x_i \quad (14)$$

The next selected feature is Shannon entropy, which is the most important feature of a particular data set. In Information theory, the entropy is used to measure the uncertainly. The more uncertainty means more information. Entropy always categorizes the amount of uncertainty associated with the value of a variable when only its distribution is unknown. The Shannon entropy shown in equation 15 gives the average information of the data set:

$$H(X) = -\sum_{i=1}^{N} p(s_i)\log_2 p(s_i) \qquad (15)$$

The Log Energy Entropy of IMF is calculated as follows:

$$Energy = \sum_{i=1}^{N} \log(s_i^2) \qquad (16)$$

The quadratic mean in statistics is known as the root mean square(RMS). It is the statistical data that defined square root of the mean of squares of a sample.

$$f_{rms} = \lim_{T\to\infty} \sqrt{\frac{1}{T}\int_0^T f(t)^2\, dt} \qquad (17)$$

Another important feature is a zero crossing rate(ZCR). It is the number of the sign changes of a signal along a particular axis. It is shown in the following equation:

$$ZCR = \frac{1}{T-1}\sum_{t=1}^{T-1} \prod (s_t s_{t-1} < 0) \qquad (18)$$

Where T is the length of the signal s. $\prod(A)$ is called the indicator function. So the indicator function is 1 if its argument A is true and 0 if A is false. In many experimental cases instead of all crossing only the positive going and negative is counted. In our experiment this feature is also important because of the gait cycle that oscillatory in nature.

**Feature Extraction:**

Extraction of unique feature always helps to enhance the classification performance. In this study, we have used six basic statistical measurement of the signal and found unique patterns among the Users. As described earlier Empirical mode decomposition methodology for joint angle signals to split it into number of intrinsic mode, we have extracted statistical features of the IMFs. After decomposition of original signal into different number of IMFs we have extracted the statistical features of all IMFs to give the input to classifier for subject identification. The algorithm 4 is based moving average filter in which we are calculating the least square error up to=.0001 and number of step initial length/2. The times series data is input and it will return the smooth data.

| Algorithm4: Moving Average Filter algorithm |
| --- |
| **Input**: InputData (x[n]), Length OfData ($l_e$), RootMeanSquare(rms)<br>**Output**: SmoothedData($S_1$[n]); CalcRootMeanSquare(rms1)<br>Initial: $l_t \leftarrow x[n]$; rms1 = rms; |
| Begin |

```
l_e = Length(x[n])
 x = (x(1: l_e − 2) + x(2: l_e − 1) + x(3: l_e))/3;
      rms1=rms(x)
While rms > 0.0001 & length(x)>l_e/2
                    x = (x(1: l_e − 2) + x(2: l_e − 1) + x(3: l_e))/3;
     //                                                          where
x is value, we have performed the moving average with window size 3
       rms = rms(x)
End while
   x= Imresize (x,l_e);
End Begin
```

The performance of machine learning algorithms depends on the several factors. The selection of accurate features from right problem is very important for better performance. We need the algorithms to automatically find the 'interesting' features that disentangle the data [89]. Deep learning is a step towards identifying these 'interesting' representations

### 6.7 Deep Learning Process Model:

Multilayer Feed forward Neural Networks (MFNN) designed for the problem uses multiple neurons interconnected in basic format of input layer, multiple hidden layers and an output layer respectively. The basic functionality of multi-layer neural networks is the same as that of a traditional one hidden layer neural network [20]. The extracted features are fed into the input layer, each of the hidden layers processes the features and generates a set of projected features which are more compact and represent better attributes to classify the data. The final layer is the output layer which processes the data from the last hidden layer and classifies the data into one of the several possible classes.

In multi layer back propagation based neural network with more hidden layer lead for poor learning rate. As if the depth of network increases the error gradients become very small for back propagation and diminish eventually. So it does not contribute the starting layer of the network. Due to poor learning of network it leads to under fitting and as we increase the size of network the parameter i.e. bias and weight also increases. So we explore the a space of very complex function, which needs large amount of data for training else it will lead for over fitting (curse of dimensionality) .The positive results are obtained only up to 2 or maximum 3 layers. More than 3 layers the network will give poor results. Then in 2006 Hinton et al. introduced Deep Belief Networks (DBNs) which was trained greedily, i.e one layer at a time in an unsupervised manner and with unlabelled data using the Restricted Boltzmann machine (RBM). Shortly after, another method based on auto-encoders was designed.

**Auto Encoder:** The fig 6.7 shows the architecture of the auto encoder. An auto encoder neural network is an unsupervised learning algorithm that applies back propagation, setting the target values to be equal to the inputs. It works best when the training data is unlabeled. The auto encoder tries to learn a function$H_{W,b}(x) \approx x$ . In other words, it is trying to learn an approximation to the identity function, so as to output x output that is similar to x input. The human push recovery data is unlabeled so we have opted DNN.

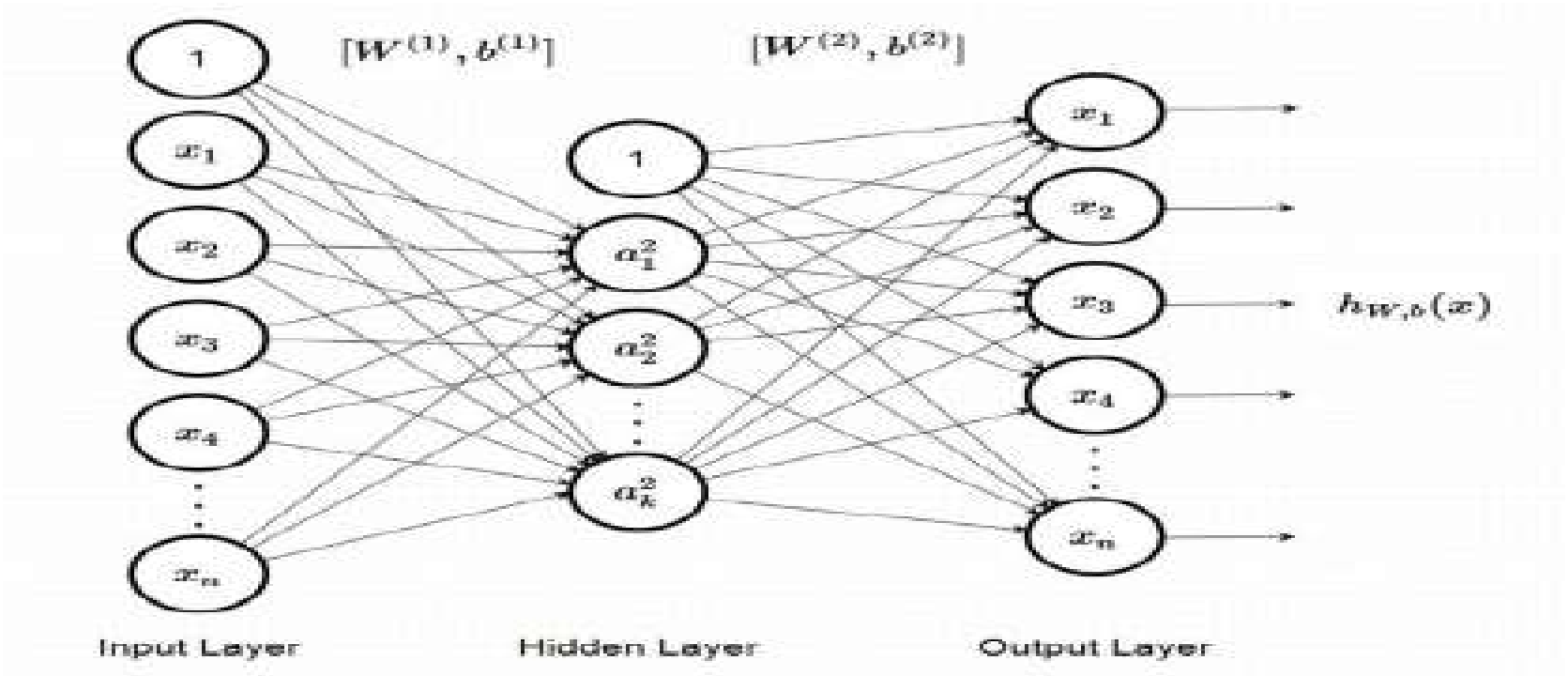

**Figure 6-7: Auto Encoder**

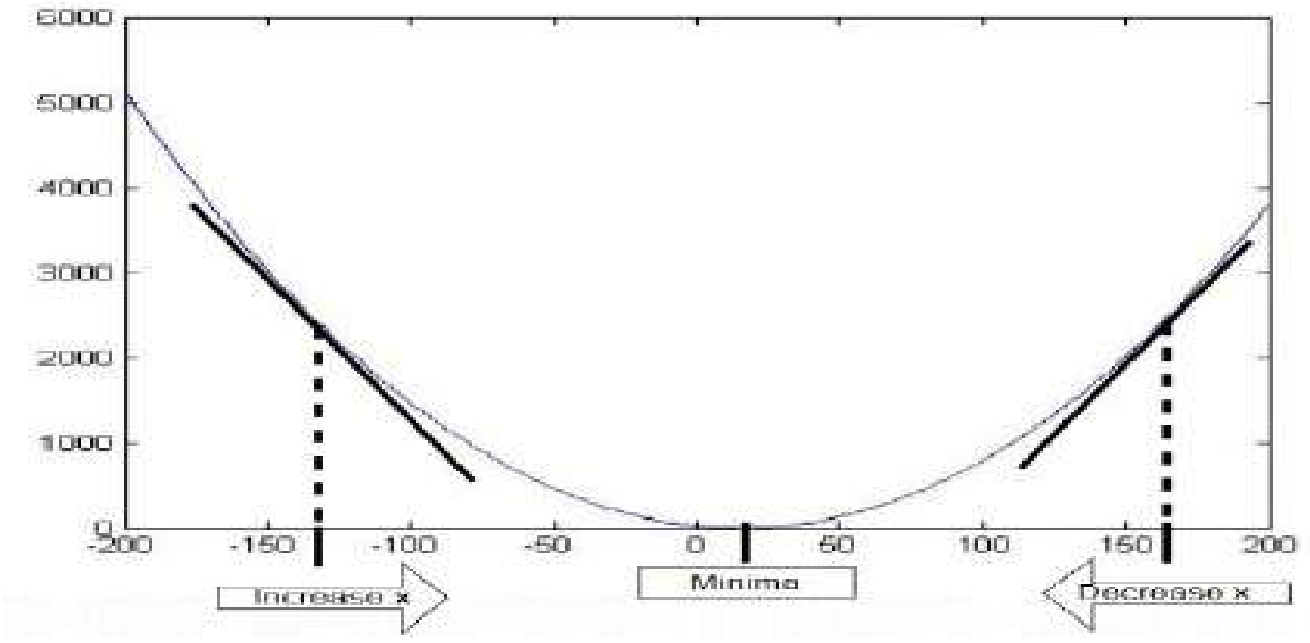

**Figure 6-8: SGD (stochastic gradient descent) Process**

### 6.7.1 SGD (stochastic gradient descent)

One of the fundamental differences between deep learning and a general multi-layered neural network is learning. In deep learning the initial training is done layer-by-layer, starting from the first hidden layer to the last, ensuring that each layer represents vital features useful for classification. The weight adjustment and learning rate are adjusted with bias $b$ value. We need to minimize the errors equation 19 representing the loss function. Let's assume $\omega$ is collection of weight matrix$\{\omega_i\}, 1 \leq i \leq N$, where $\omega_i$ represents weights connecting $i$ layer to $i+1$ and $N$ represents the number of layers;

similarly $\beta$ a collection of biases $\{\beta_i\}, 1 \leq i \leq N$, where $\beta_i$ is collection of biases at layer $i$.

$$\Lambda(\omega, \mathrm{B} \mid j) \qquad (19)$$

Deep learning is hierarchical feature extraction architecture for multilayer basic framework having a major advantage that it can handle nonlinearity in data for training and testing purposes. We use stochastic gradient descent (SGD) which is memory-efficient and very fast with HOGWILD for supporting shared memory model and therefore training and testing data is able to execute on multimodal system. The general procedure is shown in Algorithm 5. The architecture of deep learning is shown in fig.6.9 .The general procedure of using DNN is shown in Figure 6.10.

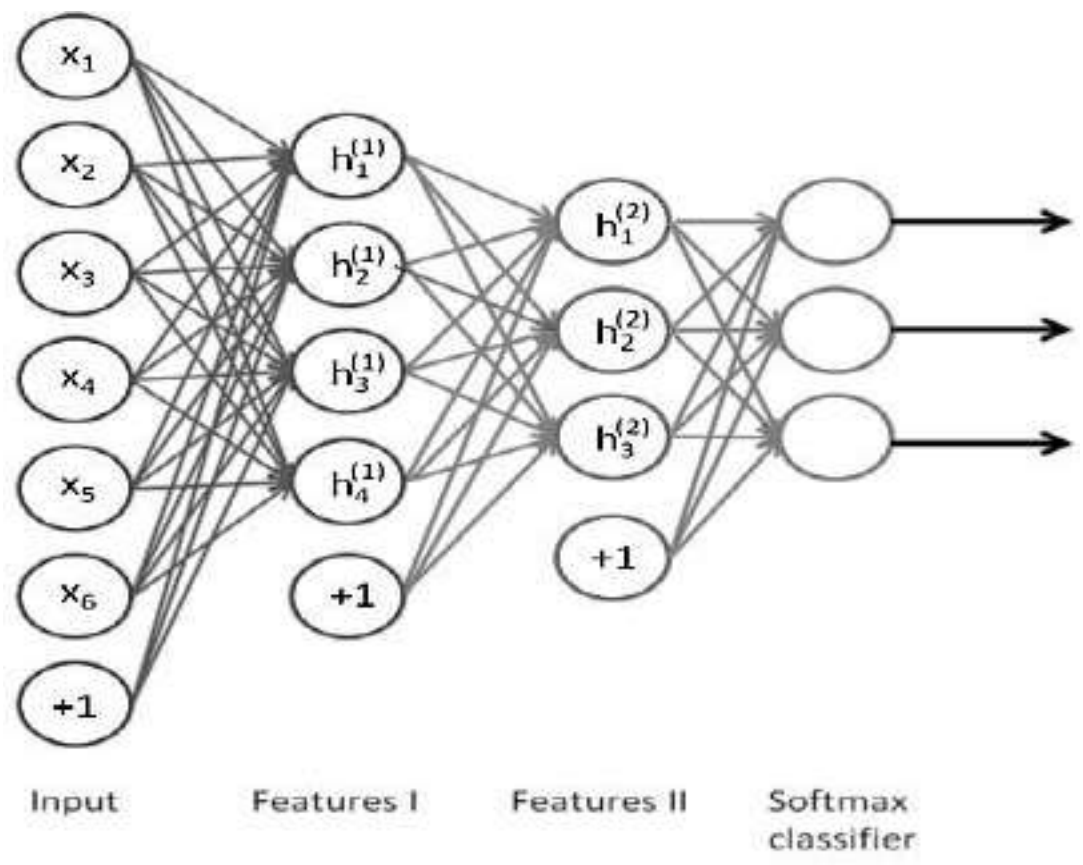

**Figure 6-9: Deep Learning Architecture**

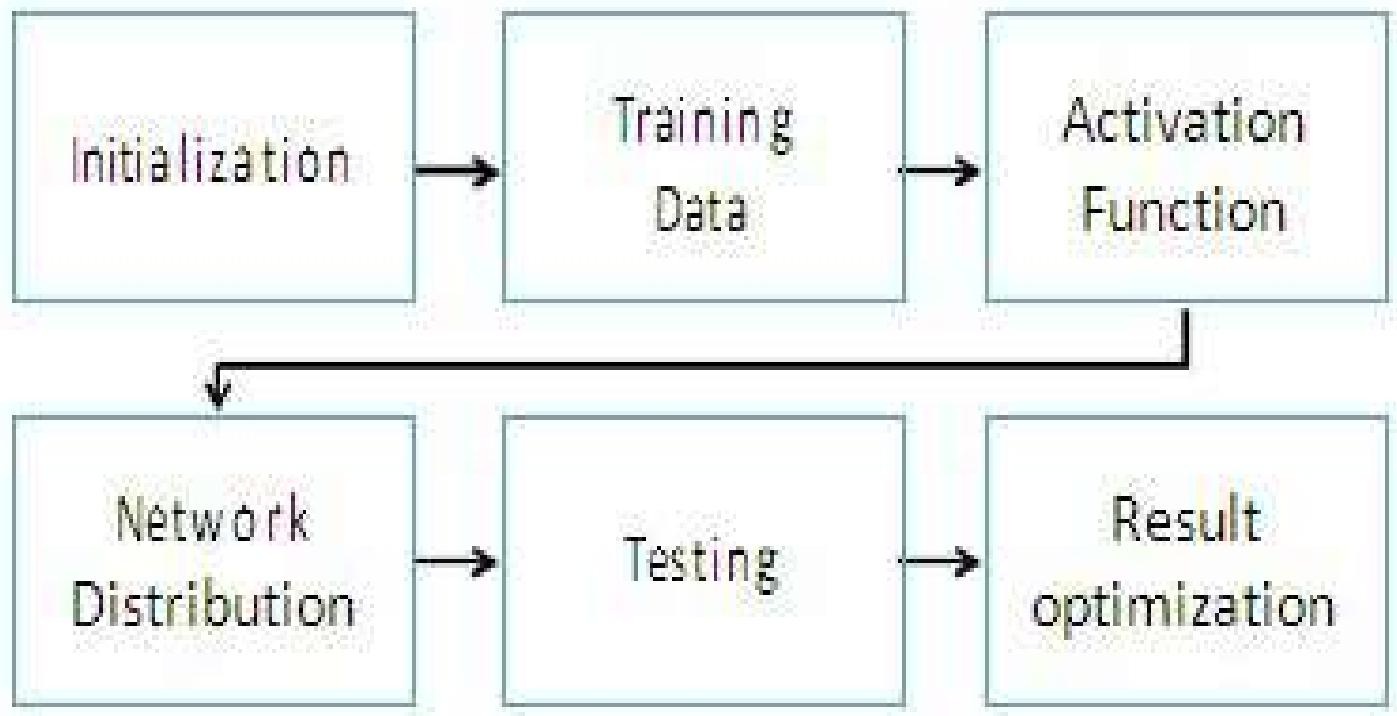

**Figure 6-10: Deep Learning Process Model**

### 6.7.2 Deep Learning Process Model

The classification is then performed using these statistical features using deep neural network (DNN). The network we proposed here having 5 hidden layer with 100, 50, 25, 10, 10 neurons in it. The input dropout ratio is set to 0.2 and hidden dropout ratio is set to 0.5 for every layer. The activation function is used here is tanh(x). The network is trained in supervised manner, because the pushes that were applied with known range of force parameter [24]. The training of the deep neural network architecture is done by gradient decent method in equation-20. The back propagation algorithm didn't help because it doesn't work when the number of hidden layers is large. As the depth of the network increases the error gradient that is back-propagated become very small and diminish eventually.

$$\Delta\omega_{ij}(t+1) = \Delta\omega_{ij}(t) + \eta\frac{\delta C}{\delta\omega_{ij}} \quad (20)$$

Where $\eta$ learning rate and C is cost function. The cost function is chosen based on learning type (supervised or unsupervised) and the activation function.

Algorithm 5: **Initialize**$W, B$ for h data load training data $T$ for all nodes not converge criteria satisfied for each node $n$ training$T_k$, do parallel

$\omega_n, \beta_n \leftarrow$ Global model (seed value, will update after)

Select$T_{k\alpha} \subset F_k$ ($F_k$ is samples for all iteration)

Divide $T_{k\alpha}$into $T_{k\alpha c}$ from $k_c$ (number of cores$k_c$)

For $T_{k\alpha c}$ on node $k$, do parallel

Initialize training $i \in T_{k\alpha c}$

Update $\omega_{ij} \in \omega_n$ and biases $b_{ij} \in \beta_n$

$$\omega_{ij} := \omega_{ij} - \alpha \times \frac{\delta L(w,\beta|j)}{\delta\omega_{ij}}$$

$$b_{ij} := b_{ij} - \alpha \times \frac{\delta L(w,\beta|j)}{\delta b_{ij}}$$

$$W := Average_k \times W_k$$

$$B := Average_k \times B_k$$

## 6.8 5- fold cross validation:

It is a technique for validation. In the next step, 5-fold cross validation was performed to analyze the validity and performance of the classifiers. During the 5-fold cross validation process, the total dataset was divided into 5 equal parts. Among the total 5 parts of the dataset, one part will be used as testing set and remaining will be used as training set in every phase of the experiment. Thus, a total of 5 experiments will be performed and accordingly 5 results will be generated. Finally, the average of the 5 results will provide us the final accuracy of the classifier used in our system.

---

Algorithm 6: 5-fold cross validation

---

Begin

Place the training sample into some random order.

Divied the smaples into 5 fold. i. e. 5 chunks of approminately n/5 Size each

for i = 1: 5

Train the all sample which do not belong to $i^{th}$ fold

Test the classifer on all the smaple of $i^{th}$ fold.

Compute the $n_i$, no of smaples which are misclafied into $i^{th}$ fold.

end for

Compute $E = \frac{\sum_{i=1}^{5} n_i}{n}$

To achieve the good accuracy of classifier, the 5 − fold cross validation is run many times.

let $E_1, E_2, \ldots, E_p$ be the accuracy estimate obtained in p run

$$e = \frac{\sum_{j=1}^{p} E_j}{p}, V = \frac{\sum_{j=1}^{p} (E_j - e)^2}{p-1}, \sigma = \sqrt{v}$$

Where is e estimated error and σ is stnadar deviation

End Begin

---

**Multi-layer neural networks:** The multi layer neural network [34] consists of an input layer, a hidden layer, and an output layer. Multi-layer neural networks can have several output units. The units of the hidden layer unction as input units to the next layer. However, multiple layers of linear units still produce only linear functions. The step function in perceptions is another choice, but it is not differentiable, and therefore not suitable for gradient descent search. The solution is the sigmoid function, a non-linear, differentiable threshold function. Fig.6.11 is the model of our multi layer propagation ANN. PCA used for selection of major feature. The units of the hidden layer function as input units to the next layer. However, multiple layers of linear units still produce only linear functions. The step function in perceptions is another choice, but it is not differentiable, and therefore not suitable for gradient descent search. The Solution: the sigmoid function, a non-linear, differentiable threshold function. Figure 6.11 is the overall model of back propagation based neural network.

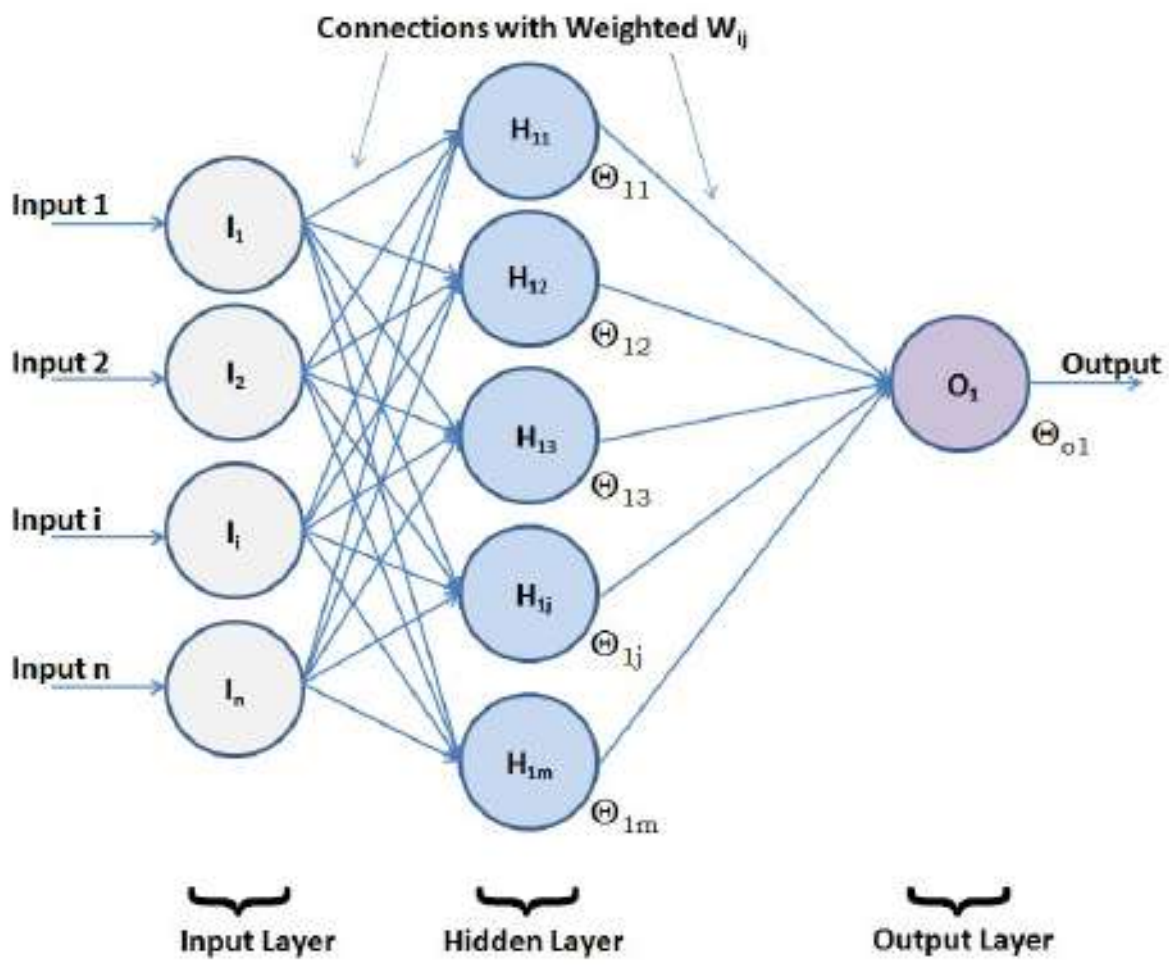

**Figure 6-11:Multi-Layer Back Propagation ANN**

### 6.8.1 Algorithm for human gait classification

1. Input is fed to the system as a feature of different gait.
2. Select the principal component, which are used for further processing.
3. Reduce the dimensionality.
4. Apply various technique for classification ANN,KNN,K-mean
5. KNN and K-mean used for classification; based on the similarities in the features.
6. The features depend on the following criteria: Walking, Jumping, Jogging and Running.
7. Compare the all type of gait

Artificial Neural network (ANN) outperforms the K-nearest neighbor (KNN), K-mean and other existing methods for classification. The algorithm for started with initializing weight of all node, then select the data point and calculate the output for each point, computer the error and propagate it back. Finally update the weight and keep this loop run till error should be not be below threshold.

Algorithm 7: Back propagation Algorithm Neural Networks

**Gradient-descent** (I, Iteration Count)

**Initialize** all weights

**For** i=1:1: Iterations Count

**do**

**select** a data point $D_k = [i, j]$ from matrix D

**set** learning rate α € [0,1]

**calculate** outputs o (p ) for each data point

**calculate** Error $\delta_{ij}$ using back propagation

**update** all weights (in parallel)

$\mathbf{w_{ij}} \leftarrow \mathbf{w_{ij}} - \boldsymbol{\alpha} * \Theta_{ij} * x_j(k-1)$

**end for**
**update** all threshold $\Theta_{ij}$ (in parallel)
**return** weights **w**
**end**

### 6.9 Results and Discussion

Ideally the human joint angle oscillates with respect to the time when the body in motion. The joint angle (knee) obtained from the fundamental experiment is different than ideal one. Researchers already worked on the analysis of variation in leg joint angle in different environment using different data capturing technique. The idea is proposed here is totally innovative and reliable. The all confusion matrix obtained from the experiment as output is shown below in fig6.12:

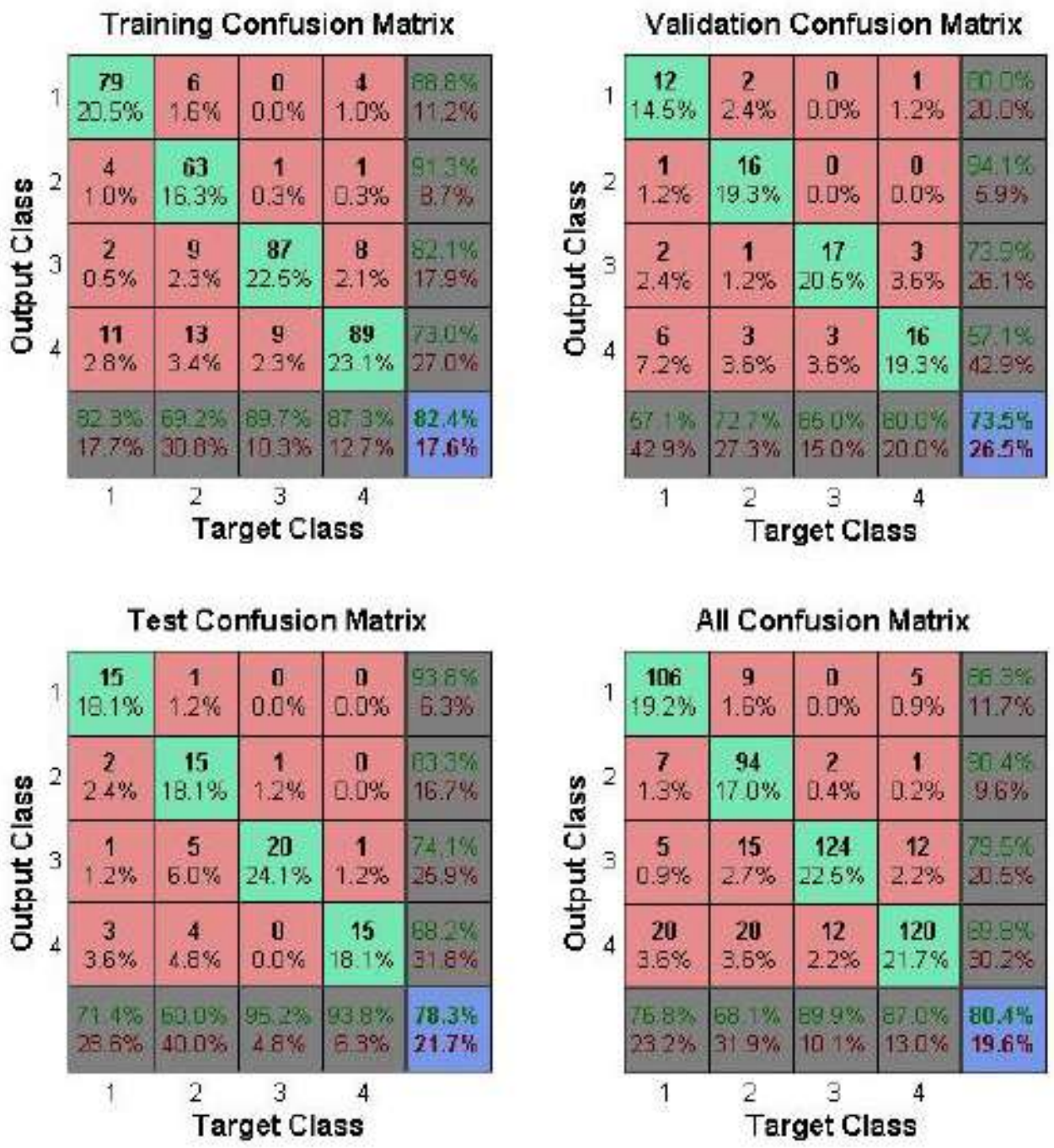

**Figure 6-12: Confusion Matrix**

### 6.10 Confusion Matrix

The accuracy of the individual class is calculated using following equation:

$$Accuracy(ACC) = \frac{TP + TN}{TP + FP + FN + TN} \qquad (21)$$

**Table 6:1: Accuracy of the Individual Class**

| | True Positives (TP) | False Positives (FP) | False Negatives (FN) | True Negative (TN) | Accuracy (ACC) |
|---|---|---|---|---|---|
| small class | 99 | 39 | 38 | 360 | 85.63% |
| Medium class | 90 | 48 | 6 | 389 | 89.86% |
| Moderate High class | 128 | 10 | 33 | 381 | 92.2% |
| High class | 106 | 32 | 52 | 362 | 84.7% |
| overall accuracy | 76.63% | | | | |

Table 6.1 is the accuracy of the individual class. The Table 6.2 corresponds to classification accuracy obtained using ANN on different chosen parameters. Table 6.3 is the result of performance using ANN using different no. of neuron and epoch and table 6.4 is Performance Matrix using Deep Neural Network.

**Table 6:2: Overall Confusion matrix of ANN using different no. of neuron and epochs**

| Hidden Neurons | Epoch | Training Accuracy | Testing Accuracy | Overall Accuracy |
|---|---|---|---|---|
| 30 | 20 | 82.4% | 78.3% | 80.4% |
| 20 | 15 | 80.6% | 75.9% | 76.6% |
| 15 | 10 | 79.9% | 74.3% | 75.5% |
| 10 | 10 | 79.0% | 74.0% | 75.0% |

The classification results are tabulated in Table 6.2 using 5 fold cross validation for ANN and DNN classifier.5 fold cross validation result using artificial neural network (ANN) classifier. The Table 6.3 corresponds to classification accuracy obtained using DNN on different chosen parameters and Table 6.4 is the Comparison of success rate by different classifier using 5 –fold cross validation using deep neural network (DNN) classifier using chosen parameter

**Table 6:3: Performance Matrix using Deep Neural Network**

| Hidden Neurons | No. of Hidden layer | Epoch | Training Accuracy | Testing Accuracy | Overall Accuracy |
|---|---|---|---|---|---|
| 150,100,60,40,20 | 5 | 20 | 90.6% | 85.71% | 87.4% |
| 100,60,40,20 | 4 | 15 | 87.4% | 80.14% | 82.2% |
| 60,40,30 | 3 | 10 | 92.4% | 85.71% | 89.28% |
| 50,30,20 | 3 | 10 | 89% | 82.14% | 84.28% |

**Table 6:4: 5 fold cross validation result using**

| Exp# | Accuracy (%) |
|---|---|
| Subset 1 | 90% |
| Subset 2 | 87% |
| Subset 3 | 86% |
| Subset 4 | 89% |
| Subset 5 | 90% |
| Overall | 88.4% |

**Table 6:5: Comparison of success rate by different classifier using 5 –fold cross validation**

| Classifier | No of Sample | No of Misclassified | Success rate (%) |
|---|---|---|---|
| ANN Classifier | 1000 | 170 | 83 % |
| DNN Classifier(Proposed) | 1000 | 110 | 89% |

From results in Table 6.5 it can be concluded that, the proposed technique achieves the highest classification rate of 89% with only 130 misclassifications.

Table 6.6 shows the result obtained after performing the one way-ANOVA (Analysis of variance). The ANOVA is performed on the result obtained from 5 fold cross validation of both classifier. As here only2classifier so t-test and ANOVA will give same results. The DNN based classification is compared with ANN in order to see where the difference is significant enough to reject hypothesis (H1). The null hypothesis will be (H0) there is no statistical variance between DNN and ANN classifier (H0). To verify the statistical significant improvement in hypothesis H1 an ANOVA (analysis of variance) is no statistical variance between DNN and ANN classifier (H0) whereas alternative hypothesis there is significant difference between classification accuracy (H1).

### 6.11 ANOVA (Analysis of Variance):

The value of p =.67 indicates that one should reject the null hypothesis in favor of the alternative. It shows there is signifies difference as null hypothesis has rejected (H0) and alternative hypothesis accepted (H1).

ANOVA analysis details:

**Table 6:6: ANOVA: Single factor Group variation**

| Group | Count | Sum | Average | Variance($\sigma$) |
|---|---|---|---|---|
| ANN Classifier | 5 | 395 | 79 | 6.5 |
| DNN Classifier(Proposed) | 5 | 434 | 86.8 | 3.3 |

| Source of variation | Sum of Square | Degree of freedom | MS | F | p-value | f-critic |
|---|---|---|---|---|---|---|
| Within | 20.43 | 1 | 20.43 | .89 | .67 | 4.2 |
| Between | 182.23 | 8 | 22.77 | | | |
| Total | 202.66 | 9 | | | | |

### 6.12 Performance Analysis of Gait Data:

This research reveals about the main gate classification approach. Here we performed the classification by the different machine learning technique i.e KNN and K-mean for GAIT data [46]. The performance table shows that the ANN is better performing technique [47].

We used OpenSim data set for Gait of four categories Normal, Crouch2, Crouch3 and Crouch4. We used 30 samples of each category of data for training and 20 data set points for testing. D = {Normal, Crouch2, Crouch3, Crouch4}, D€ $R^d$ , D dimension data set of each class and training data 30 data point and testing data point 20.

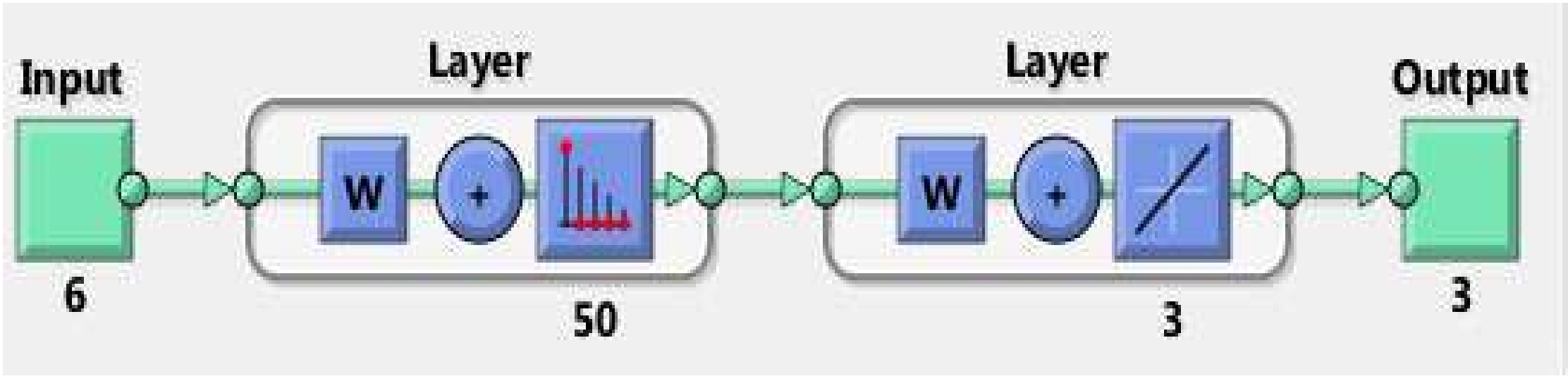

**Figure 6-13: Multi-Layer Back Propagation ANN**

Figure 6.14 is the performance curve which is related to training of model, when the mean square error will reach up to threshold in which no of epoch. As we set the threshold .001. We achieved the target in 11 epochs.

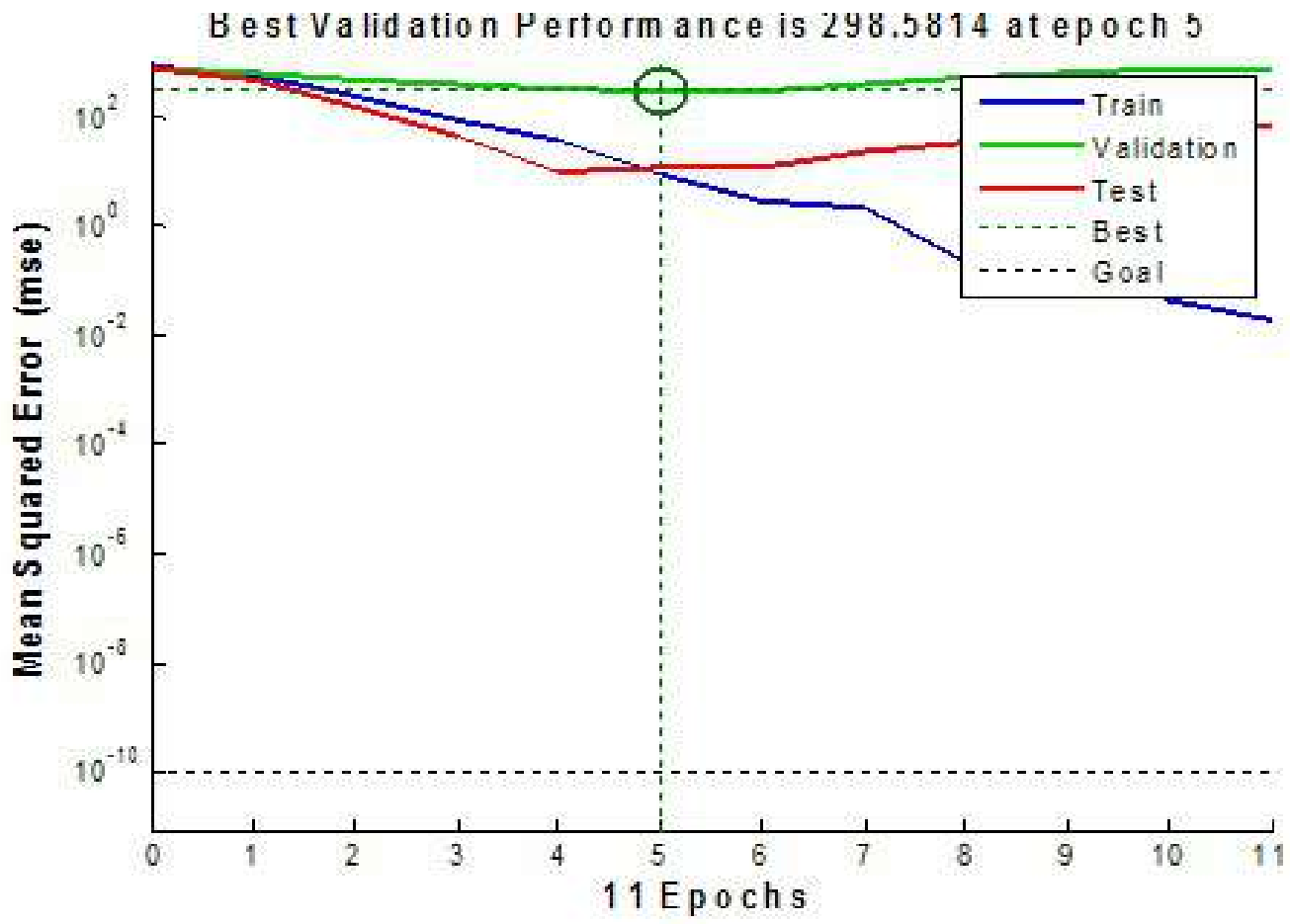
Best Validation Performance is 298.5814 at epoch 5
Mean Squared Error (mse)
11 Epochs
Train
Validation
Test
Best
Goal
(a)

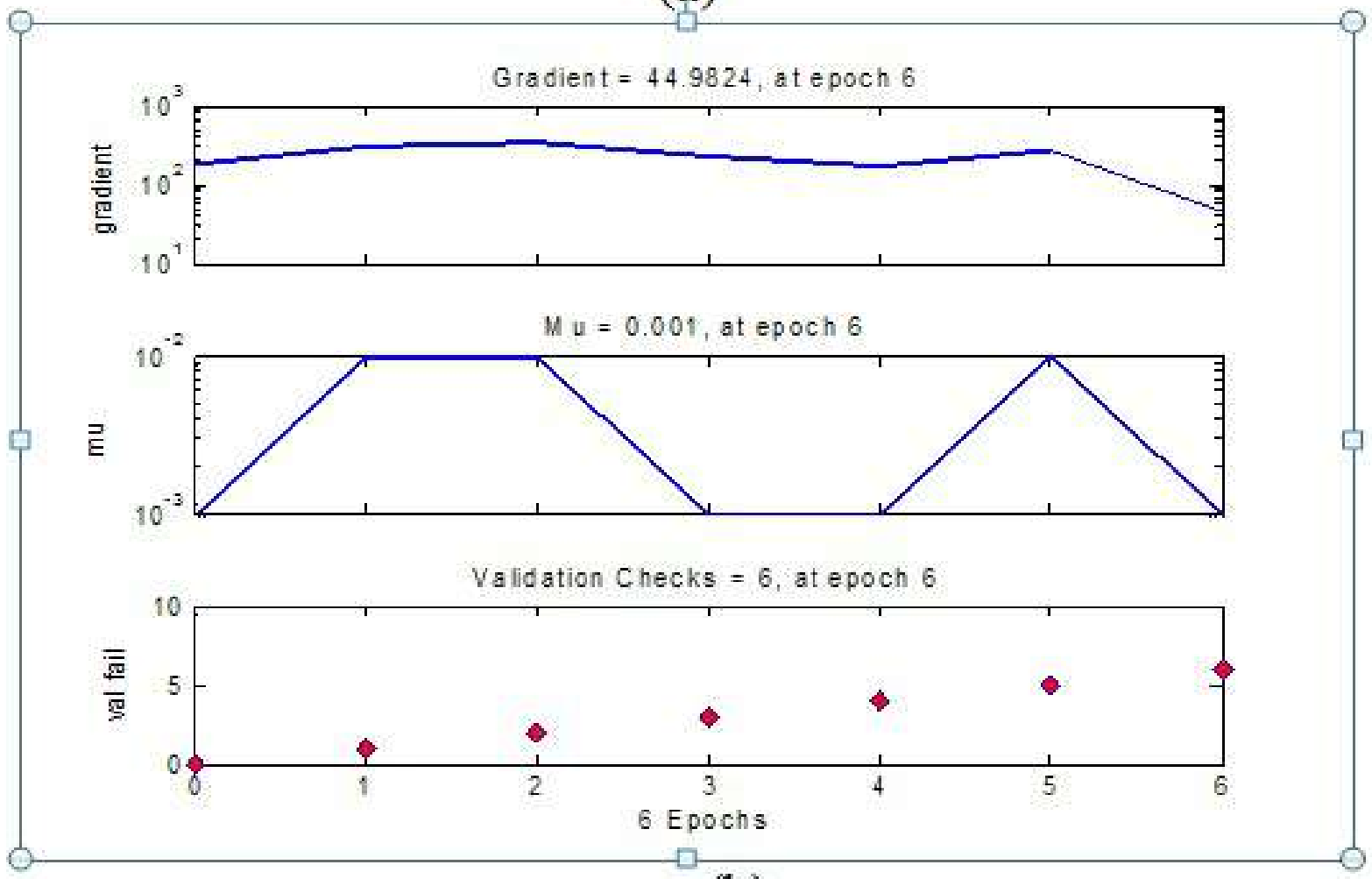
Gradient = 44.9824, at epoch 6
gradient
Mu = 0.001, at epoch 6
mu
Validation Checks = 6, at epoch 6
val fail
6 Epochs
(b)

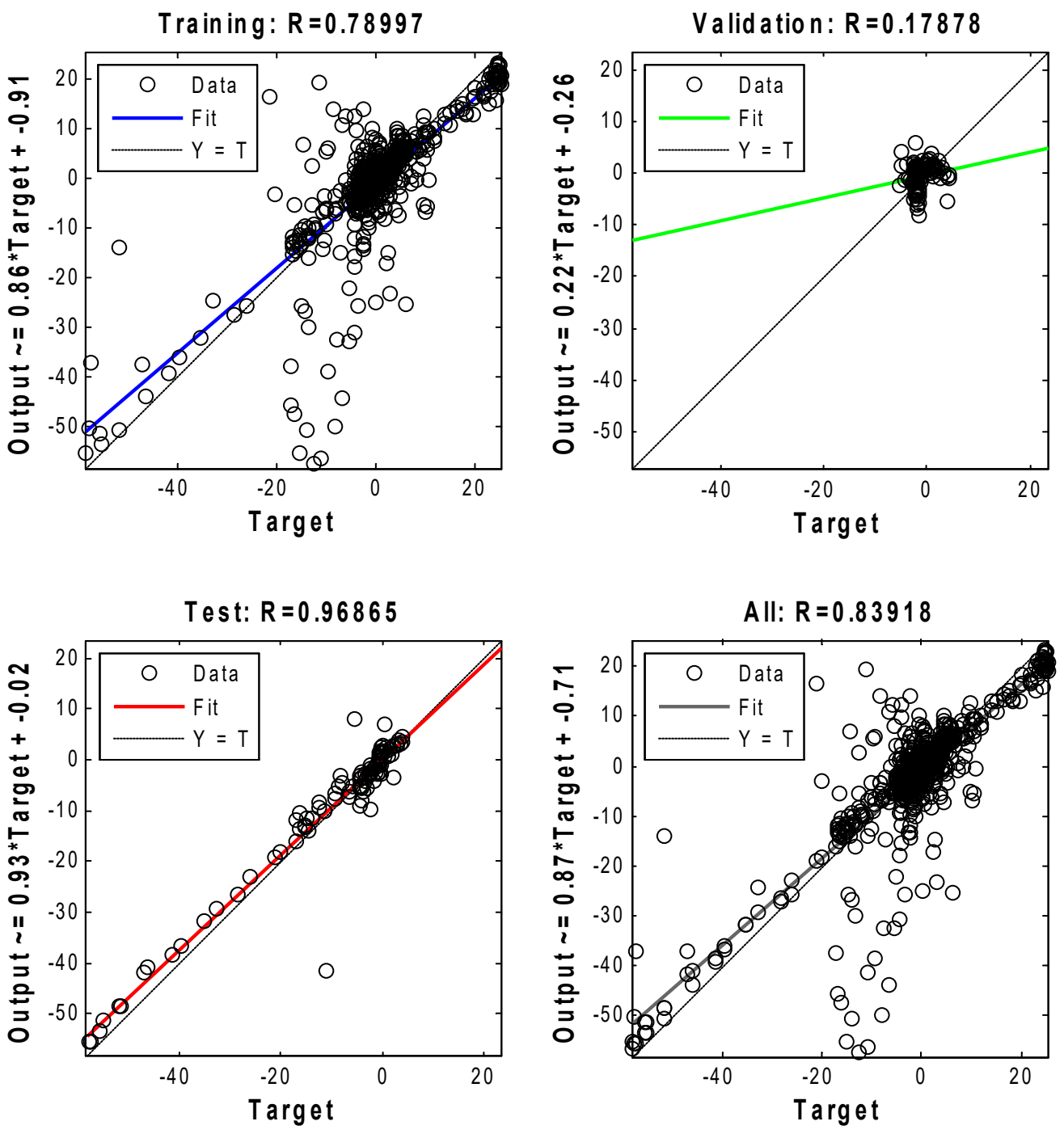

**Figure 6-14: A) Performance curve mean square error B)Training State c) Regression**

The accuracy rate (percentage) of gait classifications using K-mean where k=1. We calculated error i.e. total misclassification rate using formula (22).

$$\text{Error (Total Misclassification rate)} = \text{misclassified/ test sample} \quad (22)$$

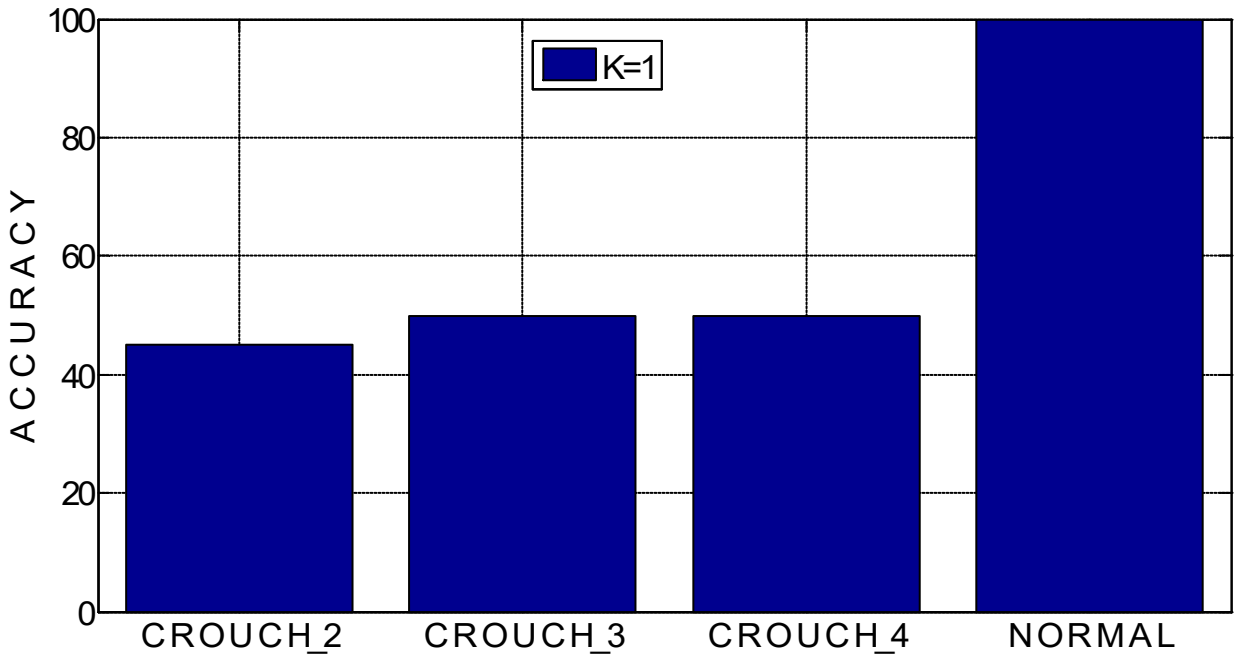

**Figure 6-15:The Accuracy classification rate of Different GAIT pattern using K-mean**

Fig.6.15 shows the accuracy bar chart of gait classification using K-mean and Fig. 6.16 the accuracy KNN for different majority value of K. Fig. 6.16 is the classification of normal gait using KNN with different majority vote.

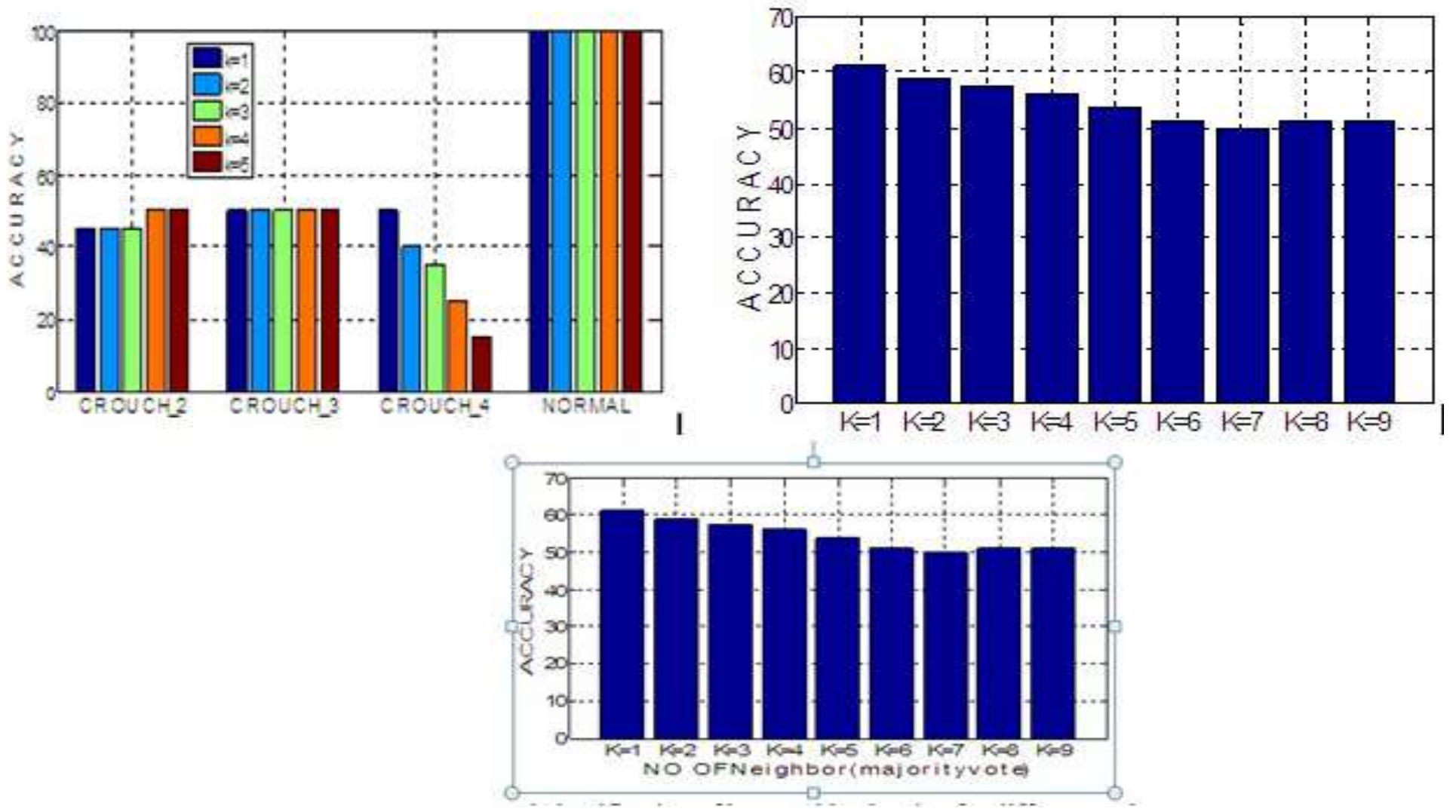

**Figure 6-16: classification of normal gait using KNN**

The graph represent the normal data is better classification rate. The gradient descent based ANN used for classification. The above table shows the ANN based classification technique outperform all the previously existed classification technique. The ANN based model propagates the input forward through the network and propagate the errors backward through the network is similar to the delta rule in gradient descent. Finally the sums over the errors of all output units Influence by a given hidden unit (this is because the training data only provides direct feedback for the output which information can used in security check and further used to identify pre identification of disease. The result shown in figure describes that ANN is better performer other machine learning technique. In the flow diagram it is shown that for a user based query to recognize and identify a human gait. It can be easily done by indexing the gait data in the database. In next step, the matching between the feature extracted by the system and that available in the database is taken into account for human gait classification.

### 6.12.1 Performance Matrix

The main performance indicator for any classification or biometric identification system is (receiver operating characteristic) ROC. It is basically, the curve of true acceptance rate (TAR) against false acceptance rate (FAR), which is the measure of no of false instance classified as positive among all intruder and imposter cases.

**FRR (False Rejection Rate)** - The probability of the legitimate claim when biometric system will fail to identify. It is a statistic biometric performance during verification task.

**TRR (True Reject Rate) -** Biometric performance in verification task

**TAR (True Acceptance Rate)** – It is count of true claim of identity when a system correctly verifies.

**FAR (False Acceptance Rate)** – The false acceptance percentage of system..

In the biometric literature, FAR is sometimes defined such that the "impostor" makes zero effort to obtain a match.

TAR=1-FRR – (23)

Where TAR-true acceptance rate and FRR-false rejection rate

To verify the result of biometric system we have four matrix terms True Acceptance Rate (FAR), True Rejection Rate (FAR), False Acceptance Rate (FAR), False Rejection Rate (FAR).

Verification results are reported in terms of the True Acceptance Rate (TAR), False Accept Rate (FAR), and ROC. The TAR is measured as the number of occurrences when genuine biometric identity is matched correctly, whereas, FAR is the measurement of the number of occurrences when imposter or intruder identity is matched falsely. EER is the point where FAR and FRR are equal, where FRR is the False Reject Rate and measured on the basis of number of false rejections of genuine matches and also given as,

FRR = 1 – TAR – (24)

ROC = TAR vs FAR – (25)

**Table 6:7: Confusion matrix for K-Mean{ Data Size 30 training and 20 testing }**

| | Normal | Crouch2 | Crouch3 | Crouch4 | TAR |
|---|---|---|---|---|---|
| Normal | 17 | 0 | 2 | 1 | 17/20=0.85 |
| Crouch2 | 0 | 17 | 3 | 0 | 17/20=0.85 |
| Crouch3 | 1 | 2 | 15 | 2 | 15/20=0.75 |
| Crouch4 | 0 | 1 | 0 | 19 | 19/20=0.95 |
| FAR | 1/18=.055 | 3/20=0.15 | 5/20=0.25 | 3/22=0.13 | |

$$TAR = \frac{\left(\frac{17}{20} + \frac{17}{20} + \frac{15}{20} + \frac{19}{20}\right)}{4} \times 100 = 85\%$$

FRR=1-TAR

$$FAR = \frac{\left(\frac{1}{18} + \frac{3}{20} + \frac{5}{20} + \frac{3}{22}\right)}{4} \times 100 = 14.79\%$$

Confusion matrix for KNN (K=1) {Data Size 30 training and 20 testing }

**Table 6:8: Confusion matrix for KNN (K=1)**

| | Normal | Crouch2 | Crouch3 | Crouch4 | TAR |
|---|---|---|---|---|---|
| Normal | 19 | 1 | 0 | 0 | 19/20=0.95 |
| Crouch2 | 0 | 17 | 3 | 0 | 17/20=0.85 |
| Crouch3 | 1 | 2 | 17 | 0 | 17/20=0.85 |
| Crouch4 | 0 | 1 | 0 | 19 | 19/20=0.95 |
| FAR | 1/20=0.05 | 4/21=0.19 | 3/20=0.15 | 0 | |

$$TAR = \frac{\left(\frac{19}{20} + \frac{17}{20} + \frac{17}{20} + \frac{19}{20}\right)}{4} \times 100 = 90\%$$

FRR=1-TAR

$$FAR = \frac{\left(\frac{1}{20} + \frac{4}{21} + \frac{3}{20}\right)}{4} \times 100 = 39.04\%$$

**Table 6:9: Confusion matrix for ANN {Data Size 30 training and 20 testing}**

| | Normal | Crouch2 | Crouch3 | Crouch4 | TAR |
|---|---|---|---|---|---|
| Normal | 20 | 0 | 0 | 0 | 20/20=1 |
| Crouch2 | 1 | 19 | 0 | 0 | 19/20=0.95 |
| Crouch3 | 1 | 1 | 18 | 0 | 18/20=0.90 |
| Crouch4 | 2 | 0 | 1 | 17 | 17/20=0.85 |
| FAR | 4/24=0.166 | 1/20=0.05 | 1/19=0.52 | 0 | |

$$TAR = \frac{\left(\frac{20}{20} + \frac{19}{20} + \frac{18}{20} + \frac{17}{20}\right)}{4} \times 100 = 92.5\%$$

FRR=1-TAR

$$FAR = \frac{\left(\frac{4}{24} + \frac{1}{20} + \frac{1}{19}\right)}{4} \times 100 = 6.73\%$$

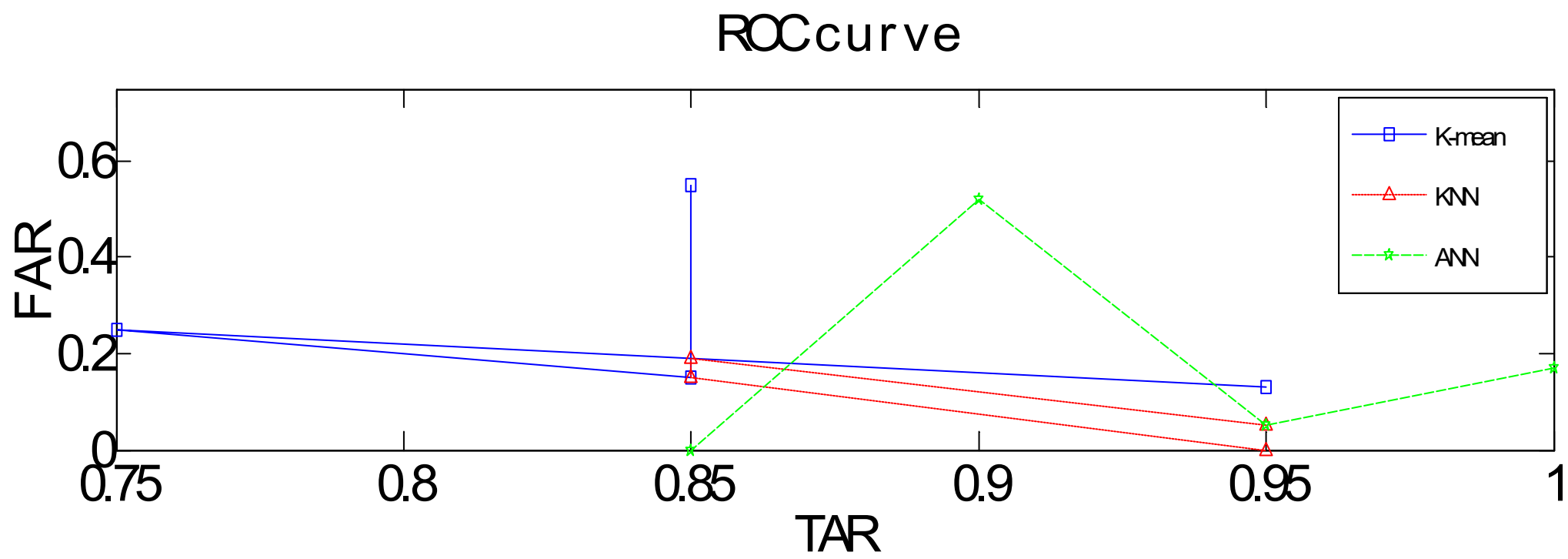

**Figure 6-17: ROC Curve**

Table 6.10 is the accuracy table which showing the ANN is better except the case of crouch 4 gait as it is more complex one gait so exact feature selection and Classification in K-mean is better. The accuracy percentage (%) of Different GAIT data classification.

**Table 6:10: Accuracy Curve**

| Method | K-Mean | KNN(n=1,2,3,4,5) | ANN |
|---|---|---|---|
| Normal Gait | 74.68 | 100,100,100,100,100 | 100 |
| Crouch 2 Gait | 51.43 | 50,50,50,55,55 | 60 |
| Crouch 3 Gait | 33.19 | 50,50,50,50,50 | 50 |
| Crouch 4 Gait | 81.80 | 50,40,35,25,15 | 55 |

## 6.13 Summary

We have classified the push recovery data collected through experiment on different subject and we have also classified the hybrid automata generated trajectories data with other class of gait. The Empirical Mode Decomposition based feature extraction technique had been shown to be effective for the classification of four different kinds of push recovery data. The experiment shows the features extracted from IMF don't degrade the accuracy of the system. These parameters are better optimized with Deep Neural Network hence we are getting more than 89.92% accuracy. Based on the experimental results we can conclude that the proposed technique is suitable for push recovery data classification and can achieve over 88.4% accuracy. An ANOVA test was performed on 5-fold cross validation results also shows that performance of DNN based classification is statistically significant rather than using ANN based. Once the pushes are classified, appropriate push recovery mechanism can be implemented on it accordingly. Similar we have classified the human gait data into four different classes' normal, crouch-2, crouch-3 and crouch-4. The result has been compared with KNN and K-mean algorithm and it has been shown that our classified outperform the existing classifiers.

# Chapter 7: Less Computationally Intensive Fuzzy Logic (Type-1) Based High level Controller for Humanoid Push Recovery

In previous chapter 6 we found the satisfactory classification results of human push recovery data which inspires us to develop the high level type-1 fuzzy logic based controller. It is sophisticated, less computationally intensive and more intuitive. The proposed controller is generic in nature and easy for training. The hierarchy fuzzy logic based controller included the decision model for three classes of forces (Small, Medium and High).

## 7.1 Fuzzy Logic based push recovery Controller

Developing a mathematical model of a humanoid bipedal robot for push recovery is extremely difficult, for the biped's inherent unstability, higher degree of nonlinearity, narrow stability zone and variability of dynamic during swing phase and stance phase. Hence we strongly believe that it will be extremely difficult, if not impossible, to develop a perfect planner for the biped push recovery controller. It should be based of learning which can accommodate imprecise logic. Therefore, this thesis presents a new type-1hierarchical fuzzy logic base controller for humanoids push recovery with an objective to develop an intelligent controller and implement biologically inspired push recovery for such robots. The work extends Gordon et al. [104] model for balancing humanoid using fuzzy logic and considering effects of roll, pitch and yaw.

## 7.2 Deliverable

To simulate the push recovery behavior on bipedal robot we have proposed the fuzzy logic based controller [105]. Fuzzy logic actually captures the fuzziness and vagueness exists in the environment [106].

The important Results & Observations:

- Introduces an intuitive fuzzy logic controller for bipedal push recovery.
- The hierarchical fuzzy logic based controller has been designed to reduce the computational cost incurred by large number of variables.
- It has been tested on the actual data and generalized the hierarchical fuzzy controller for easy trainability.
- It has been verified that the hierarchical fuzzy system can simplify the complex behavior [107].

- Our developed fuzzy inference system is less computationally intensive and able to recover the forces from all the direction.
- The impact of different magnitude forces on the different joints curve has been demonstrated.

Here we have introduced an intuitive fuzzy logic based learning approach and demonstrated that it is fast and effective. The humans negotiate push using three types of push recovery strategies namely ankle, hip and knee [108]. The major challenge associated with this fuzzy system define is the interpret a linguistic statement wisely, select the linguistic variable and their values for mapping the human behavior on robot [109]. The structure of human is highly complex so avoid fall human take different primitive actions. To simulate such behavior we need to define more linguistic variable with fuzzier rule so to simplify the behavior and mapping the actual behavior we used the hierarchical fuzzy system [110]. Some fuzzy logic are computationally very complex in real cases but as the humanoid robots have more degrees of freedom finding solutions using alternating methods are not easy at all [111].

## 7.3 Methodology

### 7.3.1 Proposed Hierarchical Fuzzy System (Fuzzy Logic controller)

To express the non linear behavior of human [112] in robot we are defining the fuzzy rule and expressing the information in term of linguistic variable [113] and value. We have designed hierarchical fuzzy logic system using FIS1 and FIS2. We are defining the rule based on the linguistic variable Force applied and the direction of motion (DoM) in roll and pitch. Based on the output of FIS1 (Fuzzy Inference System1) we combine it to the linguistic variable reaction on the input variables for FIS2, the input variable force with small, average, large and direction of motion (DoM) designed the FIS1, which will actually conclude the output in term of small and large roll, pitch, yaw. Further we extended the one more layer of hierarchy which is actually exhibit the behavior of robot behaviors based on output of FIS1 i.e. which strategy will robot will apply and will fall or recover. Finally to avoid certain action we will take primitive action. When boundary is not sharp we used fuzzy logic. A fuzzy set have degree of membership between 0 and 1.

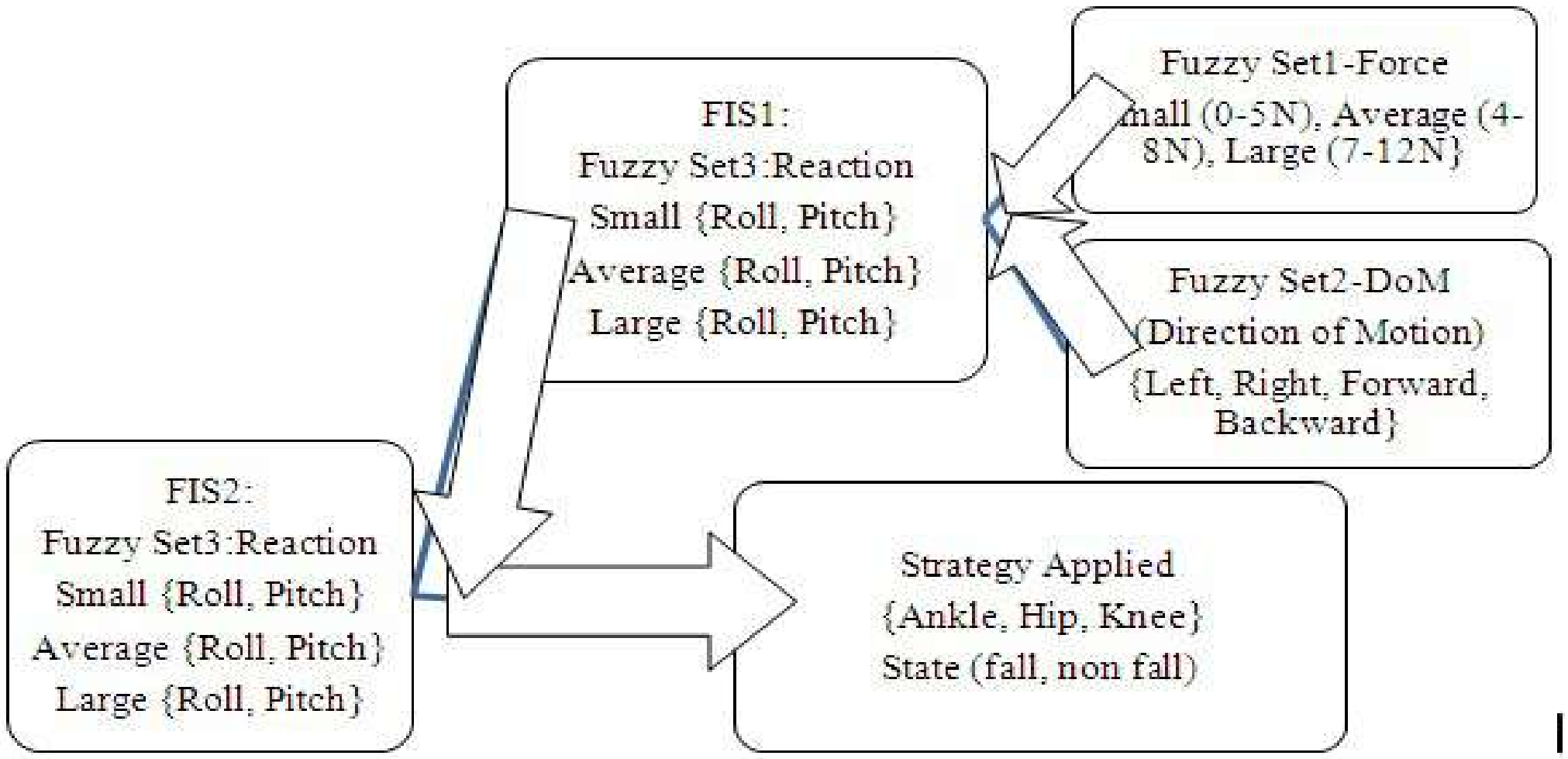

**Figure 7-1: Hierarchical Fuzzy Controller design for humanoid Push Recovery**

## 7.4 Fuzzy Inference System Design [114][115]

Each fuzzy system was built with the following three basic steps:

Identify the nouns or variables (both input and output) of the system.

Identify fuzzy sets for the variables used to generate the membership functions and their shape.

Identify the fuzzy rules.

### 7.4.1 Fuzzy Inference System 1 (FIS1) Design

The proposed Fuzzy Inference System 1(FIS1) using Force (small, average, large) and DoM (left, right, forward, backward) as input variable and gives the output reaction in term of roll and pitch. To train our controller to work fine in uneven terrain we applied fuzzy rules to our controller. The steps are

Steps1- Convert the crisp set into fuzzy set.

Step2- As there are two input variable so two crisp variables to convert into fuzzy value.

The two inputs variables are Force and Direction of Moment (DOM). The corresponding membership function for above two set are following:

Fuzzy Set1-The fuzzy value range for linguistic variable Force:

μForce=Small (x) = {0-5N}, μForce=Medium (x) ={4-9N} μForce=Large (x) ={8-12N}.

Fuzzy Set2-The fuzzy value range for linguistic variable DOM:

μDoM=Left (x), μDoM=Right (x), μDoM=Forward (x), μDoM=Backward (x)

### 7.4.2 Fuzzy Inference System 2 (FIS2) Design.

The FIS 2 is using output of FIS1 as input linguistic variables.

FIS2 has output is combination of force and direction applied. Small {Roll, Pitch}, Average {Roll, Pitch}, Large {Roll, Pitch}

Fuzzy Set3: defines a linguistic variable Reaction has values Small {Roll, Pitch} Average {Roll, Pitch} Large {Roll, Pitch}. FIS2 have output value in term of whether the robot will able to recover or not and which strategy the robot will apply for recovery. The set for FIS2 output Strategy Applied {Ankle, Hip, Knee}

And State {fall, non fall}.

### 7.4.3 Algorithm for FIS1 and FIS2 for Push Recovery Model

Algorithm8 for FIS1

```
BEGIN
        /* Input parameter for FIS1*/
        Force = {small, average, large};  DoM = {left, right, forward, backward};
        where small, average, large, left, right, forward, backward Є [0, 1]
/* FIS1 output = {Small roll, small pitch, average roll, average pitch, large roll, large pitch} */
REPEAT /* New Input */
If (Force = small && (DoM = left|| DoM= Right))
Reaction=small roll;
If (Force = small && (DoM = left|| DoM= Right))
Reaction =small pitch;
If (Force = average && (DoM = Forward|| DoM= Backward))
Reaction =average roll;
If (Force = average && (DoM = Forward|| DoM= Backward))
Reaction =average pitch;
/*for rest rule refer Table1*/
END LOOP
END;
```

| Algorithm 9 for FIS2 |
|---|

BEGIN
Use the output of FIS1 as input of FIS2;
/* FIS1 output = **{Small roll, small pitch, average roll, average pitch, large roll, large pitch} */**
REPEAT
If *REACTION* is small roll and small pitch
Then ankle strategy
If *REACTION* is small roll and average pitch
Then knee strategy
If *REACTION* is average roll and small pitch
Then knee strategy
……
//For rest rule refer table 2
UNTILL the recovery strategy applied and state determined /*Strategy applied- Ankle, Hip, Knee and State Fall and not Fall */
END LOOP
END

Table 7.1 and 7.2 defined the rule corresponding FIS1 & FIS2.

**Table 7:1: Fuzzy rule set FIS-I for learning Controller**

| ***DoM*** / ***Force*** | ***Left/Right*** | ***Forward/Backward*** |
|---|---|---|
| ***Small*** | Small Roll | Small Pitch |
| ***Average*** | Average Roll | Average Pitch |
| ***Large*** | Large Roll | Large Pitch |

**Table 7:2: Fuzzy rule set FIS-II for learning Controller**

| ***Pitch*** / ***Roll*** | ***Small Pitch*** | ***Average Pitch*** | ***Large Pitch*** |
|---|---|---|---|
| ***Small Roll*** | Ankle Strategy | Knee Strategy | Hip Strategy |
| ***Average Roll*** | Knee Strategy | Hip Strategy | Falls in frontal plane |
| ***Large Roll*** | Hip Strategy | Falls sideways | Falls |

## 7.5 Fuzzy Rule Result & Surface View Of Hierarchical Based Rule

FIS1 Rule: depending on magnitude of force applied on the model and direction of motion on model, defines the effect on the body as a result of push. The rule (33) to (36) used to predict the reaction, which is output of FIS1. Which we further used as input for FSI2.

Rule 1: $\mu_{action=small\ roll} = \max[\mu_{Force=Small}(x), \min[\mu_{DOM=Left}(x), \mu_{DOM=Right}(x)]]$-(33)

Rule 2: $\mu_{action=small\ pitch} = \max[\mu_{Force=Small}(x), \min[\mu_{DOM=Forward}(x), \mu_{DOM=Backward}(x)]]$-(34)

Rule 3: $\mu_{action=large\ roll} = \max[\mu_{Force=Large}(x), \min[\mu_{DOM=Left}(x), \mu_{DOM=Right}(x)]]$-(35)

Rule 4: $\mu_{action=large\ pitch} = \max[\mu_{Force=\ Large}(x), \min[\mu_{DOM=Forward}(x), \mu_{DOM=Backward}(x)]]$-(36)

FIS2 Rule: To avoid a fall and recover from push, as a reactive measure one applies counter force at any of three joints of lower human body. The magnitude of external force applied makes control action to zero in for ankle, hip or knee strategy. If the magnitude of push is beyond the above strategy the robot will not able to recover. The rules no (37) to (41) is the output of FIS2. This is the resultant strategy for recovery.

Rule 5: $\mu_{reaction=ankle\ strategy,\ not\ falling} = \max[\mu_{Action=small\ roll}(x), \mu_{Action=small\ pitch}(x)]$-(37)

Rule 6: $\mu_{reaction=knee\ strategy,\ not\ falling} = \max[\min[\mu_{Action=average\ roll}(x), \mu_{Action=small\ pitch}(x)], \min[\mu_{Action=average\ roll}(x), \mu_{Action=average\ pitch}(x)]]$-(38)

Rule 7: $\mu_{reaction=hip\ strategy,\ not\ falling} = \max[\min[\mu_{Action=small\ roll}(x), \mu_{Action=large\ pitch}(x)], \min[\mu_{Action=large\ roll}(x), \mu_{Action=small\ pitch}(x)], \min[\mu_{Action=average\ roll}(x), \mu_{Action=average\ pitch}(x)]]$-(39)

Rule 8: $\mu_{reaction=\ falling\ F/B} = \max[\mu_{Action=large\ roll}(x), \mu_{Action=largesmall\ pitch}(x)]$-(40)

Rule 9: $\mu_{reaction=\ falling\ L/R} = \max[\mu_{Action=small\ roll}(x), \mu_{Action=small\ pitch}(x)]$- (41)

The last step is the final step is rule evaluation.

We now use the results to *scale* or *clip* the consequent membership functions. Once again for the sake of simplicity we will clip each of the functions. Fig 7.2 shows the surface view corresponding FIS1 & FIS2.

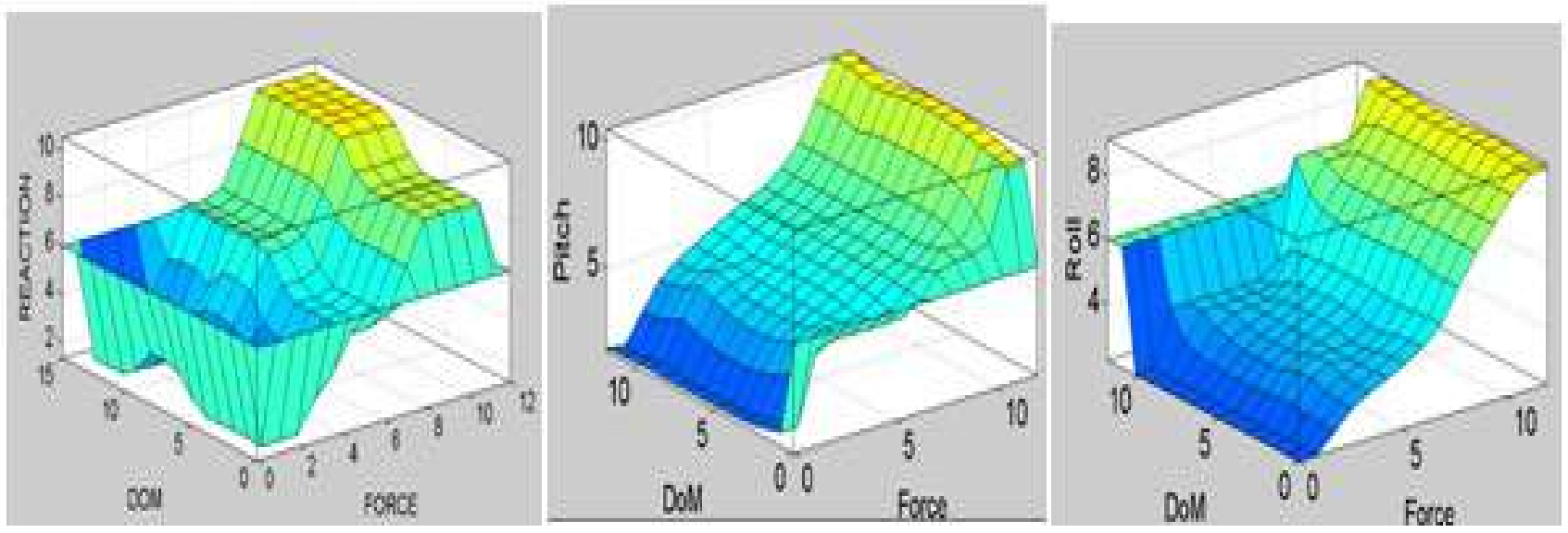

**Figure 7-2: Surface View**

Table 7.1 shows the relationship between FIS1 input variable and output. Based on the permutation of above rule set we made the following fuzzy set of rule for taking intelligent decision for controller. In this way we designed the flat set of rule for complex non linear behavior. The research introduced the new term which is defined by name auto leaning. Table 7.2shows the relationship between FIS2 input variable and output. In this way table 1 which have input as force and DOM, which will tell about the direction of magnitude in term of roll and pitch which further will use as input for FIS2. Which actually predict the type of strategy applies.

## 7.6 Validation with the simulation and experimental result

To investigate the humanoid push recovery behavior, we applied different magnitude forces of up to 12 Newton on the simulated model; it is successfully able to recovers. The simulated model is able to recover and take action fast using fuzzy logic based controller.

### 7.6.1 Verification of data

The simulated model with size and weight as described in table applied by imparting physics to it. Table 7.4 is the lookup table for real robot control, the fuzzy rules will be evaluated offline and the results are incorporated in the form of a look up table.

**Table 7:3Lookup table for real robot control offline:**

| *Force Magnitude (Newton)* | *Reaction* | *Strategy applied* |
|---|---|---|
| ***Small(0-5)*** ***Average(4-9)*** ***Large(8-12)*** | Small Roll and Small Pitch | Ankle |
| | Small Roll and Average Pitch<br>Average Pitch and Small Roll | Knee |
| | Large Roll and Small Pitch<br>Large Pitch and Small Roll<br>Average Pitch and Average Roll | Hip |
| | Large Roll and Large Pitch<br>Average Roll and Large Pitch | Fall (Not able to recover) |

## 7.7 Analysis of Curve

Genetic algorithm controller and earlier developed fuzzy logic based controller are only able to generalize the parameter in term of roll and pitch but not able to predict which particular strategy to apply. Our type-1 hierarchical fuzzy logic based controller is powerful enough to predict the required control strategy from recovery against push i.e. knee, ankle, hip strategy. Hybrid automata will help to generate the biologically inspired controller. Here we would like to analysis the performance of our simulated model, we collected the real data for five different subjects through wearable suit HMCD and same data applied to our simulated model. The look up table 3 is the data of real time simulation. Fig 7.3 (a) is the plot for left hand person when the force is small magnitude of range 0 to 5 Newton we observed the ankle joint is more active in compare to rest two joint. Fig 7.3 (b) is plot when the force is average magnitude of range 4 to 9 Newton we observed the ankle joint is more active in compare to rest two joint. Fig.7.3 (c) is plot when the force is large magnitude of range 8 to 12 Newton we observed the hip joint is more active in compare to rest two joints. When the force is more than 12 Newton the robot will not be able to recover. All Joint curves are between joint angles versus time.

Where 7.3 (a), (b) and (c) are curves for different magnitude force for left hand subject and 7.4 (a), (b) and (c) are curves for different magnitude force for right hand subject. The curve for small magnitude force shows that ankle joint have much variation, average creating more variation in ankle and knee and large have all three.

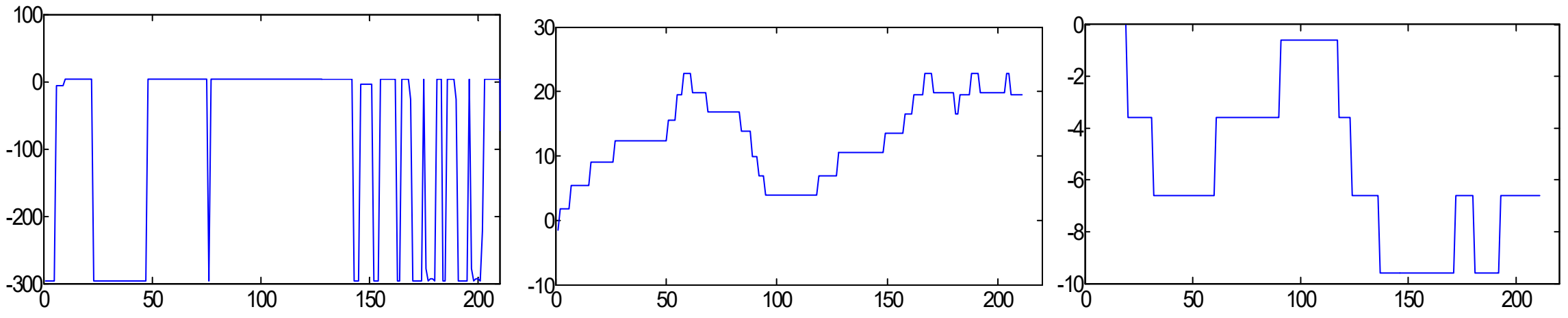

**Left Ankle** **Left Knee** **Left Hip**

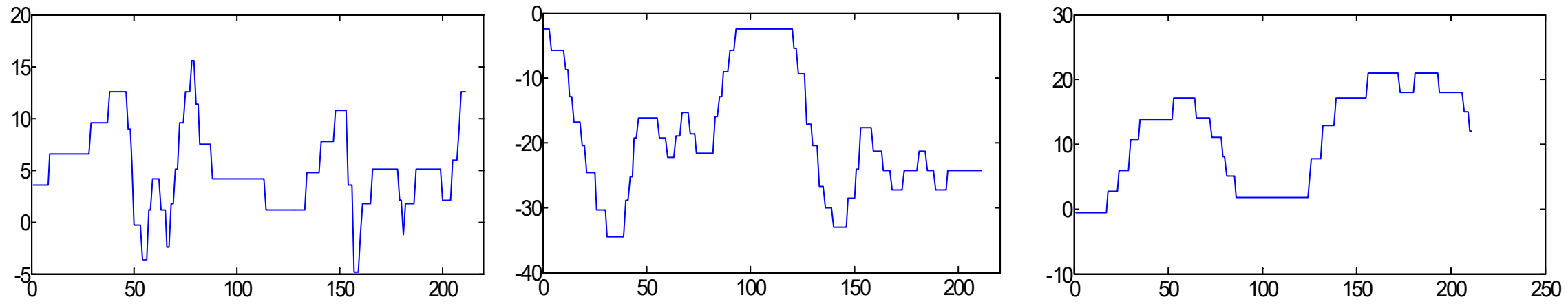

Right Ankle Right Knee Right Hip

(a) Small (0-5 Newton) Force

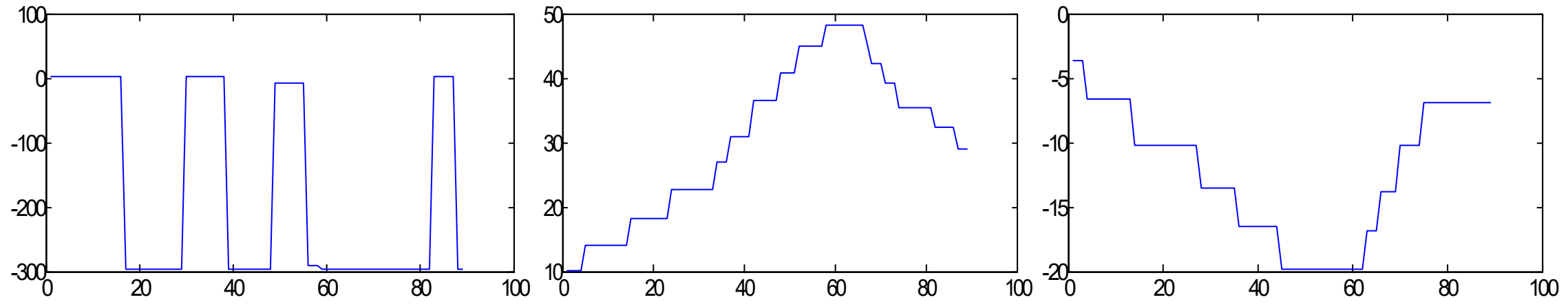

Left Ankle Left Knee Left Hip

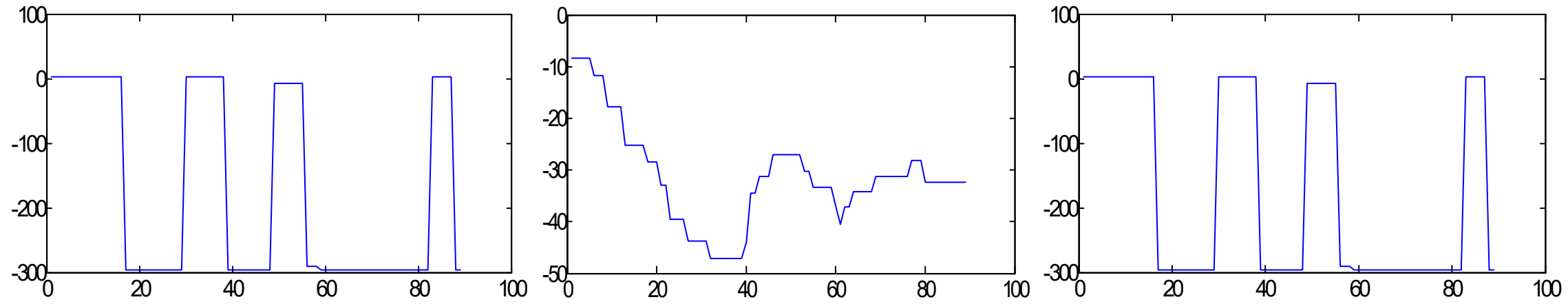

Right Ankle Right Knee Right Hip

(b) Average (4-9 Newton) Force

Left Ankle Left Knee Left Hip

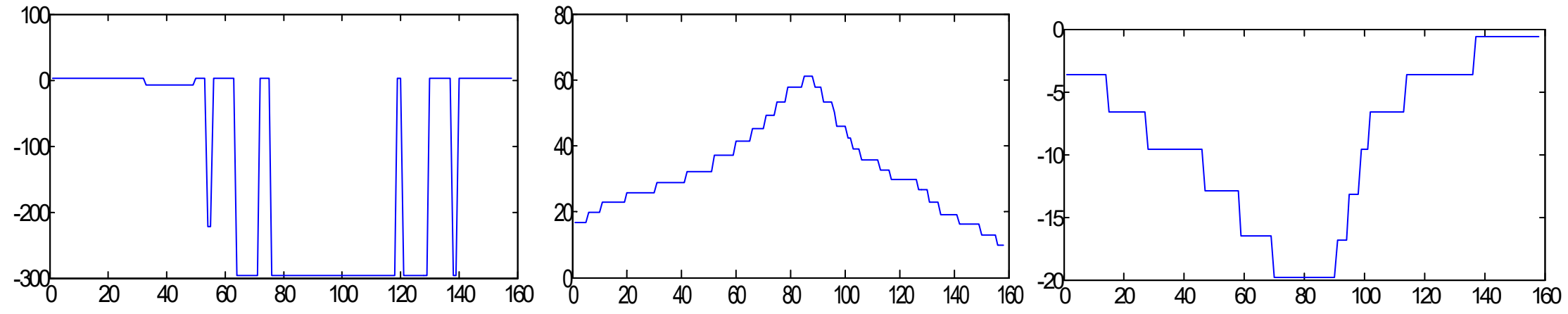

Right Ankle Right Knee Right Hip

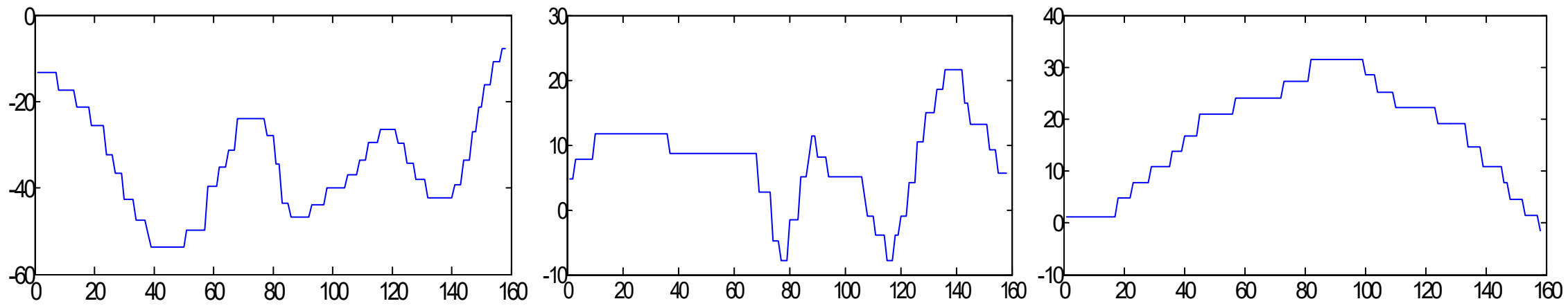

(c) Large (8-12 Newton) Force

**Figure 7-3: Observed Leg Joint Curve for Right and Left Leg of Left Hand subject**

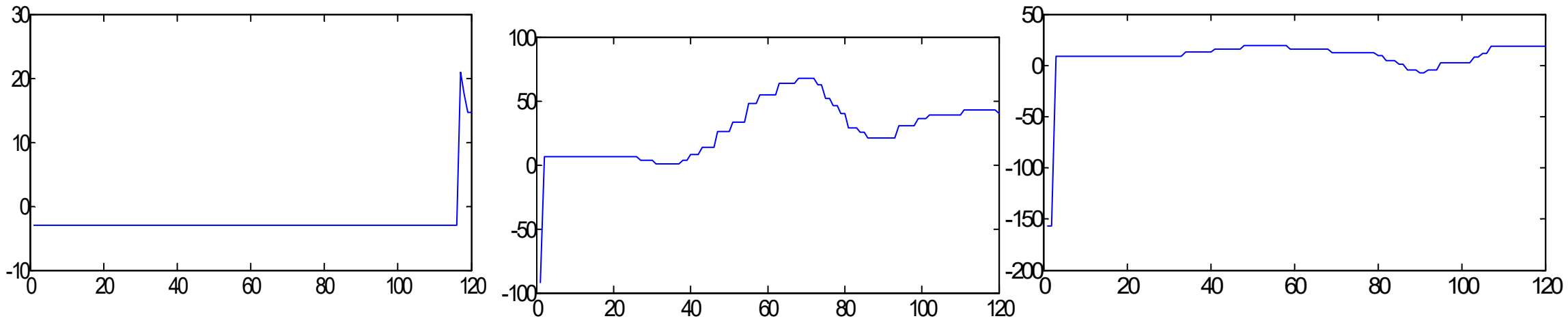

Right Ankle Right Knee Right Hip

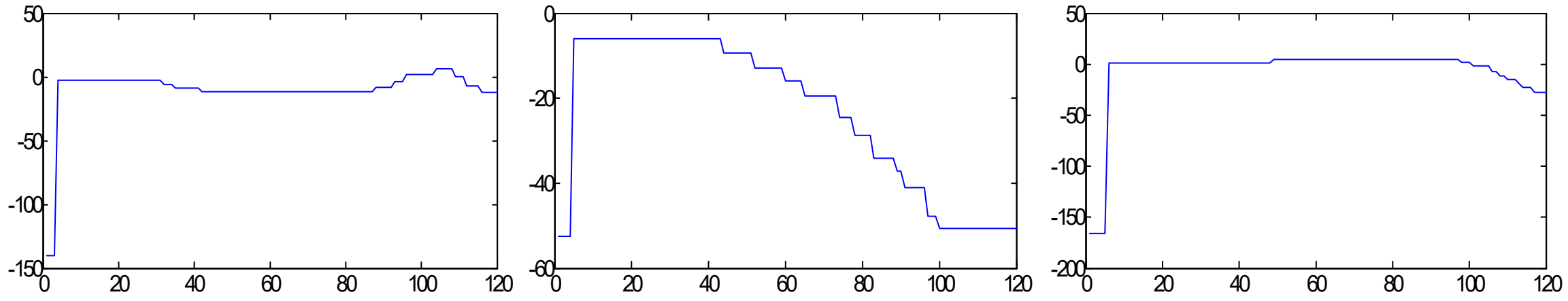

Left Ankle Left Knee Left Hip

Small (0-5 Newton) Force

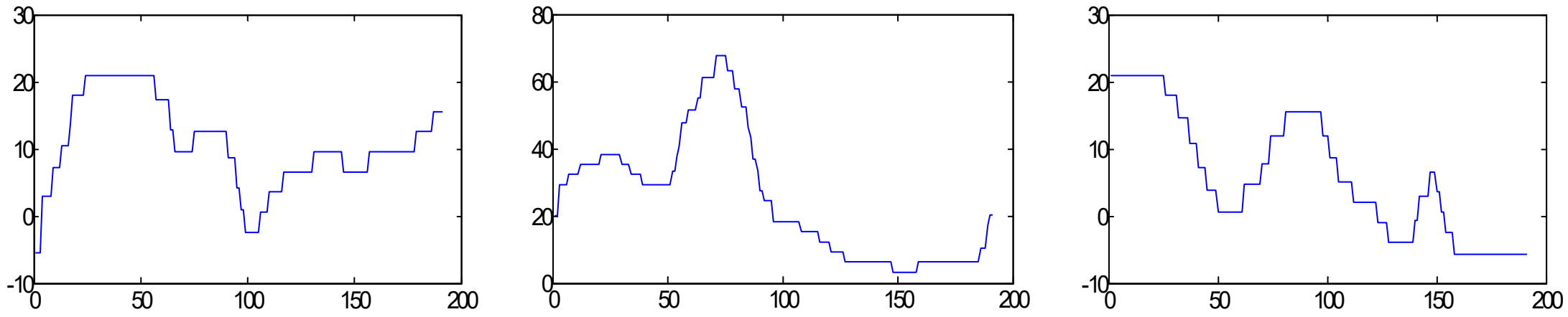

Right Ankle Right Knee Right Hip

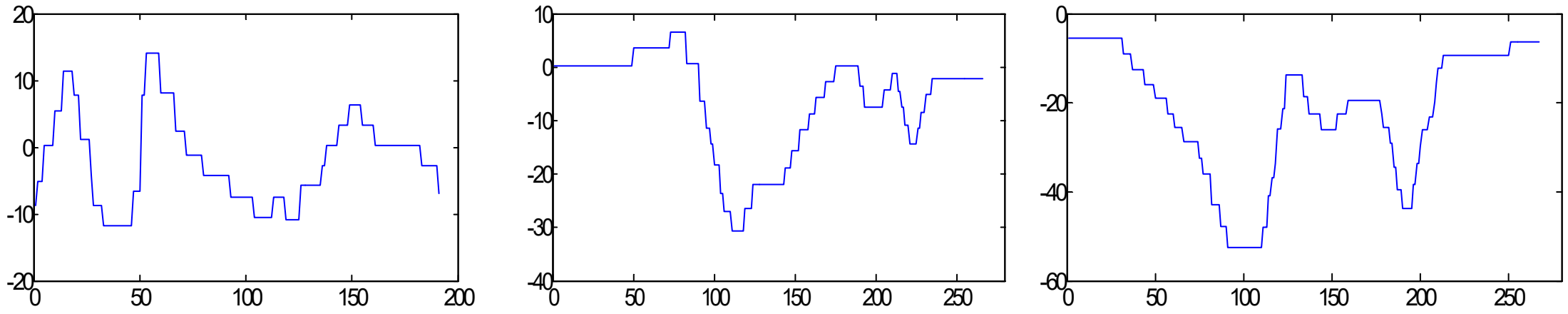

Left Ankle Left Knee Left Hip

Average (4-8 Newton) Force

Right Ankle Right Knee Right Hip

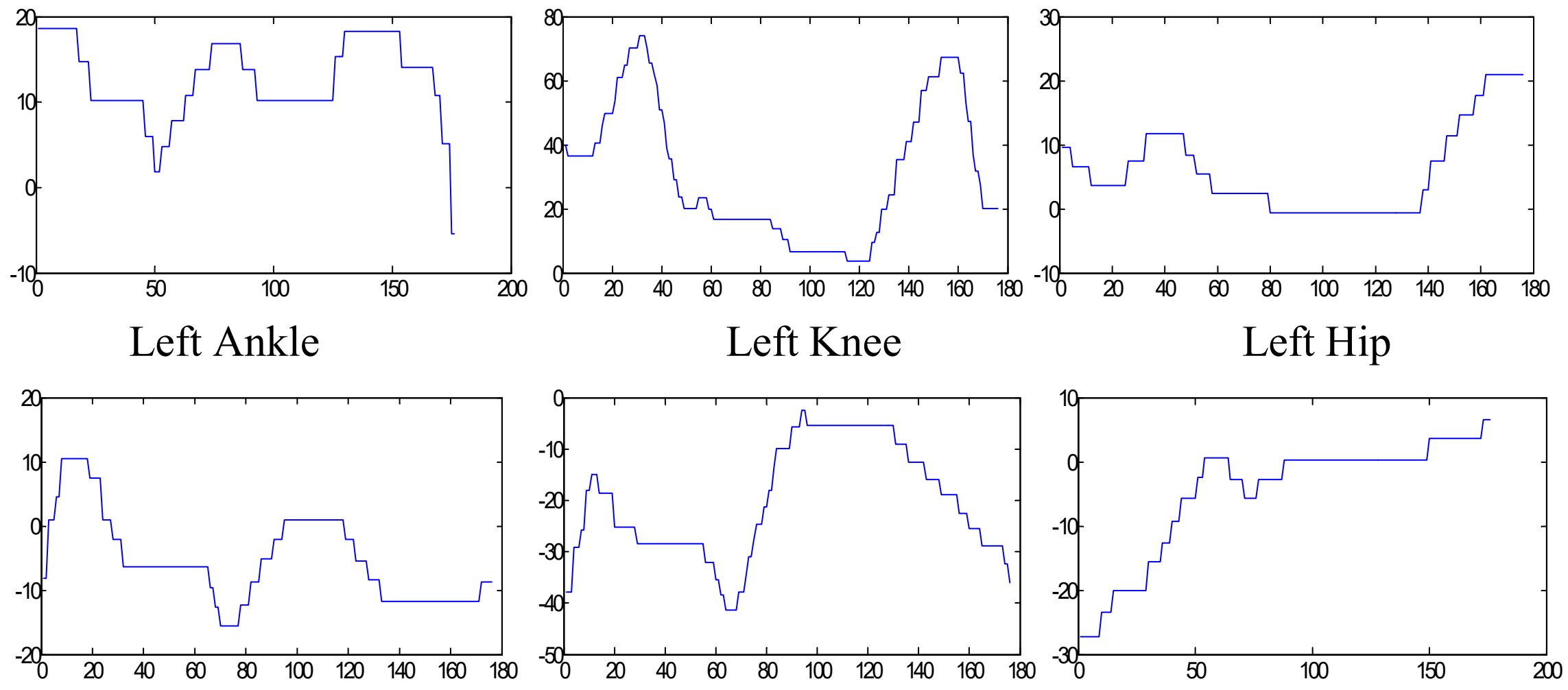

(C) Large (8-12 Newton) Force

**Figure 7-4: Observed Leg Joint Curve for Right and Left Leg of Right Hand Subject**

Table 7.4 is the different joints angle range for different magnitude of force. The different joint angles values captured through HMCD. Our model is able to perform the required strategy at given magnitude force and angle.

**Table 7:4: Validation table for our simulated model**

| *Force (Newton)* | *Joint Angle values for 6 joint (degree)* | | | | | | *Strategy applied* |
|---|---|---|---|---|---|---|---|
| | *Left* | | | *Right* | | | |
| | $\theta_H$ | $\theta_K$ | $\theta_A$ | $\theta_H$ | $\theta_K$ | $\theta_A$ | |
| ***Small (0-5)*** | **{7.2, 3.9}** | **{4.5-7.5}** | **{0.9, 3.9}** | **{-8.4,-5.4}** | **{-4.8,-1.8}** | **{-4.8,-1.8}** | Ankle |
| ***Average (4-9)*** | **{3.9, 17.1}** | **{7.5-20.1}** | **{3.9,-0.9)}** | **{-5.4,-2.4}** | **{-1.8,-11.4}** | **{-8.7,-11.4}** | Knee |
| ***Large (8-12)*** | **{14.1,-0.9}** | **{16.8,-1.8}** | **{-0.9,-0.8}** | **{-2.7,0.3}** | **{-11.4,-2.4}** | **{-11.4,-5.4}** | Hip |

## 7.8 Summary

Here we have introduced an intuitive fuzzy logic controller for bipedal push recovery and demonstrated that it is fast and effective. The hierarchical fuzzy logic based controller has been designed to reduce the computational cost incurred by large number of variables. It was very difficult to design the fuzzy rules with less number of variables for human push recovery. To overcome this problem, we have designed the hierarchical fuzzy logic controller. It has been tested on the actual data and generalized the hierarchical fuzzy controller for easy trainability. It has been verified that the hierarchical fuzzy system can simplify the complex behavior. We have introduced auto leaning term to define human nature. Our developed fuzzy inference system is less computationally intensive and able to recover the forces from all the direction. The impact of different magnitude forces on the different joints curve has been demonstrated. The fuzzy logic based controller is able to predict which particular strategy is required.

# Chapter 8: Conclusion and Future Recommendations

## 8.1 Summary of the research

### 8.1.1 Major Contributions of the thesis:

We have developed the computational model based on hybrid and cellular automata for prediction, formal verification and analyses of joint trajectories of bipedal locomotion using theoretically enriched hybrid automata technique for modelling. The chapter wise major contributions of the thesis are:

- In the first part of the thesis, we have presented the essence of bipedal robot for modern societies. We have also discussed the inherent challenges associated with bipedal robot. The major reason of instability of the existing kinematic model is the limitation of making perfect biped model with all correct structural, frictional and other nonlinear parameters. We have also presented the how computational model are suitable then kinematics based model.
- In the second chapter of the thesis, we have presented the analysis of the available bipedal robot technology and bipedal model.
- In chapter third of the thesis, we have given an overview of the bipedal technology with necessary fundamentals. We have presented all the important terminologies used with bipedal gait and push recovery.
- In chapter 4 we have presented the development of sophisticated Human Motion Capture Device (HMCD) and Human Locomotion and Push Recovery Data Capture Device (HLPRCD) devices to capture the human gait and push recovery data.
- In chapter 5 we have presented computations data driven model based on hybrid automata. We have presented the vector fields and development of hybrid automata model for bipedal walk and generation of joints trajectories. The generated joints trajectories are verified using opensim model gait 2354 and HOAP2 simulated model.
- We have also developed the unique approach to model the human gait state using cellular automata. It is able to model the normal human gait within a negotiable degree of error. Here we have written 16 CA rules to determine the state of atomic components of one leg with the help of second leg. All the states are represented using 4-bit stream.
- In chapter -6, we presented the classification of push recovery data using deep neural learning network and comparison using other machine learning techniques [20].
- In next chapter -7 we have presented the Development of a fuzzy logic based push recovery capable controller [21].

### 8.1.2 Limitations& Future Research

We have developed the computational data driven model which is not considering any physical parameters like length, mass etc. and we have assumed that the trajectory data generated form a human/robot can be adopted by their counterpart which is having similar morphological structure ( similar leg length and mass etc.). So the traditional kinematics and dynamic based model need to synergy with computational model. We have taken consideration the real human data from real environment. So we believe that it is asymptotically stable which is needed to verify through the model. The perfect model is not available.Future research may be directed towards calculating joint torques with the help of inverse dynamics with these joint trajectories as inputs (desired trajectories) and develop an execution level controller for a hardware biped robots resembling with normal human walkers.

## References

1. Saab, Layale, Philippe Souères, Nicolas Mansard, and Jean-Yves Fourquet. "Generation of human-like motion anthropomorphic systems using inverse dynamics." Computer methods in biomechanics and biomedical engineering15, no. sup1 (2012): 156-158.
2. Brian R. Duffy, “Anthropomorphism and the social robot”, Robotics and Autonomous Systems, Volume 42, Issues 3–4, 31 March 2003, Pages 177-190.
3. Fong, Terrence, Illah Nourbakhsh, and Kerstin Dautenhahn. "A survey of socially interactive robots." Robotics and autonomous systems 42.3 (2003): 143-166.
4. Semwal, Vijay Bhaskar, et al. "Biped model based on human Gait pattern parameters for sagittal plane movement." IEEE International Conference on Control, Automation, Robotics and Embedded Systems (CARE), 2013.
5. Sajid Iqbal, Xizhe Zang, Yanhe Zhu,Hanadi Mohammed Abass Ali Saad, Jie Zha, "Nonlinear Time-Series Analysis of Different Human Walking Gaits",2015 IEEE International Conference on Electro/Information Technology, At Naperville, IL, USA.
6. Shibendu Shekhar Roy, Dilip Kumar Pratihar, Effects of turning gait parameters on energy consumption and stability of a six-legged walking robot, Robotics and Autonomous Systems, Volume 60, Issue 1, January 2012, Pages 72-82.
7. Thomas A. Henzinger , “The Theory of Hybrid Automata” , Electrical Engineering and Computer Sciences.
8. Harcourt-Smith, WilliamE.H., “ The Origins of Bipedal Locomotion” , Springer Berlin Heidelberg Handbook of Paleoanthropology, 2007.
9. Sajid Iqbala,*, Xi-Zhe ZANG, Yan-He ZHU, Dong-Yang BIE, Xiao-Lu WANG and Jie ZHAO, “Nonlinear Time-Series Analysis of Human Gaits in Aging and Parkinson's Disease” 2015 International Conference on Mechanics and Control Engineering (MCE 2015).
10. Shahid Hussain, Sheng Q. Xie, Prashant K. Jamwal, Control of a robotic prosthesis for gait rehabilitation, Robotics and Autonomous Systems, Volume 61, Issue 9, September 2013, Pages 911-919.
11. B.Stephens “Humanoid push recovery,” in Proceedings of the IEEE-RAS International Conference on Humanoid Robots, pp. 589-595, 2007.
12. Zhe Tang; Meng Joo Er; Chien, C.-J., "Analysis of human gait using an Inverted Pendulum Model," IEEE International Conference on Fuzzy Systems, pp.1174,1178, 1-6 June 2008.

13. Benjamin Stephens ,Christopher Atkeson, “Modeling and Control of Periodic Humanoid Balance using the Linear Biped Model “, in*9th IEEE-RAS International Conference on Humanoid Robots* December 7-10,2009 Paris.
14. H. Hemami and P. Camana, “Nonlinear feedback in simple locomotion systems,” In IEEE Transactions on Automatic Control, vol. 21, no. 6, pp. 855–860, December 1976.
15. M.Vukobratovic,A.Frank,andD.Juricic,“Onthestabilityofbipedlocomotion,”InIEEETransactionsonBiomedicalEngineering,Vol.BME-17,no1,pp.25–36,January1970.
16. Chris Iverach-Brereton, Jacky Baltes, John Anderson, Andrew Winton, Diana Carrier, Gait design for an ice skating humanoid robot, Robotics and Autonomous Systems, Volume 62, Issue 3, March 2014, Pages 306-318.
17. Dr. MahboubBaccouch, Stephen Dodds, “A two-link manipulator: simulation and control design” in University of Nebraska at Omaha,March 2012.
18. Sejdic, Ervin, et al. "A comprehensive assessment of gait accelerometry signals in time, frequency and time-frequency domains." In IEEE Transactions on Neural Systems and Rehabilitation Engineering, 22.3 (2014): 603-612.
19. Semwal, V.B.; Bhushan, A.; Nandi, G.C., "Study of humanoid Push recovery based on experiments," Control, Automation, Robotics and Embedded Systems (CARE), 2013 International Conference on , vol., no., pp.1,6, 16-18 Dec. 2013.
20. Semwal, Vijay Bhaskar, and Gora Chand Nandi. "Toward Developing a Computational Model for Bipedal Push Recovery-A Brief." Sensors Journal, IEEE 15.4 (2015): 2021-2022.
21. Semwal, Vijay Bhaskar, et al. "Biologically-inspired push recovery capable bipedal locomotion modeling through hybrid automata." Robotics and Autonomous Systems 70 (2015): 181-190.
22. Vijay Bhaskar Semwal, Kaushik Mondal and G.C. Nandi, “Robust and Accurate Feature Selection for Humanoid Push Recovery and Classification: Deep learning Approach”, Neural Computing and Application, Springer.
23. Semwal, Vijay Bhaskar, Pavan Chakraborty, and G. C. Nandi. "Less computationally intensive fuzzy logic (type-1)-based controller for humanoid push recovery." Robotics and Autonomous Systems 63 (2015): 122-135.
24. R. W., Powell, M. J., Shah, R. P., & Ames, A. D. “human-inspired hybrid controls approach to bipedal robotic walking”. In 18th IFAC World Congress (pp. 6904-11), 2011
25. Sinnet, Ryan W., Shu Jiang, and Aaron D. Ames. "A human-inspired framework for bipedal robotic walking design." *International Journal of Biomechatronics and Biomedical Robotics* 3.1 (2014): 20-41.
26. Lack, Jordan, Matthew J. Powell, and Aaron D. Ames. "Planar multi-contact bipedal walking using hybrid zero dynamics" in Robotics and Automation (ICRA), 2014 IEEE International Conference on. IEEE, 2014.

27. Nurfarahin Onn, Mohamed Hussein, Collin Howe Hing Tang, MohdZarhamdyMd Zain, Maziah Mohamad And Lai Wei Ying ‘Motion Control of Seven-Link Human Bipedal Model’ by Faculty of Mechanical Engineering University Technology Malaysia.
28. Stephens, Benjamin J., and Christopher G. Atkeson. "Dynamic balance force control for compliant humanoid robots." Intelligent Robots and Systems (IROS), 2010 IEEE/RSJ International Conference on. IEEE, 2010.
29. David E Orin, Ambarish Goswami, Sung-Hee Lee,"Centroidal dynamics of a humanoid robot",Autonomous RobotsVolume 35, Issue 2-3,Pages,161-176, Publisher, SpringerUS.
30. Sato, Tomoya, ShoSakaino, and Kouhei Ohnishi. "Real-time walking trajectory generation method at constant body height in single support phase for three-dimensional biped robot." in *IEEE International Conference on*, *Industrial Technology,. ICIT 2009.*
31. Eric R. Westervelt, Jessy W. Grizzle, Christine Chevallereau “Feedback Control of Dynamic Bipedal Robot Locomotion’ Taylor & Francis/CRC (2007)
32. Henzinger, Thomas A. "Hybrid automata with finite bisimulations." Automata, Languages and Programming. Springer Berlin Heidelberg, 1995. 324-335.
33. Semwal, Vijay Bhaskar, et al. "An optimized feature selection technique based on incremental feature analysis for bio-metric gait data classification." Multimedia Tools and Applications (2016): 1-19.
34. Raj, Manish, Vijay BhaskarSemwal and G. C. Nandi. “Multiobjective optimized bipedal locomotion,” International Journal of Machine Learning and Cybernetics, pp. 1-17, 2017.
35. Wentao Mao; Guihe Qin; Ju-Jang Lee, "Humanoid push recovery strategy for unknown input forces,"International Conference on Mechatronics and Automation, 2009. ICMA 2009., vol., no., pp.1904,1909, 9-12 Aug. 2009.
36. Stephens, B.J.; Atkeson, C.G., "Push Recovery by stepping for humanoid robots with force controlled joints," in Humanoid Robots (Humanoids), 2010 10th IEEE-RAS International Conference on , vol., no., pp.52,59, 6-8 Dec. 2010
37. Nandi, Gora Chand, et al. "Development of Adaptive Modular Active Leg (AMAL) using bipedal robotics technology." Robotics and Autonomous Systems57.6 (2009): 603-616.
38. Lim, Hun-ok, and Atsuo Takanishi. "Biped walking robots created at Waseda University: WL and WABIAN family." *Philosophical Transactions of the Royal Society of London A: Mathematical, Physical and Engineering Sciences*365.1850 (2007): 49-64.
39. Olivares, Manuel, and Pedro Albertos. "Linear control of the flywheel inverted pendulum." *ISA transactions* 53.5 (2014): 1396-1403.

40. Coleman, Michael J., Anindya Chatterjee, and Andy Ruina. "Motions of a rimless spoked wheel: a simple three-dimensional system with impacts."*Dynamics and stability of systems* 12.3 (1997): 139-159.
41. Sinnet, R. W., Powell, M. J., Shah, R. P., & Ames, A. D. A human-inspired hybrid control approach to bipedal robotic walking. In 18th IFAC World Congress (pp. 6904-11), 2011.
42. Xianye Ben,Peng Zhang1,Rui Yan,Mingqiang Yang, Guodong Ge "Gait recognition and micro-expression recognition based on maximum margin projection with tensor representation" Neural Computing & Application 2015 .
43. M. Hobon, N. Lakbakbi Elyaaqoubi, G. Abba, Quasi optimal sagittal gait of a biped robot with a new structure of knee joint, Robotics and Autonomous Systems, Volume 62, Issue 4, April 2014, Pages 436-445.
44. Semwal, Vijay Bhaskar, Manish Raj, and G. C. Nandi. "Biometric gait identification based on a multilayer perceptron." Robotics and Autonomous Systems 65 (2015): 65-75.
45. Stevenage, Sarah V., Mark S. Nixon, and Kate Vince. "Visual analysis of gait as a cue to identity." Applied cognitive psychology 13.6 (1999): 513-526.
46. Dragomir N. Nenchev and Akinori Nishio “Ankle and hip strategies for balance recovery of a biped subjected to an impact”, Robotica, volume 26, pp. 643–653, 2008.
47. Jerry E.Pratt, JohnCarff, Sergey Drakunov and Ambarish Goswami. “Capture point: A step toward humanoid push recovery”. Proceedings of the IEEE-RAS International Conference on Humanoid Robots, pp. 200-207, 4-6dec. 2006.
48. Wentao Mao; Jeong-Jung Kim; Ju-Jang Lee, "Continuous steps toward humanoid push recovery," Automation and Logistics, 2009. ICAL ’09. IEEE International Conference on , vol., no., pp.7,12, 5-7 Aug. 2009.
49. Goswami, Ambarish, and Vinutha Kallem. "Rate of change of angular momentum and balance maintenance of biped robots." Robotics and Automation, 2004. Proceedings. ICRA'04. 2004 IEEE International Conference on. Vol. 4. IEEE, 2004.
50. B.Stephens “Humanoid push recovery,” in Proceedings of the IEEE-RAS International Conference on Humanoid Robots, pp. 589-595, 2007.
51. Jun Nakanishi, Jun Morimoto, Gen Endo , Gordon Cheng, Stefan Schaal, Mitsuo Kawato, Learning from demonstration and adaptation of biped locomotion, Robotics and Autonomous Systems 47 (2004) 79–91.
52. Ude, Aleš, Christopher G. Atkeson, and Marcia Riley. "Planning of joint trajectories for humanoid robots using B-spline wavelets." Robotics and Automation, 2000. Proceedings. ICRA'00. IEEE International Conference on. Vol. 3. IEEE, 2000.
53. Sanchez, N.; Acosta, A.M.; Stienen, A.H.A.; Dewald, J.P.A., "A Multiple Degree of Freedom Lower Extremity Isometric Device to Simultaneously Quantify Hip, Knee, and Ankle Torques," in IEEE Transactions on Neural Systems and Rehabilitation Engineering, vol.23, no.5, pp.765-775, Sept. 2015.

54. Semwal, Vijay Bhaskar, and Gora Chand Nandi. "Generation of Joint Trajectories Using Hybrid Automate-Based Model: A Rocking Block-Based Approach." *IEEE Sensors Journal* 16.14 (2016): 5805-5816.
55. Semwal, Vijay Bhaskar, et al. "Design of Vector Field for Different Subphases of Gait and Regeneration of Gait Pattern." *IEEE Transactions on Automation Science and Engineering* (2016).
56. Sajid Iqbal, Xizhe Zang, Yanhe Zhu, Jie Zhao, Bifurcations and chaos in passive dynamic walking: A review, Robotics and Autonomous Systems, Volume 62, Issue 6, June 2014, Pages 889-909.
57. Lee, Sung-Hee, and AmbarishGoswami. "Ground reaction force control at each foot: A momentum-based humanoid balance controller for non-level and non-stationary ground." in *Intelligent Robots and Systems (IROS), 2010 IEEE/RSJ International Conference on*. IEEE, 2010.
58. Cunado, David, Mark S. Nixon, and John N. Carter. "Using gait as a biometric, via phase-weighted magnitude spectra." Audio-and Video-based Biometric Person Authentication. Springer Berlin Heidelberg, 1997.
59. Knaepen, K.; Beyl, P.; Duerinck, S.; Hagman, F.; Lefeber, D.; Meeusen, R., "Human–Robot Interaction: Kinematics and Muscle Activity Inside a Powered Compliant Knee Exoskeleton," IEEE Transactions on Neural Systems and Rehabilitation Engineering, , vol.22, no.6, pp.1128,1137, Nov. 2014.
60. F.; Yamakita, M.; Kamamichi, N.; Zhi-Wei Luo, "A novel gait generation for biped walking robots based on mechanical energy constraint," IEEE Transactions on Robotics and Automation, , vol.20, no.3, pp.565,573, June 2004.
61. Wang, L., Ning, H., Tan, T., & Hu, W. (2004). Fusion of static and dynamic body biometrics for gait recognition. IEEE Transactions on Circuits and Systems for Video Technology, 14(2), 149-158.
62. Yildirim, Şahin, İkbal Eski, and Yahya Polat. "Design of adaptive neural predictor for failure analysis on hip and knee joints of humans." Neural Computing and Applications 23.1 (2013): 73-87.
63. Kale, A., Sundaresan, A., Rajagopalan, A. N., Cuntoor, N. P., Roy-Chowdhury, A. K., Kruger, V., & Chellappa, R. (2004). Identification of humans using gait IEEE Transactions on Image Processing, 13(9), 1163-1173.
64. Su-Bin Joo, Seung Eel Oh, Taeyong Sim, Hyunggun Kim, Chang Hyun Choi, Hyeran Koo, Joung Hwan Mun, "Prediction of gait speed from plantar pressure using artificial neural networks", Expert Systems with Applications, Volume 41, Issue 16, 15 November 2014, Pages 7398-7405.

65. Sepulveda, F., Wells, D. M., & Vaughan, C. L. (1993). A neural network representation of electromyography and joint dynamics in human gait Journal of biomechanics, 26(2), 101-109.
66. S.Kajita, F.Kanehiro, K.Kaneko, K.Yokoi and H.Hirukawa. “The 3d linear inverted pendulum mode: a simple modeling for a biped walking pattern generation”. Proceedings of the IEEE/RSJ International Conference on Intelligent Robots and Systems, vol. 1, pp. 239-246, 2001.
67. S Kajita and K Tani, ”Study of Dynamic Biped Locomotion on Rugged Terrain - Derivation and Application of the Linear Inverted Pendulum”. Proceedings of the 1991 IEEE International Conference on Robotics and Automation, vol. 2,pp. 1405-1411, 1991.
68. Agostini, V.; Balestra, G.; Knaflitz, M., "Segmentation and Classification of Gait Cycles," in Neural Systems and Rehabilitation Engineering, IEEE Transactions, vol.22, no.5, pp.946,952, Sept. 2014.
69. Hyunglae Lee; Hogan, N., "Time-Varying Ankle Mechanical Impedance During Human Locomotion," in IEEE Transactions on Neural Systems and Rehabilitation Engineering, vol.23, no.5, pp.755-764, Sept. 2015
70. Alur, Rajeev, et al. "The algorithmic analysis of hybrid systems." Theoretical computer science 138.1 (1995): 3-34.
71. Jeong, Man-Yong, and In-Young Yang. "Characterization on the rocking vibration of rigid blocks under horizontal harmonic excitations." International Journal of Precision Engineering and Manufacturing 13.2 (2012): 229-236.
72. Peña, Fernando, et al. "On the dynamics of rocking motion of single rigid–block structures." Earthquake Engineering and Structural Dynamics , Wiley InterScience ;(2007) 36:2383–2399.
73. Lygeros, J.; Johansson, K.H.; Simic, S.N.; Jun Zhang; Sastry, S.S., "Dynamical properties of hybrid automata," IEEE Transactions on Automatic Control, , vol.48, no.1, pp.2,17, Jan 2003 .
74. Nakaoka, Shin'ichiro, et al. "Learning from observation paradigm: Leg task models for enabling a biped humanoid robot to imitate human dances." The International Journal of Robotics Research 26.8 (2007): 829-844.
75. C. Tomlin, G. Pappas, and S. Sastry, “Conflict resolution for air traffic management: A study in multiagent hybrid systems,” IEEE Trans. Automat. Contr., vol. 43, pp. 509–521, Apr. 1998.
76. A. Balluchi, L. Benvenuti, M. Di Benedetto, C. Pinello, and A. Sangiovanni-Vincentelli, “Automotive engine control and hybrid systems: Challenges and opportunities,” Proc. IEEE, vol. 7, pp. 888–912, July 2000.
77. R. Horowitz and P. Varaiya, “Control design of an automated highway system,” Proc. IEEE, vol. 88, no. 7, pp. 913–925, July 2000.

78. S. Engell, S. Kowalewski, C. Schultz, and O. Strusberg, “Continuous discrete interactions in chemical process plants,” Proc. IEEE, vol. 7, pp. 1050–1068, July 2000.
79. B. Lennartsson, M. Tittus, B. Egardt, and S. Pettersson, “Hybrid systems in process control,” Control Syst. Mag., vol. 16, no. 5, pp. 45–56, 1996.
80. S. J. Hogan, “On the dynamics of a rigid-block motion under harmonic forcing,” Proc. Royal Society, A, vol. 425, pp. 441–476, 1989.
81. Henzinger, Thomas A., et al. "What's decidable about hybrid automata?." Proceedings of the twenty-seventh annual ACM symposium on Theory of computing. ACM, 1995.
82. Nurfarahin Onn, Mohamed Hussein, Collin Howe Hing Tang, MohdZarhamdyMd Zain, Maziah Mohamad And Lai Wei Ying ‘Motion Control of Seven-Link Human Bipedal Model’ by Faculty of Mechanical Engineering University Technology Malaysia.
83. Wolfram, Stephen. "Cellular automata." Los Alamos Science 9 (1983): 2-27.
84. Weisstein, Eric W. "Elementary cellular automaton." (2002).
85. McGeer, Tad. "Passive dynamic walking." the international journal of robotics research 9, no. 2 (1990): 62-82.
86. S.T. Venkataraman, A simple legged locomotion gait model, Robotics and Autonomous Systems, Volume 22, Issue 1, 10 November 1997, Pages 75-85.
87. Nakaoka, Shin'ichiro, et al. "Learning from observation paradigm: Leg task models for enabling a biped humanoid robot to imitate human dances." The International Journal of Robotics Research 26.8 (2007): 829-844.
88. Qi, Y.; Soh, Cheong-Boon; Gunawan, E.; Low, Kay-Soon; Thomas, R., "Assessment of Foot Trajectory for Human Gait Phase Detection Using Wireless Ultrasonic Sensor Network," Neural Systems and Rehabilitation Engineering, IEEE Transactions on , vol.PP, no.99, pp.1,1.
89. M. Hoffmann, K. Štěpánová, M. Reinstein, The effect of motor action and different sensory modalities on terrain classification in a quadruped robot running with multiple gaits, Robotics and Autonomous Systems, Available online 24 July 2014.
90. Iamsa-at, S.; Horata, P., "Handwritten Character Recognition Using Histograms of Oriented Gradient Features in Deep Learning of Artificial Neural Network," IT Convergence and Security (ICITCS), 2013 International Conference on , vol., no., pp.1,5, 16-18 Dec. 2013.
91. Baptista, Darío, and Fernando Morgado-Dias. "A survey of artificial neural network training tools." Neural Computing and Applications 23.3-4 (2013): 609-615.
92. Gao, S.; Zhang, Y.; Jia, K.; Lu, J.; Zhang, Y., "Single Sample Face Recognition via Learning Deep Supervised Auto-Encoders," Information Forensics and Security, IEEE Transactions on , vol.PP, no.99, pp.1,1.

93. Ibrahim, Ronny K., et al. "Gait pattern classification using compact features extracted from intrinsic mode functions." Engineering in Medicine and Biology Society, 2008. EMBS 2008. 30th Annual IEEE International Conference, 2008.
94. Wang, Liang, Huazhong Ning, Tieniu Tan, and Weiming Hu. "Fusion of static and dynamic body biometrics for gait recognition." IEEE Transactions on Circuits and Systems for Video Technology, 14, no. 2 (2004): 149-158.
95. Vega, I. R., & Sarkar, S. (2003). Statistical motion model based on the change of feature relationships: human gait-based recognition. IEEE Transactions on Pattern Analysis and Machine Intelligence, 25(10), 1323-1328.
96. Kankal, Murat, and Ömer Yüksek. "Artificial neural network for estimation of harbor oscillation in a cargo harbor basin." Neural Computing and Applications 25.1 (2014): 95-103.
97. Huang, P. S., C. J. Harris, and Mark S. Nixon. "Human gait recognition in canonical space using temporal templates." IEE Proceedings-Vision, Image and Signal Processing 146.2 (1999): 93-100.
98. Wang, C., Zhang, J., Wang, L., Pu, J., & Yuan, X. (2012). Human identification using temporal information preserving gait template. IEEE Transactions on, Pattern Analysis and Machine Intelligence, 4(11), 2164-2176.
99. Joachim Steigenberger, Carsten Behn, Gait generation considering dynamics for artificial segmented worms, Robotics and Autonomous Systems, Volume 59, Issues 7–8, July–August 2011, Pages 555-562.
100. Keller, J. M., Gray, M. R., & Givens, J. A. (1985). A fuzzy k-nearest neighbor algorithm. , IEEE Transactions on Systems, Man and Cybernetics, (4), 580-585.
101. Selim, S. Z., & Ismail, M. A. (1984). K-means-type algorithms: a generalized convergence theorem and characterization of local optimality. , IEEE Transactions on Pattern Analysis and Machine Intelligence, (1), 81-87.
102. Stepien, Pawel. "Sliding window empirical mode decomposition-its performance and quality." *EPJ Nonlinear Biomedical Physics* 2.1 (2014): 1.
103. Tolwinski, Susan. "The Hilbert Transform and Empirical Mode Decomposition as Tools for Data Analysis." *Tucson: University of Arizona* (2007).
104. Gordon, Sean W., and Napoleon H. Reyes. "A Method for computing the Balancing Positions of a Humanoid Robot." NZCSRSC 2008, April 2008.
105. Lian, Ruey-Jing. "Design of an enhanced adaptive self-organizing fuzzy sliding-mode controller for robotic systems." Expert Systems with Applications 39, no. 1 (2012): 1545-1554.
106. J. L. Castro ,"Fuzzy Logic Controllers Are Universal Approximates" IEEE Transactions On Systems, Man, And Cybernetics, Vol 25, No. 4,Pp.629-635, April 1995.
107. L. Zadeh, "Fuzzy sets," Information Control, vol. 8, pp. 338–353, 1965.

108. Nilesh N. Karnik, Jerry M. Mendel, Qilian Liang, “Type-2 Fuzzy Logic Systems” IEEE Transactions On Fuzzy Systems, Vol. 7, No. 6, December 1999, pp 643-657.
109. C. C. Lee, “Fuzzy logic in control systems: Fuzzy logic controller: Part I,” IEEE Trans. Syst. Man Cyber, vol. SMC-20, no. 2, pp. 404418, 1990.
110. Edward Tunstel, Marco A. A. de Oliveira, Sigal Berman,"Fuzzy Behavior Hierarchies for Multi-Robot Control"International Journal Of Intelligent Systems, VOL. 17, 449–470 -2002.
111. Chen, J.Y.C.; Barnes, M.J., "Human–Agent Teaming for Multirobot Control: A Review of Human Factors Issues," Human-Machine Systems, IEEE Transactions on , vol.44, no.1, pp.13,29, Feb. 2014.
112. Yesil, Engin. "Interval type-2 fuzzy PID load frequency controller using Big Bang–Big Crunch optimization." Applied Soft Computing 15 (2014): 100-112.
113. Michio Sugeno and Takahiro Yasukawa, “A Fuzzy-Logic-Based Approach to QualitativeModeling” ,IEEE Transactions On Systems, Man, And Cybernetics—Part C: Applications And Reviews, Vol. 42, No. 5, September 2012 pp728-742.
114. Gasilov, N. A., A. G. Fatullayev, Ş. E. Amrahov, and A. Khastan. "A new approach to fuzzy initial value problem." Soft Computing 18, no. 2 (2014): 217-225.
115. Mendel, J.M.; Xinwang Liu, "Simplified Interval Type-2 Fuzzy Logic Systems," Fuzzy Systems, IEEE Transactions on , vol.21, no.6, pp.1056,1069, Dec. 2013.
116. Harmati, István, and Krzysztof Skrzypczyk. "Robot team coordination for target tracking using fuzzy logic controller in game theoretic framework." Robotics and Autonomous Systems 57.1 (2009): 75-86.
117. M. Raj, V. Bhaskar Semwal, G.C.Nandi. Hybrid Model for Passive Locomotion Control of a Biped Humanoid:The Artificial Neural Network Approach, International Journal of Interactive Multimedia and Artificial Intelligence, (2017).
118. Nandi, Gora Chand, et al. "Modeling bipedal locomotion trajectories using hybrid automata." *Region 10 Conference (TENCON), 2016 IEEE*. IEEE, 2016.
119. Raj, Manish, Vijay Bhaskar Semwal, and G. C. Nandi. "Bidirectional association of joint angle trajectories for humanoid locomotion: the restricted Boltzmann machine approach." *Neural Computing and Applications* (2016): 1-9.

## Publications

## Journals:

1. Vijay Bhaskar Semwal and G.C. Nandi, "Toward Developing a Computational Model for Bipedal Push Recovery–A Brief," in IEEE Sensors Journal, , vol.15, no.4, pp.2021-2022, April 2015 .

2. Vijay Bhaskar Semwal and G.C. Nandi., “Design of vector field for different subphases of gait and regeneration of gait pattern” IEEE Transactions on Automation Science and Engineering .
3. Vijay Bhaskar Semwal and G.C. Nandi.,“Generation of Joints Trajectories for Bipedal locomotion using Hybrid automata model: A Rocking block based Approach” IEEE sensor Journals.
4. Semwal, Vijay Bhaskar, Pavan Chakraborty, and Gora Chand Nandi. "Less computationally intensive fuzzy logic (type-1)-based controller for humanoid push recovery." *Robotics and Autonomous Systems* 63 (2015): 122-135.
5. Semwal, Vijay Bhaskar, Manish Raj, and Gora Chand Nandi. "Biometric gait identification based on a multilayer perceptron." *Robotics and Autonomous Systems* 65 (2015): 65-75.
6. Semwal, Vijay Bhaskar, et al. "Biologically-inspired push recovery capable bipedal locomotion modeling through hybrid automata." *Robotics and Autonomous Systems* 70 (2015): 181-190.
7. Semwal, Vijay Bhaskar, Kaushik Mondal, and G. C. Nandi. "Robust and accurate feature selection for humanoid push recovery and classification: deep learning approach." Neural Computing and Applications (2015): 1-10.

## Conferences:

8. Semwal, Vijay Bhaskar, Aparajita Bhushan, and G. C. Nandi. "Study of humanoid Push recovery based on experiments." *Control, Automation, Robotics and Embedded Systems (CARE), 2013 International Conference on*. IEEE, 2013.
9. Semwal, Vijay Bhaskar, et al. "Biped model based on human Gait pattern parameters for sagittal plane movement." *Control, Automation, Robotics and Embedded Systems (CARE), 2013 International Conference on*. IEEE, 2013.
10. Vijay Bhaskar Semwal and G.C. Nandi., "Modeling Bipedal Locomotion Trajectories Using Hybrid Automata," IEEE International Conference Tencon 16.